%% file: main.tex
\documentclass[10pt,a4paper,twocolumn,logo]{nvidiatechreport}

\usepackage[numbers,sort&compress]{natbib}
\usepackage{subcaption}
\usepackage{placeins}
\usepackage[normalem]{ulem}
\usepackage{colortbl}
\definecolor{linkblue}{rgb}{0.21,0.49,0.74}
\definecolor{resultbest}{rgb}{0.00,0.38,0.25}
\definecolor{resultsecond}{rgb}{0.00,0.30,0.65}

\definecolor{rank1}{HTML}{66CC80}
\definecolor{rank2}{HTML}{8CD98B}
\definecolor{rank3}{HTML}{B3E596}
\definecolor{rank4}{HTML}{D9F2A0}
\definecolor{rank5}{HTML}{FFFFAB}
\newcommand{\rankcell}[2]{\cellcolor{rank#1}#2}
\newcommand{\ranklegend}{\colorbox{rank1}{1st}\,\colorbox{rank2}{2nd}\,\colorbox{rank3}{3rd}\,\colorbox{rank4}{4th}\,\colorbox{rank5}{5th}}
\newsavebox{\validationtablebox}
\newsavebox{\validationentrybox}
\makeatletter
\newcommand{\objectsubsection}[3]{%
  \begingroup
  \renewcommand{\subsection}{\@startsection{subsection}{2}{\z@}{0pt}{2pt}{\normalfont\small\bfseries\raggedright}}%
  \subsection{#1 object #2}\label{sec:object_#3_#2}%
  \endgroup
}
\makeatother

\newcommand{\supp}[1]{Appendix~\ref{#1}}
\newcommand{\Supp}[1]{Appendix~\ref{#1}}
\newcommand{\supplink}[2]{\hyperref[#1]{#2}}

\hypersetup{pdftitle={DOPER: Deep Object Pose Estimation Revisited},pdfauthor={Hugo Hadfield, Alperen Degirmenci, Jonathan Tremblay}}

\title{DOPER: Deep Object Pose Estimation Revisited}

\author{Hugo Hadfield, Alperen Degirmenci, and Jonathan Tremblay}
\affil{NVIDIA}
\newcommand{\paperabstract}{%
We present DOPER, an RGB pose estimation framework that shares a visual backbone across object detection, segmentation, and keypoint prediction to recover the 6D pose of known objects through Perspective-$n$-Point.
Starting from synthetic training, DOPER adapts to real observations using partial supervision while retaining synthetic keypoint labels.
Calibrated multi-view observations provide geometric constraints for keypoint supervision.
During inference, optional heatmap refinement fits a rigid pose directly to the predicted keypoint distributions, resolving ambiguous correspondences without depth input or a learned pose refiner.
Starting from underspecified CAD models, DOPER progressively refines the object representation and pose estimator. Results on SPEED+ and SHIRT further demonstrate effective adaptation to illumination changes and accurate temporal pose estimation.
Across the evaluated single-view and multi-view configurations, DOPER achieves state-of-the-art detection AP results on HANDAL, HOPEv2, HOT3D, and T-LESS while using only RGB
and running in real time (35\,Hz per object model on one GB300).
These results demonstrate the effectiveness of jointly learning detection and pose keypoints, with real observations and geometric constraints. 
}

\newcommand{\objectlabel}[3]{%
  \par\noindent{\small\bfseries #1 object #2}\par\vspace{2pt}%
}

\begin{abstract}
\textbf{Abstract:} \paperabstract
\newline
\textbf{Project page:} \href{https://nvlabs.github.io/DOPER/}{nvlabs.github.io/DOPER}\quad
\textbf{Code:} \href{https://github.com/NVlabs/DOPER}{github.com/NVlabs/DOPER}
\end{abstract}
\begin{document}
\maketitle

\suppressfloats[t]
\begin{figure}[t]
  \centering
  \includegraphics[width=\columnwidth]{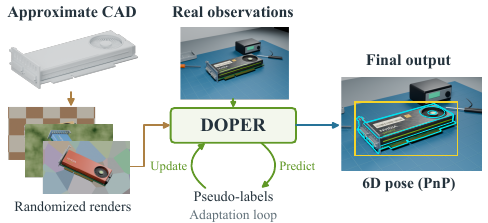}
  \caption{
  An approximate CAD model supplies DR training data. Real observations drive iterative refinement.
  The updated estimator predicts bounding box, mask, and keypoints for P$n$P pose recovery.}
  \label{fig:doper_overview}
\end{figure}

\section{Introduction}
\label{sec:intro}

Object pose estimation is fundamental to robotic manipulation. 
Given a 3D model, the goal is to first discover the object in the image then recover 3D position and orientation relative to a reference coordinate frame. 
Accurate estimates enable tasks such as grasping a small cable connector and inserting it into a tight-fitting receptacle. 
Existing approaches range from classical geometric methods, such as Perspective-$n$-Point (P$n$P), to learning-based render-and-compare methods. 
In practice, however, the available 3D model often differs from the physical object: textures may be inaccurate, geometry may be incomplete, and external components such as tags may be absent. These discrepancies challenge pose estimation in real-world settings.

Bridging this model-to-reality gap requires robustness to both appearance variation and geometric discrepancies.
One way to bridge this gap is to expose a pose estimator to sufficient variation during training, so that it learns to tolerate differences between synthetic images and real observations~\cite{tremblay2018plans,tremblay2018dope}.
Scaling this strategy to diverse object collections has enabled pose estimation for previously unseen objects, substantially reducing the need for object-specific training~\cite{labbe2023megapose,wen2024foundationpose}. Unfortunately, generalization to new objects does not by itself resolve errors in the supplied model: rendering that model under more viewpoints or lighting conditions cannot recover a missing component.
Real observations therefore provide information that synthetic training alone may lack. 
Even a few reference images can support pose estimation for a new object~\cite{he2022fs6d,geng2025oneview}, while test-time adaptation allows an estimator to learn from its deployment environment without requiring annotated target data~\cite{huang2025ttapose,shi2025crisp,lee2023ttacope}. 

Despite recent advances, the precision required for robotic manipulation remains challenging.
We introduce Deep Object Pose Estimation Revisited (DOPER), a pose estimation framework that uses real-world observations to progressively improve accuracy.
Starting from an incomplete 3D model, DOPER is initially trained on synthetic images generated using domain randomization~\cite{tremblay2018dope}. 
It jointly predicts object bounding boxes, segmentation masks, quantized orientations, and keypoint heatmaps, with the latter used to recover the final pose through P$n$P. 
The initial estimator lacks the desired precision because it uses an underspecified 3D model. However, it provides a useful starting point for adaptation (Figure~\ref{fig:doper_overview}).
We introduce two complementary test-time adaptation mechanisms that build on these predictions, enabling DOPER to achieve competitive results.

We use \textit{monocular} or \textit{multi-view} adaptation to generate supervision from the deployment environment.
For monocular observations, we keep only labels that do not require a precise metric pose: bounding boxes, segmentation masks, and quantized orientations.
With multi-view observations, obtained from stereo cameras or a moving camera,
geometric constraints additionally support reliable keypoint supervision through triangulation. 
We then fine-tune the estimator using the resulting supervision. 
The updated model can generate further pseudo-labels or be deployed. This iterative process is key for real-world deployment. 
We make the following contributions: 

\begin{itemize}
    \item We introduce DOPER, an object-specific framework that jointly predicts detections and pose keypoints supporting real-time 6D pose estimation.
    \item We propose an adaptation protocol from noisy pseudo-labels (single-view) to geometric supervision (multi-view) observations to refine underspecified 3D models, and improve pose detection and estimation.
    \item We demonstrate state-of-the-art results across HANDAL~\cite{guo2023handal}, HOT3D~\cite{banerjee2025hot3d}, HOPEv2~\cite{tyree2022hope,nguyen2025bop2024}, T-LESS~\cite{hodan2017tless}, and the satellite pose benchmarks SPEED+~\cite{park2022speedplus} and SHIRT~\cite{park2023shirt}.
\end{itemize}

\section{Related Work}

\noindent\textbf{Object pose estimation.}
Learning-based object pose estimators can be organized by how they recover pose: geometric estimation from correspondences, direct pose regression, and render-and-compare alignment.
\emph{Correspondence-based methods} predict keypoints or dense matches and recover pose using P$n$P for 2D--3D correspondences, or point-set alignment for 3D--3D correspondences~\cite{tremblay2018dope,peng2019pvnet,rad2017bb8,tekin2018seamless,he2020pvn3d,he2021ffb6d,hodan2020epos,su2022zebrapose}.
HccePose(BF)~\cite{wang2025hccepose} predicts front and back surface coordinates from detected object crops and samples points between them to construct ultra-dense correspondences for P$n$P, while DOPER jointly predicts detections and sparse keypoints.
Recent methods extend correspondence learning to unseen objects using reference observations~\cite{sun2022onepose,he2022oneposepp}: COG~\cite{che2026cog} estimates confidence-aware soft correspondences between a reference and query RGB-D image, PANY~\cite{xu2026pany} combines multi-view geometry and cross-view matching, recovering pose through P$n$P or point-set alignment depending on depth availability.
\emph{Direct regression methods} learn the mapping to pose parameters~\cite{wang2021gdrnet,xiang2018posecnn,wang2019densefusion,di2021sopose,jiang2022uni6d,ha2026race6d}. 
\emph{Render-and-compare methods} align rendered object views with observations to score and refine pose hypotheses~\cite{labbe2023megapose,wen2024foundationpose,li2018deepim,labbe2020cosypose,iwase2021repose,wang2025scflow2}.
DOPER follows the keypoint-and-P$n$P approach and specializes it to the deployment environment through iterative learning from real observations and refinement of an initially incomplete object representation.

\noindent\textbf{Self-supervised learning and test-time adaptation.}
Real observations can improve synthetically trained pose estimators without requiring manual pose annotations~\cite{wang2021self6dpp,lee2023ttacope,huang2025ttapose,wang2020self6d,sock2020poseconsistency,lee2022udacope,yang2021dscposenet,fu2022wild6d,wu2024rkhspose}.
Adaptation has also been coupled with object reconstruction~\cite{shi2025crisp,tan2025ondapose,chen2023texpose,manhardt2020cpspp,wang2026gfreedet2,floros2026novasplat}. 
DOPER builds on these directions through complementary monocular and multi-view supervision, and uses an object reconstruction from onboarding videos to correct an underspecified CAD model when rendering training data. 
Signals from pretrained models support repeated training on real observations and randomized CAD renders, targeting the precision required for robotic manipulation.

\section{Method}
\label{sec:method}

\begin{figure*}[!t]
  \centering
  \includegraphics[width=0.90\textwidth]{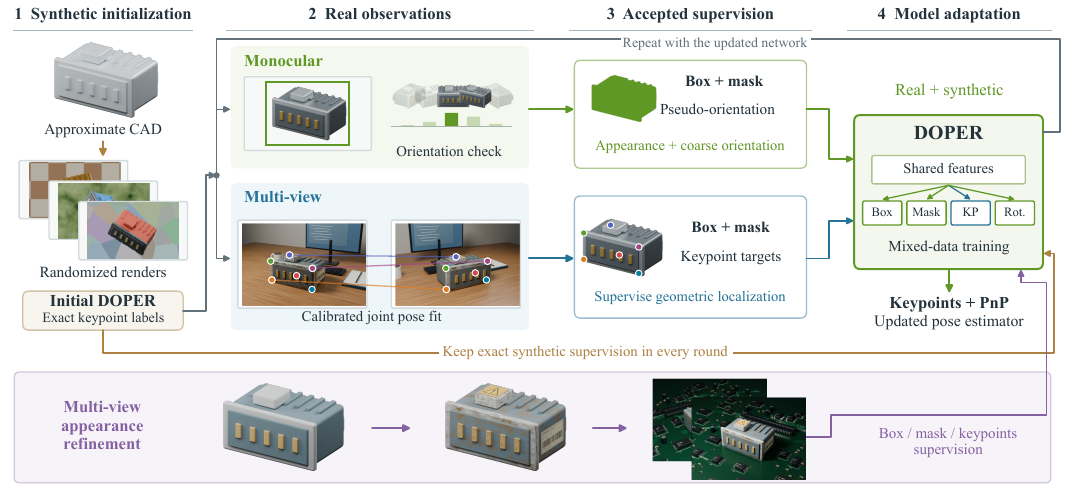}
  \caption{
  Monocular observations provide box, mask, and coarse-orientation targets (top row).
  Multi-view observations additionally provide keypoint targets and refine object appearance for new training renders (bottom row).
  Mixed real and synthetic supervision updates the estimator,
  which labels subsequent observations.
  }
  \label{fig:doper_method}
\end{figure*}

\noindent\textbf{Overview and problem formulation.}
DOPER combines synthetic supervision and real observations to adapt an object-specific pose estimator (Figure~\ref{fig:doper_method}).
Given a calibrated RGB image, an approximate CAD model, and ordered object-frame keypoints, we estimate object rotation and translation relative to the camera.
The CAD appearance and surface details may be incomplete, but metric scale must be known. 
Keypoint identities and their coordinate frame remain fixed throughout adaptation.

\medskip
\noindent\textbf{Shared visual representation and geometric pose estimation.}
Following the correspondence-based formulation of DOPE~\cite{tremblay2018dope}, we predict image locations of known object points and recover pose with P$n$P.
Our implementation shares a DINOv3-pretrained ConvNeXt backbone~\cite{simeoni2025dinov3} between a full-image branch and an object-crop branch.
The full-image branch combines multi-scale features at stride four and predicts object centers, bounding-box dimensions, center offsets, and a foreground mask.
Detection follows the CenterNet formulation~\cite{zhou2019objects}.
For each detected box, the crop branch resamples the original image at higher resolution and applies the same backbone weights, followed by a keypoint head.
This preserves small-object detail while sharing detection and segmentation supervision with keypoint localization.
The head predicts a spatial distribution per keypoint; decoded coordinates are mapped to the original image by the inverse crop transform.
Confidence filtering and robust P$n$P select consistent 2D--3D correspondences and minimize squared reprojection error, accounting for camera calibration and lens distortion.

\medskip
\noindent\textbf{Synthetic initialization and partial supervision.}
We render the initial CAD model under variations in viewpoint, distance, lighting, material appearance, backgrounds, and distractors, following domain randomization~\cite{tremblay2018plans,tremblay2018dope}.
Each synthetic sample provides a bounding box, a visible-object mask, and projected keypoints in the fixed object frame.
Photometric augmentation models additional appearance variation, while crop translation, scale, and rotation augmentation expose the keypoint branch to imperfect detections.
All geometric augmentations are applied consistently to the keypoint targets.
We optimize a weighted sum of detection, mask, and keypoint objectives jointly.
The detection term combines a center-heatmap focal loss with absolute-error losses for box dimensions and offsets; the mask term combines binary cross-entropy and soft Dice losses~\cite{milletari2016vnet}.
For keypoints, we use cross-entropy between normalized Gaussian target heatmaps and spatially normalized predictions, equivalent to minimizing their KL divergence up to a target-dependent constant.
Losses are normalized over available supervision, with validity-weighted keypoint terms and zero contribution from missing annotations.
This allows fully labeled synthetic and partially labeled real images to train one model.

\medskip
\noindent\textbf{Monocular adaptation.}
We retain high-confidence detections from unlabeled real images and use their bounding boxes to prompt a general-purpose segmentation model~\cite{ravi2025sam2}.
Alternatively, a reliable pose supplies a projected CAD mask when the model adequately represents the object and external occlusions are absent.

Optionally, a rotation quantization head provides additional supervision from
monocular observations by predicting a distribution over reference
orientations, with soft targets reflecting their angular proximity.
Agreement between this distribution and the P$n$P estimate helps select
reliable pseudo-orientations for adapting the shared representation.

Monocular targets comprise boxes, masks, and accepted coarse orientations.
Although the teacher uses keypoints and P$n$P to assess predictions, we exclude its projected keypoints from real training labels: low reprojection error can conceal an incorrect symmetric pose or localization bias.
Auxiliary targets adapt the shared representation to real appearance, while synthetic samples retain keypoint supervision.

\medskip
   \noindent\textbf{Multi-view supervision.}
We jointly fit one object pose to confident teacher keypoints in paired images with known intrinsics and fixed relative camera poses, using a robust reprojection objective.
Fits must satisfy per-view reprojection-error and correspondence-count thresholds.
Reprojecting accepted model keypoints into each camera gives geometrically consistent targets instead of copying independent teacher predictions.
For mesh-based exports, CAD projection also supplies silhouette masks and enclosing boxes, supervising detection, segmentation, and keypoint prediction.

\medskip
\noindent\textbf{Iterative adaptation and deployment.}
Each round fine-tunes the current estimator with a fresh optimizer and a fixed mixture of synthetic, monocular, and multi-view data.
Every batch retains exact synthetic keypoint labels; other sources contribute only their available targets.
The backbone uses a learning rate one tenth that of the prediction heads in the reference training recipe.
The updated model can then label additional captures for a subsequent round or be deployed using the same detection--crop--keypoint--P$n$P pipeline.
Unlike source-free online adaptation~\cite{lee2023ttacope}, this procedure retains synthetic rehearsal and includes review of generated labels.

\begin{figure}[t]
  \centering
  \includegraphics[width=0.86\columnwidth]{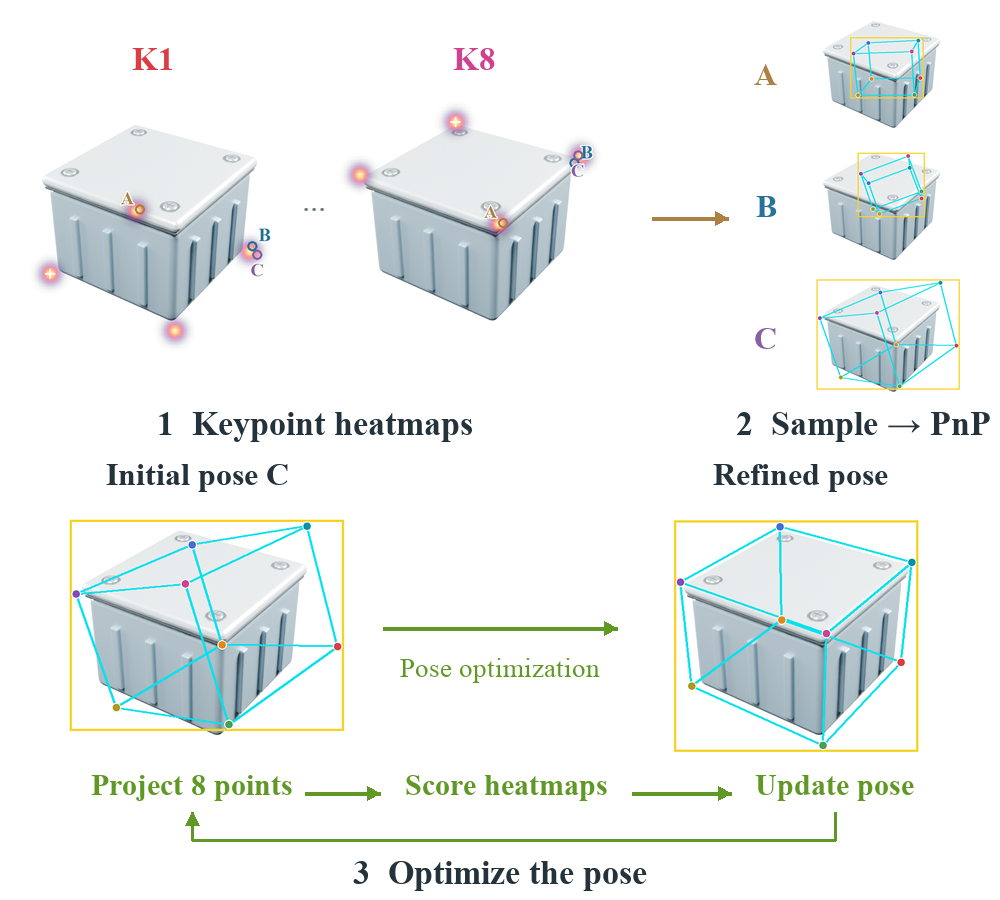}
  \caption{Heatmap refinement: ambiguous heatmaps for eight CAD corners (1), sampled P$n$P hypotheses (2), and pose optimization from hypothesis C (3). The optimizer scores projected keypoints jointly. Cyan cuboids and yellow boxes show pose projections.}
  \label{fig:heatmap_refinement}
\end{figure}

\medskip
\noindent\textbf{Heatmap keypoint selection.}
Symmetries and uncertain image evidence can produce multiple peaks in a keypoint heatmap, making independent selection of image locations ambiguous, see Figure~\ref{fig:heatmap_refinement}.
Using heatmap optimization, we can project the CAD keypoints into their corresponding heatmaps and bilinearly sample the spatial log probabilities.
Bounded non-linear optimization adjusts the object's rotation and translation to minimize the summed negative log probabilities, moving all keypoints together through one rigid pose.
For multiple views, we sum the same objective over associated observations while keeping the cameras fixed.
We accept an update only if it improves the objective and keeps the keypoints in front of every participating camera.

\begin{table*}[!t]
  \centering
  \caption{HANDAL, HOPEv2, HOT3D, and T-LESS test results (\%, higher is better). \ranklegend{} ranks the five highest distinct values in each accuracy column; ties share a color. DOPER time is mean seconds per image with all object models of the dataset (40, 28, 33, and 30 models) on one GB300; timings are unshaded.
  $\dagger$: On HANDAL, HOPEv2, and HOT3D, HccePose was trained on PBR images and used BOP-provided CNOS/FastSAM detections at inference using recommended method settings.
  $\ddagger$: The same PBR-trained pose models use DOPER detections (see \supp{sec:hccepose_training}).
  }
  \label{tab:BOP_results}
  \label{tab:tless_test}
  \footnotesize
  \setlength{\tabcolsep}{1.8pt}
  \resizebox{\textwidth}{!}{%
  \begin{tabular}{@{}ll|rrrr|rrrr|rrrr|rrrr@{}}
    \toprule
    & & \multicolumn{4}{c|}{HANDAL~\cite{guo2023handal}} & \multicolumn{4}{c|}{HOPEv2~\cite{tyree2022hope,nguyen2025bop2024}} & \multicolumn{4}{c|}{HOT3D~\cite{banerjee2025hot3d}} & \multicolumn{4}{c}{T-LESS~\cite{hodan2017tless}} \\
    \cmidrule(lr){3-6}\cmidrule(lr){7-10}\cmidrule(lr){11-14}\cmidrule(lr){15-18}
    Method & Input & AP & MSSD & MSPD & Time (s) & AP & MSSD & MSPD & Time (s) & AP & MSSD & MSPD & Time (s) & AP & MSSD & MSPD & Time (s) \\
    \midrule
    \multicolumn{18}{@{}l}{\textit{Model-based, unseen objects}} \\
    FoundPose~\cite{ornek2024foundpose} & RGB & -- & -- & -- & -- & -- & -- & -- & -- & -- & -- & -- & -- & -- & -- & -- & -- \\
    GigaPose~\cite{nguyen2024gigapose} & RGB & 4.1 & 1.8 & 6.5 & 0.82 & 16.7 & 4.6 & 28.8 & 1.14 & 7.2 & 1.6 & 12.9 & 0.65 & 25.5 & 10.0 & 41.0 & 0.96 \\
    GigaPose + GenFlow~\cite{moon2024genflow} & RGB & 25.6 & 23.9 & 27.3 & 4.36 & 41.1 & 34.6 & 47.5 & 6.73 & 26.8 & 24.2 & 29.3 & 4.72 & 43.6 & 40.5 & 46.7 & 5.99 \\
    GigaPose + GenFlow (5 hypo)~\cite{nguyen2024gigapose,moon2024genflow} & RGB & -- & -- & -- & -- & -- & -- & -- & -- & -- & -- & -- & -- & 56.5 & 52.6 & 60.3 & 22.03 \\
    OPFormer (CNOS)~\cite{moroz2026opformer} & RGB & 19.2 & 13.4 & 25.0 & 0.48 & 35.1 & 27.2 & 43.1 & 1.19 & -- & -- & -- & -- & -- & -- & -- & -- \\
    OPFormer + MegaPose~\cite{moroz2026opformer} & RGB & 26.2 & 23.0 & 29.4 & 2.33 & 39.2 & 33.7 & 44.7 & 5.77 & -- & -- & -- & -- & -- & -- & -- & -- \\
    Co-op (CNOS, coarse)~\cite{moon2025coop} & RGB & 39.3 & 34.8 & 43.8 & 2.41 & 46.8 & 39.1 & 54.5 & 3.25 & 32.6 & 28.4 & 36.7 & 2.35 & -- & -- & -- & -- \\
    Co-op (CNOS, 1 hypo)~\cite{moon2025coop} & RGB & 45.4 & 43.2 & 47.6 & 4.78 & 47.7 & 40.5 & 54.9 & 7.48 & 38.8 & 37.1 & 40.4 & 4.15 & 58.8 & 54.9 & 62.7 & 9.09 \\
    Co-op (MUSE, 1 hypo)~\cite{moon2025coop} & RGB & 47.7 & 45.0 & 50.4 & 3.24 & 51.4 & 43.5 & 59.3 & 4.86 & 40.1 & 38.3 & 41.9 & 1.96 & -- & -- & -- & -- \\
    Co-op (F3DT2D, 5 hypo)~\cite{moon2025coop} & RGB & -- & -- & -- & -- & -- & -- & -- & -- & -- & -- & -- & -- & 61.6 & 57.5 & 65.7 & 9.29 \\
    3PT-Pose (Nov.\ 2025)~\cite{kalra2026threept} & MV RGB & \rankcell{5}{65.7} & \rankcell{5}{59.5} & 71.8 & 14.52 & 55.1 & 45.6 & 64.5 & 15.29 & 55.3 & \rankcell{5}{49.8} & 60.8 & 18.22 & 77.5 & 70.3 & 84.7 & 14.08 \\
    3PT-Pose-H3~\cite{kalra2026threept} & RGB-D & 64.6 & 56.9 & \rankcell{5}{72.4} & 38.14 & 59.9 & 57.2 & 62.6 & 50.83 & 51.3 & 45.6 & 57.0 & 126.20 & -- & -- & -- & -- \\
    FoundationPose (SAM-FP)~\cite{wen2024foundationpose,bop2026samfp} & RGB-D & -- & -- & -- & -- & \rankcell{5}{66.8} & \rankcell{1}{65.3} & 68.3 & 477.73 & -- & -- & -- & -- & -- & -- & -- & -- \\
    FreeZeV2.2~\cite{caraffa2025freezev2} & RGB-D & -- & -- & -- & -- & -- & -- & -- & -- & -- & -- & -- & -- & 77.7 & 76.8 & 78.7 & 30.06 \\
    FRTPose-WAPR.v2 (multi-detectors)~\cite{wang2026wapr} & RGB-D & -- & -- & -- & -- & -- & -- & -- & -- & -- & -- & -- & -- & \rankcell{5}{83.1} & \rankcell{3}{82.2} & 84.0 & 556.69 \\
    \midrule
    \multicolumn{18}{@{}l}{\textit{Model-free, unseen objects}} \\
    OPFormer (CNOS)~\cite{moroz2026opformer} & RGB & 20.4 & 13.0 & 27.7 & 0.60 & 33.5 & 24.3 & 42.7 & 1.16 & -- & -- & -- & -- & -- & -- & -- & -- \\
    GFreeDet2~\cite{wang2026gfreedet2} & RGB & 45.8 & 40.1 & 51.4 & 38.37 & 45.2 & 36.5 & 53.8 & 37.90 & 48.9 & 43.9 & 53.8 & 42.41 & -- & -- & -- & -- \\
    GFreeDet2 (default 2D)~\cite{wang2026gfreedet2} & RGB & 42.2 & 37.3 & 47.0 & 28.22 & 41.1 & 33.9 & 48.2 & 36.76 & 48.3 & 43.4 & 53.2 & 24.67 & -- & -- & -- & -- \\
    GFreeDet2 (lite)~\cite{wang2026gfreedet2} & RGB & 40.0 & 35.1 & 44.9 & 9.03 & 39.6 & 32.8 & 46.3 & 12.74 & 42.3 & 37.4 & 47.2 & 10.39 & -- & -- & -- & -- \\
    X0 (Dynamic-0)~\cite{greco2026dynamic0} & RGB & 52.1 & 46.0 & 58.1 & 38.40 & 47.3 & 38.5 & 56.0 & 40.98 & 53.1 & 47.7 & 58.4 & 42.19 & -- & -- & -- & -- \\
    NovaSplat (CNOS25)~\cite{floros2026novasplat} & MV RGB & -- & -- & -- & -- & 52.9 & 50.9 & 54.8 & 56.30 & -- & -- & -- & -- & -- & -- & -- & -- \\
    NeMO-Pose~\cite{jung2026nemo} & RGB-D & 23.5 & 18.1 & 29.0 & 0.95 & 30.2 & 29.1 & 31.2 & 5.15 & -- & -- & -- & -- & -- & -- & -- & -- \\
    \midrule
    \multicolumn{18}{@{}l}{\textit{Model-based, seen objects}} \\
    HccePose(BF)~\cite{wang2025hccepose}$^{\dagger}$ & RGB & 33.2 & 28.0 & 38.3 & -- & 41.4 & 31.7 & 51.1 & -- & 32.9 & 26.4 & 39.5 & -- & \rankcell{3}{87.3} & \rankcell{4}{81.5} & \rankcell{1}{93.0} & 0.10 \\
    HccePose + DOPER boxes$^{\ddagger}$ & RGB & 55.1 & 45.2 & 65.1 & -- & 56.6 & 41.2 & \rankcell{5}{72.1} & -- & 45.7 & 35.6 & 55.7 & -- & -- & -- & -- & -- \\
    \midrule
    \multicolumn{18}{@{}l}{\textit{DOPER: model-based, seen objects}} \\
    DOPER base & RGB & \rankcell{4}{76.1} & \rankcell{4}{68.6} & \rankcell{4}{83.5} & 1.20 & 63.0 & 47.6 & \rankcell{2}{78.3} & 1.09 & 55.2 & 46.3 & 64.1 & 1.03 & 74.7 & 65.1 & 84.2 & 0.85 \\
    DOPER base + heatmap & RGB & \rankcell{3}{77.0} & \rankcell{3}{69.7} & \rankcell{3}{84.2} & 1.23 & 62.9 & 47.4 & \rankcell{2}{78.3} & 1.13 & \rankcell{5}{57.2} & 49.2 & \rankcell{5}{65.2} & 1.06 & 76.3 & 67.7 & 85.0 & 0.89 \\
    DOPER base + MV & MV RGB & -- & -- & -- & -- & \rankcell{2}{71.4} & \rankcell{3}{64.2} & \rankcell{1}{78.7} & 2.99 & 57.1 & 49.6 & 64.6 & 2.48 & 83.0 & 80.8 & 85.2 & 3.43 \\
    DOPER base + MV + heatmap & MV RGB & -- & -- & -- & -- & \rankcell{1}{71.7} & \rankcell{2}{64.7} & \rankcell{1}{78.7} & 3.16 & \rankcell{4}{57.3} & \rankcell{5}{49.8} & 64.7 & 2.62 & \rankcell{4}{83.3} & \rankcell{5}{81.4} & 85.3 & 3.63 \\
    DOPER adapted & RGB & \rankcell{2}{80.5} & \rankcell{2}{73.5} & \rankcell{2}{87.5} & 1.18 & 62.3 & 46.7 & \rankcell{3}{78.0} & 1.07 & \rankcell{3}{63.5} & \rankcell{4}{55.0} & \rankcell{4}{72.0} & 1.11 & 81.4 & 74.6 & \rankcell{5}{88.2} & 0.79 \\
    DOPER adapted + heatmap & RGB & \rankcell{1}{81.4} & \rankcell{1}{74.6} & \rankcell{1}{88.2} & 1.21 & 62.2 & 46.4 & \rankcell{4}{77.9} & 1.11 & \rankcell{2}{65.3} & \rankcell{3}{57.7} & \rankcell{1}{73.0} & 1.15 & 82.8 & 77.0 & \rankcell{4}{88.6} & 0.82 \\
    DOPER adapted + MV & MV RGB & -- & -- & -- & -- & \rankcell{4}{70.9} & \rankcell{5}{63.4} & \rankcell{2}{78.3} & 2.95 & \rankcell{2}{65.3} & \rankcell{2}{58.2} & \rankcell{3}{72.4} & 2.69 & \rankcell{2}{87.5} & \rankcell{2}{86.1} & \rankcell{3}{88.8} & 3.21 \\
    DOPER adapted + MV + heatmap & MV RGB & -- & -- & -- & -- & \rankcell{3}{71.2} & \rankcell{4}{64.0} & \rankcell{2}{78.3} & 3.13 & \rankcell{1}{65.5} & \rankcell{1}{58.5} & \rankcell{2}{72.5} & 2.84 & \rankcell{1}{87.7} & \rankcell{1}{86.5} & \rankcell{2}{88.9} & 3.40 \\
    \bottomrule
  \end{tabular}%
  }
\end{table*}

\section{Experiments}

We evaluate DOPER on challenging model-based object detection and pose estimation tasks. We first consider BOP datasets~\cite{nguyen2025bop2024} with data available for adaptation, followed by SPEED+~\cite{park2022speedplus} to assess pose estimation from underspecified 3D models. We also evaluate temporal pose estimation on SHIRT, study how keypoint count with placement affect accuracy, and how model performance can be recovered with adaptation. 

\label{sec:experiments}

\subsection{Object Detection \& Pose on BOP datasets}
\label{sec:handal_hope}

\begin{figure*}[!t]
  \centering
  \includegraphics[width=\textwidth,height=0.43\textheight,keepaspectratio]{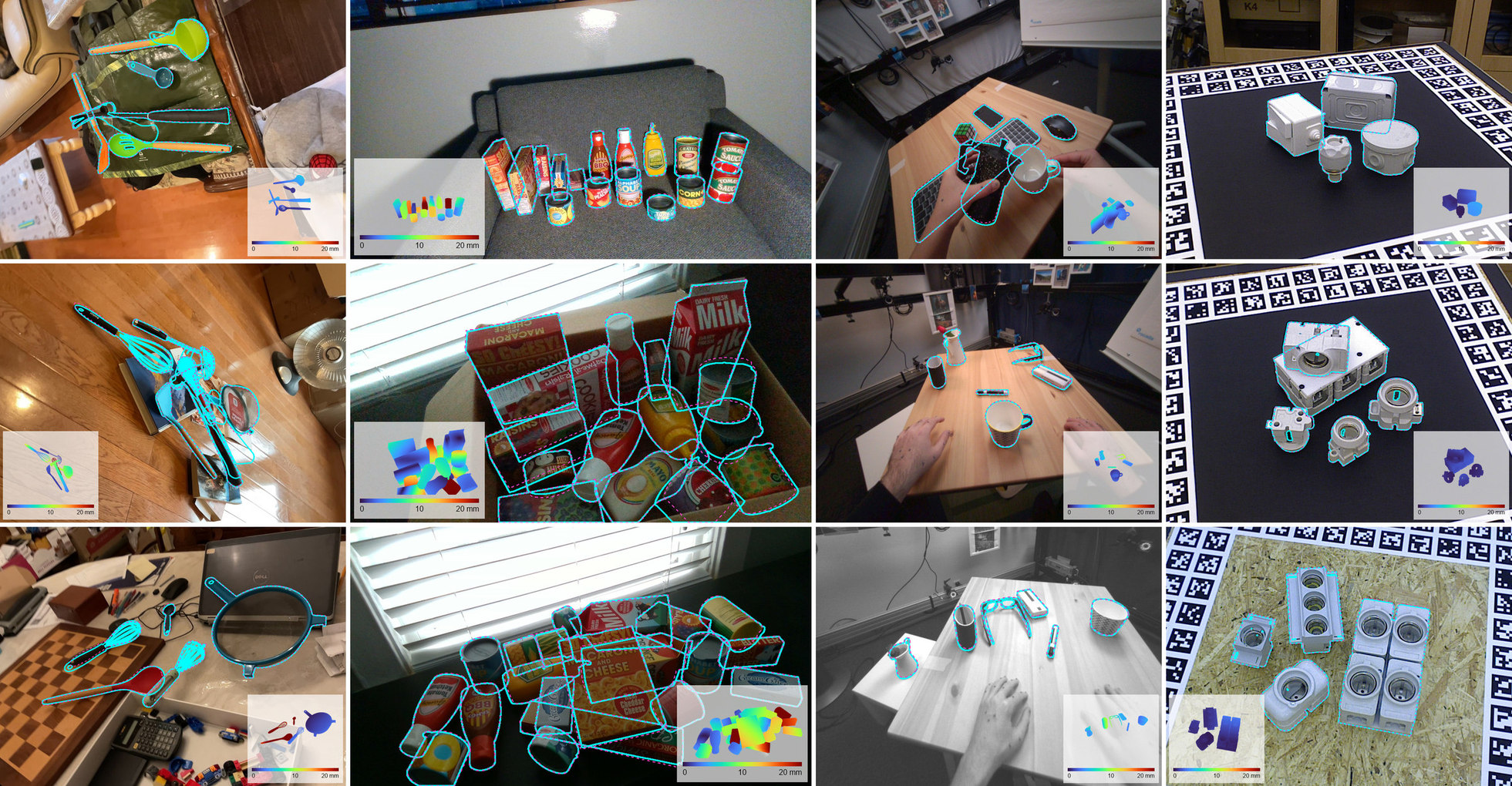}
  \caption{
  DOPER 6D pose predictions from RGB images (cyan CAD overlays) on HANDAL, HOPEv2, HOT3D, and T-LESS (left to right). The examples illustrate model adaptation on HOPEv2 and HOT3D, and heatmap-based pose optimization on HANDAL and T-LESS.  
  The first three columns show validation examples, and T-LESS shows test predictions.  
  Dashed magenta contours show ground truth; insets show symmetry-aware surface error (mm): \mbox{0\,\protect\raisebox{-0.1ex}{\protect\includegraphics[width=1.3in,height=1.1ex]{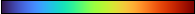}}\,20}.
  }
  \label{fig:handal_hope_qualitative}
\end{figure*}

We evaluate synthetic-to-real transfer on HANDAL~\cite{guo2023handal},  HOPEv2~\cite{tyree2022hope,nguyen2025bop2024}, and HOT3D~\cite{banerjee2025hot3d} using the BOP releases and their object onboarding sequences.
We report controlled onboarding experiments on official BOP test evaluations of the synthetic base and adapted on onboarding data model set, see Table~\ref{tab:BOP_results}, Figure~\ref{fig:handal_hope_qualitative} and \supplink{sec:supplemental_material}{supplemental material} with per object break down.  

\noindent\textbf{Metrics.}
We evaluate pose accuracy using BOP’s symmetry-aware MSSD and MSPD errors~\cite{nguyen2025bop2024}. 
Intuitively, MSSD measures how far the estimated object is displaced in 3D, whereas MSPD measures how far its projection is displaced in the image. Both use the largest pointwise discrepancy after accounting for object symmetries.
MSPD is normalized to a 640-pixel image width. For all four datasets, including T-LESS, we report official BOP test detection AP, computed as the mean of the MSSD and MSPD AP scores.
We report mean runtime per image across all object models, including detection, pose estimation, and any reported refinement, weighted by the number of evaluated images.

\noindent\textbf{Implementation.}
For the controlled onboarding experiments, we train one DINOv3 ConvNeXt-Small model per object for 25 epochs on BOP physically based renders, using 640-by-640 detection images, 512-by-512 crops, and 64 fixed keypoints (see Sec.~\ref{sec:keypoint_placement}).
Each adaptation experiment starts from the same synthetic model and runs for three epochs with a fresh optimizer, 
mixing twelve synthetic and four real images per batch, except for T-LESS, which uses ten synthetic and six real images: two from each of its three image sensors (PrimeSense, Kinect, and Canon).
We use the final checkpoint without selecting epochs on real validation scores; for the pseudo-label experiments, labels are generated once (see \supp{sec:appendix_adaptation}).
The pseudo-label experiments use real boxes and masks only; synthetic images retain keypoint supervision.
\textit{Box/mask supervision} and \textit{box/mask/keypoint supervision} use ground-truth boxes, masks and/or keypoints projected from their annotated poses.
Supervised adaptation uses static onboarding annotations for HOPEv2, HANDAL and HOT3D, and annotated training images from the three T-LESS sensors.
We also train, as a strong baseline, the released ResNet34 HccePose~\cite{wang2025hccepose} implementation on the same PBR images as DOPER under synthetic-only supervision. We allocate 50k training steps, twice the number used for DOPER. 
We omit HccePose runtimes because its inference implementation was not optimized.
We evaluate HccePose using detections from CNOS and DOPER, applying box jitter to DOPER detections as a runtime augmentation to select the best bounding box.

DOPER table timings use one NVIDIA GB300 with all object models resident and decoded images in host memory.
A CUDA-synchronized wall-clock timer covers preprocessing, transfers, every required object/view pass, P$n$P, association, heatmap refinement when enabled, and depth correction when enabled. Model loading, disk reads, and warmup are excluded.

\noindent\textbf{Results.}
Table~\ref{tab:BOP_results} reports test results for 40 HANDAL objects on 1,684 images, 28 HOPEv2 objects on 457 images, 33 HOT3D objects on 5,140 images, and 30 T-LESS objects on 1,000 images.
In single-view evaluation, every object model scans every image, retaining up to five detections above confidence 0.15 before P$n$P; no GT boxes or object-presence information are used.
Adaptation improves performance across datasets, with at most an improvement of AP of 7.5 percentage points. 

\noindent\textbf{Depth.}
With only RGB images and no separate pose refiner, DOPER still leads on two of the four benchmarks. 
With optional depth correction, it achieves the best results across all four, see Table~\ref{tab:doper_depth}.
For example, depth-based translation correction improves AP by up to 12.9 percentage points, from 62.9 to 75.8 on HOPEv2.

\noindent\textbf{Multi-view inference.}
For the reference images with camera calibration, we select the three nearest support cameras by camera-center distance and annotations of the support views are never used.
We associate detections of the same object across views using projected keypoints and a robust pairwise fit, accepting matches with a maximum per-view median reprojection error of 6 pixels.
Starting from the reference P$n$P estimate, we refine a single object pose using keypoints from multiple views while keeping the cameras fixed. Keypoint confidence and RANSAC filter unreliable observations, and a Huber loss reduces the influence of remaining outliers.

\begin{table}[!htbp]
\small
  \centering
  \caption{DOPER with depth correction (\%, higher is better). All variants use registered reference-frame depth; MV denotes multi-view inference. \ranklegend{} ranks accuracy within each dataset; ties share a color. Time is seconds per image with all object models of the dataset (28 for HOPEv2, 30 for T-LESS) on one GB300, including depth correction.}
  \label{tab:doper_depth}
  \footnotesize
  \setlength{\tabcolsep}{3pt}
  \resizebox{\columnwidth}{!}{%
  \begin{tabular}{@{}l|rrrr@{}}
    \toprule
    Variant & AP & MSSD & MSPD & Time (s) \\
    \midrule
    \multicolumn{5}{@{}l}{\textit{HOPEv2~\cite{tyree2022hope,nguyen2025bop2024}}} \\
    Base & \rankcell{5}{74.6} & 70.6 & \rankcell{2}{78.5} & 1.25 \\
    Base + heatmap & \rankcell{5}{74.6} & \rankcell{5}{70.7} & \rankcell{2}{78.5} & 1.29 \\
    Base + MV & \rankcell{2}{75.7} & \rankcell{2}{72.7} & \rankcell{1}{78.7} & 3.15 \\
    Base + MV + heatmap & \rankcell{1}{75.8} & \rankcell{1}{72.9} & \rankcell{1}{78.7} & 3.32 \\
    Adapted & 74.1 & 70.1 & \rankcell{5}{78.1} & 1.23 \\
    Adapted + heatmap & 74.1 & 70.0 & \rankcell{5}{78.1} & 1.27 \\
    Adapted + MV & \rankcell{4}{75.3} & \rankcell{4}{72.3} & \rankcell{4}{78.3} & 3.12 \\
    Adapted + MV + heatmap & \rankcell{3}{75.5} & \rankcell{3}{72.6} & \rankcell{3}{78.4} & 3.30 \\
    \midrule
    \multicolumn{5}{@{}l}{\textit{T-LESS~\cite{hodan2017tless}}} \\
    Base & 82.4 & 80.3 & 84.4 & 0.94 \\
    Base + heatmap & 83.3 & 81.5 & 85.2 & 0.97 \\
    Base + MV & 83.8 & 82.4 & 85.2 & 3.52 \\
    Base + MV + heatmap & \rankcell{5}{84.0} & \rankcell{5}{82.6} & \rankcell{5}{85.4} & 3.71 \\
    Adapted & \rankcell{4}{86.4} & \rankcell{4}{84.6} & \rankcell{4}{88.2} & 0.88 \\
    Adapted + heatmap & \rankcell{3}{87.0} & \rankcell{3}{85.4} & \rankcell{3}{88.7} & 0.91 \\
    Adapted + MV & \rankcell{2}{87.6} & \rankcell{2}{86.3} & \rankcell{2}{88.8} & 3.30 \\
    Adapted + MV + heatmap & \rankcell{1}{87.7} & \rankcell{1}{86.5} & \rankcell{1}{88.9} & 3.50 \\
    \bottomrule
  \end{tabular}%
  }
\end{table}

\begin{figure}[t]
  \centering
  \includegraphics[width=\columnwidth]{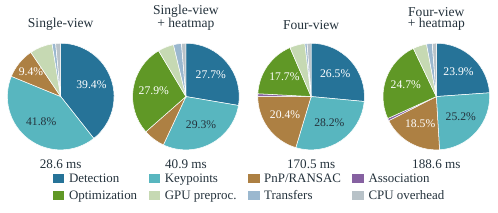}
  \caption{Runtime components for one T-LESS object model on one GB300.
  }
  \label{fig:runtime_components}
\end{figure}

\noindent\textbf{Inference performance.}
We profile one DOPER model for T-LESS object~1 on the GB300 used for Table~\ref{tab:BOP_results}, with input images resident on the GPU.
We decode keypoint heatmaps on the GPU and minimize the heatmap objective (Fig.~\ref{fig:heatmap_refinement}) with analytic gradients.
Mean latency is 28.6\,ms for single-view inference (35\,Hz per object model) and 40.9\,ms with heatmap refinement; four views take 170.5 and 188.6\,ms, respectively.
Without these two steps, inference takes 1.26$\times$, 4.51$\times$, 1.18$\times$, and 2.34$\times$ longer.
Figure~\ref{fig:runtime_components} shows the component costs; the table timings use the same implementation (see \supp{sec:appendix_inference_performance}).

\subsection{Monocular pose estimation on SPEED+}
\input{sec/experiments_qualitative_figure}
\input{sec/experiments_comparison_table}

We evaluate DOPER on SPEED+~\cite{park2022speedplus}, which tests transfer from synthetic training images to photographs of the Tango spacecraft mockup under two illumination conditions.

\noindent\textbf{Metrics.}
We evaluate all 6,740 Lightbox and 2,791 Sunlamp images using DOPER's predicted bounding boxes.
Adaptation uses these same images without their pose labels, following the transductive setting of SPEC2021~\cite{park2023spec2021}.
Scores are the mean sum of rotation error in radians and translation error normalized by ground-truth distance, where lower is better.

\noindent\textbf{Implementation.}
The SPEED+ model uses a DINOv3-pretrained ConvNeXt-Small backbone and 75 keypoints, 64 selected automatically and 11 reference keypoints of the SPEED+ dataset.
We extend the original synthetic data with 10,000 calibrated Blender renders, sampling new rotations and translations with randomized directional lighting.
The renders use equal proportions of Earth and noisy-space backgrounds, together with achromatic flare augmentation.
Each adaptation step pairs eight accepted real samples with the same eight-sample synthetic mixture, each generation runs for ten epochs and passes its final checkpoint to the next generation.

Table~\ref{tab:speedplus_recent} shows that both additional synthetic renders and real-image adaptation improve performance on SPEED+, while Figure~\ref{fig:speedplus_qualitative} shows training data and qualitative results.
Additional renders reduce the Lightbox and Sunlamp errors from 0.2013 to 0.12273 and from 0.2018 to 0.10342, respectively.
Adaptation then uses quantized orientation targets, followed by projected-box supervision, without real keypoint or mask labels.
By generation~7, recall reaches 85.33\% on Lightbox and 62.67\% on Sunlamp, while gross failures fall from 461 to 14 and from 153 to 2.
We report generation~7 on both domains, although Sunlamp reaches its lowest score at generation~6.
These results evaluate adaptation to the observed image pool.

\begin{samepage}
\subsection{Temporal pose estimation on SHIRT}
\label{sec:shirt}

We evaluate the frozen generation~7 model on both SHIRT Lightbox trajectories~\cite{park2023shirt}, using all 2,371 frames per sequence.
\end{samepage}
Table~\ref{tab:shirt_tracking} reports framewise predictions, forward unscented Kalman filtering (UKF), and offline Rauch--Tung--Striebel (RTS) smoothing, which incorporates future observations.
Both UKF variants reduce mean rotation error below one degree, demonstrating that DOPER predictions support accurate temporal pose estimation.
DOPER's P$n$P UKF achieves lower second-orbit mean axial and lateral position errors than SPNv2 UKF+ASNC~\cite{park2023shirt} on both trajectories.

\begin{table}[!htbp]
  \centering
  \caption{SHIRT tracking pose errors ($\downarrow$ best). \ranklegend{} marks the five lowest distinct errors per column among SPNv2 and DOPER; ties share a color.
  ROE1 (hold) and ROE2 (approach) refer to two spacecraft-motion sequences.
  }
  \label{tab:shirt_tracking}
  \small
  \setlength{\tabcolsep}{3pt}
  \resizebox{\columnwidth}{!}{%
\begin{tabular}{@{}lrrrr@{}}
    \toprule
    & \multicolumn{2}{c}{ROE1} & \multicolumn{2}{c}{ROE2} \\

    \cmidrule(lr){2-3}\cmidrule(l){4-5}
    Method & $R\,(^\circ)$ & $t\,(\mathrm{cm})$ & $R\,(^\circ)$ & $t\,(\mathrm{cm})$ \\
    \midrule
    U-AUKF~\cite{cassinis2023uncertainty} & 14\phantom{.000} & 25\phantom{.00} & 9\phantom{.000} & 38\phantom{.00} \\
    TTA + aug.~\cite{bravo2024keypointtta} & 3\phantom{.000} & 11\phantom{.00} & 5\phantom{.000} & 9\phantom{.00} \\
    CAPLR$_m$~\cite{chenni2026caplr} & 10.65\phantom{0} & 26.6\phantom{0} & 4.49\phantom{0} & 8.7\phantom{0} \\
    SPNv2 (heatmap)~\cite{park2023shirt} & \rankcell{5}{17.168} & 30.3\phantom{0} & \rankcell{5}{4.669} & 20.3\phantom{0} \\
    SPNv2 (direct)~\cite{park2023shirt} & 17.584 & \rankcell{4}{17.5\phantom{0}} & 5.484 & \rankcell{5}{10.0\phantom{0}} \\
    \addlinespace[2pt]
    DOPER framewise & \rankcell{4}{7.245} & \rankcell{5}{17.8\phantom{0}} & \rankcell{4}{2.351} & \rankcell{4}{7.48} \\
    DOPER P$n$P UKF & \rankcell{3}{0.955} & \rankcell{2}{12.90} & \rankcell{3}{0.749} & \rankcell{2}{3.95} \\
    DOPER Keypoint UKF & \rankcell{2}{0.758} & \rankcell{3}{13.90} & \rankcell{2}{0.676} & \rankcell{3}{5.15} \\
    DOPER UKF + RTS (offline) & \rankcell{1}{0.576} & \rankcell{1}{12.86} & \rankcell{1}{0.547} & \rankcell{1}{3.87} \\
    \midrule
    \multicolumn{5}{@{}l}{\textit{Second-orbit mean position errors (cm)}} \\
    & \multicolumn{2}{c}{ROE1} & \multicolumn{2}{c}{ROE2} \\
    \cmidrule(lr){2-3}\cmidrule(l){4-5}
    Method & Axial & Lateral & Axial & Lateral \\
    \midrule
    SPNv2 UKF+ASNC~\cite{park2023shirt} & 13.42 & 1.63 & 4.90 & 1.05 \\
    DOPER P$n$P UKF & \rankcell{1}{12.78} & \rankcell{1}{1.00} & \rankcell{1}{2.52} & \rankcell{1}{0.54} \\
    \bottomrule
  \end{tabular}}

\end{table}

\subsection{Keypoint count and placement}
\label{sec:keypoint_placement}

We study six keypoint placement strategies at 8, 16, 32, 64, and 128 points on one object per dataset: HANDAL~8, HOT3D~2, HOPEv2~24, and T-LESS~9.
All 120 models use ConvNeXt-Tiny, the same frozen PBR training populations, and 25 epochs with seed 42. 
Table~\ref{tab:keypoint_placement_mean} averages detector-based MSPD recall equally across the four objects, using real validation images for HANDAL, HOT3D, and HOPEv2 and test images for T-LESS.

\begin{figure}[!ht]
  \ifdefined\nolinenumbers\nolinenumbers\fi
  \centering
  \includegraphics[width=0.94\columnwidth]{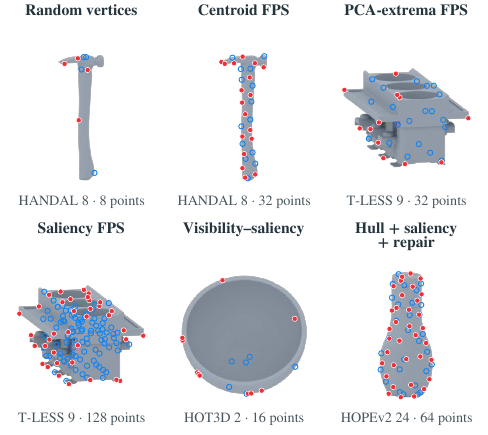}
  \caption{Keypoint placement examples from the ablation. Red dots mark visible points; blue rings mark self-occluded points. Examples use 8, 16, 32, 64, and 128 points, as labeled; the 64-point hull-based layout matches the main BOP recipe.}
  \label{fig:keypoint_placement_examples}
\end{figure}

\noindent\textbf{Placement strategies.}
The strategies balance coverage of the object's shape 
(see Figure~\ref{fig:keypoint_placement_examples}), 
distinctive geometric features, and visibility.
Farthest-point sampling (FPS) spreads points by repeatedly selecting the point farthest from its nearest selected point.
\textit{Random vertices} samples mesh vertices uniformly, while \textit{Centroid FPS} starts near the mesh-vertex centroid and \textit{PCA-extrema FPS} starts at the extremes along the object's principal axes.
\textit{Saliency FPS} favors sharp edges, corners, and open boundaries before spreading the points. 
\textit{Visibility--saliency} also favors points visible from many viewing directions.
\textit{Hull + saliency + repair} starts with eight visible anchors around the object's outer shape, adds points using saliency FPS, and replaces rarely visible points.
The main BOP experiments use this last strategy with 64 keypoints.

\begin{table}[t]
\small
  \ifdefined\nolinenumbers\nolinenumbers\fi
  \centering
  \caption{Keypoint placement and count. Each entry is MSPD AR (\%) averaged equally over four objects; columns give keypoint counts. These are selected-object results on mixed held-out splits, not full-dataset AP.}
  \label{tab:keypoint_placement_mean}
  \footnotesize
  \setlength{\tabcolsep}{3pt}
  \resizebox{\columnwidth}{!}{%
  \begin{tabular}{@{}l|rrrrr@{}}
    \toprule
    Placement & 8 & 16 & 32 & 64 & 128 \\
    \midrule
    Random vertices & 63.91 & 69.82 & 73.28 & 73.78 & 75.07 \\
    Centroid FPS & 66.22 & 72.15 & 75.90 & 77.66 & 77.39 \\
    PCA-extrema FPS & 65.64 & 71.08 & 74.81 & 77.00 & 77.34 \\
    Saliency FPS & 68.42 & 71.43 & 74.89 & 77.04 & 76.69 \\
    Visibility--saliency & 57.73 & 61.86 & 61.04 & 62.50 & 65.36 \\
    Hull + saliency + repair & 65.23 & 72.00 & 75.71 & 77.01 & 76.92 \\
    \bottomrule
  \end{tabular}%
  }
\end{table}

\noindent\textbf{Results.}
\textit{Centroid FPS} gives the highest mean MSPD AR at 16--128 points, peaking at 77.66 with 64.
Increasing the count to 128 reduces its mean by 0.27 points.
\textit{Saliency FPS} leads at eight points, so the preferred placement depends on the keypoint budget.
\Supp{sec:keypoint_placement_details} provides per-object scores, training details, and layout visualizations.

\subsection{Underspecified CAD}
\label{sec:underspecified_cad}

\begin{figure}[!ht]
  \ifdefined\nolinenumbers\nolinenumbers\fi
  \centering
  \includegraphics[width=\columnwidth]{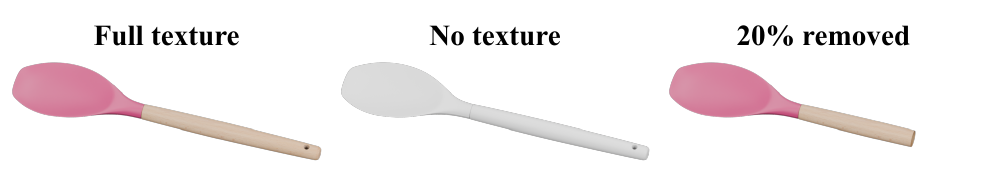}
  \caption{Underspecified CAD variants of HANDAL object 27 (spatula)~\cite{guo2023handal}: original textured model (left), texture removed (middle), and 20\% removed from one end along the principal axis (right).
  }
  \label{fig:handal_cad_ablation}
\end{figure}

To test whether real observations can compensate for CAD deficiencies, we remove texture or part of the geometry from 19 HANDAL objects (Figure~\ref{fig:handal_cad_ablation}). 
We reconstruct the objects from onboarding videos with NuRec~\cite{nvidia2026nurec} and replace half of the synthetic training images with NuRec composites. 
Compared with matched composites of the degraded CAD model, NuRec improves validation AR by an estimated 23.6 and 16.3 points in the two settings, respectively. These gains are consistent across most objects; \supp{sec:appendix_underspecified_cad} details the training protocol, statistical analysis, and results.

\FloatBarrier

\section{Conclusion}
We present DOPER, which learns detection and keypoints jointly and uses geometric fitting to estimate object pose. The same keypoints support multi-view estimation and heatmap refinement, while synthetic initialization and partial real supervision enable adaptation. DOPER achieves strong RGB results across four object-pose benchmarks and supports illumination adaptation and temporal pose estimation on SPEED+ and SHIRT. Our HANDAL study further shows that reconstructions from onboarding videos can compensate for missing texture and incomplete CAD geometry.

These results leave several challenges open: reducing per-object training and refinement costs, and making reconstruction-based adaptation independent of annotated poses. 
Furthermore, a universal metric keypoint detector could help streamline object onboarding and reduce reliance on object-specific training data generation.

\section{Acknowledgment}

The authors would like to thank, in no particular order, Stan Birchfield, Van Nguyen Nguyen, and Tomáš Hodaň for their manuscript feedback. 
We would also like to thank Iretiayo Akinola, Toya Takahashi, ABB Robotics, Alex Zook, Valts Blukis, Andrew Dahdouh, Zheng Wang, Yichao Pan, Leif Jentoft, Zeo Liu, and Yashraj Narang for their constant encouragement and positive feedback.

{
    \small
    \bibliographystyle{ieeenat_fullname}
    \bibliography{main}
}

\onecolumn
\appendix
\phantomsection
\section*{Supplemental Material}
\label{sec:supplemental_material}
\section{Adaptation protocol and validation metric}
\label{sec:appendix_adaptation}

The tables below give per-object results for the adaptation experiments in the main paper.
Unlike the main paper, which reports BOP detection AP, they report validation AR, the average of the MSSD and MSPD recall rates.

Each row is a different way of adapting the synthetic base model.
The pseudo-label rows train on labels that the base model generates from the static or dynamic onboarding videos, or from both.
The two supervised rows use ground-truth annotations, as described in the main paper.
The Labels column counts the pseudo-labels that pass our filter, and a dash means fewer than 20 passed, so we did not train that model.
The three score columns evaluate the same model in three ways: on its own detections (Single), with box jitter at test time (Jitter), and with ground-truth boxes (GT-box), which shows how much accuracy is lost to detection.

Each image grid shows up to 16 frames with the lowest error for that object, so the grids show what successful predictions look like rather than typical errors.
The T-LESS tables report test AP instead of validation AR.

\clearpage
\noindent
\begin{minipage}[t][0.30\textheight][t]{0.32\textwidth}
\vspace{0pt}\centering
\begin{lrbox}{\validationentrybox}
\begin{minipage}{\linewidth}
  \vspace{0pt}
  \centering
  \objectlabel{HANDAL}{02}{handal}
  \captionsetup{font=scriptsize,skip=2pt,hypcap=false}
  \captionof{table}{Validation AR (\%). 662 validation target frames.}
  \label{tab:app_handal_02}
  \fontsize{6.3}{7.0}\selectfont
  \renewcommand{\arraystretch}{0.95}
  \setlength{\tabcolsep}{1.6pt}
  \begin{lrbox}{\validationtablebox}
  \begin{tabular}{@{}lrrrr@{}}
    \toprule
    Variant & Labels & Single & Jitter & GT-box \\
    \midrule
    Synthetic base & -- & \rankcell{5}{85.22} & \rankcell{5}{85.63} & \rankcell{5}{89.26} \\
    Static pseudo & 340 & \rankcell{4}{86.00} & \rankcell{4}{86.34} & \rankcell{2}{89.78} \\
    Box/mask supervision & 230 & \rankcell{3}{86.28} & \rankcell{3}{86.78} & \rankcell{4}{89.43} \\
    Box/mask/keypoint supervision & 230 & \rankcell{1}{86.87} & \rankcell{1}{87.28} & \rankcell{1}{90.32} \\
    Test-input pseudo & 106 & \rankcell{2}{86.53} & \rankcell{2}{86.92} & \rankcell{3}{89.65} \\
    \bottomrule
  \end{tabular}
  \end{lrbox}
  \includegraphics[width=\linewidth,keepaspectratio]{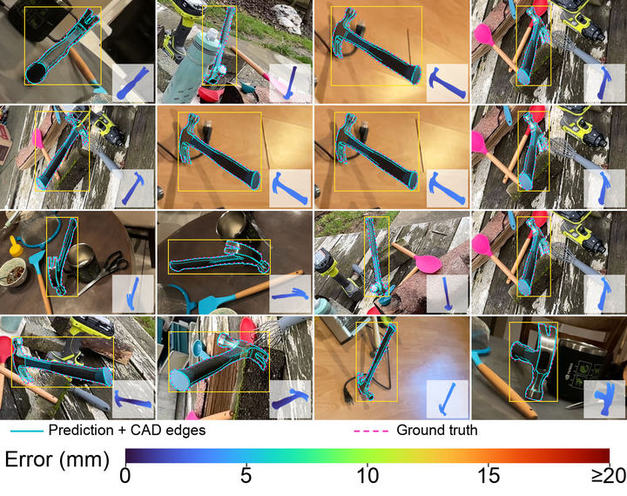}
  \par\vspace{2pt}
  \ifdim\wd\validationtablebox>\linewidth
    \resizebox{\linewidth}{!}{\usebox{\validationtablebox}}
  \else
    \usebox{\validationtablebox}
  \fi
\end{minipage}
\end{lrbox}
\ifdim\dimexpr\ht\validationentrybox+\dp\validationentrybox\relax>0.30\textheight
  \resizebox*{!}{0.30\textheight}{\usebox{\validationentrybox}}
\else
  \usebox{\validationentrybox}
\fi
\end{minipage}
\hfill%
\begin{minipage}[t][0.30\textheight][t]{0.32\textwidth}
\vspace{0pt}\centering
\begin{lrbox}{\validationentrybox}
\begin{minipage}{\linewidth}
  \vspace{0pt}
  \centering
  \objectlabel{HANDAL}{06}{handal}
  \captionsetup{font=scriptsize,skip=2pt,hypcap=false}
  \captionof{table}{Validation AR (\%). 666 validation target frames.}
  \label{tab:app_handal_06}
  \fontsize{6.3}{7.0}\selectfont
  \renewcommand{\arraystretch}{0.95}
  \setlength{\tabcolsep}{1.6pt}
  \begin{lrbox}{\validationtablebox}
  \begin{tabular}{@{}lrrrr@{}}
    \toprule
    Variant & Labels & Single & Jitter & GT-box \\
    \midrule
    Synthetic base & -- & \rankcell{5}{95.59} & \rankcell{4}{96.09} & \rankcell{5}{96.58} \\
    Static pseudo & 175 & \rankcell{3}{96.31} & \rankcell{2}{96.58} & \rankcell{4}{96.70} \\
    Box/mask supervision & 229 & \rankcell{2}{96.46} & \rankcell{1}{96.76} & \rankcell{2}{96.91} \\
    Box/mask/keypoint supervision & 229 & \rankcell{1}{96.55} & \rankcell{1}{96.76} & \rankcell{1}{97.22} \\
    Test-input pseudo & 100 & \rankcell{4}{96.09} & \rankcell{3}{96.34} & \rankcell{3}{96.88} \\
    \bottomrule
  \end{tabular}
  \end{lrbox}
  \includegraphics[width=\linewidth,keepaspectratio]{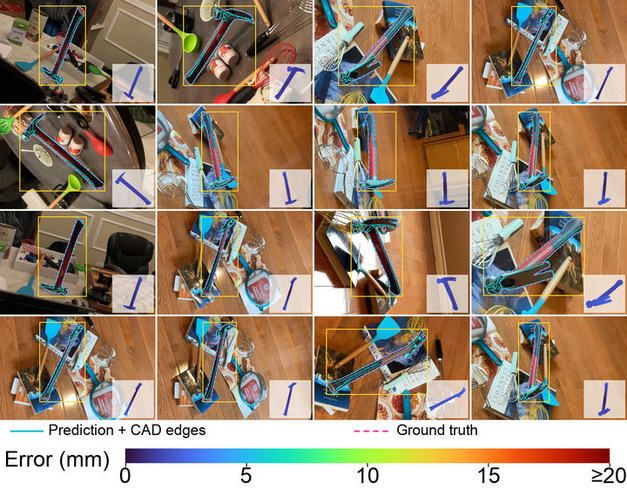}
  \par\vspace{2pt}
  \ifdim\wd\validationtablebox>\linewidth
    \resizebox{\linewidth}{!}{\usebox{\validationtablebox}}
  \else
    \usebox{\validationtablebox}
  \fi
\end{minipage}
\end{lrbox}
\ifdim\dimexpr\ht\validationentrybox+\dp\validationentrybox\relax>0.30\textheight
  \resizebox*{!}{0.30\textheight}{\usebox{\validationentrybox}}
\else
  \usebox{\validationentrybox}
\fi
\end{minipage}
\hfill%
\begin{minipage}[t][0.30\textheight][t]{0.32\textwidth}
\vspace{0pt}\centering
\begin{lrbox}{\validationentrybox}
\begin{minipage}{\linewidth}
  \vspace{0pt}
  \centering
  \objectlabel{HANDAL}{08}{handal}
  \captionsetup{font=scriptsize,skip=2pt,hypcap=false}
  \captionof{table}{Validation AR (\%). 233 validation target frames.}
  \label{tab:app_handal_08}
  \fontsize{6.3}{7.0}\selectfont
  \renewcommand{\arraystretch}{0.95}
  \setlength{\tabcolsep}{1.6pt}
  \begin{lrbox}{\validationtablebox}
  \begin{tabular}{@{}lrrrr@{}}
    \toprule
    Variant & Labels & Single & Jitter & GT-box \\
    \midrule
    Synthetic base & -- & \rankcell{4}{70.04} & \rankcell{4}{72.47} & \rankcell{4}{81.57} \\
    Static pseudo & 6 & -- & -- & -- \\
    Box/mask supervision & 228 & \rankcell{2}{78.73} & \rankcell{2}{80.73} & \rankcell{2}{87.04} \\
    Box/mask/keypoint supervision & 228 & \rankcell{1}{79.72} & \rankcell{1}{81.09} & \rankcell{1}{88.28} \\
    Test-input pseudo & 65 & \rankcell{3}{71.67} & \rankcell{3}{74.29} & \rankcell{3}{82.32} \\
    \bottomrule
  \end{tabular}
  \end{lrbox}
  \includegraphics[width=\linewidth,keepaspectratio]{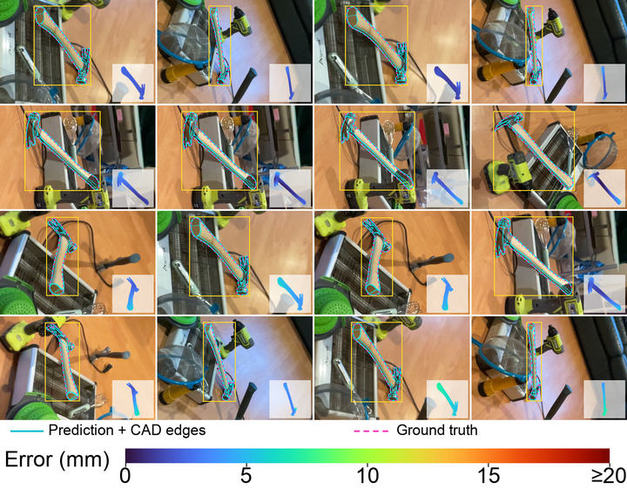}
  \par\vspace{2pt}
  \ifdim\wd\validationtablebox>\linewidth
    \resizebox{\linewidth}{!}{\usebox{\validationtablebox}}
  \else
    \usebox{\validationtablebox}
  \fi
\end{minipage}
\end{lrbox}
\ifdim\dimexpr\ht\validationentrybox+\dp\validationentrybox\relax>0.30\textheight
  \resizebox*{!}{0.30\textheight}{\usebox{\validationentrybox}}
\else
  \usebox{\validationentrybox}
\fi
\end{minipage}
\par\vspace{0.012\textheight}\noindent
\begin{minipage}[t][0.30\textheight][t]{0.32\textwidth}
\vspace{0pt}\centering
\begin{lrbox}{\validationentrybox}
\begin{minipage}{\linewidth}
  \vspace{0pt}
  \centering
  \objectlabel{HANDAL}{09}{handal}
  \captionsetup{font=scriptsize,skip=2pt,hypcap=false}
  \captionof{table}{Validation AR (\%). 225 validation target frames.}
  \label{tab:app_handal_09}
  \fontsize{6.3}{7.0}\selectfont
  \renewcommand{\arraystretch}{0.95}
  \setlength{\tabcolsep}{1.6pt}
  \begin{lrbox}{\validationtablebox}
  \begin{tabular}{@{}lrrrr@{}}
    \toprule
    Variant & Labels & Single & Jitter & GT-box \\
    \midrule
    Synthetic base & -- & \rankcell{4}{93.58} & \rankcell{4}{95.73} & \rankcell{5}{94.18} \\
    Static pseudo & 508 & \rankcell{5}{93.44} & \rankcell{5}{94.84} & \rankcell{4}{94.36} \\
    Box/mask supervision & 247 & \rankcell{3}{94.60} & \rankcell{2}{96.22} & \rankcell{3}{94.80} \\
    Box/mask/keypoint supervision & 247 & \rankcell{2}{94.67} & \rankcell{1}{96.40} & \rankcell{2}{95.00} \\
    Test-input pseudo & 122 & \rankcell{1}{95.07} & \rankcell{3}{95.93} & \rankcell{1}{95.42} \\
    \bottomrule
  \end{tabular}
  \end{lrbox}
  \includegraphics[width=\linewidth,keepaspectratio]{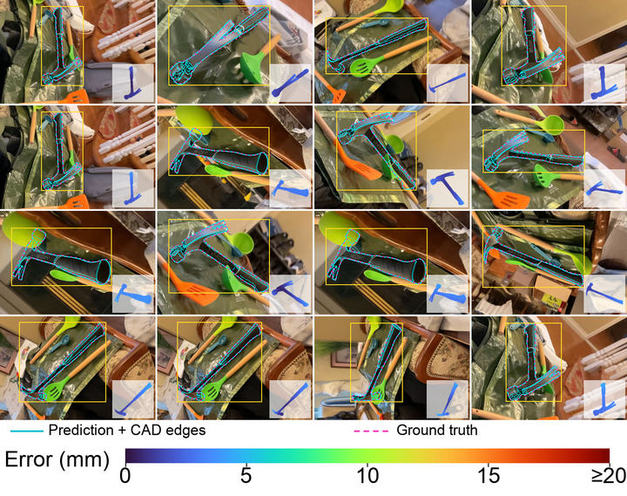}
  \par\vspace{2pt}
  \ifdim\wd\validationtablebox>\linewidth
    \resizebox{\linewidth}{!}{\usebox{\validationtablebox}}
  \else
    \usebox{\validationtablebox}
  \fi
\end{minipage}
\end{lrbox}
\ifdim\dimexpr\ht\validationentrybox+\dp\validationentrybox\relax>0.30\textheight
  \resizebox*{!}{0.30\textheight}{\usebox{\validationentrybox}}
\else
  \usebox{\validationentrybox}
\fi
\end{minipage}
\hfill%
\begin{minipage}[t][0.30\textheight][t]{0.32\textwidth}
\vspace{0pt}\centering
\begin{lrbox}{\validationentrybox}
\begin{minipage}{\linewidth}
  \vspace{0pt}
  \centering
  \objectlabel{HANDAL}{10}{handal}
  \captionsetup{font=scriptsize,skip=2pt,hypcap=false}
  \captionof{table}{Validation AR (\%). 1,102 validation target frames.}
  \label{tab:app_handal_10}
  \fontsize{6.3}{7.0}\selectfont
  \renewcommand{\arraystretch}{0.95}
  \setlength{\tabcolsep}{1.6pt}
  \begin{lrbox}{\validationtablebox}
  \begin{tabular}{@{}lrrrr@{}}
    \toprule
    Variant & Labels & Single & Jitter & GT-box \\
    \midrule
    Synthetic base & -- & \rankcell{5}{86.86} & \rankcell{5}{87.13} & \rankcell{4}{87.40} \\
    Static pseudo & 874 & \rankcell{3}{87.21} & \rankcell{3}{87.71} & \rankcell{2}{88.01} \\
    Box/mask supervision & 151 & \rankcell{4}{86.87} & \rankcell{4}{87.47} & \rankcell{3}{87.86} \\
    Box/mask/keypoint supervision & 151 & \rankcell{2}{87.55} & \rankcell{1}{87.98} & \rankcell{1}{88.18} \\
    Test-input pseudo & 94 & \rankcell{1}{87.70} & \rankcell{2}{87.94} & \rankcell{3}{87.86} \\
    \bottomrule
  \end{tabular}
  \end{lrbox}
  \includegraphics[width=\linewidth,keepaspectratio]{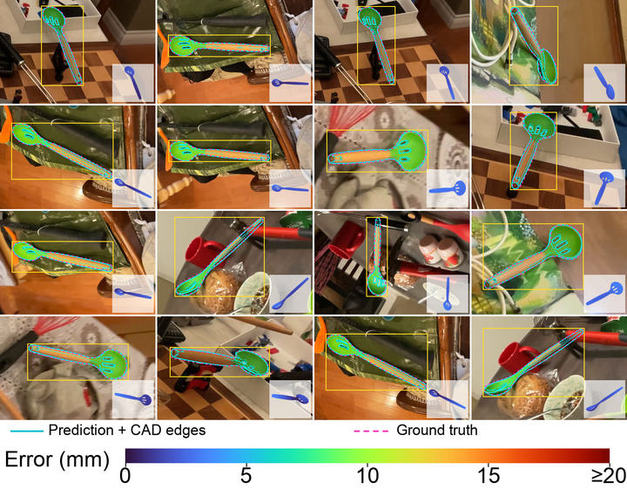}
  \par\vspace{2pt}
  \ifdim\wd\validationtablebox>\linewidth
    \resizebox{\linewidth}{!}{\usebox{\validationtablebox}}
  \else
    \usebox{\validationtablebox}
  \fi
\end{minipage}
\end{lrbox}
\ifdim\dimexpr\ht\validationentrybox+\dp\validationentrybox\relax>0.30\textheight
  \resizebox*{!}{0.30\textheight}{\usebox{\validationentrybox}}
\else
  \usebox{\validationentrybox}
\fi
\end{minipage}
\hfill%
\begin{minipage}[t][0.30\textheight][t]{0.32\textwidth}
\vspace{0pt}\centering
\begin{lrbox}{\validationentrybox}
\begin{minipage}{\linewidth}
  \vspace{0pt}
  \centering
  \objectlabel{HANDAL}{11}{handal}
  \captionsetup{font=scriptsize,skip=2pt,hypcap=false}
  \captionof{table}{Validation AR (\%). 879 validation target frames.}
  \label{tab:app_handal_11}
  \fontsize{6.3}{7.0}\selectfont
  \renewcommand{\arraystretch}{0.95}
  \setlength{\tabcolsep}{1.6pt}
  \begin{lrbox}{\validationtablebox}
  \begin{tabular}{@{}lrrrr@{}}
    \toprule
    Variant & Labels & Single & Jitter & GT-box \\
    \midrule
    Synthetic base & -- & \rankcell{4}{72.66} & \rankcell{4}{73.39} & \rankcell{5}{77.34} \\
    Static pseudo & 728 & \rankcell{5}{72.54} & \rankcell{5}{73.30} & \rankcell{2}{77.75} \\
    Box/mask supervision & 259 & \rankcell{3}{74.15} & \rankcell{2}{74.77} & \rankcell{4}{77.65} \\
    Box/mask/keypoint supervision & 259 & \rankcell{1}{75.43} & \rankcell{1}{76.07} & \rankcell{1}{78.78} \\
    Test-input pseudo & 112 & \rankcell{2}{74.18} & \rankcell{3}{74.76} & \rankcell{3}{77.68} \\
    \bottomrule
  \end{tabular}
  \end{lrbox}
  \includegraphics[width=\linewidth,keepaspectratio]{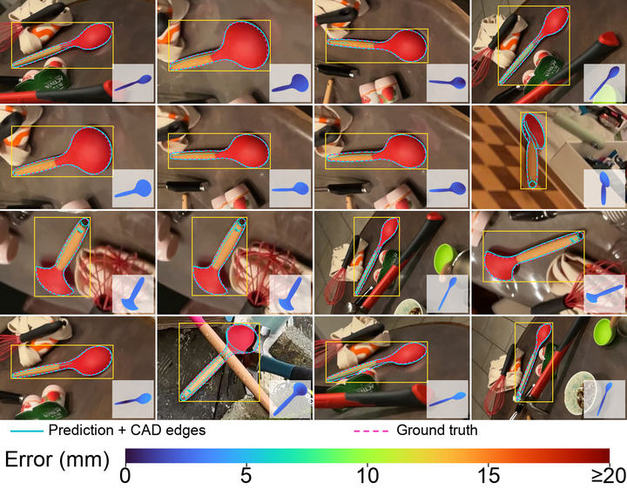}
  \par\vspace{2pt}
  \ifdim\wd\validationtablebox>\linewidth
    \resizebox{\linewidth}{!}{\usebox{\validationtablebox}}
  \else
    \usebox{\validationtablebox}
  \fi
\end{minipage}
\end{lrbox}
\ifdim\dimexpr\ht\validationentrybox+\dp\validationentrybox\relax>0.30\textheight
  \resizebox*{!}{0.30\textheight}{\usebox{\validationentrybox}}
\else
  \usebox{\validationentrybox}
\fi
\end{minipage}
\par\vspace{0.012\textheight}\noindent
\begin{minipage}[t][0.30\textheight][t]{0.32\textwidth}
\vspace{0pt}\centering
\begin{lrbox}{\validationentrybox}
\begin{minipage}{\linewidth}
  \vspace{0pt}
  \centering
  \objectlabel{HANDAL}{12}{handal}
  \captionsetup{font=scriptsize,skip=2pt,hypcap=false}
  \captionof{table}{Validation AR (\%). 442 validation target frames.}
  \label{tab:app_handal_12}
  \fontsize{6.3}{7.0}\selectfont
  \renewcommand{\arraystretch}{0.95}
  \setlength{\tabcolsep}{1.6pt}
  \begin{lrbox}{\validationtablebox}
  \begin{tabular}{@{}lrrrr@{}}
    \toprule
    Variant & Labels & Single & Jitter & GT-box \\
    \midrule
    Synthetic base & -- & \rankcell{5}{59.63} & \rankcell{5}{60.67} & \rankcell{5}{86.98} \\
    Static pseudo & 659 & \rankcell{4}{81.31} & \rankcell{3}{81.91} & \rankcell{4}{89.64} \\
    Box/mask supervision & 253 & \rankcell{3}{81.49} & \rankcell{4}{81.89} & \rankcell{3}{89.80} \\
    Box/mask/keypoint supervision & 253 & \rankcell{2}{83.16} & \rankcell{2}{83.62} & \rankcell{1}{90.32} \\
    Test-input pseudo & 125 & \rankcell{1}{83.81} & \rankcell{1}{83.99} & \rankcell{2}{89.85} \\
    \bottomrule
  \end{tabular}
  \end{lrbox}
  \includegraphics[width=\linewidth,keepaspectratio]{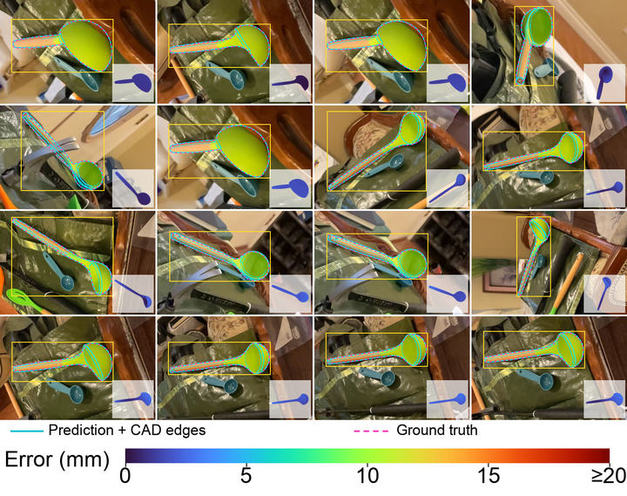}
  \par\vspace{2pt}
  \ifdim\wd\validationtablebox>\linewidth
    \resizebox{\linewidth}{!}{\usebox{\validationtablebox}}
  \else
    \usebox{\validationtablebox}
  \fi
\end{minipage}
\end{lrbox}
\ifdim\dimexpr\ht\validationentrybox+\dp\validationentrybox\relax>0.30\textheight
  \resizebox*{!}{0.30\textheight}{\usebox{\validationentrybox}}
\else
  \usebox{\validationentrybox}
\fi
\end{minipage}
\hfill%
\begin{minipage}[t][0.30\textheight][t]{0.32\textwidth}
\vspace{0pt}\centering
\begin{lrbox}{\validationentrybox}
\begin{minipage}{\linewidth}
  \vspace{0pt}
  \centering
  \objectlabel{HANDAL}{16}{handal}
  \captionsetup{font=scriptsize,skip=2pt,hypcap=false}
  \captionof{table}{Validation AR (\%). 662 validation target frames.}
  \label{tab:app_handal_16}
  \fontsize{6.3}{7.0}\selectfont
  \renewcommand{\arraystretch}{0.95}
  \setlength{\tabcolsep}{1.6pt}
  \begin{lrbox}{\validationtablebox}
  \begin{tabular}{@{}lrrrr@{}}
    \toprule
    Variant & Labels & Single & Jitter & GT-box \\
    \midrule
    Synthetic base & -- & \rankcell{4}{68.21} & \rankcell{4}{68.55} & \rankcell{4}{70.24} \\
    Static pseudo & 1263 & \rankcell{2}{69.27} & \rankcell{2}{69.58} & \rankcell{3}{71.00} \\
    Box/mask supervision & 226 & \rankcell{3}{68.68} & \rankcell{3}{68.96} & \rankcell{2}{71.03} \\
    Box/mask/keypoint supervision & 226 & \rankcell{1}{70.36} & \rankcell{1}{70.45} & \rankcell{1}{72.94} \\
    Test-input pseudo & 61 & \rankcell{5}{67.95} & \rankcell{5}{68.22} & \rankcell{5}{70.18} \\
    \bottomrule
  \end{tabular}
  \end{lrbox}
  \includegraphics[width=\linewidth,keepaspectratio]{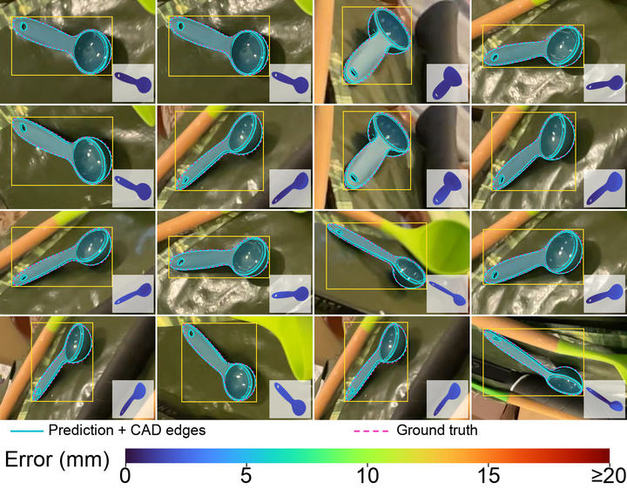}
  \par\vspace{2pt}
  \ifdim\wd\validationtablebox>\linewidth
    \resizebox{\linewidth}{!}{\usebox{\validationtablebox}}
  \else
    \usebox{\validationtablebox}
  \fi
\end{minipage}
\end{lrbox}
\ifdim\dimexpr\ht\validationentrybox+\dp\validationentrybox\relax>0.30\textheight
  \resizebox*{!}{0.30\textheight}{\usebox{\validationentrybox}}
\else
  \usebox{\validationentrybox}
\fi
\end{minipage}
\hfill%
\begin{minipage}[t][0.30\textheight][t]{0.32\textwidth}
\vspace{0pt}\centering
\begin{lrbox}{\validationentrybox}
\begin{minipage}{\linewidth}
  \vspace{0pt}
  \centering
  \objectlabel{HANDAL}{17}{handal}
  \captionsetup{font=scriptsize,skip=2pt,hypcap=false}
  \captionof{table}{Validation AR (\%). 441 validation target frames.}
  \label{tab:app_handal_17}
  \fontsize{6.3}{7.0}\selectfont
  \renewcommand{\arraystretch}{0.95}
  \setlength{\tabcolsep}{1.6pt}
  \begin{lrbox}{\validationtablebox}
  \begin{tabular}{@{}lrrrr@{}}
    \toprule
    Variant & Labels & Single & Jitter & GT-box \\
    \midrule
    Synthetic base & -- & \rankcell{3}{57.83} & \rankcell{3}{58.78} & \rankcell{5}{55.69} \\
    Static pseudo & 120 & \rankcell{4}{57.79} & \rankcell{4}{58.55} & \rankcell{3}{56.12} \\
    Box/mask supervision & 254 & \rankcell{5}{57.00} & \rankcell{5}{57.23} & \rankcell{4}{55.76} \\
    Box/mask/keypoint supervision & 254 & \rankcell{1}{61.58} & \rankcell{1}{62.15} & \rankcell{1}{62.04} \\
    Test-input pseudo & 86 & \rankcell{2}{58.89} & \rankcell{2}{59.24} & \rankcell{2}{56.36} \\
    \bottomrule
  \end{tabular}
  \end{lrbox}
  \includegraphics[width=\linewidth,keepaspectratio]{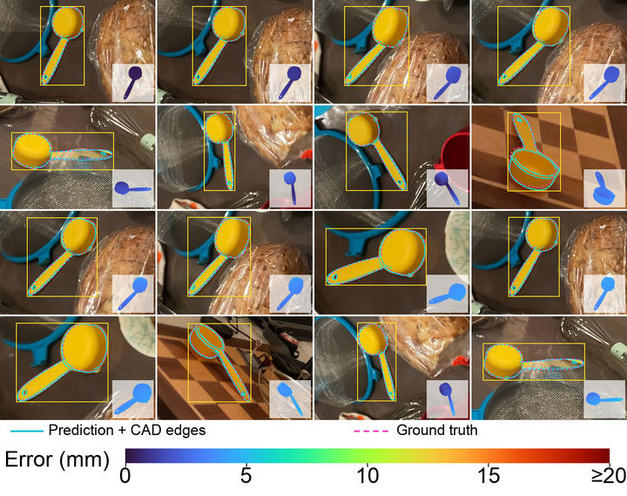}
  \par\vspace{2pt}
  \ifdim\wd\validationtablebox>\linewidth
    \resizebox{\linewidth}{!}{\usebox{\validationtablebox}}
  \else
    \usebox{\validationtablebox}
  \fi
\end{minipage}
\end{lrbox}
\ifdim\dimexpr\ht\validationentrybox+\dp\validationentrybox\relax>0.30\textheight
  \resizebox*{!}{0.30\textheight}{\usebox{\validationentrybox}}
\else
  \usebox{\validationentrybox}
\fi
\end{minipage}
\par
\newpage
\noindent
\begin{minipage}[t][0.30\textheight][t]{0.32\textwidth}
\vspace{0pt}\centering
\begin{lrbox}{\validationentrybox}
\begin{minipage}{\linewidth}
  \vspace{0pt}
  \centering
  \objectlabel{HANDAL}{20}{handal}
  \captionsetup{font=scriptsize,skip=2pt,hypcap=false}
  \captionof{table}{Validation AR (\%). 448 validation target frames.}
  \label{tab:app_handal_20}
  \fontsize{6.3}{7.0}\selectfont
  \renewcommand{\arraystretch}{0.95}
  \setlength{\tabcolsep}{1.6pt}
  \begin{lrbox}{\validationtablebox}
  \begin{tabular}{@{}lrrrr@{}}
    \toprule
    Variant & Labels & Single & Jitter & GT-box \\
    \midrule
    Synthetic base & -- & \rankcell{1}{98.35} & \rankcell{1}{98.36} & \rankcell{2}{98.10} \\
    Static pseudo & 318 & \rankcell{4}{98.09} & \rankcell{2}{98.15} & \rankcell{3}{98.08} \\
    Box/mask supervision & 247 & \rankcell{3}{98.11} & \rankcell{3}{98.14} & \rankcell{4}{97.61} \\
    Box/mask/keypoint supervision & 247 & \rankcell{5}{97.89} & \rankcell{5}{97.91} & \rankcell{5}{97.51} \\
    Test-input pseudo & 137 & \rankcell{2}{98.14} & \rankcell{4}{98.11} & \rankcell{1}{98.14} \\
    \bottomrule
  \end{tabular}
  \end{lrbox}
  \includegraphics[width=\linewidth,keepaspectratio]{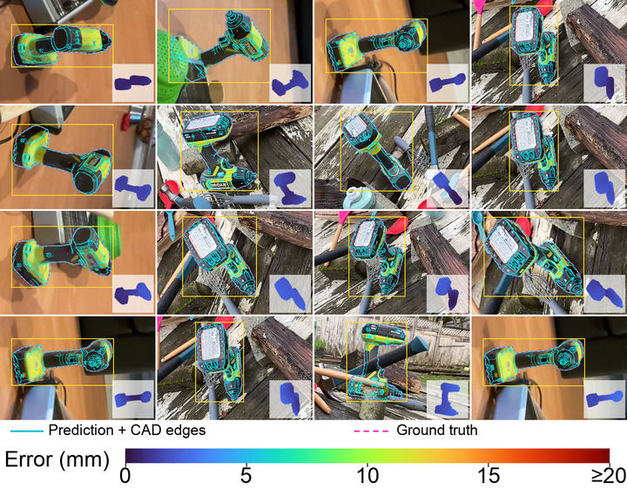}
  \par\vspace{2pt}
  \ifdim\wd\validationtablebox>\linewidth
    \resizebox{\linewidth}{!}{\usebox{\validationtablebox}}
  \else
    \usebox{\validationtablebox}
  \fi
\end{minipage}
\end{lrbox}
\ifdim\dimexpr\ht\validationentrybox+\dp\validationentrybox\relax>0.30\textheight
  \resizebox*{!}{0.30\textheight}{\usebox{\validationentrybox}}
\else
  \usebox{\validationentrybox}
\fi
\end{minipage}
\hfill%
\begin{minipage}[t][0.30\textheight][t]{0.32\textwidth}
\vspace{0pt}\centering
\begin{lrbox}{\validationentrybox}
\begin{minipage}{\linewidth}
  \vspace{0pt}
  \centering
  \objectlabel{HANDAL}{27}{handal}
  \captionsetup{font=scriptsize,skip=2pt,hypcap=false}
  \captionof{table}{Validation AR (\%). 423 validation target frames.}
  \label{tab:app_handal_27}
  \fontsize{6.3}{7.0}\selectfont
  \renewcommand{\arraystretch}{0.95}
  \setlength{\tabcolsep}{1.6pt}
  \begin{lrbox}{\validationtablebox}
  \begin{tabular}{@{}lrrrr@{}}
    \toprule
    Variant & Labels & Single & Jitter & GT-box \\
    \midrule
    Synthetic base & -- & \rankcell{5}{76.44} & \rankcell{5}{77.46} & \rankcell{5}{79.16} \\
    Static pseudo & 644 & \rankcell{4}{78.37} & \rankcell{4}{79.62} & \rankcell{3}{81.69} \\
    Box/mask supervision & 256 & \rankcell{2}{81.03} & \rankcell{2}{82.10} & \rankcell{2}{82.83} \\
    Box/mask/keypoint supervision & 256 & \rankcell{1}{82.07} & \rankcell{1}{82.62} & \rankcell{1}{83.83} \\
    Test-input pseudo & 122 & \rankcell{3}{80.80} & \rankcell{3}{81.47} & \rankcell{4}{81.25} \\
    \bottomrule
  \end{tabular}
  \end{lrbox}
  \includegraphics[width=\linewidth,keepaspectratio]{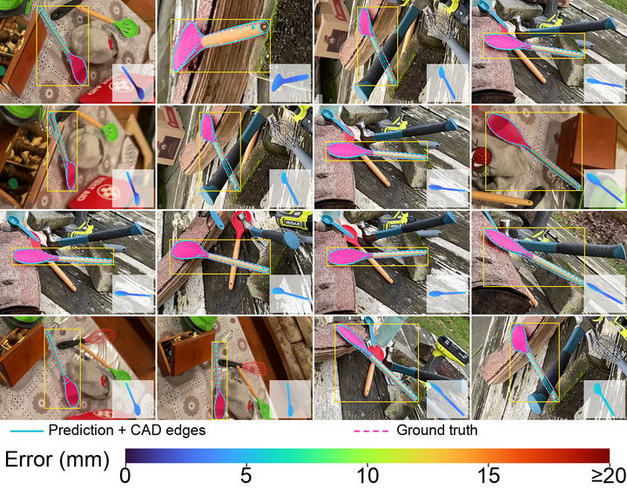}
  \par\vspace{2pt}
  \ifdim\wd\validationtablebox>\linewidth
    \resizebox{\linewidth}{!}{\usebox{\validationtablebox}}
  \else
    \usebox{\validationtablebox}
  \fi
\end{minipage}
\end{lrbox}
\ifdim\dimexpr\ht\validationentrybox+\dp\validationentrybox\relax>0.30\textheight
  \resizebox*{!}{0.30\textheight}{\usebox{\validationentrybox}}
\else
  \usebox{\validationentrybox}
\fi
\end{minipage}
\hfill%
\begin{minipage}[t][0.30\textheight][t]{0.32\textwidth}
\vspace{0pt}\centering
\begin{lrbox}{\validationentrybox}
\begin{minipage}{\linewidth}
  \vspace{0pt}
  \centering
  \objectlabel{HANDAL}{28}{handal}
  \captionsetup{font=scriptsize,skip=2pt,hypcap=false}
  \captionof{table}{Validation AR (\%). 450 validation target frames.}
  \label{tab:app_handal_28}
  \fontsize{6.3}{7.0}\selectfont
  \renewcommand{\arraystretch}{0.95}
  \setlength{\tabcolsep}{1.6pt}
  \begin{lrbox}{\validationtablebox}
  \begin{tabular}{@{}lrrrr@{}}
    \toprule
    Variant & Labels & Single & Jitter & GT-box \\
    \midrule
    Synthetic base & -- & \rankcell{3}{90.44} & \rankcell{4}{90.47} & \rankcell{3}{90.64} \\
    Static pseudo & 770 & \rankcell{1}{91.04} & \rankcell{1}{91.03} & \rankcell{5}{90.17} \\
    Box/mask supervision & 274 & \rankcell{5}{90.12} & \rankcell{3}{90.49} & \rankcell{2}{90.71} \\
    Box/mask/keypoint supervision & 274 & \rankcell{4}{90.28} & \rankcell{5}{90.11} & \rankcell{1}{91.04} \\
    Test-input pseudo & 127 & \rankcell{2}{90.68} & \rankcell{2}{90.79} & \rankcell{4}{90.31} \\
    \bottomrule
  \end{tabular}
  \end{lrbox}
  \includegraphics[width=\linewidth,keepaspectratio]{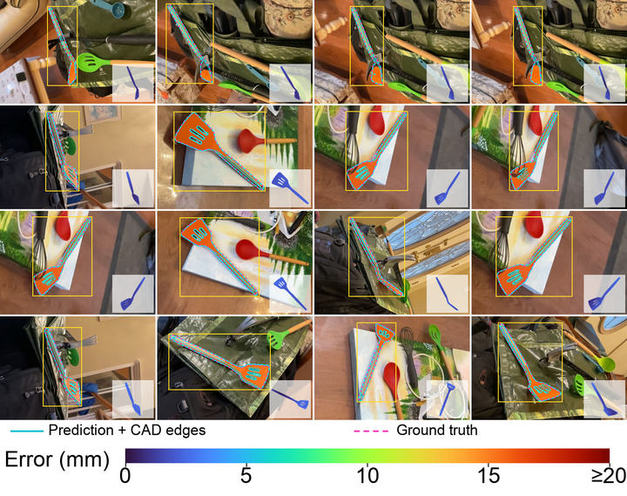}
  \par\vspace{2pt}
  \ifdim\wd\validationtablebox>\linewidth
    \resizebox{\linewidth}{!}{\usebox{\validationtablebox}}
  \else
    \usebox{\validationtablebox}
  \fi
\end{minipage}
\end{lrbox}
\ifdim\dimexpr\ht\validationentrybox+\dp\validationentrybox\relax>0.30\textheight
  \resizebox*{!}{0.30\textheight}{\usebox{\validationentrybox}}
\else
  \usebox{\validationentrybox}
\fi
\end{minipage}
\par\vspace{0.012\textheight}\noindent
\begin{minipage}[t][0.30\textheight][t]{0.32\textwidth}
\vspace{0pt}\centering
\begin{lrbox}{\validationentrybox}
\begin{minipage}{\linewidth}
  \vspace{0pt}
  \centering
  \objectlabel{HANDAL}{29}{handal}
  \captionsetup{font=scriptsize,skip=2pt,hypcap=false}
  \captionof{table}{Validation AR (\%). 663 validation target frames.}
  \label{tab:app_handal_29}
  \fontsize{6.3}{7.0}\selectfont
  \renewcommand{\arraystretch}{0.95}
  \setlength{\tabcolsep}{1.6pt}
  \begin{lrbox}{\validationtablebox}
  \begin{tabular}{@{}lrrrr@{}}
    \toprule
    Variant & Labels & Single & Jitter & GT-box \\
    \midrule
    Synthetic base & -- & \rankcell{5}{78.40} & \rankcell{5}{79.01} & \rankcell{5}{80.72} \\
    Static pseudo & 607 & \rankcell{2}{79.43} & \rankcell{3}{80.07} & \rankcell{2}{81.88} \\
    Box/mask supervision & 272 & \rankcell{4}{78.62} & \rankcell{4}{79.29} & \rankcell{3}{81.72} \\
    Box/mask/keypoint supervision & 272 & \rankcell{1}{79.92} & \rankcell{1}{80.63} & \rankcell{1}{81.95} \\
    Test-input pseudo & 113 & \rankcell{3}{79.38} & \rankcell{2}{80.21} & \rankcell{4}{81.16} \\
    \bottomrule
  \end{tabular}
  \end{lrbox}
  \includegraphics[width=\linewidth,keepaspectratio]{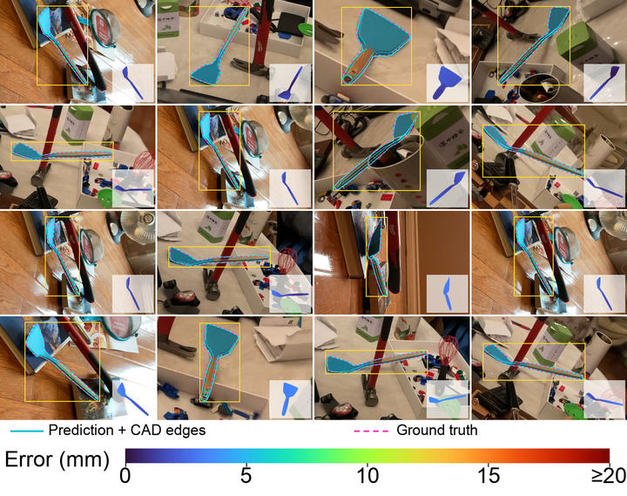}
  \par\vspace{2pt}
  \ifdim\wd\validationtablebox>\linewidth
    \resizebox{\linewidth}{!}{\usebox{\validationtablebox}}
  \else
    \usebox{\validationtablebox}
  \fi
\end{minipage}
\end{lrbox}
\ifdim\dimexpr\ht\validationentrybox+\dp\validationentrybox\relax>0.30\textheight
  \resizebox*{!}{0.30\textheight}{\usebox{\validationentrybox}}
\else
  \usebox{\validationentrybox}
\fi
\end{minipage}
\hfill%
\begin{minipage}[t][0.30\textheight][t]{0.32\textwidth}
\vspace{0pt}\centering
\begin{lrbox}{\validationentrybox}
\begin{minipage}{\linewidth}
  \vspace{0pt}
  \centering
  \objectlabel{HANDAL}{31}{handal}
  \captionsetup{font=scriptsize,skip=2pt,hypcap=false}
  \captionof{table}{Validation AR (\%). 666 validation target frames.}
  \label{tab:app_handal_31}
  \fontsize{6.3}{7.0}\selectfont
  \renewcommand{\arraystretch}{0.95}
  \setlength{\tabcolsep}{1.6pt}
  \begin{lrbox}{\validationtablebox}
  \begin{tabular}{@{}lrrrr@{}}
    \toprule
    Variant & Labels & Single & Jitter & GT-box \\
    \midrule
    Synthetic base & -- & \rankcell{4}{82.60} & \rankcell{4}{83.71} & \rankcell{4}{81.29} \\
    Static pseudo & 863 & \rankcell{3}{82.99} & \rankcell{3}{83.86} & \rankcell{2}{81.51} \\
    Box/mask supervision & 251 & \rankcell{2}{83.33} & \rankcell{2}{84.15} & \rankcell{3}{81.35} \\
    Box/mask/keypoint supervision & 251 & \rankcell{5}{82.58} & \rankcell{5}{83.63} & \rankcell{5}{81.01} \\
    Test-input pseudo & 100 & \rankcell{1}{83.51} & \rankcell{1}{84.88} & \rankcell{1}{81.73} \\
    \bottomrule
  \end{tabular}
  \end{lrbox}
  \includegraphics[width=\linewidth,keepaspectratio]{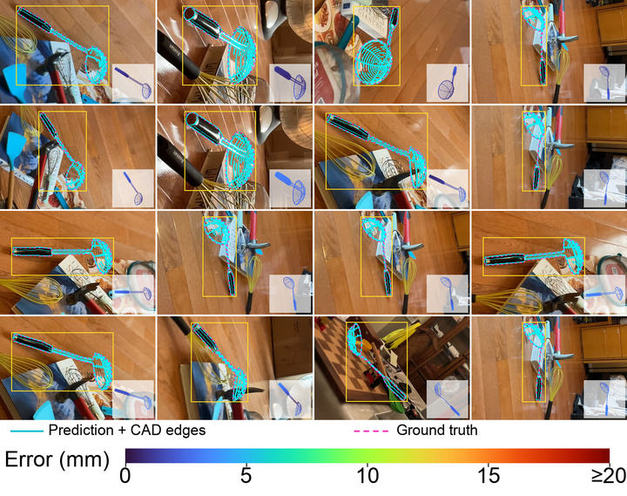}
  \par\vspace{2pt}
  \ifdim\wd\validationtablebox>\linewidth
    \resizebox{\linewidth}{!}{\usebox{\validationtablebox}}
  \else
    \usebox{\validationtablebox}
  \fi
\end{minipage}
\end{lrbox}
\ifdim\dimexpr\ht\validationentrybox+\dp\validationentrybox\relax>0.30\textheight
  \resizebox*{!}{0.30\textheight}{\usebox{\validationentrybox}}
\else
  \usebox{\validationentrybox}
\fi
\end{minipage}
\hfill%
\begin{minipage}[t][0.30\textheight][t]{0.32\textwidth}
\vspace{0pt}\centering
\begin{lrbox}{\validationentrybox}
\begin{minipage}{\linewidth}
  \vspace{0pt}
  \centering
  \objectlabel{HANDAL}{33}{handal}
  \captionsetup{font=scriptsize,skip=2pt,hypcap=false}
  \captionof{table}{Validation AR (\%). 891 validation target frames.}
  \label{tab:app_handal_33}
  \fontsize{6.3}{7.0}\selectfont
  \renewcommand{\arraystretch}{0.95}
  \setlength{\tabcolsep}{1.6pt}
  \begin{lrbox}{\validationtablebox}
  \begin{tabular}{@{}lrrrr@{}}
    \toprule
    Variant & Labels & Single & Jitter & GT-box \\
    \midrule
    Synthetic base & -- & \rankcell{5}{60.22} & \rankcell{5}{60.74} & \rankcell{5}{64.52} \\
    Static pseudo & 259 & \rankcell{4}{63.42} & \rankcell{4}{63.40} & \rankcell{4}{67.32} \\
    Box/mask supervision & 264 & \rankcell{2}{77.01} & \rankcell{2}{77.97} & \rankcell{2}{79.29} \\
    Box/mask/keypoint supervision & 264 & \rankcell{1}{83.96} & \rankcell{1}{84.74} & \rankcell{1}{86.14} \\
    Test-input pseudo & 48 & \rankcell{3}{64.12} & \rankcell{3}{64.45} & \rankcell{3}{67.46} \\
    \bottomrule
  \end{tabular}
  \end{lrbox}
  \includegraphics[width=\linewidth,keepaspectratio]{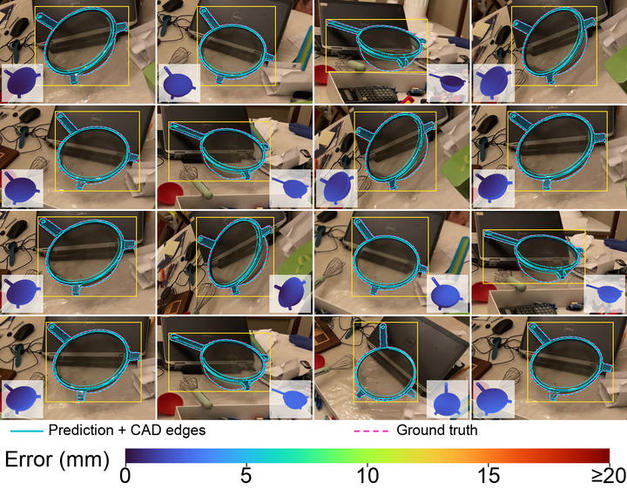}
  \par\vspace{2pt}
  \ifdim\wd\validationtablebox>\linewidth
    \resizebox{\linewidth}{!}{\usebox{\validationtablebox}}
  \else
    \usebox{\validationtablebox}
  \fi
\end{minipage}
\end{lrbox}
\ifdim\dimexpr\ht\validationentrybox+\dp\validationentrybox\relax>0.30\textheight
  \resizebox*{!}{0.30\textheight}{\usebox{\validationentrybox}}
\else
  \usebox{\validationentrybox}
\fi
\end{minipage}
\par\vspace{0.012\textheight}\noindent
\begin{minipage}[t][0.30\textheight][t]{0.32\textwidth}
\vspace{0pt}\centering
\begin{lrbox}{\validationentrybox}
\begin{minipage}{\linewidth}
  \vspace{0pt}
  \centering
  \objectlabel{HANDAL}{34}{handal}
  \captionsetup{font=scriptsize,skip=2pt,hypcap=false}
  \captionof{table}{Validation AR (\%). 441 validation target frames.}
  \label{tab:app_handal_34}
  \fontsize{6.3}{7.0}\selectfont
  \renewcommand{\arraystretch}{0.95}
  \setlength{\tabcolsep}{1.6pt}
  \begin{lrbox}{\validationtablebox}
  \begin{tabular}{@{}lrrrr@{}}
    \toprule
    Variant & Labels & Single & Jitter & GT-box \\
    \midrule
    Synthetic base & -- & \rankcell{4}{93.99} & \rankcell{5}{94.33} & \rankcell{5}{94.25} \\
    Static pseudo & 668 & \rankcell{2}{94.38} & \rankcell{1}{94.60} & \rankcell{3}{94.52} \\
    Box/mask supervision & 275 & \rankcell{3}{94.33} & \rankcell{3}{94.50} & \rankcell{2}{94.59} \\
    Box/mask/keypoint supervision & 275 & \rankcell{2}{94.38} & \rankcell{2}{94.59} & \rankcell{4}{94.44} \\
    Test-input pseudo & 139 & \rankcell{1}{94.52} & \rankcell{4}{94.39} & \rankcell{1}{94.90} \\
    \bottomrule
  \end{tabular}
  \end{lrbox}
  \includegraphics[width=\linewidth,keepaspectratio]{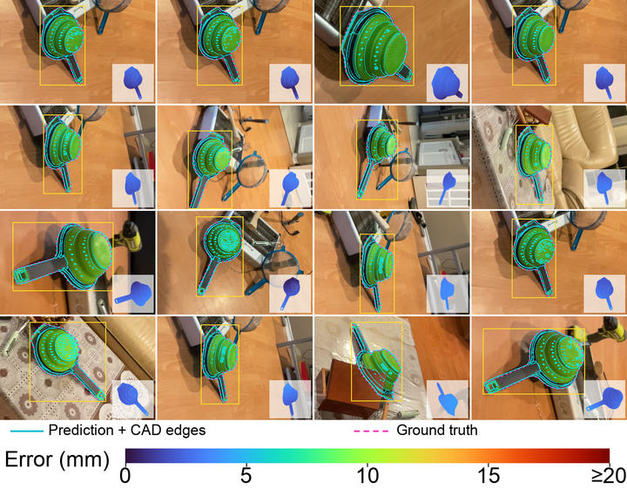}
  \par\vspace{2pt}
  \ifdim\wd\validationtablebox>\linewidth
    \resizebox{\linewidth}{!}{\usebox{\validationtablebox}}
  \else
    \usebox{\validationtablebox}
  \fi
\end{minipage}
\end{lrbox}
\ifdim\dimexpr\ht\validationentrybox+\dp\validationentrybox\relax>0.30\textheight
  \resizebox*{!}{0.30\textheight}{\usebox{\validationentrybox}}
\else
  \usebox{\validationentrybox}
\fi
\end{minipage}
\hfill%
\begin{minipage}[t][0.30\textheight][t]{0.32\textwidth}
\vspace{0pt}\centering
\begin{lrbox}{\validationentrybox}
\begin{minipage}{\linewidth}
  \vspace{0pt}
  \centering
  \objectlabel{HANDAL}{36}{handal}
  \captionsetup{font=scriptsize,skip=2pt,hypcap=false}
  \captionof{table}{Validation AR (\%). 662 validation target frames.}
  \label{tab:app_handal_36}
  \fontsize{6.3}{7.0}\selectfont
  \renewcommand{\arraystretch}{0.95}
  \setlength{\tabcolsep}{1.6pt}
  \begin{lrbox}{\validationtablebox}
  \begin{tabular}{@{}lrrrr@{}}
    \toprule
    Variant & Labels & Single & Jitter & GT-box \\
    \midrule
    Synthetic base & -- & \rankcell{5}{67.52} & \rankcell{4}{68.75} & \rankcell{3}{68.03} \\
    Static pseudo & 28 & \rankcell{4}{67.60} & \rankcell{5}{68.56} & \rankcell{4}{67.94} \\
    Box/mask supervision & 126 & \rankcell{1}{68.06} & \rankcell{1}{68.86} & \rankcell{2}{68.32} \\
    Box/mask/keypoint supervision & 126 & \rankcell{3}{67.63} & \rankcell{2}{68.78} & \rankcell{5}{66.75} \\
    Test-input pseudo & 57 & \rankcell{2}{67.90} & \rankcell{3}{68.76} & \rankcell{1}{68.66} \\
    \bottomrule
  \end{tabular}
  \end{lrbox}
  \includegraphics[width=\linewidth,keepaspectratio]{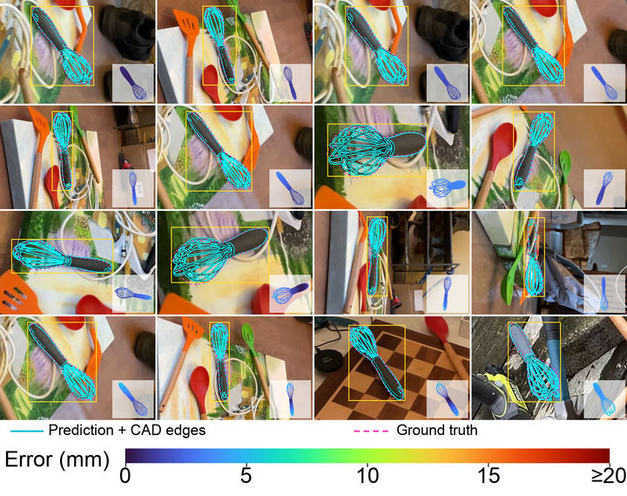}
  \par\vspace{2pt}
  \ifdim\wd\validationtablebox>\linewidth
    \resizebox{\linewidth}{!}{\usebox{\validationtablebox}}
  \else
    \usebox{\validationtablebox}
  \fi
\end{minipage}
\end{lrbox}
\ifdim\dimexpr\ht\validationentrybox+\dp\validationentrybox\relax>0.30\textheight
  \resizebox*{!}{0.30\textheight}{\usebox{\validationentrybox}}
\else
  \usebox{\validationentrybox}
\fi
\end{minipage}
\hfill%
\begin{minipage}[t][0.30\textheight][t]{0.32\textwidth}
\vspace{0pt}\centering
\begin{lrbox}{\validationentrybox}
\begin{minipage}{\linewidth}
  \vspace{0pt}
  \centering
  \objectlabel{HANDAL}{37}{handal}
  \captionsetup{font=scriptsize,skip=2pt,hypcap=false}
  \captionof{table}{Validation AR (\%). 1,099 validation target frames.}
  \label{tab:app_handal_37}
  \fontsize{6.3}{7.0}\selectfont
  \renewcommand{\arraystretch}{0.95}
  \setlength{\tabcolsep}{1.6pt}
  \begin{lrbox}{\validationtablebox}
  \begin{tabular}{@{}lrrrr@{}}
    \toprule
    Variant & Labels & Single & Jitter & GT-box \\
    \midrule
    Synthetic base & -- & \rankcell{5}{65.80} & \rankcell{4}{68.25} & \rankcell{4}{67.27} \\
    Static pseudo & 188 & \rankcell{1}{67.88} & \rankcell{1}{69.37} & \rankcell{1}{68.53} \\
    Box/mask supervision & 122 & \rankcell{3}{66.77} & \rankcell{3}{68.50} & \rankcell{3}{67.74} \\
    Box/mask/keypoint supervision & 122 & \rankcell{2}{67.40} & \rankcell{2}{68.90} & \rankcell{2}{68.51} \\
    Test-input pseudo & 72 & \rankcell{4}{66.14} & \rankcell{5}{67.67} & \rankcell{5}{67.02} \\
    \bottomrule
  \end{tabular}
  \end{lrbox}
  \includegraphics[width=\linewidth,keepaspectratio]{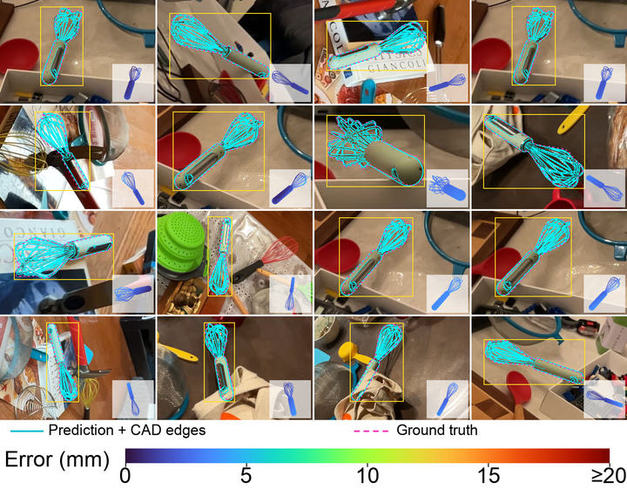}
  \par\vspace{2pt}
  \ifdim\wd\validationtablebox>\linewidth
    \resizebox{\linewidth}{!}{\usebox{\validationtablebox}}
  \else
    \usebox{\validationtablebox}
  \fi
\end{minipage}
\end{lrbox}
\ifdim\dimexpr\ht\validationentrybox+\dp\validationentrybox\relax>0.30\textheight
  \resizebox*{!}{0.30\textheight}{\usebox{\validationentrybox}}
\else
  \usebox{\validationentrybox}
\fi
\end{minipage}
\par
\newpage
\noindent
\begin{minipage}[t][0.30\textheight][t]{0.32\textwidth}
\vspace{0pt}\centering
\begin{lrbox}{\validationentrybox}
\begin{minipage}{\linewidth}
  \vspace{0pt}
  \centering
  \objectlabel{HANDAL}{39}{handal}
  \captionsetup{font=scriptsize,skip=2pt,hypcap=false}
  \captionof{table}{Validation AR (\%). 652 validation target frames.}
  \label{tab:app_handal_39}
  \fontsize{6.3}{7.0}\selectfont
  \renewcommand{\arraystretch}{0.95}
  \setlength{\tabcolsep}{1.6pt}
  \begin{lrbox}{\validationtablebox}
  \begin{tabular}{@{}lrrrr@{}}
    \toprule
    Variant & Labels & Single & Jitter & GT-box \\
    \midrule
    Synthetic base & -- & \rankcell{5}{75.17} & \rankcell{3}{76.90} & \rankcell{3}{74.30} \\
    Static pseudo & 97 & \rankcell{3}{75.32} & \rankcell{2}{77.14} & \rankcell{4}{74.16} \\
    Box/mask supervision & 127 & \rankcell{2}{75.62} & \rankcell{3}{76.90} & \rankcell{2}{74.64} \\
    Box/mask/keypoint supervision & 127 & \rankcell{1}{76.14} & \rankcell{1}{77.38} & \rankcell{1}{74.82} \\
    Test-input pseudo & 50 & \rankcell{4}{75.23} & \rankcell{4}{76.71} & \rankcell{5}{74.10} \\
    \bottomrule
  \end{tabular}
  \end{lrbox}
  \includegraphics[width=\linewidth,keepaspectratio]{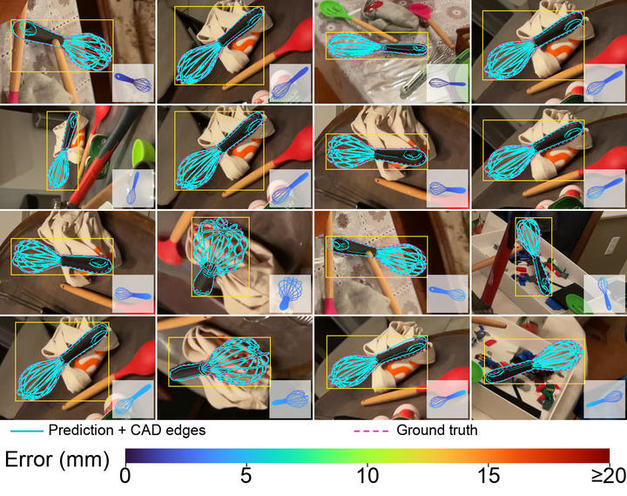}
  \par\vspace{2pt}
  \ifdim\wd\validationtablebox>\linewidth
    \resizebox{\linewidth}{!}{\usebox{\validationtablebox}}
  \else
    \usebox{\validationtablebox}
  \fi
\end{minipage}
\end{lrbox}
\ifdim\dimexpr\ht\validationentrybox+\dp\validationentrybox\relax>0.30\textheight
  \resizebox*{!}{0.30\textheight}{\usebox{\validationentrybox}}
\else
  \usebox{\validationentrybox}
\fi
\end{minipage}
\hfill%
\begin{minipage}[t][0.30\textheight][t]{0.32\textwidth}
\vspace{0pt}\centering
\begin{lrbox}{\validationentrybox}
\begin{minipage}{\linewidth}
  \vspace{0pt}
  \centering
  \objectlabel{HANDAL}{40}{handal}
  \captionsetup{font=scriptsize,skip=2pt,hypcap=false}
  \captionof{table}{Validation AR (\%). 447 validation target frames.}
  \label{tab:app_handal_40}
  \fontsize{6.3}{7.0}\selectfont
  \renewcommand{\arraystretch}{0.95}
  \setlength{\tabcolsep}{1.6pt}
  \begin{lrbox}{\validationtablebox}
  \begin{tabular}{@{}lrrrr@{}}
    \toprule
    Variant & Labels & Single & Jitter & GT-box \\
    \midrule
    Synthetic base & -- & \rankcell{4}{88.13} & 88.91 & \rankcell{3}{87.18} \\
    Static pseudo & 40 & \rankcell{3}{88.29} & \rankcell{5}{89.07} & \rankcell{5}{86.99} \\
    Dynamic pseudo & 0 & -- & -- & -- \\
    Static + dynamic & 40 & \rankcell{2}{88.40} & \rankcell{4}{89.09} & \rankcell{4}{87.15} \\
    Box/mask supervision & 136 & \rankcell{4}{88.13} & \rankcell{3}{89.23} & 86.91 \\
    Box/mask/keypoint supervision & 136 & \rankcell{3}{88.29} & \rankcell{2}{89.30} & \rankcell{2}{87.35} \\
    Test-input pseudo & 76 & \rankcell{1}{88.48} & \rankcell{1}{89.69} & \rankcell{1}{87.43} \\
    \bottomrule
  \end{tabular}
  \end{lrbox}
  \includegraphics[width=\linewidth,keepaspectratio]{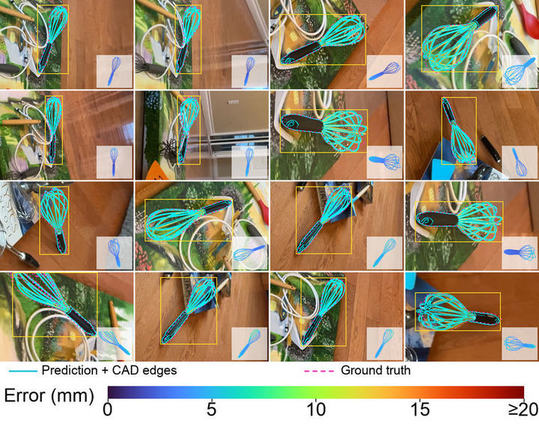}
  \par\vspace{2pt}
  \ifdim\wd\validationtablebox>\linewidth
    \resizebox{\linewidth}{!}{\usebox{\validationtablebox}}
  \else
    \usebox{\validationtablebox}
  \fi
  \par\vspace{2pt}\raggedright All rows: ConvNeXt-Small. Test-input pseudo is transductive. Object 40 remains outside the original macro-average population. Grid: box/mask/keypoint adaptation.
\end{minipage}
\end{lrbox}
\ifdim\dimexpr\ht\validationentrybox+\dp\validationentrybox\relax>0.30\textheight
  \resizebox*{!}{0.30\textheight}{\usebox{\validationentrybox}}
\else
  \usebox{\validationentrybox}
\fi
\end{minipage}
\hfill%
\begin{minipage}[t]{0.32\textwidth}\end{minipage}
\clearpage
\noindent
\begin{minipage}[t][0.30\textheight][t]{0.32\textwidth}
\vspace{0pt}\centering
\begin{lrbox}{\validationentrybox}
\begin{minipage}{\linewidth}
  \vspace{0pt}
  \centering
  \objectlabel{HOPE}{01}{hope}
  \captionsetup{font=scriptsize,skip=2pt,hypcap=false}
  \captionof{table}{Validation AR (\%).}
  \label{tab:app_hope_01}
  \fontsize{6.3}{7.0}\selectfont
  \renewcommand{\arraystretch}{0.95}
  \setlength{\tabcolsep}{1.6pt}
  \begin{lrbox}{\validationtablebox}
  \begin{tabular}{@{}lrrrr@{}}
    \toprule
    Variant & Labels & Single & Jitter & GT-box \\
    \midrule
    Synthetic base & -- & \rankcell{5}{77.11} & \rankcell{5}{77.11} & \rankcell{3}{88.33} \\
    Static pseudo & 7 & -- & -- & -- \\
    Dynamic pseudo & 193 & \rankcell{4}{77.22} & \rankcell{4}{77.22} & \rankcell{1}{90.22} \\
    Static + dynamic & 200 & \rankcell{3}{77.44} & \rankcell{3}{77.44} & \rankcell{2}{89.44} \\
    Box/mask supervision & 347 & \rankcell{2}{81.00} & \rankcell{2}{81.00} & \rankcell{4}{87.44} \\
    Box/mask/keypoint supervision & 347 & \rankcell{1}{81.56} & \rankcell{1}{81.56} & \rankcell{5}{86.78} \\
    \bottomrule
  \end{tabular}
  \end{lrbox}
  \includegraphics[width=\linewidth,keepaspectratio]{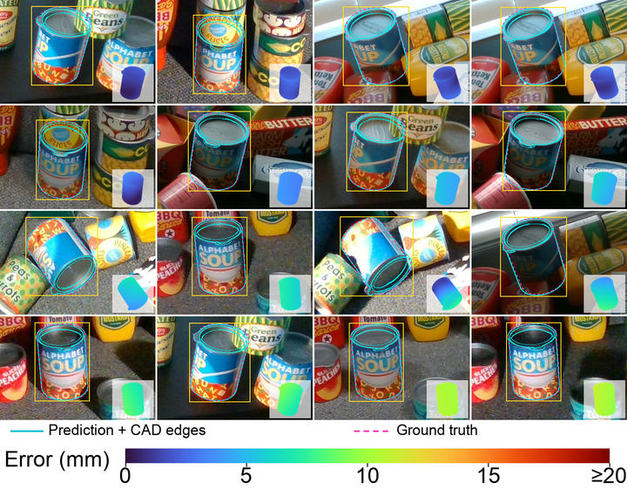}
  \par\vspace{2pt}
  \ifdim\wd\validationtablebox>\linewidth
    \resizebox{\linewidth}{!}{\usebox{\validationtablebox}}
  \else
    \usebox{\validationtablebox}
  \fi
\end{minipage}
\end{lrbox}
\ifdim\dimexpr\ht\validationentrybox+\dp\validationentrybox\relax>0.30\textheight
  \resizebox*{!}{0.30\textheight}{\usebox{\validationentrybox}}
\else
  \usebox{\validationentrybox}
\fi
\end{minipage}
\hfill%
\begin{minipage}[t][0.30\textheight][t]{0.32\textwidth}
\vspace{0pt}\centering
\begin{lrbox}{\validationentrybox}
\begin{minipage}{\linewidth}
  \vspace{0pt}
  \centering
  \objectlabel{HOPE}{02}{hope}
  \captionsetup{font=scriptsize,skip=2pt,hypcap=false}
  \captionof{table}{Validation AR (\%).}
  \label{tab:app_hope_02}
  \fontsize{6.3}{7.0}\selectfont
  \renewcommand{\arraystretch}{0.95}
  \setlength{\tabcolsep}{1.6pt}
  \begin{lrbox}{\validationtablebox}
  \begin{tabular}{@{}lrrrr@{}}
    \toprule
    Variant & Labels & Single & Jitter & GT-box \\
    \midrule
    Synthetic base & -- & \rankcell{2}{76.11} & \rankcell{2}{76.11} & \rankcell{3}{89.22} \\
    Static pseudo & 6 & -- & -- & -- \\
    Dynamic pseudo & 130 & \rankcell{2}{76.11} & \rankcell{2}{76.11} & \rankcell{5}{87.44} \\
    Static + dynamic & 136 & \rankcell{3}{75.78} & \rankcell{3}{75.78} & \rankcell{4}{87.67} \\
    Box/mask supervision & 376 & \rankcell{4}{75.67} & \rankcell{4}{75.67} & \rankcell{2}{89.56} \\
    Box/mask/keypoint supervision & 376 & \rankcell{1}{76.33} & \rankcell{1}{76.33} & \rankcell{1}{89.67} \\
    \bottomrule
  \end{tabular}
  \end{lrbox}
  \includegraphics[width=\linewidth,keepaspectratio]{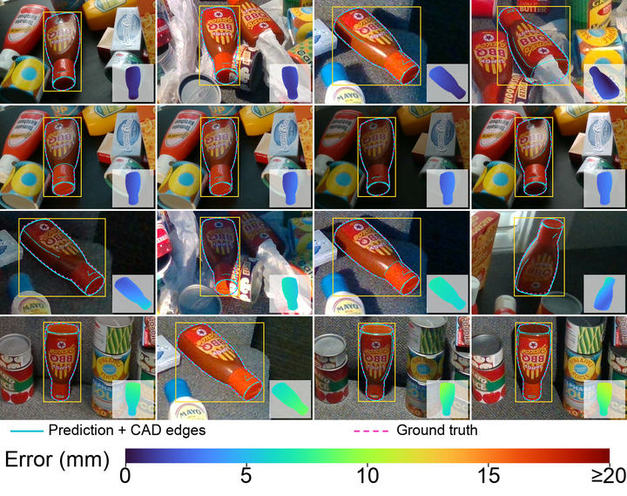}
  \par\vspace{2pt}
  \ifdim\wd\validationtablebox>\linewidth
    \resizebox{\linewidth}{!}{\usebox{\validationtablebox}}
  \else
    \usebox{\validationtablebox}
  \fi
\end{minipage}
\end{lrbox}
\ifdim\dimexpr\ht\validationentrybox+\dp\validationentrybox\relax>0.30\textheight
  \resizebox*{!}{0.30\textheight}{\usebox{\validationentrybox}}
\else
  \usebox{\validationentrybox}
\fi
\end{minipage}
\hfill%
\begin{minipage}[t][0.30\textheight][t]{0.32\textwidth}
\vspace{0pt}\centering
\begin{lrbox}{\validationentrybox}
\begin{minipage}{\linewidth}
  \vspace{0pt}
  \centering
  \objectlabel{HOPE}{03}{hope}
  \captionsetup{font=scriptsize,skip=2pt,hypcap=false}
  \captionof{table}{Validation AR (\%).}
  \label{tab:app_hope_03}
  \fontsize{6.3}{7.0}\selectfont
  \renewcommand{\arraystretch}{0.95}
  \setlength{\tabcolsep}{1.6pt}
  \begin{lrbox}{\validationtablebox}
  \begin{tabular}{@{}lrrrr@{}}
    \toprule
    Variant & Labels & Single & Jitter & GT-box \\
    \midrule
    Synthetic base & -- & \rankcell{3}{91.44} & \rankcell{3}{91.44} & \rankcell{1}{92.00} \\
    Static pseudo & 85 & \rankcell{1}{92.33} & \rankcell{1}{92.33} & \rankcell{3}{91.44} \\
    Dynamic pseudo & 183 & \rankcell{4}{91.00} & \rankcell{4}{91.00} & \rankcell{4}{91.33} \\
    Static + dynamic & 268 & \rankcell{3}{91.44} & \rankcell{3}{91.44} & \rankcell{2}{91.56} \\
    Box/mask supervision & 375 & \rankcell{2}{91.78} & \rankcell{2}{91.78} & \rankcell{5}{91.22} \\
    Box/mask/keypoint supervision & 375 & \rankcell{4}{91.00} & \rankcell{4}{91.00} & \rankcell{3}{91.44} \\
    \bottomrule
  \end{tabular}
  \end{lrbox}
  \includegraphics[width=\linewidth,keepaspectratio]{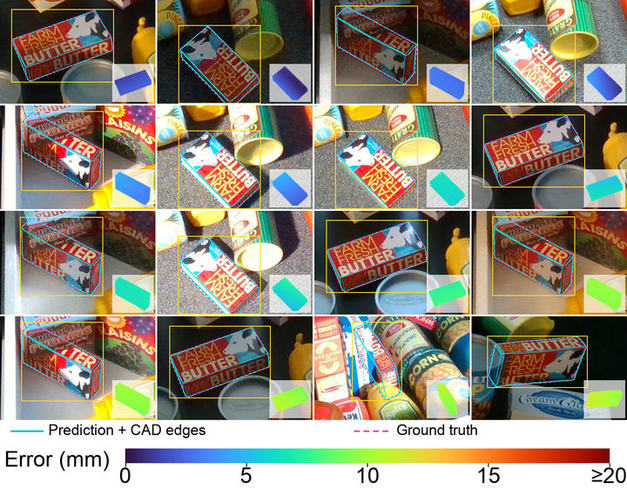}
  \par\vspace{2pt}
  \ifdim\wd\validationtablebox>\linewidth
    \resizebox{\linewidth}{!}{\usebox{\validationtablebox}}
  \else
    \usebox{\validationtablebox}
  \fi
\end{minipage}
\end{lrbox}
\ifdim\dimexpr\ht\validationentrybox+\dp\validationentrybox\relax>0.30\textheight
  \resizebox*{!}{0.30\textheight}{\usebox{\validationentrybox}}
\else
  \usebox{\validationentrybox}
\fi
\end{minipage}
\par\vspace{0.012\textheight}\noindent
\begin{minipage}[t][0.30\textheight][t]{0.32\textwidth}
\vspace{0pt}\centering
\begin{lrbox}{\validationentrybox}
\begin{minipage}{\linewidth}
  \vspace{0pt}
  \centering
  \objectlabel{HOPE}{04}{hope}
  \captionsetup{font=scriptsize,skip=2pt,hypcap=false}
  \captionof{table}{Validation AR (\%).}
  \label{tab:app_hope_04}
  \fontsize{6.3}{7.0}\selectfont
  \renewcommand{\arraystretch}{0.95}
  \setlength{\tabcolsep}{1.6pt}
  \begin{lrbox}{\validationtablebox}
  \begin{tabular}{@{}lrrrr@{}}
    \toprule
    Variant & Labels & Single & Jitter & GT-box \\
    \midrule
    Synthetic base & -- & \rankcell{2}{80.50} & \rankcell{2}{80.50} & \rankcell{1}{82.50} \\
    Static pseudo & 53 & \rankcell{1}{82.00} & \rankcell{1}{82.00} & \rankcell{1}{82.50} \\
    Dynamic pseudo & 179 & \rankcell{4}{79.25} & \rankcell{3}{79.75} & \rankcell{1}{82.50} \\
    Static + dynamic & 232 & \rankcell{3}{79.75} & \rankcell{3}{79.75} & \rankcell{2}{81.50} \\
    Box/mask supervision & 390 & \rankcell{4}{79.25} & \rankcell{4}{79.25} & \rankcell{3}{80.50} \\
    Box/mask/keypoint supervision & 390 & \rankcell{5}{78.25} & \rankcell{5}{78.50} & \rankcell{4}{80.25} \\
    \bottomrule
  \end{tabular}
  \end{lrbox}
  \includegraphics[width=\linewidth,keepaspectratio]{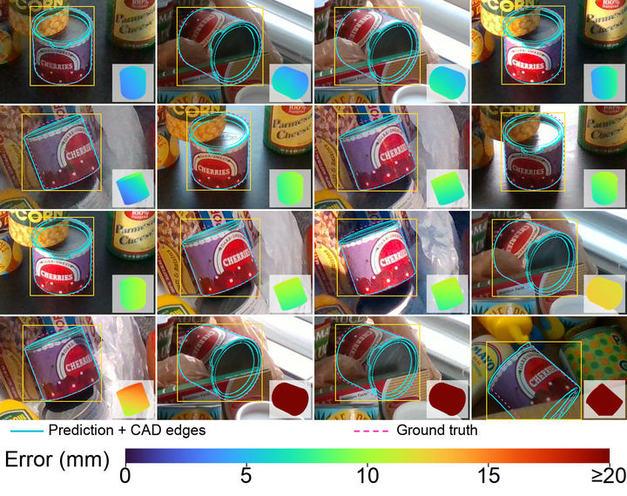}
  \par\vspace{2pt}
  \ifdim\wd\validationtablebox>\linewidth
    \resizebox{\linewidth}{!}{\usebox{\validationtablebox}}
  \else
    \usebox{\validationtablebox}
  \fi
\end{minipage}
\end{lrbox}
\ifdim\dimexpr\ht\validationentrybox+\dp\validationentrybox\relax>0.30\textheight
  \resizebox*{!}{0.30\textheight}{\usebox{\validationentrybox}}
\else
  \usebox{\validationentrybox}
\fi
\end{minipage}
\hfill%
\begin{minipage}[t][0.30\textheight][t]{0.32\textwidth}
\vspace{0pt}\centering
\begin{lrbox}{\validationentrybox}
\begin{minipage}{\linewidth}
  \vspace{0pt}
  \centering
  \objectlabel{HOPE}{05}{hope}
  \captionsetup{font=scriptsize,skip=2pt,hypcap=false}
  \captionof{table}{Validation AR (\%).}
  \label{tab:app_hope_05}
  \fontsize{6.3}{7.0}\selectfont
  \renewcommand{\arraystretch}{0.95}
  \setlength{\tabcolsep}{1.6pt}
  \begin{lrbox}{\validationtablebox}
  \begin{tabular}{@{}lrrrr@{}}
    \toprule
    Variant & Labels & Single & Jitter & GT-box \\
    \midrule
    Synthetic base & -- & 80.67 & \rankcell{4}{83.83} & \rankcell{1}{84.00} \\
    Static pseudo & 151 & \rankcell{5}{82.00} & \rankcell{1}{87.17} & \rankcell{3}{83.50} \\
    Dynamic pseudo & 231 & \rankcell{1}{84.67} & \rankcell{3}{85.17} & \rankcell{2}{83.83} \\
    Static + dynamic & 382 & \rankcell{3}{83.17} & \rankcell{2}{86.67} & \rankcell{4}{82.83} \\
    Box/mask supervision & 361 & \rankcell{4}{82.67} & \rankcell{5}{83.17} & \rankcell{5}{82.33} \\
    Box/mask/keypoint supervision & 361 & \rankcell{2}{83.50} & \rankcell{4}{83.83} & 81.67 \\
    \bottomrule
  \end{tabular}
  \end{lrbox}
  \includegraphics[width=\linewidth,keepaspectratio]{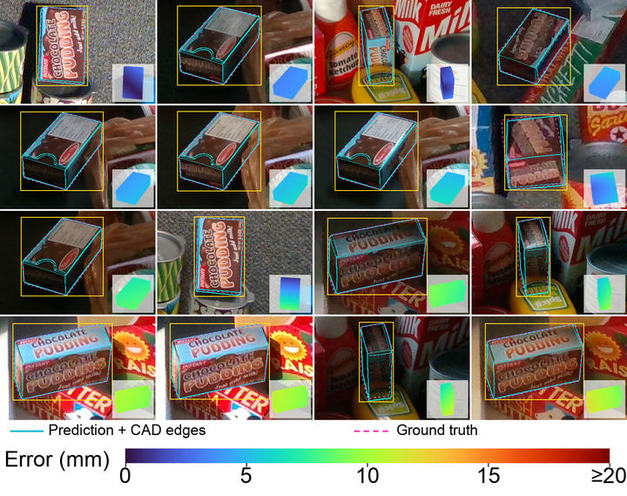}
  \par\vspace{2pt}
  \ifdim\wd\validationtablebox>\linewidth
    \resizebox{\linewidth}{!}{\usebox{\validationtablebox}}
  \else
    \usebox{\validationtablebox}
  \fi
\end{minipage}
\end{lrbox}
\ifdim\dimexpr\ht\validationentrybox+\dp\validationentrybox\relax>0.30\textheight
  \resizebox*{!}{0.30\textheight}{\usebox{\validationentrybox}}
\else
  \usebox{\validationentrybox}
\fi
\end{minipage}
\hfill%
\begin{minipage}[t][0.30\textheight][t]{0.32\textwidth}
\vspace{0pt}\centering
\begin{lrbox}{\validationentrybox}
\begin{minipage}{\linewidth}
  \vspace{0pt}
  \centering
  \objectlabel{HOPE}{06}{hope}
  \captionsetup{font=scriptsize,skip=2pt,hypcap=false}
  \captionof{table}{Validation AR (\%).}
  \label{tab:app_hope_06}
  \fontsize{6.3}{7.0}\selectfont
  \renewcommand{\arraystretch}{0.95}
  \setlength{\tabcolsep}{1.6pt}
  \begin{lrbox}{\validationtablebox}
  \begin{tabular}{@{}lrrrr@{}}
    \toprule
    Variant & Labels & Single & Jitter & GT-box \\
    \midrule
    Synthetic base & -- & \rankcell{2}{85.20} & \rankcell{2}{85.20} & \rankcell{3}{85.20} \\
    Static pseudo & 12 & -- & -- & -- \\
    Dynamic pseudo & 34 & \rankcell{3}{84.80} & \rankcell{3}{84.80} & \rankcell{2}{85.40} \\
    Static + dynamic & 46 & \rankcell{4}{84.40} & \rankcell{4}{84.40} & \rankcell{3}{85.20} \\
    Box/mask supervision & 384 & \rankcell{5}{83.20} & \rankcell{5}{83.00} & \rankcell{4}{84.20} \\
    Box/mask/keypoint supervision & 384 & \rankcell{1}{87.40} & \rankcell{1}{87.40} & \rankcell{1}{88.40} \\
    \bottomrule
  \end{tabular}
  \end{lrbox}
  \includegraphics[width=\linewidth,keepaspectratio]{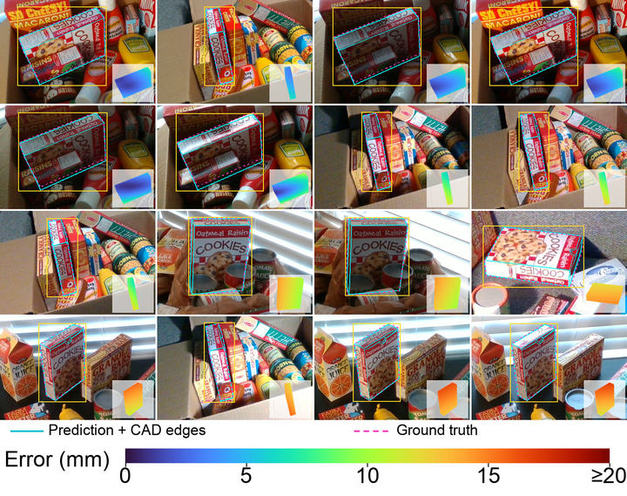}
  \par\vspace{2pt}
  \ifdim\wd\validationtablebox>\linewidth
    \resizebox{\linewidth}{!}{\usebox{\validationtablebox}}
  \else
    \usebox{\validationtablebox}
  \fi
\end{minipage}
\end{lrbox}
\ifdim\dimexpr\ht\validationentrybox+\dp\validationentrybox\relax>0.30\textheight
  \resizebox*{!}{0.30\textheight}{\usebox{\validationentrybox}}
\else
  \usebox{\validationentrybox}
\fi
\end{minipage}
\par\vspace{0.012\textheight}\noindent
\begin{minipage}[t][0.30\textheight][t]{0.32\textwidth}
\vspace{0pt}\centering
\begin{lrbox}{\validationentrybox}
\begin{minipage}{\linewidth}
  \vspace{0pt}
  \centering
  \objectlabel{HOPE}{07}{hope}
  \captionsetup{font=scriptsize,skip=2pt,hypcap=false}
  \captionof{table}{Validation AR (\%).}
  \label{tab:app_hope_07}
  \fontsize{6.3}{7.0}\selectfont
  \renewcommand{\arraystretch}{0.95}
  \setlength{\tabcolsep}{1.6pt}
  \begin{lrbox}{\validationtablebox}
  \begin{tabular}{@{}lrrrr@{}}
    \toprule
    Variant & Labels & Single & Jitter & GT-box \\
    \midrule
    Synthetic base & -- & \rankcell{2}{59.88} & \rankcell{2}{62.00} & \rankcell{4}{77.88} \\
    Static pseudo & 42 & \rankcell{5}{57.12} & \rankcell{5}{58.38} & 76.00 \\
    Dynamic pseudo & 139 & \rankcell{5}{57.12} & \rankcell{3}{60.75} & \rankcell{3}{78.12} \\
    Static + dynamic & 181 & \rankcell{4}{58.12} & \rankcell{4}{59.12} & \rankcell{1}{79.50} \\
    Box/mask supervision & 326 & \rankcell{3}{58.62} & \rankcell{1}{62.25} & \rankcell{5}{77.12} \\
    Box/mask/keypoint supervision & 326 & \rankcell{1}{60.88} & \rankcell{2}{62.00} & \rankcell{2}{79.00} \\
    \bottomrule
  \end{tabular}
  \end{lrbox}
  \includegraphics[width=\linewidth,keepaspectratio]{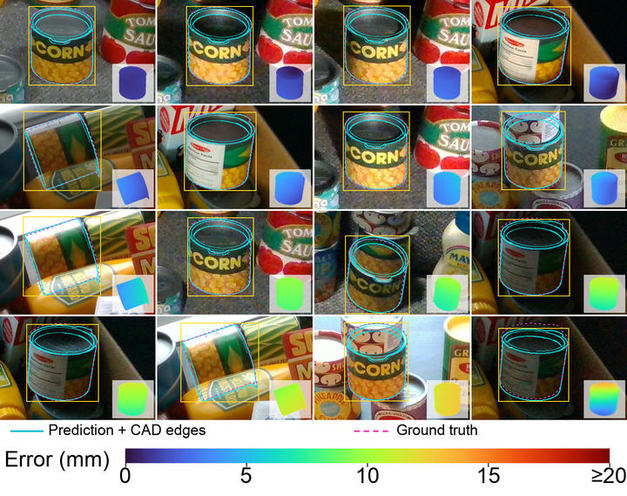}
  \par\vspace{2pt}
  \ifdim\wd\validationtablebox>\linewidth
    \resizebox{\linewidth}{!}{\usebox{\validationtablebox}}
  \else
    \usebox{\validationtablebox}
  \fi
\end{minipage}
\end{lrbox}
\ifdim\dimexpr\ht\validationentrybox+\dp\validationentrybox\relax>0.30\textheight
  \resizebox*{!}{0.30\textheight}{\usebox{\validationentrybox}}
\else
  \usebox{\validationentrybox}
\fi
\end{minipage}
\hfill%
\begin{minipage}[t][0.30\textheight][t]{0.32\textwidth}
\vspace{0pt}\centering
\begin{lrbox}{\validationentrybox}
\begin{minipage}{\linewidth}
  \vspace{0pt}
  \centering
  \objectlabel{HOPE}{08}{hope}
  \captionsetup{font=scriptsize,skip=2pt,hypcap=false}
  \captionof{table}{Validation AR (\%).}
  \label{tab:app_hope_08}
  \fontsize{6.3}{7.0}\selectfont
  \renewcommand{\arraystretch}{0.95}
  \setlength{\tabcolsep}{1.6pt}
  \begin{lrbox}{\validationtablebox}
  \begin{tabular}{@{}lrrrr@{}}
    \toprule
    Variant & Labels & Single & Jitter & GT-box \\
    \midrule
    Synthetic base & -- & \rankcell{2}{61.40} & \rankcell{1}{61.80} & \rankcell{2}{61.40} \\
    Static pseudo & 161 & \rankcell{2}{61.40} & \rankcell{2}{61.60} & \rankcell{5}{60.40} \\
    Dynamic pseudo & 221 & \rankcell{4}{61.00} & \rankcell{3}{61.00} & \rankcell{1}{61.60} \\
    Static + dynamic & 382 & \rankcell{1}{61.60} & \rankcell{1}{61.80} & \rankcell{3}{60.80} \\
    Box/mask supervision & 370 & \rankcell{3}{61.20} & \rankcell{2}{61.60} & \rankcell{4}{60.60} \\
    Box/mask/keypoint supervision & 370 & \rankcell{5}{58.00} & \rankcell{4}{58.20} & 57.60 \\
    \bottomrule
  \end{tabular}
  \end{lrbox}
  \includegraphics[width=\linewidth,keepaspectratio]{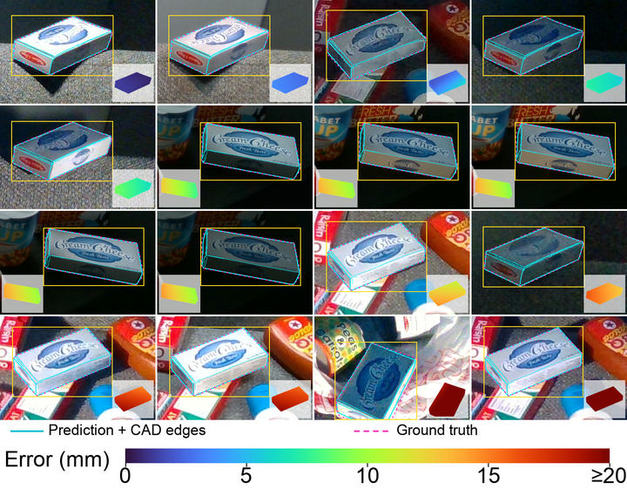}
  \par\vspace{2pt}
  \ifdim\wd\validationtablebox>\linewidth
    \resizebox{\linewidth}{!}{\usebox{\validationtablebox}}
  \else
    \usebox{\validationtablebox}
  \fi
\end{minipage}
\end{lrbox}
\ifdim\dimexpr\ht\validationentrybox+\dp\validationentrybox\relax>0.30\textheight
  \resizebox*{!}{0.30\textheight}{\usebox{\validationentrybox}}
\else
  \usebox{\validationentrybox}
\fi
\end{minipage}
\hfill%
\begin{minipage}[t][0.30\textheight][t]{0.32\textwidth}
\vspace{0pt}\centering
\begin{lrbox}{\validationentrybox}
\begin{minipage}{\linewidth}
  \vspace{0pt}
  \centering
  \objectlabel{HOPE}{09}{hope}
  \captionsetup{font=scriptsize,skip=2pt,hypcap=false}
  \captionof{table}{Validation AR (\%).}
  \label{tab:app_hope_09}
  \fontsize{6.3}{7.0}\selectfont
  \renewcommand{\arraystretch}{0.95}
  \setlength{\tabcolsep}{1.6pt}
  \begin{lrbox}{\validationtablebox}
  \begin{tabular}{@{}lrrrr@{}}
    \toprule
    Variant & Labels & Single & Jitter & GT-box \\
    \midrule
    Synthetic base & -- & \rankcell{4}{94.80} & \rankcell{4}{94.80} & \rankcell{4}{94.80} \\
    Static pseudo & 21 & \rankcell{3}{95.00} & \rankcell{3}{95.00} & \rankcell{3}{95.40} \\
    Dynamic pseudo & 80 & \rankcell{2}{95.20} & \rankcell{2}{95.20} & \rankcell{2}{96.20} \\
    Static + dynamic & 101 & \rankcell{2}{95.20} & \rankcell{2}{95.20} & \rankcell{1}{96.80} \\
    Box/mask supervision & 352 & \rankcell{5}{94.20} & \rankcell{5}{94.20} & \rankcell{5}{93.80} \\
    Box/mask/keypoint supervision & 352 & \rankcell{1}{96.00} & \rankcell{1}{96.00} & \rankcell{2}{96.20} \\
    \bottomrule
  \end{tabular}
  \end{lrbox}
  \includegraphics[width=\linewidth,keepaspectratio]{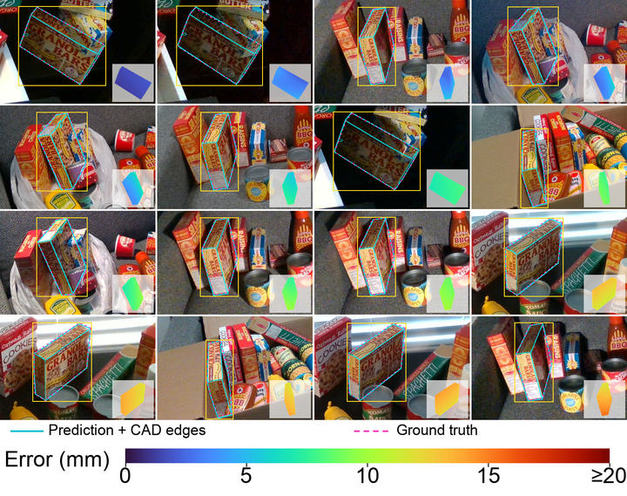}
  \par\vspace{2pt}
  \ifdim\wd\validationtablebox>\linewidth
    \resizebox{\linewidth}{!}{\usebox{\validationtablebox}}
  \else
    \usebox{\validationtablebox}
  \fi
\end{minipage}
\end{lrbox}
\ifdim\dimexpr\ht\validationentrybox+\dp\validationentrybox\relax>0.30\textheight
  \resizebox*{!}{0.30\textheight}{\usebox{\validationentrybox}}
\else
  \usebox{\validationentrybox}
\fi
\end{minipage}
\par
\newpage
\noindent
\begin{minipage}[t][0.30\textheight][t]{0.32\textwidth}
\vspace{0pt}\centering
\begin{lrbox}{\validationentrybox}
\begin{minipage}{\linewidth}
  \vspace{0pt}
  \centering
  \objectlabel{HOPE}{10}{hope}
  \captionsetup{font=scriptsize,skip=2pt,hypcap=false}
  \captionof{table}{Validation AR (\%).}
  \label{tab:app_hope_10}
  \fontsize{6.3}{7.0}\selectfont
  \renewcommand{\arraystretch}{0.95}
  \setlength{\tabcolsep}{1.6pt}
  \begin{lrbox}{\validationtablebox}
  \begin{tabular}{@{}lrrrr@{}}
    \toprule
    Variant & Labels & Single & Jitter & GT-box \\
    \midrule
    Synthetic base & -- & \rankcell{4}{84.14} & \rankcell{4}{84.29} & \rankcell{2}{86.14} \\
    Static pseudo & 26 & \rankcell{3}{84.71} & \rankcell{3}{84.71} & \rankcell{1}{86.86} \\
    Dynamic pseudo & 269 & \rankcell{1}{85.00} & \rankcell{1}{85.00} & \rankcell{3}{86.00} \\
    Static + dynamic & 295 & \rankcell{2}{84.86} & \rankcell{2}{84.86} & \rankcell{4}{85.71} \\
    Box/mask supervision & 363 & \rankcell{5}{82.86} & \rankcell{5}{83.29} & \rankcell{5}{85.00} \\
    Box/mask/keypoint supervision & 363 & 81.43 & 81.29 & 83.00 \\
    \bottomrule
  \end{tabular}
  \end{lrbox}
  \includegraphics[width=\linewidth,keepaspectratio]{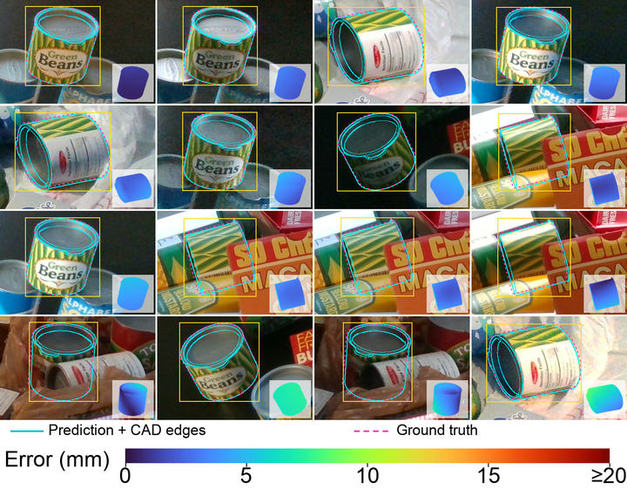}
  \par\vspace{2pt}
  \ifdim\wd\validationtablebox>\linewidth
    \resizebox{\linewidth}{!}{\usebox{\validationtablebox}}
  \else
    \usebox{\validationtablebox}
  \fi
\end{minipage}
\end{lrbox}
\ifdim\dimexpr\ht\validationentrybox+\dp\validationentrybox\relax>0.30\textheight
  \resizebox*{!}{0.30\textheight}{\usebox{\validationentrybox}}
\else
  \usebox{\validationentrybox}
\fi
\end{minipage}
\hfill%
\begin{minipage}[t][0.30\textheight][t]{0.32\textwidth}
\vspace{0pt}\centering
\begin{lrbox}{\validationentrybox}
\begin{minipage}{\linewidth}
  \vspace{0pt}
  \centering
  \objectlabel{HOPE}{11}{hope}
  \captionsetup{font=scriptsize,skip=2pt,hypcap=false}
  \captionof{table}{Validation AR (\%).}
  \label{tab:app_hope_11}
  \fontsize{6.3}{7.0}\selectfont
  \renewcommand{\arraystretch}{0.95}
  \setlength{\tabcolsep}{1.6pt}
  \begin{lrbox}{\validationtablebox}
  \begin{tabular}{@{}lrrrr@{}}
    \toprule
    Variant & Labels & Single & Jitter & GT-box \\
    \midrule
    Synthetic base & -- & \rankcell{2}{81.25} & \rankcell{2}{81.25} & \rankcell{3}{86.88} \\
    Static pseudo & 26 & \rankcell{4}{80.75} & \rankcell{4}{80.75} & \rankcell{3}{86.88} \\
    Dynamic pseudo & 224 & \rankcell{2}{81.25} & \rankcell{5}{80.50} & \rankcell{2}{87.38} \\
    Static + dynamic & 250 & \rankcell{3}{81.12} & \rankcell{3}{81.12} & \rankcell{1}{87.50} \\
    Box/mask supervision & 374 & \rankcell{1}{81.62} & \rankcell{1}{81.62} & \rankcell{4}{86.38} \\
    Box/mask/keypoint supervision & 374 & \rankcell{5}{80.50} & \rankcell{5}{80.50} & \rankcell{1}{87.50} \\
    \bottomrule
  \end{tabular}
  \end{lrbox}
  \includegraphics[width=\linewidth,keepaspectratio]{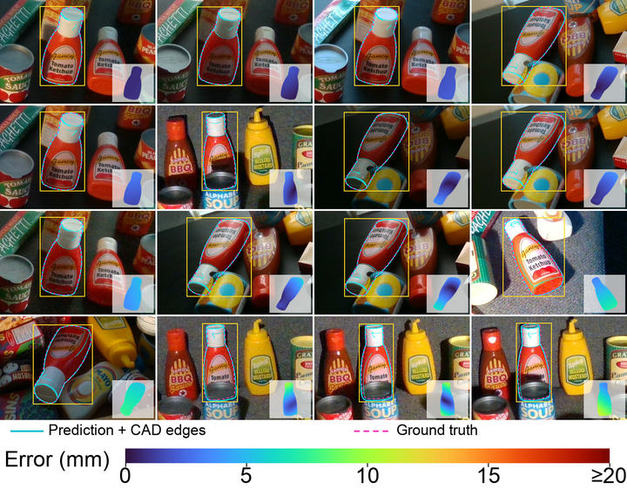}
  \par\vspace{2pt}
  \ifdim\wd\validationtablebox>\linewidth
    \resizebox{\linewidth}{!}{\usebox{\validationtablebox}}
  \else
    \usebox{\validationtablebox}
  \fi
\end{minipage}
\end{lrbox}
\ifdim\dimexpr\ht\validationentrybox+\dp\validationentrybox\relax>0.30\textheight
  \resizebox*{!}{0.30\textheight}{\usebox{\validationentrybox}}
\else
  \usebox{\validationentrybox}
\fi
\end{minipage}
\hfill%
\begin{minipage}[t][0.30\textheight][t]{0.32\textwidth}
\vspace{0pt}\centering
\begin{lrbox}{\validationentrybox}
\begin{minipage}{\linewidth}
  \vspace{0pt}
  \centering
  \objectlabel{HOPE}{12}{hope}
  \captionsetup{font=scriptsize,skip=2pt,hypcap=false}
  \captionof{table}{Validation AR (\%).}
  \label{tab:app_hope_12}
  \fontsize{6.3}{7.0}\selectfont
  \renewcommand{\arraystretch}{0.95}
  \setlength{\tabcolsep}{1.6pt}
  \begin{lrbox}{\validationtablebox}
  \begin{tabular}{@{}lrrrr@{}}
    \toprule
    Variant & Labels & Single & Jitter & GT-box \\
    \midrule
    Synthetic base & -- & \rankcell{2}{78.10} & \rankcell{2}{78.10} & \rankcell{2}{86.10} \\
    Static pseudo & 25 & \rankcell{5}{75.50} & \rankcell{4}{77.50} & \rankcell{4}{85.90} \\
    Dynamic pseudo & 41 & \rankcell{3}{78.00} & \rankcell{3}{78.00} & \rankcell{5}{85.60} \\
    Static + dynamic & 66 & \rankcell{3}{78.00} & \rankcell{3}{78.00} & \rankcell{1}{86.70} \\
    Box/mask supervision & 372 & \rankcell{4}{77.50} & \rankcell{4}{77.50} & \rankcell{3}{86.00} \\
    Box/mask/keypoint supervision & 372 & \rankcell{1}{79.70} & \rankcell{1}{79.80} & \rankcell{4}{85.90} \\
    \bottomrule
  \end{tabular}
  \end{lrbox}
  \includegraphics[width=\linewidth,keepaspectratio]{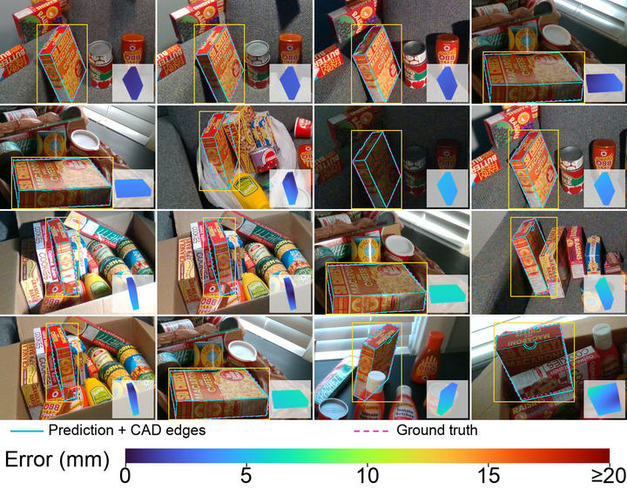}
  \par\vspace{2pt}
  \ifdim\wd\validationtablebox>\linewidth
    \resizebox{\linewidth}{!}{\usebox{\validationtablebox}}
  \else
    \usebox{\validationtablebox}
  \fi
\end{minipage}
\end{lrbox}
\ifdim\dimexpr\ht\validationentrybox+\dp\validationentrybox\relax>0.30\textheight
  \resizebox*{!}{0.30\textheight}{\usebox{\validationentrybox}}
\else
  \usebox{\validationentrybox}
\fi
\end{minipage}
\par\vspace{0.012\textheight}\noindent
\begin{minipage}[t][0.30\textheight][t]{0.32\textwidth}
\vspace{0pt}\centering
\begin{lrbox}{\validationentrybox}
\begin{minipage}{\linewidth}
  \vspace{0pt}
  \centering
  \objectlabel{HOPE}{13}{hope}
  \captionsetup{font=scriptsize,skip=2pt,hypcap=false}
  \captionof{table}{Validation AR (\%).}
  \label{tab:app_hope_13}
  \fontsize{6.3}{7.0}\selectfont
  \renewcommand{\arraystretch}{0.95}
  \setlength{\tabcolsep}{1.6pt}
  \begin{lrbox}{\validationtablebox}
  \begin{tabular}{@{}lrrrr@{}}
    \toprule
    Variant & Labels & Single & Jitter & GT-box \\
    \midrule
    Synthetic base & -- & \rankcell{1}{95.50} & \rankcell{1}{95.50} & \rankcell{3}{95.67} \\
    Static pseudo & 54 & \rankcell{3}{94.83} & \rankcell{3}{94.83} & \rankcell{4}{95.50} \\
    Dynamic pseudo & 268 & \rankcell{3}{94.83} & \rankcell{3}{94.83} & \rankcell{2}{95.83} \\
    Static + dynamic & 322 & \rankcell{4}{94.67} & \rankcell{4}{94.67} & \rankcell{1}{96.50} \\
    Box/mask supervision & 328 & \rankcell{2}{95.17} & \rankcell{2}{95.17} & \rankcell{4}{95.50} \\
    Box/mask/keypoint supervision & 328 & \rankcell{5}{94.50} & \rankcell{5}{94.50} & \rankcell{4}{95.50} \\
    \bottomrule
  \end{tabular}
  \end{lrbox}
  \includegraphics[width=\linewidth,keepaspectratio]{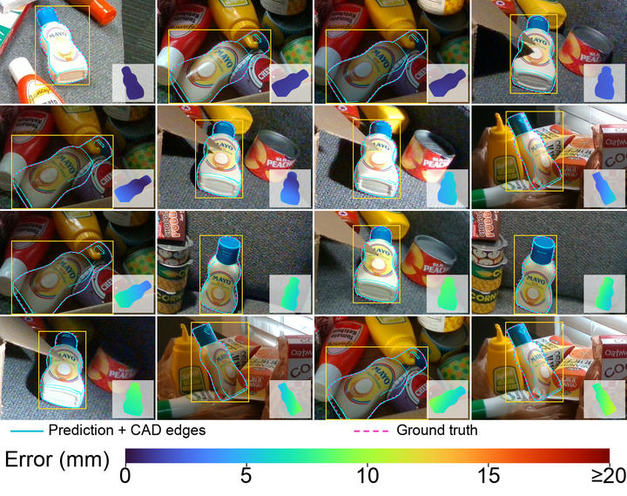}
  \par\vspace{2pt}
  \ifdim\wd\validationtablebox>\linewidth
    \resizebox{\linewidth}{!}{\usebox{\validationtablebox}}
  \else
    \usebox{\validationtablebox}
  \fi
\end{minipage}
\end{lrbox}
\ifdim\dimexpr\ht\validationentrybox+\dp\validationentrybox\relax>0.30\textheight
  \resizebox*{!}{0.30\textheight}{\usebox{\validationentrybox}}
\else
  \usebox{\validationentrybox}
\fi
\end{minipage}
\hfill%
\begin{minipage}[t][0.30\textheight][t]{0.32\textwidth}
\vspace{0pt}\centering
\begin{lrbox}{\validationentrybox}
\begin{minipage}{\linewidth}
  \vspace{0pt}
  \centering
  \objectlabel{HOPE}{14}{hope}
  \captionsetup{font=scriptsize,skip=2pt,hypcap=false}
  \captionof{table}{Validation AR (\%).}
  \label{tab:app_hope_14}
  \fontsize{6.3}{7.0}\selectfont
  \renewcommand{\arraystretch}{0.95}
  \setlength{\tabcolsep}{1.6pt}
  \begin{lrbox}{\validationtablebox}
  \begin{tabular}{@{}lrrrr@{}}
    \toprule
    Variant & Labels & Single & Jitter & GT-box \\
    \midrule
    Synthetic base & -- & \rankcell{1}{98.60} & \rankcell{1}{98.20} & \rankcell{1}{98.20} \\
    Static pseudo & 0 & -- & -- & -- \\
    Dynamic pseudo & 60 & \rankcell{2}{97.40} & \rankcell{3}{96.80} & \rankcell{2}{97.80} \\
    Static + dynamic & 60 & \rankcell{2}{97.40} & \rankcell{5}{96.40} & \rankcell{3}{97.60} \\
    Box/mask supervision & 382 & \rankcell{4}{96.80} & \rankcell{4}{96.60} & \rankcell{3}{97.60} \\
    Box/mask/keypoint supervision & 382 & \rankcell{3}{97.00} & \rankcell{2}{97.00} & \rankcell{4}{97.20} \\
    \bottomrule
  \end{tabular}
  \end{lrbox}
  \includegraphics[width=\linewidth,keepaspectratio]{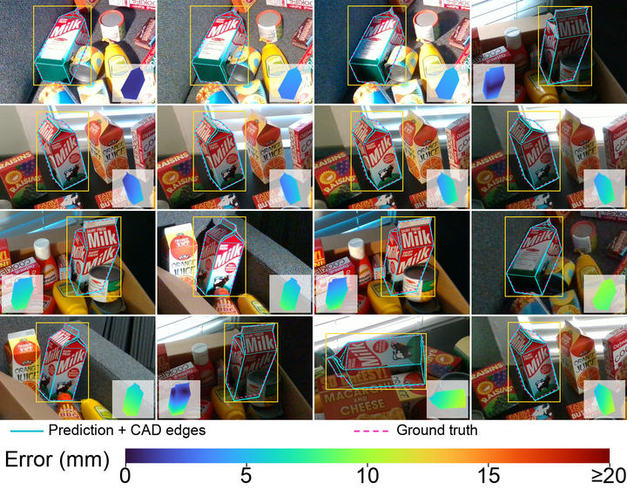}
  \par\vspace{2pt}
  \ifdim\wd\validationtablebox>\linewidth
    \resizebox{\linewidth}{!}{\usebox{\validationtablebox}}
  \else
    \usebox{\validationtablebox}
  \fi
\end{minipage}
\end{lrbox}
\ifdim\dimexpr\ht\validationentrybox+\dp\validationentrybox\relax>0.30\textheight
  \resizebox*{!}{0.30\textheight}{\usebox{\validationentrybox}}
\else
  \usebox{\validationentrybox}
\fi
\end{minipage}
\hfill%
\begin{minipage}[t][0.30\textheight][t]{0.32\textwidth}
\vspace{0pt}\centering
\begin{lrbox}{\validationentrybox}
\begin{minipage}{\linewidth}
  \vspace{0pt}
  \centering
  \objectlabel{HOPE}{15}{hope}
  \captionsetup{font=scriptsize,skip=2pt,hypcap=false}
  \captionof{table}{Validation AR (\%).}
  \label{tab:app_hope_15}
  \fontsize{6.3}{7.0}\selectfont
  \renewcommand{\arraystretch}{0.95}
  \setlength{\tabcolsep}{1.6pt}
  \begin{lrbox}{\validationtablebox}
  \begin{tabular}{@{}lrrrr@{}}
    \toprule
    Variant & Labels & Single & Jitter & GT-box \\
    \midrule
    Synthetic base & -- & \rankcell{2}{57.57} & \rankcell{2}{58.29} & \rankcell{5}{83.29} \\
    Static pseudo & 25 & \rankcell{3}{56.71} & \rankcell{4}{57.71} & \rankcell{3}{84.29} \\
    Dynamic pseudo & 141 & \rankcell{1}{58.00} & \rankcell{1}{60.57} & \rankcell{4}{84.14} \\
    Static + dynamic & 166 & \rankcell{4}{56.29} & \rankcell{3}{58.00} & \rankcell{2}{84.57} \\
    Box/mask supervision & 351 & 52.57 & 53.00 & 82.57 \\
    Box/mask/keypoint supervision & 351 & \rankcell{5}{53.43} & \rankcell{5}{53.29} & \rankcell{1}{84.71} \\
    \bottomrule
  \end{tabular}
  \end{lrbox}
  \includegraphics[width=\linewidth,keepaspectratio]{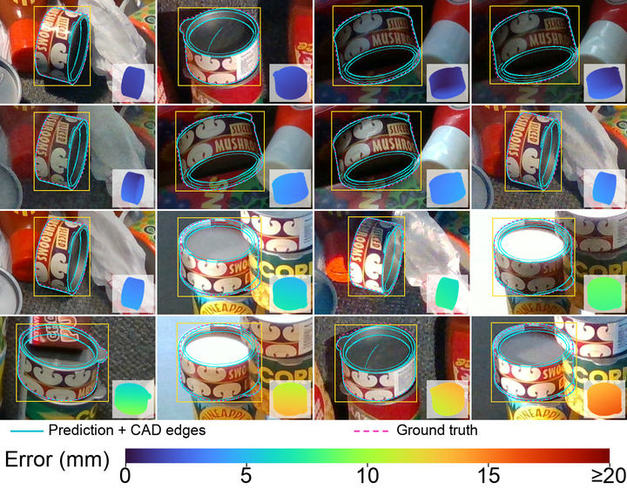}
  \par\vspace{2pt}
  \ifdim\wd\validationtablebox>\linewidth
    \resizebox{\linewidth}{!}{\usebox{\validationtablebox}}
  \else
    \usebox{\validationtablebox}
  \fi
\end{minipage}
\end{lrbox}
\ifdim\dimexpr\ht\validationentrybox+\dp\validationentrybox\relax>0.30\textheight
  \resizebox*{!}{0.30\textheight}{\usebox{\validationentrybox}}
\else
  \usebox{\validationentrybox}
\fi
\end{minipage}
\par\vspace{0.012\textheight}\noindent
\begin{minipage}[t][0.30\textheight][t]{0.32\textwidth}
\vspace{0pt}\centering
\begin{lrbox}{\validationentrybox}
\begin{minipage}{\linewidth}
  \vspace{0pt}
  \centering
  \objectlabel{HOPE}{16}{hope}
  \captionsetup{font=scriptsize,skip=2pt,hypcap=false}
  \captionof{table}{Validation AR (\%).}
  \label{tab:app_hope_16}
  \fontsize{6.3}{7.0}\selectfont
  \renewcommand{\arraystretch}{0.95}
  \setlength{\tabcolsep}{1.6pt}
  \begin{lrbox}{\validationtablebox}
  \begin{tabular}{@{}lrrrr@{}}
    \toprule
    Variant & Labels & Single & Jitter & GT-box \\
    \midrule
    Synthetic base & -- & \rankcell{1}{91.60} & \rankcell{1}{91.60} & \rankcell{4}{87.30} \\
    Static pseudo & 6 & -- & -- & -- \\
    Dynamic pseudo & 161 & \rankcell{4}{90.40} & \rankcell{4}{90.40} & \rankcell{2}{88.00} \\
    Static + dynamic & 167 & \rankcell{3}{91.10} & \rankcell{3}{91.10} & \rankcell{1}{88.80} \\
    Box/mask supervision & 347 & \rankcell{5}{90.30} & \rankcell{5}{90.30} & \rankcell{5}{85.80} \\
    Box/mask/keypoint supervision & 347 & \rankcell{2}{91.40} & \rankcell{2}{91.40} & \rankcell{3}{87.80} \\
    \bottomrule
  \end{tabular}
  \end{lrbox}
  \includegraphics[width=\linewidth,keepaspectratio]{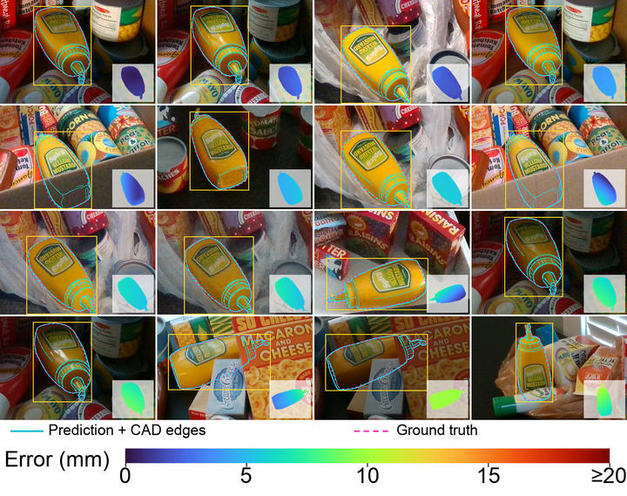}
  \par\vspace{2pt}
  \ifdim\wd\validationtablebox>\linewidth
    \resizebox{\linewidth}{!}{\usebox{\validationtablebox}}
  \else
    \usebox{\validationtablebox}
  \fi
\end{minipage}
\end{lrbox}
\ifdim\dimexpr\ht\validationentrybox+\dp\validationentrybox\relax>0.30\textheight
  \resizebox*{!}{0.30\textheight}{\usebox{\validationentrybox}}
\else
  \usebox{\validationentrybox}
\fi
\end{minipage}
\hfill%
\begin{minipage}[t][0.30\textheight][t]{0.32\textwidth}
\vspace{0pt}\centering
\begin{lrbox}{\validationentrybox}
\begin{minipage}{\linewidth}
  \vspace{0pt}
  \centering
  \objectlabel{HOPE}{17}{hope}
  \captionsetup{font=scriptsize,skip=2pt,hypcap=false}
  \captionof{table}{Validation AR (\%).}
  \label{tab:app_hope_17}
  \fontsize{6.3}{7.0}\selectfont
  \renewcommand{\arraystretch}{0.95}
  \setlength{\tabcolsep}{1.6pt}
  \begin{lrbox}{\validationtablebox}
  \begin{tabular}{@{}lrrrr@{}}
    \toprule
    Variant & Labels & Single & Jitter & GT-box \\
    \midrule
    Synthetic base & -- & \rankcell{1}{77.40} & \rankcell{2}{77.00} & \rankcell{5}{61.40} \\
    Static pseudo & 2 & -- & -- & -- \\
    Dynamic pseudo & 66 & \rankcell{4}{73.20} & \rankcell{2}{77.00} & \rankcell{3}{62.80} \\
    Static + dynamic & 68 & \rankcell{3}{76.80} & \rankcell{1}{77.60} & \rankcell{2}{63.20} \\
    Box/mask supervision & 381 & \rankcell{2}{77.00} & \rankcell{3}{76.80} & \rankcell{4}{62.40} \\
    Box/mask/keypoint supervision & 381 & \rankcell{5}{72.00} & \rankcell{4}{72.80} & \rankcell{1}{64.20} \\
    \bottomrule
  \end{tabular}
  \end{lrbox}
  \includegraphics[width=\linewidth,keepaspectratio]{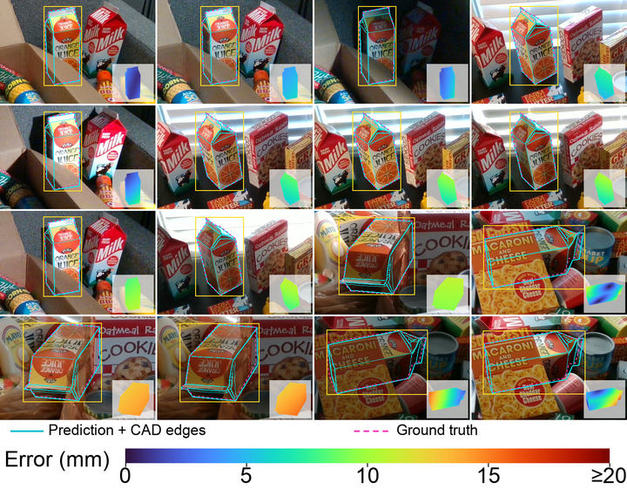}
  \par\vspace{2pt}
  \ifdim\wd\validationtablebox>\linewidth
    \resizebox{\linewidth}{!}{\usebox{\validationtablebox}}
  \else
    \usebox{\validationtablebox}
  \fi
\end{minipage}
\end{lrbox}
\ifdim\dimexpr\ht\validationentrybox+\dp\validationentrybox\relax>0.30\textheight
  \resizebox*{!}{0.30\textheight}{\usebox{\validationentrybox}}
\else
  \usebox{\validationentrybox}
\fi
\end{minipage}
\hfill%
\begin{minipage}[t][0.30\textheight][t]{0.32\textwidth}
\vspace{0pt}\centering
\begin{lrbox}{\validationentrybox}
\begin{minipage}{\linewidth}
  \vspace{0pt}
  \centering
  \objectlabel{HOPE}{18}{hope}
  \captionsetup{font=scriptsize,skip=2pt,hypcap=false}
  \captionof{table}{Validation AR (\%).}
  \label{tab:app_hope_18}
  \fontsize{6.3}{7.0}\selectfont
  \renewcommand{\arraystretch}{0.95}
  \setlength{\tabcolsep}{1.6pt}
  \begin{lrbox}{\validationtablebox}
  \begin{tabular}{@{}lrrrr@{}}
    \toprule
    Variant & Labels & Single & Jitter & GT-box \\
    \midrule
    Synthetic base & -- & \rankcell{3}{88.00} & \rankcell{3}{88.00} & \rankcell{4}{87.00} \\
    Static pseudo & 47 & \rankcell{1}{90.17} & \rankcell{1}{90.17} & \rankcell{1}{88.17} \\
    Dynamic pseudo & 247 & \rankcell{4}{87.83} & \rankcell{4}{87.83} & \rankcell{2}{87.33} \\
    Static + dynamic & 294 & \rankcell{2}{88.67} & \rankcell{2}{88.67} & \rankcell{3}{87.17} \\
    Box/mask supervision & 342 & \rankcell{2}{88.67} & \rankcell{2}{88.67} & 84.83 \\
    Box/mask/keypoint supervision & 342 & \rankcell{5}{87.67} & \rankcell{5}{87.67} & \rankcell{5}{86.50} \\
    \bottomrule
  \end{tabular}
  \end{lrbox}
  \includegraphics[width=\linewidth,keepaspectratio]{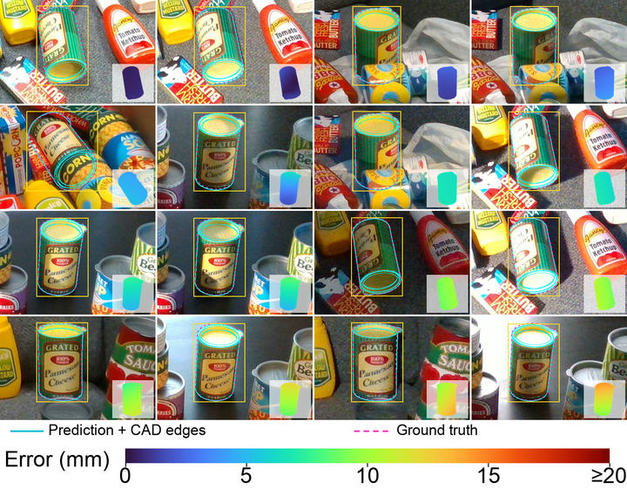}
  \par\vspace{2pt}
  \ifdim\wd\validationtablebox>\linewidth
    \resizebox{\linewidth}{!}{\usebox{\validationtablebox}}
  \else
    \usebox{\validationtablebox}
  \fi
\end{minipage}
\end{lrbox}
\ifdim\dimexpr\ht\validationentrybox+\dp\validationentrybox\relax>0.30\textheight
  \resizebox*{!}{0.30\textheight}{\usebox{\validationentrybox}}
\else
  \usebox{\validationentrybox}
\fi
\end{minipage}
\par
\newpage
\noindent
\begin{minipage}[t][0.30\textheight][t]{0.32\textwidth}
\vspace{0pt}\centering
\begin{lrbox}{\validationentrybox}
\begin{minipage}{\linewidth}
  \vspace{0pt}
  \centering
  \objectlabel{HOPE}{19}{hope}
  \captionsetup{font=scriptsize,skip=2pt,hypcap=false}
  \captionof{table}{Validation AR (\%).}
  \label{tab:app_hope_19}
  \fontsize{6.3}{7.0}\selectfont
  \renewcommand{\arraystretch}{0.95}
  \setlength{\tabcolsep}{1.6pt}
  \begin{lrbox}{\validationtablebox}
  \begin{tabular}{@{}lrrrr@{}}
    \toprule
    Variant & Labels & Single & Jitter & GT-box \\
    \midrule
    Synthetic base & -- & \rankcell{2}{71.75} & \rankcell{2}{71.75} & \rankcell{2}{90.00} \\
    Static pseudo & 135 & \rankcell{5}{70.50} & \rankcell{5}{70.50} & \rankcell{5}{89.00} \\
    Dynamic pseudo & 235 & \rankcell{1}{72.75} & \rankcell{1}{72.75} & \rankcell{3}{89.75} \\
    Static + dynamic & 370 & 69.50 & 69.50 & \rankcell{1}{91.00} \\
    Box/mask supervision & 337 & \rankcell{3}{71.50} & \rankcell{3}{71.50} & \rankcell{4}{89.50} \\
    Box/mask/keypoint supervision & 337 & \rankcell{4}{71.25} & \rankcell{4}{71.25} & 88.50 \\
    \bottomrule
  \end{tabular}
  \end{lrbox}
  \includegraphics[width=\linewidth,keepaspectratio]{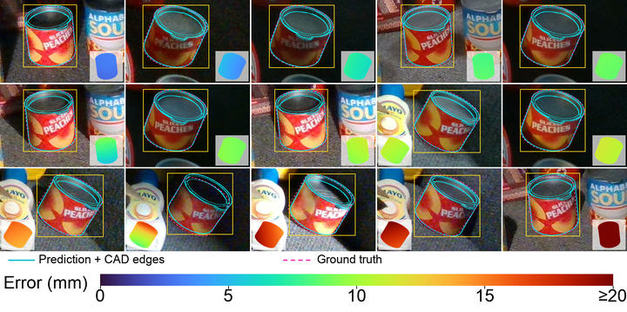}
  \par\vspace{2pt}
  \ifdim\wd\validationtablebox>\linewidth
    \resizebox{\linewidth}{!}{\usebox{\validationtablebox}}
  \else
    \usebox{\validationtablebox}
  \fi
\end{minipage}
\end{lrbox}
\ifdim\dimexpr\ht\validationentrybox+\dp\validationentrybox\relax>0.30\textheight
  \resizebox*{!}{0.30\textheight}{\usebox{\validationentrybox}}
\else
  \usebox{\validationentrybox}
\fi
\end{minipage}
\hfill%
\begin{minipage}[t][0.30\textheight][t]{0.32\textwidth}
\vspace{0pt}\centering
\begin{lrbox}{\validationentrybox}
\begin{minipage}{\linewidth}
  \vspace{0pt}
  \centering
  \objectlabel{HOPE}{20}{hope}
  \captionsetup{font=scriptsize,skip=2pt,hypcap=false}
  \captionof{table}{Validation AR (\%).}
  \label{tab:app_hope_20}
  \fontsize{6.3}{7.0}\selectfont
  \renewcommand{\arraystretch}{0.95}
  \setlength{\tabcolsep}{1.6pt}
  \begin{lrbox}{\validationtablebox}
  \begin{tabular}{@{}lrrrr@{}}
    \toprule
    Variant & Labels & Single & Jitter & GT-box \\
    \midrule
    Synthetic base & -- & \rankcell{2}{74.00} & \rankcell{2}{74.00} & \rankcell{2}{84.50} \\
    Static pseudo & 15 & -- & -- & -- \\
    Dynamic pseudo & 110 & \rankcell{4}{72.67} & \rankcell{4}{72.67} & \rankcell{3}{84.33} \\
    Static + dynamic & 125 & \rankcell{3}{73.50} & \rankcell{3}{73.50} & \rankcell{1}{84.67} \\
    Box/mask supervision & 351 & \rankcell{1}{74.17} & \rankcell{1}{74.17} & \rankcell{1}{84.67} \\
    Box/mask/keypoint supervision & 351 & \rankcell{5}{72.50} & \rankcell{5}{72.50} & \rankcell{4}{83.33} \\
    \bottomrule
  \end{tabular}
  \end{lrbox}
  \includegraphics[width=\linewidth,keepaspectratio]{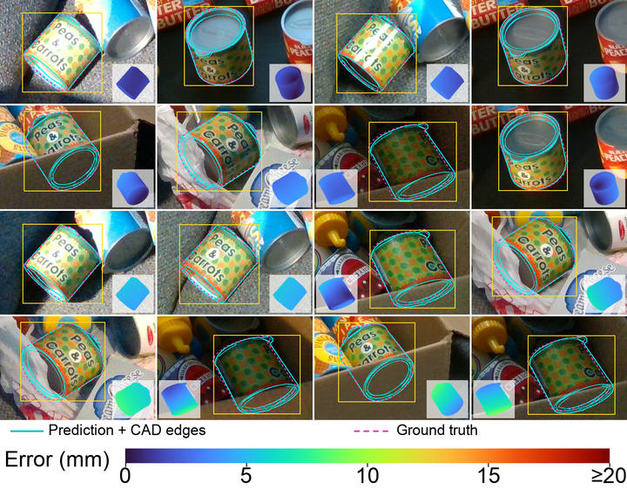}
  \par\vspace{2pt}
  \ifdim\wd\validationtablebox>\linewidth
    \resizebox{\linewidth}{!}{\usebox{\validationtablebox}}
  \else
    \usebox{\validationtablebox}
  \fi
\end{minipage}
\end{lrbox}
\ifdim\dimexpr\ht\validationentrybox+\dp\validationentrybox\relax>0.30\textheight
  \resizebox*{!}{0.30\textheight}{\usebox{\validationentrybox}}
\else
  \usebox{\validationentrybox}
\fi
\end{minipage}
\hfill%
\begin{minipage}[t][0.30\textheight][t]{0.32\textwidth}
\vspace{0pt}\centering
\begin{lrbox}{\validationentrybox}
\begin{minipage}{\linewidth}
  \vspace{0pt}
  \centering
  \objectlabel{HOPE}{21}{hope}
  \captionsetup{font=scriptsize,skip=2pt,hypcap=false}
  \captionof{table}{Validation AR (\%).}
  \label{tab:app_hope_21}
  \fontsize{6.3}{7.0}\selectfont
  \renewcommand{\arraystretch}{0.95}
  \setlength{\tabcolsep}{1.6pt}
  \begin{lrbox}{\validationtablebox}
  \begin{tabular}{@{}lrrrr@{}}
    \toprule
    Variant & Labels & Single & Jitter & GT-box \\
    \midrule
    Synthetic base & -- & \rankcell{3}{85.09} & \rankcell{3}{85.09} & \rankcell{3}{87.55} \\
    Static pseudo & 22 & \rankcell{4}{85.00} & \rankcell{4}{85.00} & \rankcell{2}{87.73} \\
    Dynamic pseudo & 211 & \rankcell{1}{85.45} & \rankcell{2}{85.18} & \rankcell{1}{88.18} \\
    Static + dynamic & 233 & \rankcell{2}{85.18} & \rankcell{1}{85.27} & \rankcell{2}{87.73} \\
    Box/mask supervision & 355 & \rankcell{5}{83.09} & \rankcell{5}{83.09} & \rankcell{4}{87.09} \\
    Box/mask/keypoint supervision & 355 & 83.00 & 82.73 & \rankcell{5}{85.73} \\
    \bottomrule
  \end{tabular}
  \end{lrbox}
  \includegraphics[width=\linewidth,keepaspectratio]{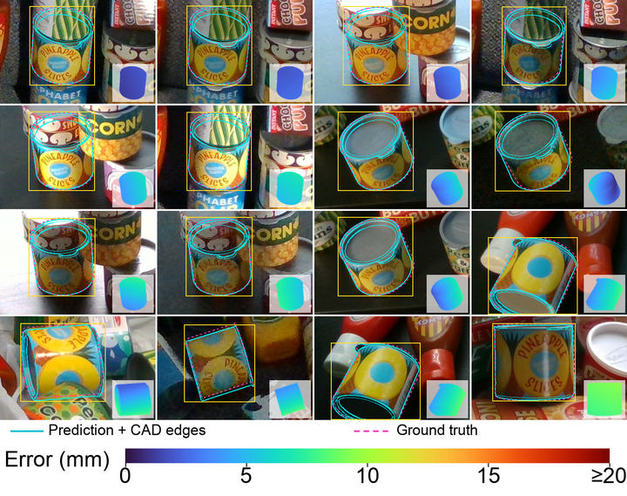}
  \par\vspace{2pt}
  \ifdim\wd\validationtablebox>\linewidth
    \resizebox{\linewidth}{!}{\usebox{\validationtablebox}}
  \else
    \usebox{\validationtablebox}
  \fi
\end{minipage}
\end{lrbox}
\ifdim\dimexpr\ht\validationentrybox+\dp\validationentrybox\relax>0.30\textheight
  \resizebox*{!}{0.30\textheight}{\usebox{\validationentrybox}}
\else
  \usebox{\validationentrybox}
\fi
\end{minipage}
\par\vspace{0.012\textheight}\noindent
\begin{minipage}[t][0.30\textheight][t]{0.32\textwidth}
\vspace{0pt}\centering
\begin{lrbox}{\validationentrybox}
\begin{minipage}{\linewidth}
  \vspace{0pt}
  \centering
  \objectlabel{HOPE}{22}{hope}
  \captionsetup{font=scriptsize,skip=2pt,hypcap=false}
  \captionof{table}{Validation AR (\%).}
  \label{tab:app_hope_22}
  \fontsize{6.3}{7.0}\selectfont
  \renewcommand{\arraystretch}{0.95}
  \setlength{\tabcolsep}{1.6pt}
  \begin{lrbox}{\validationtablebox}
  \begin{tabular}{@{}lrrrr@{}}
    \toprule
    Variant & Labels & Single & Jitter & GT-box \\
    \midrule
    Synthetic base & -- & \rankcell{2}{94.50} & \rankcell{4}{94.00} & \rankcell{3}{93.00} \\
    Static pseudo & 136 & \rankcell{5}{92.50} & 92.50 & \rankcell{4}{92.25} \\
    Dynamic pseudo & 210 & \rankcell{2}{94.50} & \rankcell{2}{94.50} & \rankcell{4}{92.25} \\
    Static + dynamic & 346 & \rankcell{3}{94.25} & \rankcell{3}{94.25} & \rankcell{1}{94.00} \\
    Box/mask supervision & 338 & \rankcell{1}{95.75} & \rankcell{1}{95.50} & \rankcell{3}{93.00} \\
    Box/mask/keypoint supervision & 338 & \rankcell{4}{93.75} & \rankcell{5}{93.75} & \rankcell{2}{93.75} \\
    \bottomrule
  \end{tabular}
  \end{lrbox}
  \includegraphics[width=\linewidth,keepaspectratio]{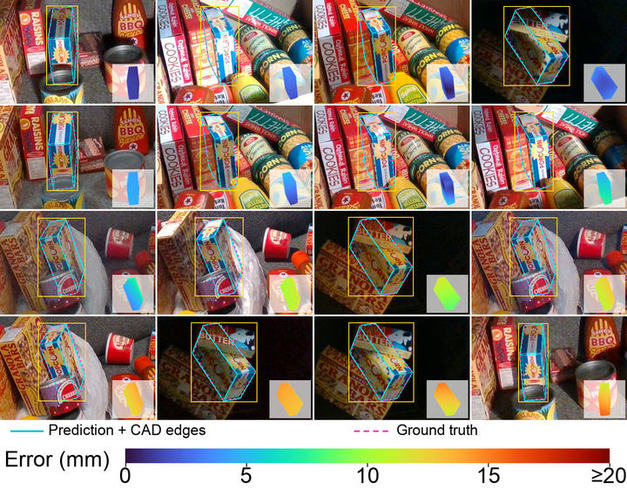}
  \par\vspace{2pt}
  \ifdim\wd\validationtablebox>\linewidth
    \resizebox{\linewidth}{!}{\usebox{\validationtablebox}}
  \else
    \usebox{\validationtablebox}
  \fi
\end{minipage}
\end{lrbox}
\ifdim\dimexpr\ht\validationentrybox+\dp\validationentrybox\relax>0.30\textheight
  \resizebox*{!}{0.30\textheight}{\usebox{\validationentrybox}}
\else
  \usebox{\validationentrybox}
\fi
\end{minipage}
\hfill%
\begin{minipage}[t][0.30\textheight][t]{0.32\textwidth}
\vspace{0pt}\centering
\begin{lrbox}{\validationentrybox}
\begin{minipage}{\linewidth}
  \vspace{0pt}
  \centering
  \objectlabel{HOPE}{23}{hope}
  \captionsetup{font=scriptsize,skip=2pt,hypcap=false}
  \captionof{table}{Validation AR (\%).}
  \label{tab:app_hope_23}
  \fontsize{6.3}{7.0}\selectfont
  \renewcommand{\arraystretch}{0.95}
  \setlength{\tabcolsep}{1.6pt}
  \begin{lrbox}{\validationtablebox}
  \begin{tabular}{@{}lrrrr@{}}
    \toprule
    Variant & Labels & Single & Jitter & GT-box \\
    \midrule
    Synthetic base & -- & \rankcell{3}{85.00} & \rankcell{3}{85.25} & \rankcell{2}{83.75} \\
    Static pseudo & 38 & \rankcell{4}{84.88} & \rankcell{4}{84.88} & \rankcell{3}{83.62} \\
    Dynamic pseudo & 168 & 84.38 & \rankcell{5}{84.50} & 82.75 \\
    Static + dynamic & 206 & \rankcell{5}{84.75} & \rankcell{5}{84.50} & \rankcell{5}{82.88} \\
    Box/mask supervision & 364 & \rankcell{1}{86.00} & \rankcell{1}{86.38} & \rankcell{1}{86.38} \\
    Box/mask/keypoint supervision & 364 & \rankcell{2}{85.50} & \rankcell{2}{85.50} & \rankcell{4}{83.25} \\
    \bottomrule
  \end{tabular}
  \end{lrbox}
  \includegraphics[width=\linewidth,keepaspectratio]{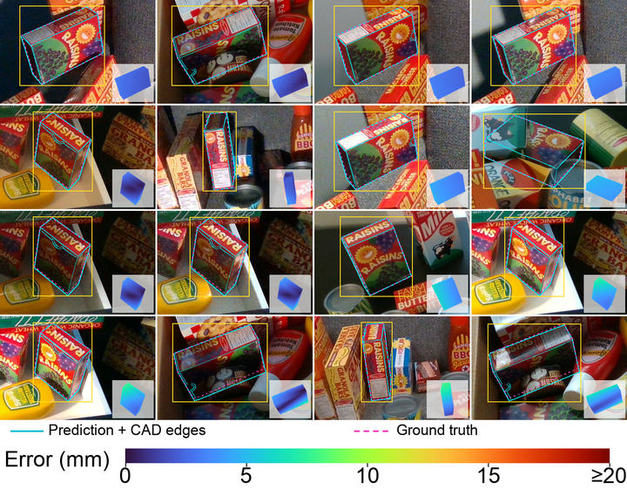}
  \par\vspace{2pt}
  \ifdim\wd\validationtablebox>\linewidth
    \resizebox{\linewidth}{!}{\usebox{\validationtablebox}}
  \else
    \usebox{\validationtablebox}
  \fi
\end{minipage}
\end{lrbox}
\ifdim\dimexpr\ht\validationentrybox+\dp\validationentrybox\relax>0.30\textheight
  \resizebox*{!}{0.30\textheight}{\usebox{\validationentrybox}}
\else
  \usebox{\validationentrybox}
\fi
\end{minipage}
\hfill%
\begin{minipage}[t][0.30\textheight][t]{0.32\textwidth}
\vspace{0pt}\centering
\begin{lrbox}{\validationentrybox}
\begin{minipage}{\linewidth}
  \vspace{0pt}
  \centering
  \objectlabel{HOPE}{24}{hope}
  \captionsetup{font=scriptsize,skip=2pt,hypcap=false}
  \captionof{table}{Validation AR (\%).}
  \label{tab:app_hope_24}
  \fontsize{6.3}{7.0}\selectfont
  \renewcommand{\arraystretch}{0.95}
  \setlength{\tabcolsep}{1.6pt}
  \begin{lrbox}{\validationtablebox}
  \begin{tabular}{@{}lrrrr@{}}
    \toprule
    Variant & Labels & Single & Jitter & GT-box \\
    \midrule
    Synthetic base & -- & \rankcell{3}{96.50} & \rankcell{3}{96.50} & \rankcell{3}{97.00} \\
    Static pseudo & 58 & \rankcell{3}{96.50} & \rankcell{3}{96.50} & \rankcell{3}{97.00} \\
    Dynamic pseudo & 281 & \rankcell{3}{96.50} & \rankcell{3}{96.50} & \rankcell{1}{98.00} \\
    Static + dynamic & 339 & \rankcell{4}{96.00} & \rankcell{4}{96.00} & \rankcell{1}{98.00} \\
    Box/mask supervision & 343 & \rankcell{2}{97.50} & \rankcell{2}{97.50} & \rankcell{1}{98.00} \\
    Box/mask/keypoint supervision & 343 & \rankcell{1}{98.00} & \rankcell{1}{98.00} & \rankcell{2}{97.50} \\
    \bottomrule
  \end{tabular}
  \end{lrbox}
  \includegraphics[width=\linewidth,keepaspectratio]{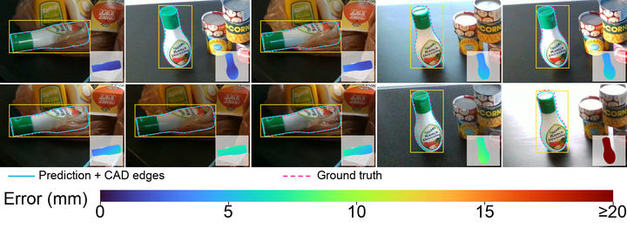}
  \par\vspace{2pt}
  \ifdim\wd\validationtablebox>\linewidth
    \resizebox{\linewidth}{!}{\usebox{\validationtablebox}}
  \else
    \usebox{\validationtablebox}
  \fi
\end{minipage}
\end{lrbox}
\ifdim\dimexpr\ht\validationentrybox+\dp\validationentrybox\relax>0.30\textheight
  \resizebox*{!}{0.30\textheight}{\usebox{\validationentrybox}}
\else
  \usebox{\validationentrybox}
\fi
\end{minipage}
\par\vspace{0.012\textheight}\noindent
\begin{minipage}[t][0.30\textheight][t]{0.32\textwidth}
\vspace{0pt}\centering
\begin{lrbox}{\validationentrybox}
\begin{minipage}{\linewidth}
  \vspace{0pt}
  \centering
  \objectlabel{HOPE}{25}{hope}
  \captionsetup{font=scriptsize,skip=2pt,hypcap=false}
  \captionof{table}{Validation AR (\%).}
  \label{tab:app_hope_25}
  \fontsize{6.3}{7.0}\selectfont
  \renewcommand{\arraystretch}{0.95}
  \setlength{\tabcolsep}{1.6pt}
  \begin{lrbox}{\validationtablebox}
  \begin{tabular}{@{}lrrrr@{}}
    \toprule
    Variant & Labels & Single & Jitter & GT-box \\
    \midrule
    Synthetic base & -- & \rankcell{1}{95.29} & \rankcell{1}{94.43} & \rankcell{3}{94.14} \\
    Static pseudo & 15 & -- & -- & -- \\
    Dynamic pseudo & 354 & \rankcell{2}{94.71} & \rankcell{4}{93.57} & \rankcell{2}{94.29} \\
    Static + dynamic & 369 & \rankcell{2}{94.71} & \rankcell{3}{93.71} & \rankcell{1}{94.57} \\
    Box/mask supervision & 386 & \rankcell{3}{94.57} & \rankcell{2}{94.29} & \rankcell{1}{94.57} \\
    Box/mask/keypoint supervision & 386 & \rankcell{4}{94.00} & \rankcell{2}{94.29} & \rankcell{2}{94.29} \\
    \bottomrule
  \end{tabular}
  \end{lrbox}
  \includegraphics[width=\linewidth,keepaspectratio]{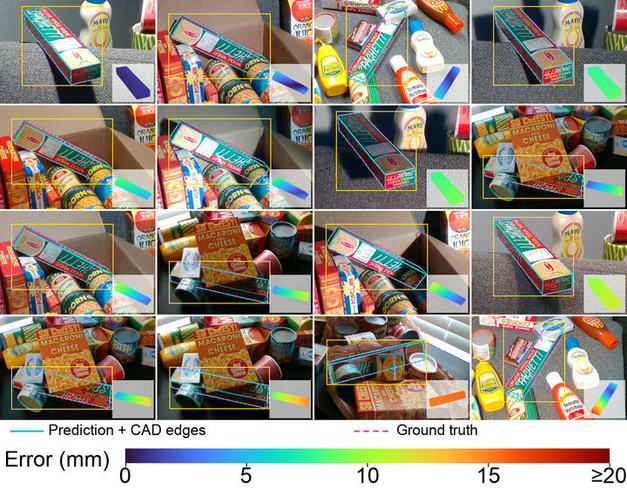}
  \par\vspace{2pt}
  \ifdim\wd\validationtablebox>\linewidth
    \resizebox{\linewidth}{!}{\usebox{\validationtablebox}}
  \else
    \usebox{\validationtablebox}
  \fi
\end{minipage}
\end{lrbox}
\ifdim\dimexpr\ht\validationentrybox+\dp\validationentrybox\relax>0.30\textheight
  \resizebox*{!}{0.30\textheight}{\usebox{\validationentrybox}}
\else
  \usebox{\validationentrybox}
\fi
\end{minipage}
\hfill%
\begin{minipage}[t][0.30\textheight][t]{0.32\textwidth}
\vspace{0pt}\centering
\begin{lrbox}{\validationentrybox}
\begin{minipage}{\linewidth}
  \vspace{0pt}
  \centering
  \objectlabel{HOPE}{26}{hope}
  \captionsetup{font=scriptsize,skip=2pt,hypcap=false}
  \captionof{table}{Validation AR (\%).}
  \label{tab:app_hope_26}
  \fontsize{6.3}{7.0}\selectfont
  \renewcommand{\arraystretch}{0.95}
  \setlength{\tabcolsep}{1.6pt}
  \begin{lrbox}{\validationtablebox}
  \begin{tabular}{@{}lrrrr@{}}
    \toprule
    Variant & Labels & Single & Jitter & GT-box \\
    \midrule
    Synthetic base & -- & \rankcell{1}{83.43} & \rankcell{1}{83.14} & \rankcell{2}{86.86} \\
    Static pseudo & 16 & -- & -- & -- \\
    Dynamic pseudo & 153 & \rankcell{2}{80.57} & \rankcell{2}{80.57} & \rankcell{1}{87.71} \\
    Static + dynamic & 169 & \rankcell{3}{79.29} & \rankcell{3}{79.29} & \rankcell{1}{87.71} \\
    Box/mask supervision & 351 & \rankcell{5}{76.86} & \rankcell{5}{75.14} & \rankcell{4}{84.29} \\
    Box/mask/keypoint supervision & 351 & \rankcell{4}{77.43} & \rankcell{4}{75.43} & \rankcell{3}{85.86} \\
    \bottomrule
  \end{tabular}
  \end{lrbox}
  \includegraphics[width=\linewidth,keepaspectratio]{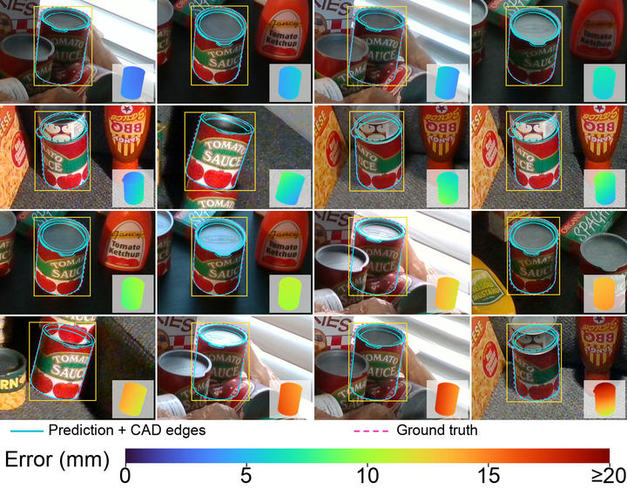}
  \par\vspace{2pt}
  \ifdim\wd\validationtablebox>\linewidth
    \resizebox{\linewidth}{!}{\usebox{\validationtablebox}}
  \else
    \usebox{\validationtablebox}
  \fi
\end{minipage}
\end{lrbox}
\ifdim\dimexpr\ht\validationentrybox+\dp\validationentrybox\relax>0.30\textheight
  \resizebox*{!}{0.30\textheight}{\usebox{\validationentrybox}}
\else
  \usebox{\validationentrybox}
\fi
\end{minipage}
\hfill%
\begin{minipage}[t][0.30\textheight][t]{0.32\textwidth}
\vspace{0pt}\centering
\begin{lrbox}{\validationentrybox}
\begin{minipage}{\linewidth}
  \vspace{0pt}
  \centering
  \objectlabel{HOPE}{27}{hope}
  \captionsetup{font=scriptsize,skip=2pt,hypcap=false}
  \captionof{table}{Validation AR (\%).}
  \label{tab:app_hope_27}
  \fontsize{6.3}{7.0}\selectfont
  \renewcommand{\arraystretch}{0.95}
  \setlength{\tabcolsep}{1.6pt}
  \begin{lrbox}{\validationtablebox}
  \begin{tabular}{@{}lrrrr@{}}
    \toprule
    Variant & Labels & Single & Jitter & GT-box \\
    \midrule
    Synthetic base & -- & 57.00 & \rankcell{4}{60.29} & \rankcell{2}{77.57} \\
    Static pseudo & 84 & \rankcell{3}{59.00} & \rankcell{2}{63.57} & \rankcell{4}{76.29} \\
    Dynamic pseudo & 297 & \rankcell{1}{63.57} & \rankcell{1}{66.43} & \rankcell{1}{78.86} \\
    Static + dynamic & 381 & \rankcell{4}{58.43} & \rankcell{2}{63.57} & \rankcell{5}{75.86} \\
    Box/mask supervision & 347 & \rankcell{5}{58.29} & \rankcell{5}{59.57} & 70.14 \\
    Box/mask/keypoint supervision & 347 & \rankcell{2}{60.00} & \rankcell{3}{61.71} & \rankcell{3}{76.57} \\
    \bottomrule
  \end{tabular}
  \end{lrbox}
  \includegraphics[width=\linewidth,keepaspectratio]{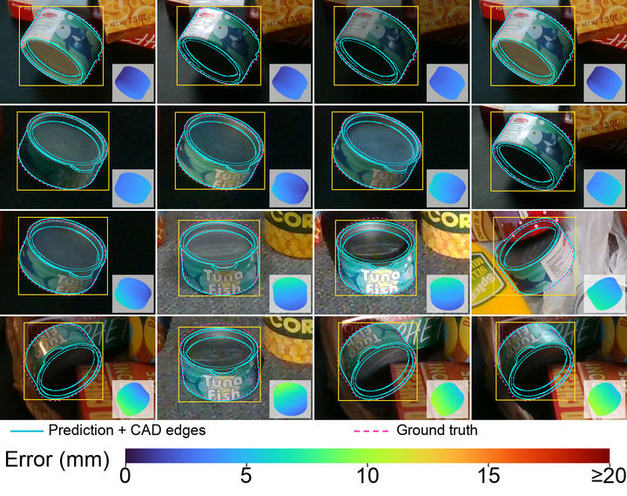}
  \par\vspace{2pt}
  \ifdim\wd\validationtablebox>\linewidth
    \resizebox{\linewidth}{!}{\usebox{\validationtablebox}}
  \else
    \usebox{\validationtablebox}
  \fi
\end{minipage}
\end{lrbox}
\ifdim\dimexpr\ht\validationentrybox+\dp\validationentrybox\relax>0.30\textheight
  \resizebox*{!}{0.30\textheight}{\usebox{\validationentrybox}}
\else
  \usebox{\validationentrybox}
\fi
\end{minipage}
\par
\newpage
\noindent
\begin{minipage}[t][0.30\textheight][t]{0.32\textwidth}
\vspace{0pt}\centering
\begin{lrbox}{\validationentrybox}
\begin{minipage}{\linewidth}
  \vspace{0pt}
  \centering
  \objectlabel{HOPE}{28}{hope}
  \captionsetup{font=scriptsize,skip=2pt,hypcap=false}
  \captionof{table}{Validation AR (\%).}
  \label{tab:app_hope_28}
  \fontsize{6.3}{7.0}\selectfont
  \renewcommand{\arraystretch}{0.95}
  \setlength{\tabcolsep}{1.6pt}
  \begin{lrbox}{\validationtablebox}
  \begin{tabular}{@{}lrrrr@{}}
    \toprule
    Variant & Labels & Single & Jitter & GT-box \\
    \midrule
    Synthetic base & -- & \rankcell{3}{89.00} & \rankcell{3}{89.00} & \rankcell{2}{89.50} \\
    Static pseudo & 0 & -- & -- & -- \\
    Dynamic pseudo & 134 & \rankcell{2}{89.25} & \rankcell{2}{89.25} & \rankcell{4}{88.75} \\
    Static + dynamic & 134 & \rankcell{4}{88.50} & \rankcell{4}{88.50} & \rankcell{3}{89.00} \\
    Box/mask supervision & 381 & \rankcell{5}{87.75} & \rankcell{5}{87.75} & \rankcell{5}{86.75} \\
    Box/mask/keypoint supervision & 381 & \rankcell{1}{90.25} & \rankcell{1}{90.25} & \rankcell{1}{92.00} \\
    \bottomrule
  \end{tabular}
  \end{lrbox}
  \includegraphics[width=\linewidth,keepaspectratio]{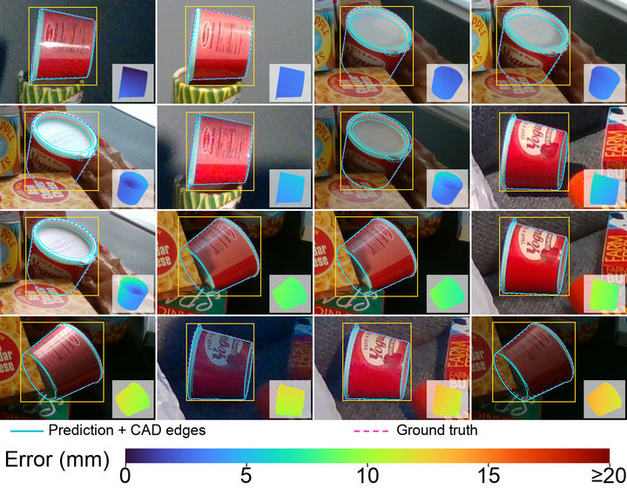}
  \par\vspace{2pt}
  \ifdim\wd\validationtablebox>\linewidth
    \resizebox{\linewidth}{!}{\usebox{\validationtablebox}}
  \else
    \usebox{\validationtablebox}
  \fi
\end{minipage}
\end{lrbox}
\ifdim\dimexpr\ht\validationentrybox+\dp\validationentrybox\relax>0.30\textheight
  \resizebox*{!}{0.30\textheight}{\usebox{\validationentrybox}}
\else
  \usebox{\validationentrybox}
\fi
\end{minipage}
\hfill%
\begin{minipage}[t]{0.32\textwidth}\end{minipage}
\hfill%
\begin{minipage}[t]{0.32\textwidth}\end{minipage}
\clearpage
\noindent
\begin{minipage}[t][0.30\textheight][t]{0.32\textwidth}
\vspace{0pt}\centering
\begin{lrbox}{\validationentrybox}
\begin{minipage}{\linewidth}
  \vspace{0pt}
  \centering
  \objectlabel{HOT3D}{01}{hot3d}
  \captionsetup{font=scriptsize,skip=2pt,hypcap=false}
  \captionof{table}{Validation AR (\%). 250 validation target frames. ConvNeXt-Small.}
  \label{tab:app_hot3d_01}
  \fontsize{6.3}{7.0}\selectfont
  \renewcommand{\arraystretch}{0.95}
  \setlength{\tabcolsep}{1.6pt}
  \begin{lrbox}{\validationtablebox}
  \begin{tabular}{@{}lrrrr@{}}
    \toprule
    Variant & Labels & Single & Jitter & GT-box \\
    \midrule
    Synthetic base & -- & 30.24 & 32.56 & 45.08 \\
    Static pseudo & 321 & \rankcell{4}{42.56} & \rankcell{3}{44.80} & \rankcell{3}{49.76} \\
    Dynamic pseudo & 33 & \rankcell{5}{33.06} & \rankcell{5}{35.44} & \rankcell{5}{47.22} \\
    Static + dynamic & 352 & \rankcell{3}{43.06} & \rankcell{4}{44.68} & \rankcell{4}{49.72} \\
    Box/mask supervision & 3510 & \rankcell{2}{46.54} & \rankcell{2}{48.40} & \rankcell{2}{52.68} \\
    Box/mask/keypoint supervision & 3510 & \rankcell{1}{50.70} & \rankcell{1}{51.44} & \rankcell{1}{58.12} \\
    \bottomrule
  \end{tabular}
  \end{lrbox}
  \includegraphics[width=\linewidth,keepaspectratio]{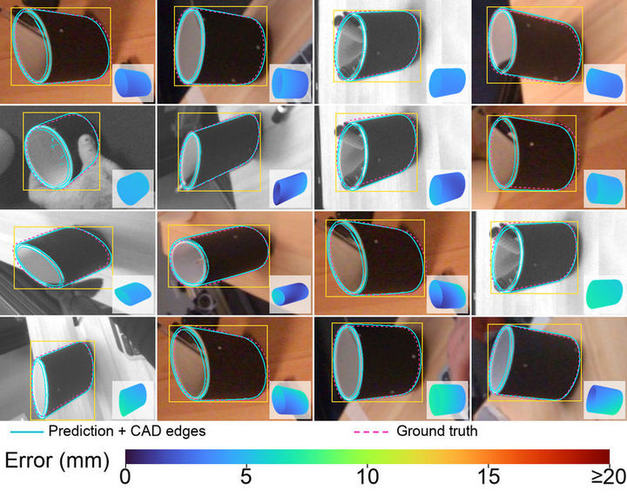}
  \par\vspace{2pt}
  \ifdim\wd\validationtablebox>\linewidth
    \resizebox{\linewidth}{!}{\usebox{\validationtablebox}}
  \else
    \usebox{\validationtablebox}
  \fi
\end{minipage}
\end{lrbox}
\ifdim\dimexpr\ht\validationentrybox+\dp\validationentrybox\relax>0.30\textheight
  \resizebox*{!}{0.30\textheight}{\usebox{\validationentrybox}}
\else
  \usebox{\validationentrybox}
\fi
\end{minipage}
\hfill%
\begin{minipage}[t][0.30\textheight][t]{0.32\textwidth}
\vspace{0pt}\centering
\begin{lrbox}{\validationentrybox}
\begin{minipage}{\linewidth}
  \vspace{0pt}
  \centering
  \objectlabel{HOT3D}{02}{hot3d}
  \captionsetup{font=scriptsize,skip=2pt,hypcap=false}
  \captionof{table}{Validation AR (\%). 265 validation target frames. ConvNeXt-Small.}
  \label{tab:app_hot3d_02}
  \fontsize{6.3}{7.0}\selectfont
  \renewcommand{\arraystretch}{0.95}
  \setlength{\tabcolsep}{1.6pt}
  \begin{lrbox}{\validationtablebox}
  \begin{tabular}{@{}lrrrr@{}}
    \toprule
    Variant & Labels & Single & Jitter & GT-box \\
    \midrule
    Synthetic base & -- & \rankcell{2}{33.57} & \rankcell{3}{35.68} & \rankcell{2}{35.92} \\
    Static pseudo & 0 & -- & -- & -- \\
    Dynamic pseudo & 0 & -- & -- & -- \\
    Static + dynamic & 0 & -- & -- & -- \\
    Box/mask supervision & 3350 & \rankcell{3}{32.43} & \rankcell{2}{36.64} & \rankcell{3}{34.74} \\
    Box/mask/keypoint supervision & 3350 & \rankcell{1}{60.66} & \rankcell{1}{65.13} & \rankcell{1}{62.19} \\
    \bottomrule
  \end{tabular}
  \end{lrbox}
  \includegraphics[width=\linewidth,keepaspectratio]{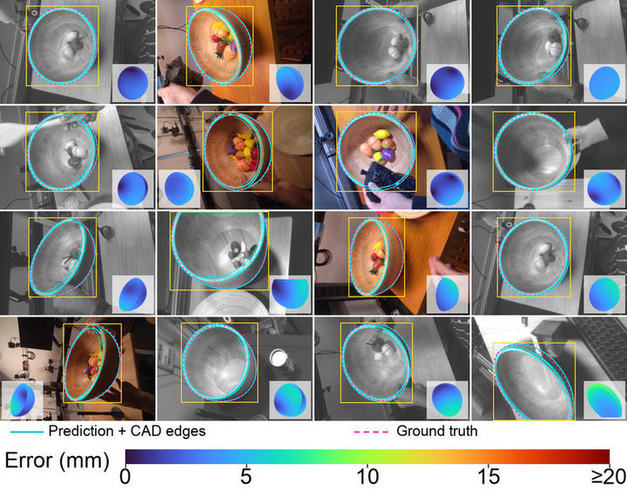}
  \par\vspace{2pt}
  \ifdim\wd\validationtablebox>\linewidth
    \resizebox{\linewidth}{!}{\usebox{\validationtablebox}}
  \else
    \usebox{\validationtablebox}
  \fi
\end{minipage}
\end{lrbox}
\ifdim\dimexpr\ht\validationentrybox+\dp\validationentrybox\relax>0.30\textheight
  \resizebox*{!}{0.30\textheight}{\usebox{\validationentrybox}}
\else
  \usebox{\validationentrybox}
\fi
\end{minipage}
\hfill%
\begin{minipage}[t][0.30\textheight][t]{0.32\textwidth}
\vspace{0pt}\centering
\begin{lrbox}{\validationentrybox}
\begin{minipage}{\linewidth}
  \vspace{0pt}
  \centering
  \objectlabel{HOT3D}{03}{hot3d}
  \captionsetup{font=scriptsize,skip=2pt,hypcap=false}
  \captionof{table}{Validation AR (\%). 295 validation target frames. ConvNeXt-Small.}
  \label{tab:app_hot3d_03}
  \fontsize{6.3}{7.0}\selectfont
  \renewcommand{\arraystretch}{0.95}
  \setlength{\tabcolsep}{1.6pt}
  \begin{lrbox}{\validationtablebox}
  \begin{tabular}{@{}lrrrr@{}}
    \toprule
    Variant & Labels & Single & Jitter & GT-box \\
    \midrule
    Synthetic base & -- & \rankcell{3}{32.92} & \rankcell{3}{38.68} & \rankcell{3}{29.32} \\
    Static pseudo & 0 & -- & -- & -- \\
    Dynamic pseudo & 0 & -- & -- & -- \\
    Static + dynamic & 0 & -- & -- & -- \\
    Box/mask supervision & 3283 & \rankcell{2}{36.00} & \rankcell{2}{40.85} & \rankcell{2}{33.88} \\
    Box/mask/keypoint supervision & 3283 & \rankcell{1}{51.73} & \rankcell{1}{55.10} & \rankcell{1}{51.27} \\
    \bottomrule
  \end{tabular}
  \end{lrbox}
  \includegraphics[width=\linewidth,keepaspectratio]{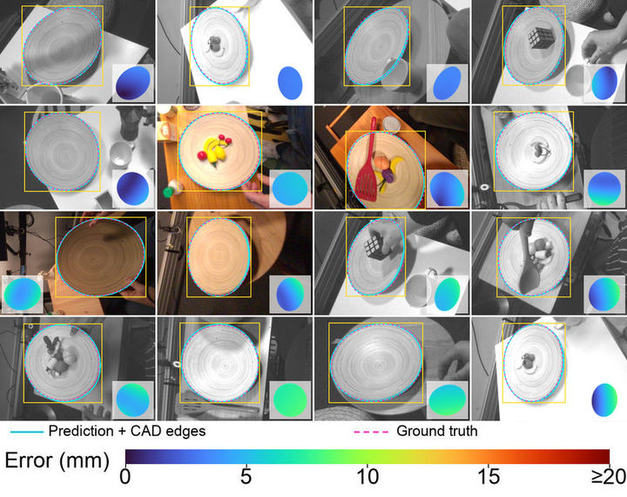}
  \par\vspace{2pt}
  \ifdim\wd\validationtablebox>\linewidth
    \resizebox{\linewidth}{!}{\usebox{\validationtablebox}}
  \else
    \usebox{\validationtablebox}
  \fi
\end{minipage}
\end{lrbox}
\ifdim\dimexpr\ht\validationentrybox+\dp\validationentrybox\relax>0.30\textheight
  \resizebox*{!}{0.30\textheight}{\usebox{\validationentrybox}}
\else
  \usebox{\validationentrybox}
\fi
\end{minipage}
\par\vspace{0.012\textheight}\noindent
\begin{minipage}[t][0.30\textheight][t]{0.32\textwidth}
\vspace{0pt}\centering
\begin{lrbox}{\validationentrybox}
\begin{minipage}{\linewidth}
  \vspace{0pt}
  \centering
  \objectlabel{HOT3D}{04}{hot3d}
  \captionsetup{font=scriptsize,skip=2pt,hypcap=false}
  \captionof{table}{Validation AR (\%). 190 validation target frames. ConvNeXt-Small.}
  \label{tab:app_hot3d_04}
  \fontsize{6.3}{7.0}\selectfont
  \renewcommand{\arraystretch}{0.95}
  \setlength{\tabcolsep}{1.6pt}
  \begin{lrbox}{\validationtablebox}
  \begin{tabular}{@{}lrrrr@{}}
    \toprule
    Variant & Labels & Single & Jitter & GT-box \\
    \midrule
    Synthetic base & -- & 55.00 & 54.66 & 65.26 \\
    Static pseudo & 2379 & \rankcell{3}{60.61} & \rankcell{3}{60.00} & \rankcell{3}{66.61} \\
    Dynamic pseudo & 110 & \rankcell{5}{58.29} & \rankcell{5}{57.18} & \rankcell{5}{66.05} \\
    Static + dynamic & 2483 & \rankcell{2}{60.66} & \rankcell{4}{59.97} & \rankcell{2}{66.79} \\
    Box/mask supervision & 3277 & \rankcell{4}{60.24} & \rankcell{2}{60.61} & \rankcell{4}{66.34} \\
    Box/mask/keypoint supervision & 3277 & \rankcell{1}{63.47} & \rankcell{1}{63.03} & \rankcell{1}{69.37} \\
    \bottomrule
  \end{tabular}
  \end{lrbox}
  \includegraphics[width=\linewidth,keepaspectratio]{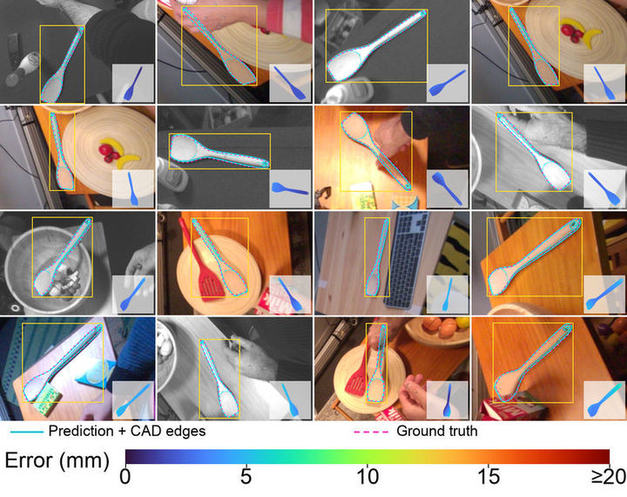}
  \par\vspace{2pt}
  \ifdim\wd\validationtablebox>\linewidth
    \resizebox{\linewidth}{!}{\usebox{\validationtablebox}}
  \else
    \usebox{\validationtablebox}
  \fi
\end{minipage}
\end{lrbox}
\ifdim\dimexpr\ht\validationentrybox+\dp\validationentrybox\relax>0.30\textheight
  \resizebox*{!}{0.30\textheight}{\usebox{\validationentrybox}}
\else
  \usebox{\validationentrybox}
\fi
\end{minipage}
\hfill%
\begin{minipage}[t][0.30\textheight][t]{0.32\textwidth}
\vspace{0pt}\centering
\begin{lrbox}{\validationentrybox}
\begin{minipage}{\linewidth}
  \vspace{0pt}
  \centering
  \objectlabel{HOT3D}{05}{hot3d}
  \captionsetup{font=scriptsize,skip=2pt,hypcap=false}
  \captionof{table}{Validation AR (\%). 179 validation target frames. ConvNeXt-Small.}
  \label{tab:app_hot3d_05}
  \fontsize{6.3}{7.0}\selectfont
  \renewcommand{\arraystretch}{0.95}
  \setlength{\tabcolsep}{1.6pt}
  \begin{lrbox}{\validationtablebox}
  \begin{tabular}{@{}lrrrr@{}}
    \toprule
    Variant & Labels & Single & Jitter & GT-box \\
    \midrule
    Synthetic base & -- & 70.73 & 71.87 & 76.51 \\
    Static pseudo & 1784 & \rankcell{4}{74.41} & \rankcell{4}{75.67} & \rankcell{3}{77.37} \\
    Dynamic pseudo & 285 & \rankcell{5}{70.89} & \rankcell{5}{72.12} & \rankcell{4}{77.12} \\
    Static + dynamic & 2060 & \rankcell{3}{74.75} & \rankcell{2}{76.09} & \rankcell{5}{76.87} \\
    Box/mask supervision & 3237 & \rankcell{2}{75.47} & \rankcell{1}{76.20} & \rankcell{2}{77.82} \\
    Box/mask/keypoint supervision & 3237 & \rankcell{1}{75.87} & \rankcell{3}{76.06} & \rankcell{1}{78.41} \\
    \bottomrule
  \end{tabular}
  \end{lrbox}
  \includegraphics[width=\linewidth,keepaspectratio]{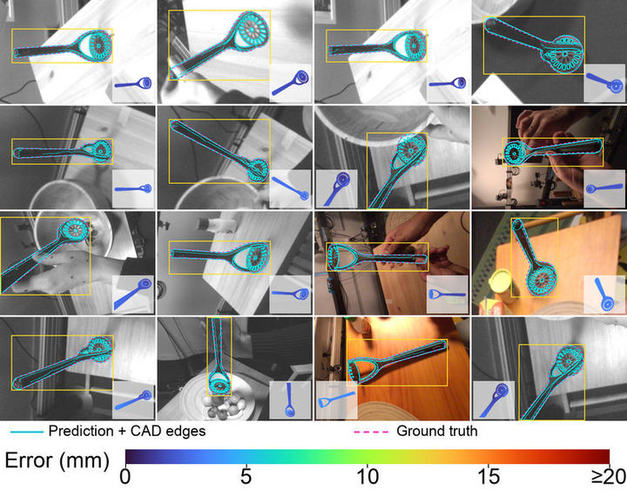}
  \par\vspace{2pt}
  \ifdim\wd\validationtablebox>\linewidth
    \resizebox{\linewidth}{!}{\usebox{\validationtablebox}}
  \else
    \usebox{\validationtablebox}
  \fi
\end{minipage}
\end{lrbox}
\ifdim\dimexpr\ht\validationentrybox+\dp\validationentrybox\relax>0.30\textheight
  \resizebox*{!}{0.30\textheight}{\usebox{\validationentrybox}}
\else
  \usebox{\validationentrybox}
\fi
\end{minipage}
\hfill%
\begin{minipage}[t][0.30\textheight][t]{0.32\textwidth}
\vspace{0pt}\centering
\begin{lrbox}{\validationentrybox}
\begin{minipage}{\linewidth}
  \vspace{0pt}
  \centering
  \objectlabel{HOT3D}{06}{hot3d}
  \captionsetup{font=scriptsize,skip=2pt,hypcap=false}
  \captionof{table}{Validation AR (\%). 162 validation target frames. ConvNeXt-Small.}
  \label{tab:app_hot3d_06}
  \fontsize{6.3}{7.0}\selectfont
  \renewcommand{\arraystretch}{0.95}
  \setlength{\tabcolsep}{1.6pt}
  \begin{lrbox}{\validationtablebox}
  \begin{tabular}{@{}lrrrr@{}}
    \toprule
    Variant & Labels & Single & Jitter & GT-box \\
    \midrule
    Synthetic base & -- & 63.67 & 64.75 & 70.22 \\
    Static pseudo & 2413 & \rankcell{4}{67.78} & \rankcell{3}{68.89} & \rankcell{4}{71.85} \\
    Dynamic pseudo & 962 & \rankcell{5}{66.08} & \rankcell{5}{66.94} & \rankcell{2}{72.16} \\
    Static + dynamic & 3369 & \rankcell{2}{68.12} & \rankcell{4}{68.80} & \rankcell{3}{72.04} \\
    Box/mask supervision & 3328 & \rankcell{3}{67.90} & \rankcell{2}{69.10} & \rankcell{5}{71.20} \\
    Box/mask/keypoint supervision & 3328 & \rankcell{1}{68.64} & \rankcell{1}{69.14} & \rankcell{1}{72.84} \\
    \bottomrule
  \end{tabular}
  \end{lrbox}
  \includegraphics[width=\linewidth,keepaspectratio]{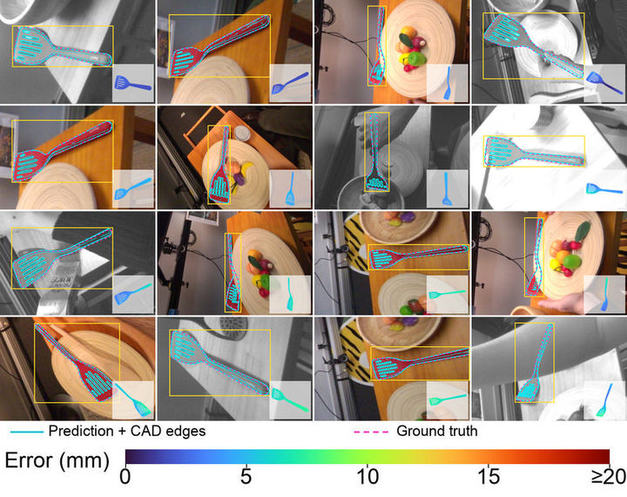}
  \par\vspace{2pt}
  \ifdim\wd\validationtablebox>\linewidth
    \resizebox{\linewidth}{!}{\usebox{\validationtablebox}}
  \else
    \usebox{\validationtablebox}
  \fi
\end{minipage}
\end{lrbox}
\ifdim\dimexpr\ht\validationentrybox+\dp\validationentrybox\relax>0.30\textheight
  \resizebox*{!}{0.30\textheight}{\usebox{\validationentrybox}}
\else
  \usebox{\validationentrybox}
\fi
\end{minipage}
\par\vspace{0.012\textheight}\noindent
\begin{minipage}[t][0.30\textheight][t]{0.32\textwidth}
\vspace{0pt}\centering
\begin{lrbox}{\validationentrybox}
\begin{minipage}{\linewidth}
  \vspace{0pt}
  \centering
  \objectlabel{HOT3D}{07}{hot3d}
  \captionsetup{font=scriptsize,skip=2pt,hypcap=false}
  \captionof{table}{Validation AR (\%). 307 validation target frames. ConvNeXt-Small.}
  \label{tab:app_hot3d_07}
  \fontsize{6.3}{7.0}\selectfont
  \renewcommand{\arraystretch}{0.95}
  \setlength{\tabcolsep}{1.6pt}
  \begin{lrbox}{\validationtablebox}
  \begin{tabular}{@{}lrrrr@{}}
    \toprule
    Variant & Labels & Single & Jitter & GT-box \\
    \midrule
    Synthetic base & -- & 60.41 & 60.90 & \rankcell{5}{61.25} \\
    Static pseudo & 1236 & \rankcell{3}{62.25} & \rankcell{3}{62.49} & \rankcell{3}{62.25} \\
    Dynamic pseudo & 70 & \rankcell{5}{61.34} & \rankcell{5}{61.95} & 61.12 \\
    Static + dynamic & 1294 & \rankcell{2}{62.26} & \rankcell{2}{62.52} & \rankcell{4}{62.23} \\
    Box/mask supervision & 3485 & \rankcell{4}{61.99} & \rankcell{4}{62.33} & \rankcell{2}{63.21} \\
    Box/mask/keypoint supervision & 3485 & \rankcell{1}{64.40} & \rankcell{1}{64.80} & \rankcell{1}{67.85} \\
    \bottomrule
  \end{tabular}
  \end{lrbox}
  \includegraphics[width=\linewidth,keepaspectratio]{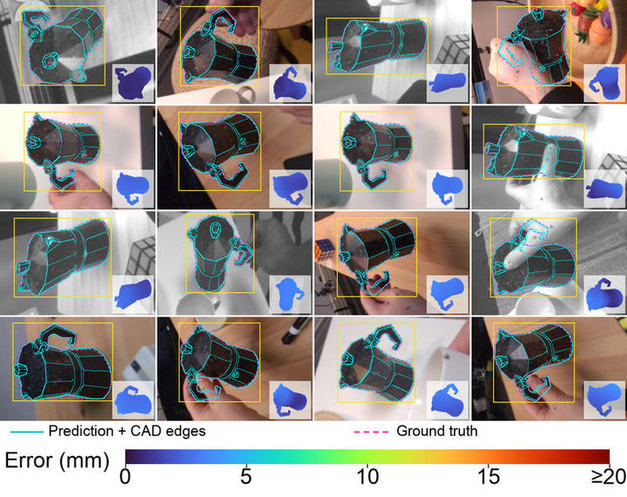}
  \par\vspace{2pt}
  \ifdim\wd\validationtablebox>\linewidth
    \resizebox{\linewidth}{!}{\usebox{\validationtablebox}}
  \else
    \usebox{\validationtablebox}
  \fi
\end{minipage}
\end{lrbox}
\ifdim\dimexpr\ht\validationentrybox+\dp\validationentrybox\relax>0.30\textheight
  \resizebox*{!}{0.30\textheight}{\usebox{\validationentrybox}}
\else
  \usebox{\validationentrybox}
\fi
\end{minipage}
\hfill%
\begin{minipage}[t][0.30\textheight][t]{0.32\textwidth}
\vspace{0pt}\centering
\begin{lrbox}{\validationentrybox}
\begin{minipage}{\linewidth}
  \vspace{0pt}
  \centering
  \objectlabel{HOT3D}{08}{hot3d}
  \captionsetup{font=scriptsize,skip=2pt,hypcap=false}
  \captionof{table}{Validation AR (\%). 263 validation target frames. ConvNeXt-Small.}
  \label{tab:app_hot3d_08}
  \fontsize{6.3}{7.0}\selectfont
  \renewcommand{\arraystretch}{0.95}
  \setlength{\tabcolsep}{1.6pt}
  \begin{lrbox}{\validationtablebox}
  \begin{tabular}{@{}lrrrr@{}}
    \toprule
    Variant & Labels & Single & Jitter & GT-box \\
    \midrule
    Synthetic base & -- & 65.99 & 66.81 & \rankcell{5}{64.81} \\
    Static pseudo & 1958 & \rankcell{2}{66.88} & \rankcell{2}{67.49} & \rankcell{4}{64.85} \\
    Dynamic pseudo & 362 & \rankcell{3}{66.52} & \rankcell{3}{67.19} & \rankcell{2}{65.44} \\
    Static + dynamic & 2309 & \rankcell{4}{66.31} & \rankcell{4}{67.17} & \rankcell{3}{65.30} \\
    Box/mask supervision & 3718 & \rankcell{5}{66.08} & \rankcell{5}{67.05} & 64.77 \\
    Box/mask/keypoint supervision & 3718 & \rankcell{1}{69.98} & \rankcell{1}{71.73} & \rankcell{1}{68.54} \\
    \bottomrule
  \end{tabular}
  \end{lrbox}
  \includegraphics[width=\linewidth,keepaspectratio]{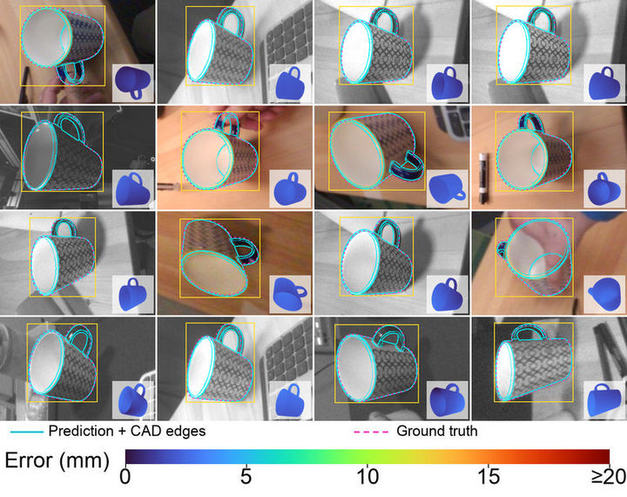}
  \par\vspace{2pt}
  \ifdim\wd\validationtablebox>\linewidth
    \resizebox{\linewidth}{!}{\usebox{\validationtablebox}}
  \else
    \usebox{\validationtablebox}
  \fi
\end{minipage}
\end{lrbox}
\ifdim\dimexpr\ht\validationentrybox+\dp\validationentrybox\relax>0.30\textheight
  \resizebox*{!}{0.30\textheight}{\usebox{\validationentrybox}}
\else
  \usebox{\validationentrybox}
\fi
\end{minipage}
\hfill%
\begin{minipage}[t][0.30\textheight][t]{0.32\textwidth}
\vspace{0pt}\centering
\begin{lrbox}{\validationentrybox}
\begin{minipage}{\linewidth}
  \vspace{0pt}
  \centering
  \objectlabel{HOT3D}{09}{hot3d}
  \captionsetup{font=scriptsize,skip=2pt,hypcap=false}
  \captionof{table}{Validation AR (\%). 307 validation target frames. ConvNeXt-Small.}
  \label{tab:app_hot3d_09}
  \fontsize{6.3}{7.0}\selectfont
  \renewcommand{\arraystretch}{0.95}
  \setlength{\tabcolsep}{1.6pt}
  \begin{lrbox}{\validationtablebox}
  \begin{tabular}{@{}lrrrr@{}}
    \toprule
    Variant & Labels & Single & Jitter & GT-box \\
    \midrule
    Synthetic base & -- & 62.23 & 63.91 & \rankcell{4}{64.32} \\
    Static pseudo & 1568 & \rankcell{1}{64.63} & \rankcell{2}{65.70} & \rankcell{3}{64.71} \\
    Dynamic pseudo & 433 & \rankcell{5}{62.33} & \rankcell{5}{64.25} & \rankcell{2}{65.34} \\
    Static + dynamic & 1985 & \rankcell{2}{64.38} & \rankcell{3}{65.37} & \rankcell{5}{64.28} \\
    Box/mask supervision & 3521 & \rankcell{4}{63.66} & \rankcell{4}{64.87} & \rankcell{3}{64.71} \\
    Box/mask/keypoint supervision & 3521 & \rankcell{3}{64.17} & \rankcell{1}{66.22} & \rankcell{1}{67.33} \\
    \bottomrule
  \end{tabular}
  \end{lrbox}
  \includegraphics[width=\linewidth,keepaspectratio]{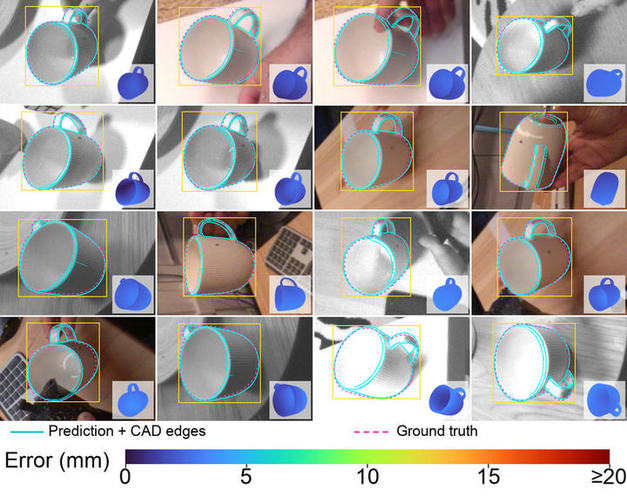}
  \par\vspace{2pt}
  \ifdim\wd\validationtablebox>\linewidth
    \resizebox{\linewidth}{!}{\usebox{\validationtablebox}}
  \else
    \usebox{\validationtablebox}
  \fi
\end{minipage}
\end{lrbox}
\ifdim\dimexpr\ht\validationentrybox+\dp\validationentrybox\relax>0.30\textheight
  \resizebox*{!}{0.30\textheight}{\usebox{\validationentrybox}}
\else
  \usebox{\validationentrybox}
\fi
\end{minipage}
\par
\newpage
\noindent
\begin{minipage}[t][0.30\textheight][t]{0.32\textwidth}
\vspace{0pt}\centering
\begin{lrbox}{\validationentrybox}
\begin{minipage}{\linewidth}
  \vspace{0pt}
  \centering
  \objectlabel{HOT3D}{10}{hot3d}
  \captionsetup{font=scriptsize,skip=2pt,hypcap=false}
  \captionof{table}{Validation AR (\%). 163 validation target frames. ConvNeXt-Small.}
  \label{tab:app_hot3d_10}
  \fontsize{6.3}{7.0}\selectfont
  \renewcommand{\arraystretch}{0.95}
  \setlength{\tabcolsep}{1.6pt}
  \begin{lrbox}{\validationtablebox}
  \begin{tabular}{@{}lrrrr@{}}
    \toprule
    Variant & Labels & Single & Jitter & GT-box \\
    \midrule
    Synthetic base & -- & \rankcell{5}{30.83} & \rankcell{5}{31.35} & 42.36 \\
    Static pseudo & 2589 & \rankcell{1}{38.53} & \rankcell{1}{39.17} & \rankcell{3}{44.29} \\
    Dynamic pseudo & 1271 & 26.07 & 26.72 & \rankcell{2}{44.63} \\
    Static + dynamic & 3848 & \rankcell{4}{35.61} & \rankcell{4}{36.07} & \rankcell{4}{43.90} \\
    Box/mask supervision & 3759 & \rankcell{3}{37.55} & \rankcell{3}{38.62} & \rankcell{5}{43.80} \\
    Box/mask/keypoint supervision & 3759 & \rankcell{2}{37.73} & \rankcell{2}{38.96} & \rankcell{1}{44.75} \\
    \bottomrule
  \end{tabular}
  \end{lrbox}
  \includegraphics[width=\linewidth,keepaspectratio]{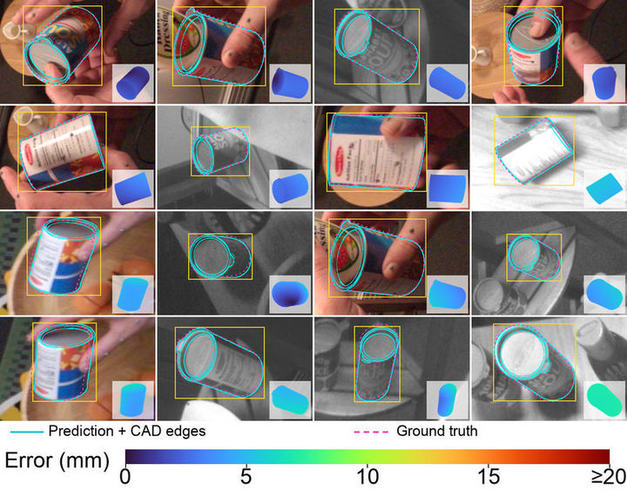}
  \par\vspace{2pt}
  \ifdim\wd\validationtablebox>\linewidth
    \resizebox{\linewidth}{!}{\usebox{\validationtablebox}}
  \else
    \usebox{\validationtablebox}
  \fi
\end{minipage}
\end{lrbox}
\ifdim\dimexpr\ht\validationentrybox+\dp\validationentrybox\relax>0.30\textheight
  \resizebox*{!}{0.30\textheight}{\usebox{\validationentrybox}}
\else
  \usebox{\validationentrybox}
\fi
\end{minipage}
\hfill%
\begin{minipage}[t][0.30\textheight][t]{0.32\textwidth}
\vspace{0pt}\centering
\begin{lrbox}{\validationentrybox}
\begin{minipage}{\linewidth}
  \vspace{0pt}
  \centering
  \objectlabel{HOT3D}{11}{hot3d}
  \captionsetup{font=scriptsize,skip=2pt,hypcap=false}
  \captionof{table}{Validation AR (\%). 173 validation target frames. ConvNeXt-Small.}
  \label{tab:app_hot3d_11}
  \fontsize{6.3}{7.0}\selectfont
  \renewcommand{\arraystretch}{0.95}
  \setlength{\tabcolsep}{1.6pt}
  \begin{lrbox}{\validationtablebox}
  \begin{tabular}{@{}lrrrr@{}}
    \toprule
    Variant & Labels & Single & Jitter & GT-box \\
    \midrule
    Synthetic base & -- & 56.85 & 57.77 & \rankcell{4}{61.82} \\
    Static pseudo & 3081 & \rankcell{3}{61.13} & \rankcell{3}{62.11} & \rankcell{3}{62.02} \\
    Dynamic pseudo & 1043 & \rankcell{5}{59.68} & \rankcell{5}{61.42} & \rankcell{2}{62.89} \\
    Static + dynamic & 4103 & \rankcell{1}{61.91} & \rankcell{1}{62.69} & \rankcell{2}{62.89} \\
    Box/mask supervision & 3444 & \rankcell{4}{60.98} & \rankcell{4}{61.94} & \rankcell{2}{62.89} \\
    Box/mask/keypoint supervision & 3444 & \rankcell{2}{61.45} & \rankcell{2}{62.49} & \rankcell{1}{64.54} \\
    \bottomrule
  \end{tabular}
  \end{lrbox}
  \includegraphics[width=\linewidth,keepaspectratio]{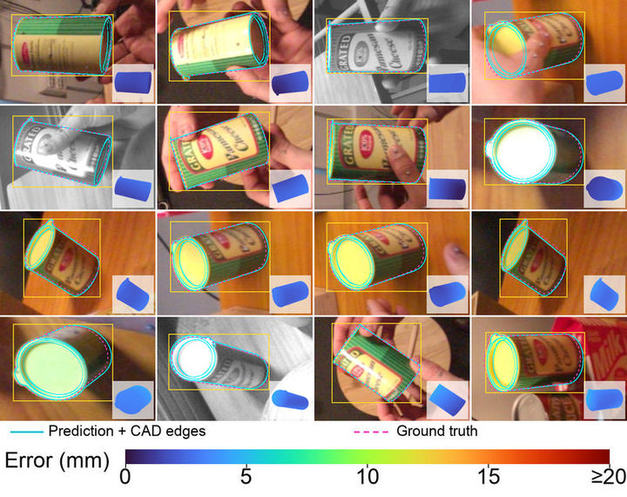}
  \par\vspace{2pt}
  \ifdim\wd\validationtablebox>\linewidth
    \resizebox{\linewidth}{!}{\usebox{\validationtablebox}}
  \else
    \usebox{\validationtablebox}
  \fi
\end{minipage}
\end{lrbox}
\ifdim\dimexpr\ht\validationentrybox+\dp\validationentrybox\relax>0.30\textheight
  \resizebox*{!}{0.30\textheight}{\usebox{\validationentrybox}}
\else
  \usebox{\validationentrybox}
\fi
\end{minipage}
\hfill%
\begin{minipage}[t][0.30\textheight][t]{0.32\textwidth}
\vspace{0pt}\centering
\begin{lrbox}{\validationentrybox}
\begin{minipage}{\linewidth}
  \vspace{0pt}
  \centering
  \objectlabel{HOT3D}{12}{hot3d}
  \captionsetup{font=scriptsize,skip=2pt,hypcap=false}
  \captionof{table}{Validation AR (\%). 169 validation target frames. ConvNeXt-Small.}
  \label{tab:app_hot3d_12}
  \fontsize{6.3}{7.0}\selectfont
  \renewcommand{\arraystretch}{0.95}
  \setlength{\tabcolsep}{1.6pt}
  \begin{lrbox}{\validationtablebox}
  \begin{tabular}{@{}lrrrr@{}}
    \toprule
    Variant & Labels & Single & Jitter & GT-box \\
    \midrule
    Synthetic base & -- & 50.47 & \rankcell{5}{51.36} & \rankcell{5}{60.77} \\
    Static pseudo & 2975 & \rankcell{3}{53.20} & \rankcell{3}{53.82} & \rankcell{2}{62.31} \\
    Dynamic pseudo & 917 & \rankcell{5}{50.74} & 51.07 & \rankcell{4}{60.95} \\
    Static + dynamic & 3872 & \rankcell{4}{52.60} & \rankcell{4}{53.17} & \rankcell{3}{61.80} \\
    Box/mask supervision & 3240 & \rankcell{2}{53.61} & \rankcell{2}{54.35} & \rankcell{3}{61.80} \\
    Box/mask/keypoint supervision & 3240 & \rankcell{1}{54.94} & \rankcell{1}{54.91} & \rankcell{1}{64.62} \\
    \bottomrule
  \end{tabular}
  \end{lrbox}
  \includegraphics[width=\linewidth,keepaspectratio]{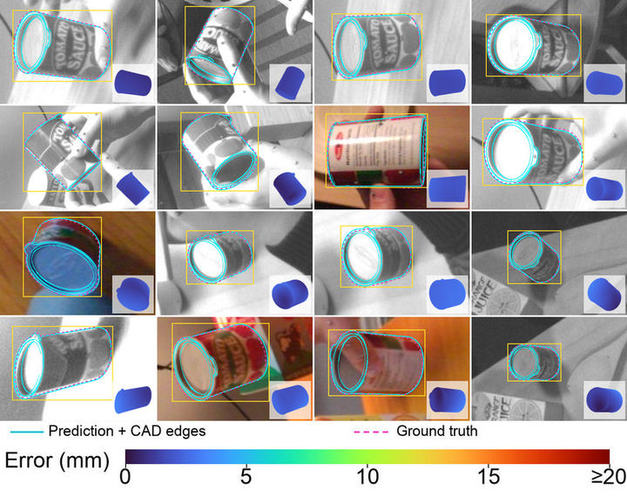}
  \par\vspace{2pt}
  \ifdim\wd\validationtablebox>\linewidth
    \resizebox{\linewidth}{!}{\usebox{\validationtablebox}}
  \else
    \usebox{\validationtablebox}
  \fi
\end{minipage}
\end{lrbox}
\ifdim\dimexpr\ht\validationentrybox+\dp\validationentrybox\relax>0.30\textheight
  \resizebox*{!}{0.30\textheight}{\usebox{\validationentrybox}}
\else
  \usebox{\validationentrybox}
\fi
\end{minipage}
\par\vspace{0.012\textheight}\noindent
\begin{minipage}[t][0.30\textheight][t]{0.32\textwidth}
\vspace{0pt}\centering
\begin{lrbox}{\validationentrybox}
\begin{minipage}{\linewidth}
  \vspace{0pt}
  \centering
  \objectlabel{HOT3D}{13}{hot3d}
  \captionsetup{font=scriptsize,skip=2pt,hypcap=false}
  \captionof{table}{Validation AR (\%). 116 validation target frames. ConvNeXt-Small.}
  \label{tab:app_hot3d_13}
  \fontsize{6.3}{7.0}\selectfont
  \renewcommand{\arraystretch}{0.95}
  \setlength{\tabcolsep}{1.6pt}
  \begin{lrbox}{\validationtablebox}
  \begin{tabular}{@{}lrrrr@{}}
    \toprule
    Variant & Labels & Single & Jitter & GT-box \\
    \midrule
    Synthetic base & -- & 53.97 & \rankcell{5}{56.90} & 59.96 \\
    Static pseudo & 2194 & \rankcell{3}{59.01} & \rankcell{1}{63.66} & \rankcell{5}{61.03} \\
    Dynamic pseudo & 510 & \rankcell{5}{55.47} & \rankcell{4}{57.59} & \rankcell{1}{62.41} \\
    Static + dynamic & 2692 & \rankcell{2}{59.53} & \rankcell{2}{63.41} & \rankcell{4}{61.68} \\
    Box/mask supervision & 3367 & \rankcell{4}{58.66} & \rankcell{2}{63.41} & \rankcell{2}{62.37} \\
    Box/mask/keypoint supervision & 3367 & \rankcell{1}{61.08} & \rankcell{3}{63.36} & \rankcell{3}{61.85} \\
    \bottomrule
  \end{tabular}
  \end{lrbox}
  \includegraphics[width=\linewidth,keepaspectratio]{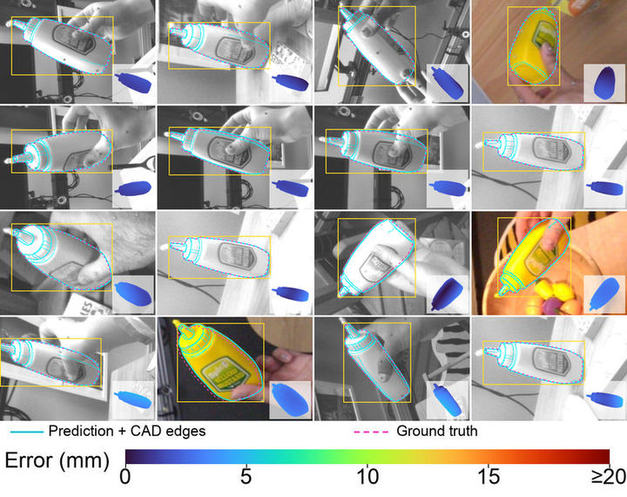}
  \par\vspace{2pt}
  \ifdim\wd\validationtablebox>\linewidth
    \resizebox{\linewidth}{!}{\usebox{\validationtablebox}}
  \else
    \usebox{\validationtablebox}
  \fi
\end{minipage}
\end{lrbox}
\ifdim\dimexpr\ht\validationentrybox+\dp\validationentrybox\relax>0.30\textheight
  \resizebox*{!}{0.30\textheight}{\usebox{\validationentrybox}}
\else
  \usebox{\validationentrybox}
\fi
\end{minipage}
\hfill%
\begin{minipage}[t][0.30\textheight][t]{0.32\textwidth}
\vspace{0pt}\centering
\begin{lrbox}{\validationentrybox}
\begin{minipage}{\linewidth}
  \vspace{0pt}
  \centering
  \objectlabel{HOT3D}{14}{hot3d}
  \captionsetup{font=scriptsize,skip=2pt,hypcap=false}
  \captionof{table}{Validation AR (\%). 108 validation target frames. ConvNeXt-Small.}
  \label{tab:app_hot3d_14}
  \fontsize{6.3}{7.0}\selectfont
  \renewcommand{\arraystretch}{0.95}
  \setlength{\tabcolsep}{1.6pt}
  \begin{lrbox}{\validationtablebox}
  \begin{tabular}{@{}lrrrr@{}}
    \toprule
    Variant & Labels & Single & Jitter & GT-box \\
    \midrule
    Synthetic base & -- & 45.14 & 47.69 & 49.72 \\
    Static pseudo & 1872 & \rankcell{4}{46.67} & \rankcell{5}{48.33} & \rankcell{3}{53.47} \\
    Dynamic pseudo & 793 & \rankcell{2}{47.31} & \rankcell{2}{51.16} & \rankcell{5}{52.92} \\
    Static + dynamic & 2658 & \rankcell{5}{46.06} & \rankcell{4}{48.66} & \rankcell{4}{53.01} \\
    Box/mask supervision & 3278 & \rankcell{3}{47.13} & \rankcell{3}{50.05} & \rankcell{2}{55.37} \\
    Box/mask/keypoint supervision & 3278 & \rankcell{1}{50.69} & \rankcell{1}{51.90} & \rankcell{1}{56.39} \\
    \bottomrule
  \end{tabular}
  \end{lrbox}
  \includegraphics[width=\linewidth,keepaspectratio]{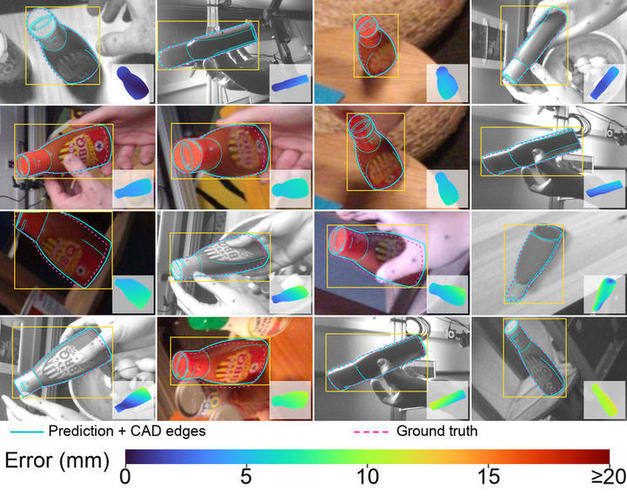}
  \par\vspace{2pt}
  \ifdim\wd\validationtablebox>\linewidth
    \resizebox{\linewidth}{!}{\usebox{\validationtablebox}}
  \else
    \usebox{\validationtablebox}
  \fi
\end{minipage}
\end{lrbox}
\ifdim\dimexpr\ht\validationentrybox+\dp\validationentrybox\relax>0.30\textheight
  \resizebox*{!}{0.30\textheight}{\usebox{\validationentrybox}}
\else
  \usebox{\validationentrybox}
\fi
\end{minipage}
\hfill%
\begin{minipage}[t][0.30\textheight][t]{0.32\textwidth}
\vspace{0pt}\centering
\begin{lrbox}{\validationentrybox}
\begin{minipage}{\linewidth}
  \vspace{0pt}
  \centering
  \objectlabel{HOT3D}{15}{hot3d}
  \captionsetup{font=scriptsize,skip=2pt,hypcap=false}
  \captionof{table}{Validation AR (\%). 176 validation target frames. ConvNeXt-Small.}
  \label{tab:app_hot3d_15}
  \fontsize{6.3}{7.0}\selectfont
  \renewcommand{\arraystretch}{0.95}
  \setlength{\tabcolsep}{1.6pt}
  \begin{lrbox}{\validationtablebox}
  \begin{tabular}{@{}lrrrr@{}}
    \toprule
    Variant & Labels & Single & Jitter & GT-box \\
    \midrule
    Synthetic base & -- & 51.70 & 52.02 & 52.87 \\
    Static pseudo & 2104 & \rankcell{5}{53.04} & \rankcell{4}{54.06} & \rankcell{4}{54.01} \\
    Dynamic pseudo & 802 & \rankcell{4}{53.24} & \rankcell{5}{53.75} & \rankcell{5}{53.61} \\
    Static + dynamic & 2899 & \rankcell{3}{53.66} & \rankcell{3}{54.43} & \rankcell{3}{54.15} \\
    Box/mask supervision & 3335 & \rankcell{2}{54.66} & \rankcell{2}{55.74} & \rankcell{1}{55.17} \\
    Box/mask/keypoint supervision & 3335 & \rankcell{1}{55.20} & \rankcell{1}{56.16} & \rankcell{2}{54.86} \\
    \bottomrule
  \end{tabular}
  \end{lrbox}
  \includegraphics[width=\linewidth,keepaspectratio]{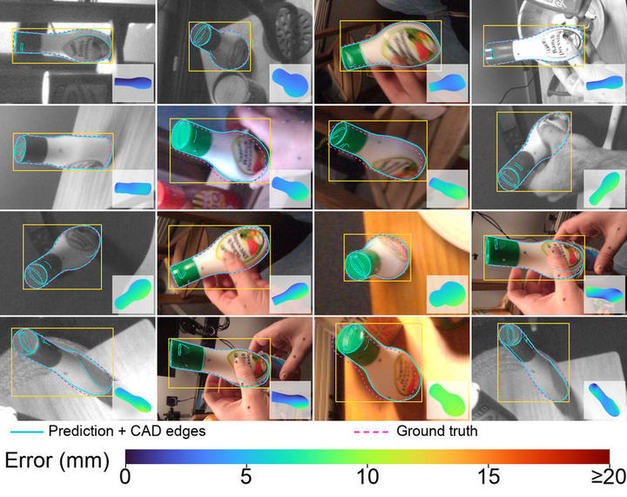}
  \par\vspace{2pt}
  \ifdim\wd\validationtablebox>\linewidth
    \resizebox{\linewidth}{!}{\usebox{\validationtablebox}}
  \else
    \usebox{\validationtablebox}
  \fi
\end{minipage}
\end{lrbox}
\ifdim\dimexpr\ht\validationentrybox+\dp\validationentrybox\relax>0.30\textheight
  \resizebox*{!}{0.30\textheight}{\usebox{\validationentrybox}}
\else
  \usebox{\validationentrybox}
\fi
\end{minipage}
\par\vspace{0.012\textheight}\noindent
\begin{minipage}[t][0.30\textheight][t]{0.32\textwidth}
\vspace{0pt}\centering
\begin{lrbox}{\validationentrybox}
\begin{minipage}{\linewidth}
  \vspace{0pt}
  \centering
  \objectlabel{HOT3D}{16}{hot3d}
  \captionsetup{font=scriptsize,skip=2pt,hypcap=false}
  \captionof{table}{Validation AR (\%). 277 validation target frames. ConvNeXt-Small.}
  \label{tab:app_hot3d_16}
  \fontsize{6.3}{7.0}\selectfont
  \renewcommand{\arraystretch}{0.95}
  \setlength{\tabcolsep}{1.6pt}
  \begin{lrbox}{\validationtablebox}
  \begin{tabular}{@{}lrrrr@{}}
    \toprule
    Variant & Labels & Single & Jitter & GT-box \\
    \midrule
    Synthetic base & -- & 60.61 & 62.20 & 70.43 \\
    Static pseudo & 2179 & \rankcell{3}{65.09} & \rankcell{3}{67.13} & \rankcell{4}{72.55} \\
    Dynamic pseudo & 107 & \rankcell{5}{63.19} & \rankcell{5}{64.75} & \rankcell{5}{71.57} \\
    Static + dynamic & 2280 & \rankcell{4}{65.04} & \rankcell{4}{67.09} & \rankcell{3}{72.62} \\
    Box/mask supervision & 3330 & \rankcell{2}{67.64} & \rankcell{2}{69.19} & \rankcell{2}{73.81} \\
    Box/mask/keypoint supervision & 3330 & \rankcell{1}{70.09} & \rankcell{1}{70.47} & \rankcell{1}{78.48} \\
    \bottomrule
  \end{tabular}
  \end{lrbox}
  \includegraphics[width=\linewidth,keepaspectratio]{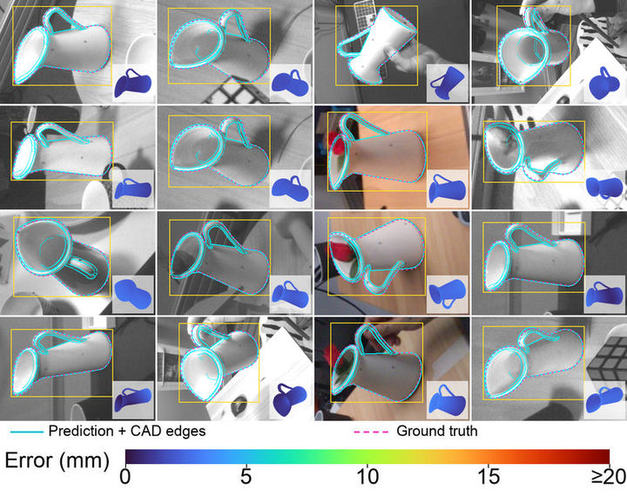}
  \par\vspace{2pt}
  \ifdim\wd\validationtablebox>\linewidth
    \resizebox{\linewidth}{!}{\usebox{\validationtablebox}}
  \else
    \usebox{\validationtablebox}
  \fi
\end{minipage}
\end{lrbox}
\ifdim\dimexpr\ht\validationentrybox+\dp\validationentrybox\relax>0.30\textheight
  \resizebox*{!}{0.30\textheight}{\usebox{\validationentrybox}}
\else
  \usebox{\validationentrybox}
\fi
\end{minipage}
\hfill%
\begin{minipage}[t][0.30\textheight][t]{0.32\textwidth}
\vspace{0pt}\centering
\begin{lrbox}{\validationentrybox}
\begin{minipage}{\linewidth}
  \vspace{0pt}
  \centering
  \objectlabel{HOT3D}{17}{hot3d}
  \captionsetup{font=scriptsize,skip=2pt,hypcap=false}
  \captionof{table}{Validation AR (\%). 179 validation target frames. ConvNeXt-Small.}
  \label{tab:app_hot3d_17}
  \fontsize{6.3}{7.0}\selectfont
  \renewcommand{\arraystretch}{0.95}
  \setlength{\tabcolsep}{1.6pt}
  \begin{lrbox}{\validationtablebox}
  \begin{tabular}{@{}lrrrr@{}}
    \toprule
    Variant & Labels & Single & Jitter & GT-box \\
    \midrule
    Synthetic base & -- & \rankcell{5}{70.42} & 70.87 & 70.06 \\
    Static pseudo & 1470 & 70.39 & \rankcell{5}{71.90} & \rankcell{5}{71.03} \\
    Dynamic pseudo & 318 & \rankcell{4}{71.12} & \rankcell{3}{72.60} & \rankcell{4}{71.40} \\
    Static + dynamic & 1777 & \rankcell{3}{71.54} & \rankcell{4}{72.12} & \rankcell{3}{71.96} \\
    Box/mask supervision & 3359 & \rankcell{2}{74.27} & \rankcell{2}{73.94} & \rankcell{2}{72.01} \\
    Box/mask/keypoint supervision & 3359 & \rankcell{1}{74.61} & \rankcell{1}{75.39} & \rankcell{1}{75.70} \\
    \bottomrule
  \end{tabular}
  \end{lrbox}
  \includegraphics[width=\linewidth,keepaspectratio]{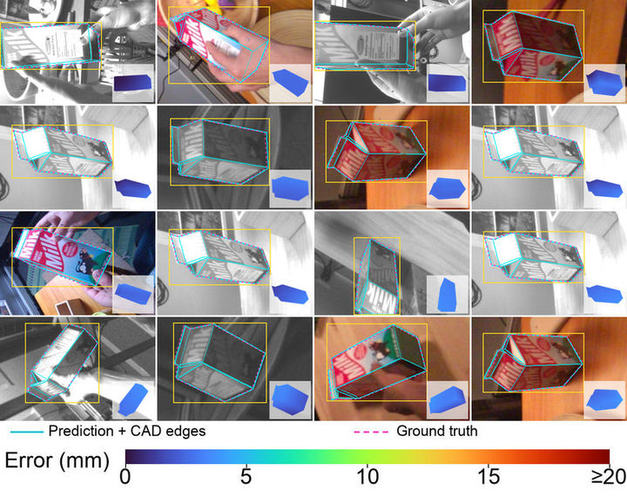}
  \par\vspace{2pt}
  \ifdim\wd\validationtablebox>\linewidth
    \resizebox{\linewidth}{!}{\usebox{\validationtablebox}}
  \else
    \usebox{\validationtablebox}
  \fi
\end{minipage}
\end{lrbox}
\ifdim\dimexpr\ht\validationentrybox+\dp\validationentrybox\relax>0.30\textheight
  \resizebox*{!}{0.30\textheight}{\usebox{\validationentrybox}}
\else
  \usebox{\validationentrybox}
\fi
\end{minipage}
\hfill%
\begin{minipage}[t][0.30\textheight][t]{0.32\textwidth}
\vspace{0pt}\centering
\begin{lrbox}{\validationentrybox}
\begin{minipage}{\linewidth}
  \vspace{0pt}
  \centering
  \objectlabel{HOT3D}{18}{hot3d}
  \captionsetup{font=scriptsize,skip=2pt,hypcap=false}
  \captionof{table}{Validation AR (\%). 184 validation target frames. ConvNeXt-Small.}
  \label{tab:app_hot3d_18}
  \fontsize{6.3}{7.0}\selectfont
  \renewcommand{\arraystretch}{0.95}
  \setlength{\tabcolsep}{1.6pt}
  \begin{lrbox}{\validationtablebox}
  \begin{tabular}{@{}lrrrr@{}}
    \toprule
    Variant & Labels & Single & Jitter & GT-box \\
    \midrule
    Synthetic base & -- & \rankcell{3}{69.51} & \rankcell{2}{70.71} & 68.83 \\
    Static pseudo & 1579 & 69.10 & \rankcell{4}{69.89} & \rankcell{5}{68.99} \\
    Dynamic pseudo & 420 & \rankcell{2}{70.54} & \rankcell{3}{70.16} & \rankcell{4}{69.02} \\
    Static + dynamic & 1993 & \rankcell{5}{69.29} & \rankcell{5}{69.46} & \rankcell{3}{69.38} \\
    Box/mask supervision & 3184 & \rankcell{4}{69.43} & \rankcell{4}{69.89} & \rankcell{2}{69.70} \\
    Box/mask/keypoint supervision & 3184 & \rankcell{1}{70.73} & \rankcell{1}{71.55} & \rankcell{1}{71.85} \\
    \bottomrule
  \end{tabular}
  \end{lrbox}
  \includegraphics[width=\linewidth,keepaspectratio]{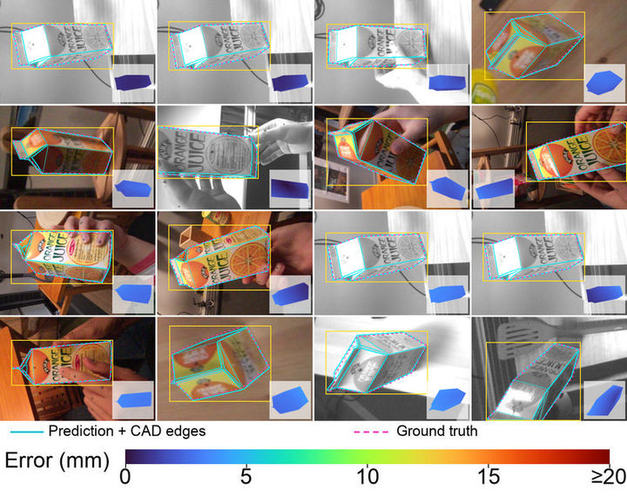}
  \par\vspace{2pt}
  \ifdim\wd\validationtablebox>\linewidth
    \resizebox{\linewidth}{!}{\usebox{\validationtablebox}}
  \else
    \usebox{\validationtablebox}
  \fi
\end{minipage}
\end{lrbox}
\ifdim\dimexpr\ht\validationentrybox+\dp\validationentrybox\relax>0.30\textheight
  \resizebox*{!}{0.30\textheight}{\usebox{\validationentrybox}}
\else
  \usebox{\validationentrybox}
\fi
\end{minipage}
\par
\newpage
\noindent
\begin{minipage}[t][0.30\textheight][t]{0.32\textwidth}
\vspace{0pt}\centering
\begin{lrbox}{\validationentrybox}
\begin{minipage}{\linewidth}
  \vspace{0pt}
  \centering
  \objectlabel{HOT3D}{19}{hot3d}
  \captionsetup{font=scriptsize,skip=2pt,hypcap=false}
  \captionof{table}{Validation AR (\%). 200 validation target frames. ConvNeXt-Small.}
  \label{tab:app_hot3d_19}
  \fontsize{6.3}{7.0}\selectfont
  \renewcommand{\arraystretch}{0.95}
  \setlength{\tabcolsep}{1.6pt}
  \begin{lrbox}{\validationtablebox}
  \begin{tabular}{@{}lrrrr@{}}
    \toprule
    Variant & Labels & Single & Jitter & GT-box \\
    \midrule
    Synthetic base & -- & \rankcell{3}{58.85} & \rankcell{3}{60.90} & \rankcell{3}{65.12} \\
    Static pseudo & 5 & -- & -- & -- \\
    Dynamic pseudo & 11 & -- & -- & -- \\
    Static + dynamic & 16 & -- & -- & -- \\
    Box/mask supervision & 3474 & \rankcell{2}{62.40} & \rankcell{2}{64.05} & \rankcell{2}{66.62} \\
    Box/mask/keypoint supervision & 3474 & \rankcell{1}{64.82} & \rankcell{1}{66.35} & \rankcell{1}{68.95} \\
    \bottomrule
  \end{tabular}
  \end{lrbox}
  \includegraphics[width=\linewidth,keepaspectratio]{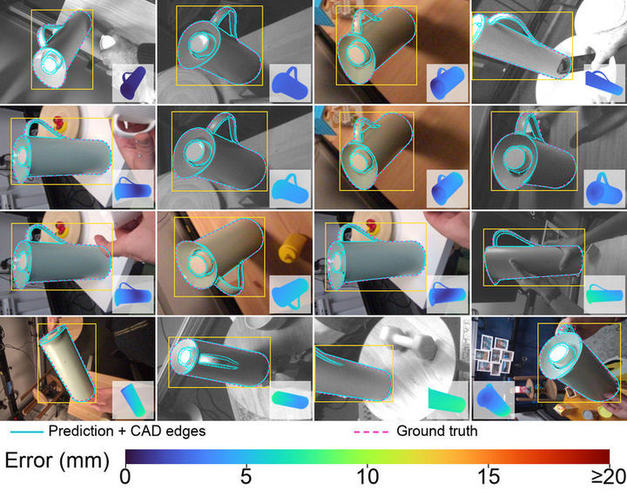}
  \par\vspace{2pt}
  \ifdim\wd\validationtablebox>\linewidth
    \resizebox{\linewidth}{!}{\usebox{\validationtablebox}}
  \else
    \usebox{\validationtablebox}
  \fi
\end{minipage}
\end{lrbox}
\ifdim\dimexpr\ht\validationentrybox+\dp\validationentrybox\relax>0.30\textheight
  \resizebox*{!}{0.30\textheight}{\usebox{\validationentrybox}}
\else
  \usebox{\validationentrybox}
\fi
\end{minipage}
\hfill%
\begin{minipage}[t][0.30\textheight][t]{0.32\textwidth}
\vspace{0pt}\centering
\begin{lrbox}{\validationentrybox}
\begin{minipage}{\linewidth}
  \vspace{0pt}
  \centering
  \objectlabel{HOT3D}{20}{hot3d}
  \captionsetup{font=scriptsize,skip=2pt,hypcap=false}
  \captionof{table}{Validation AR (\%). 120 validation target frames. ConvNeXt-Small.}
  \label{tab:app_hot3d_20}
  \fontsize{6.3}{7.0}\selectfont
  \renewcommand{\arraystretch}{0.95}
  \setlength{\tabcolsep}{1.6pt}
  \begin{lrbox}{\validationtablebox}
  \begin{tabular}{@{}lrrrr@{}}
    \toprule
    Variant & Labels & Single & Jitter & GT-box \\
    \midrule
    Synthetic base & -- & 63.96 & 63.92 & 64.58 \\
    Static pseudo & 2303 & \rankcell{4}{65.83} & \rankcell{4}{66.17} & \rankcell{3}{65.83} \\
    Dynamic pseudo & 680 & \rankcell{5}{64.71} & \rankcell{5}{64.33} & \rankcell{5}{64.88} \\
    Static + dynamic & 2976 & \rankcell{3}{66.04} & \rankcell{3}{66.46} & \rankcell{4}{65.58} \\
    Box/mask supervision & 3372 & \rankcell{2}{67.04} & \rankcell{2}{66.83} & \rankcell{2}{66.54} \\
    Box/mask/keypoint supervision & 3372 & \rankcell{1}{69.46} & \rankcell{1}{69.75} & \rankcell{1}{70.67} \\
    \bottomrule
  \end{tabular}
  \end{lrbox}
  \includegraphics[width=\linewidth,keepaspectratio]{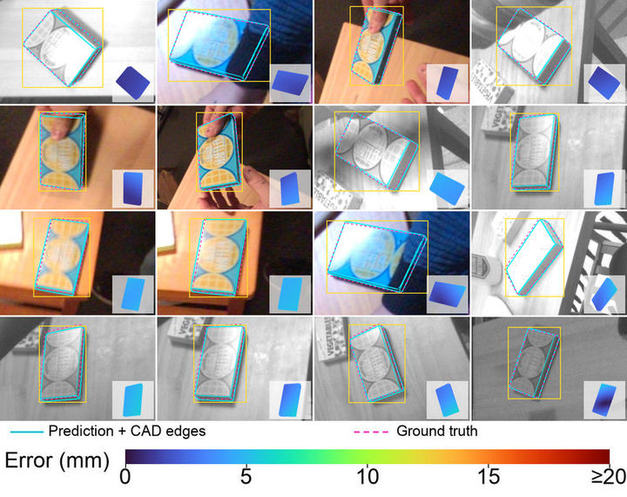}
  \par\vspace{2pt}
  \ifdim\wd\validationtablebox>\linewidth
    \resizebox{\linewidth}{!}{\usebox{\validationtablebox}}
  \else
    \usebox{\validationtablebox}
  \fi
\end{minipage}
\end{lrbox}
\ifdim\dimexpr\ht\validationentrybox+\dp\validationentrybox\relax>0.30\textheight
  \resizebox*{!}{0.30\textheight}{\usebox{\validationentrybox}}
\else
  \usebox{\validationentrybox}
\fi
\end{minipage}
\hfill%
\begin{minipage}[t][0.30\textheight][t]{0.32\textwidth}
\vspace{0pt}\centering
\begin{lrbox}{\validationentrybox}
\begin{minipage}{\linewidth}
  \vspace{0pt}
  \centering
  \objectlabel{HOT3D}{21}{hot3d}
  \captionsetup{font=scriptsize,skip=2pt,hypcap=false}
  \captionof{table}{Validation AR (\%). 121 validation target frames. ConvNeXt-Small.}
  \label{tab:app_hot3d_21}
  \fontsize{6.3}{7.0}\selectfont
  \renewcommand{\arraystretch}{0.95}
  \setlength{\tabcolsep}{1.6pt}
  \begin{lrbox}{\validationtablebox}
  \begin{tabular}{@{}lrrrr@{}}
    \toprule
    Variant & Labels & Single & Jitter & GT-box \\
    \midrule
    Synthetic base & -- & 53.55 & 53.76 & 55.00 \\
    Static pseudo & 2253 & \rankcell{3}{56.12} & \rankcell{2}{56.24} & \rankcell{2}{57.11} \\
    Dynamic pseudo & 506 & \rankcell{5}{53.80} & \rankcell{5}{54.05} & \rankcell{4}{56.61} \\
    Static + dynamic & 2747 & \rankcell{2}{56.16} & \rankcell{3}{56.20} & \rankcell{5}{56.49} \\
    Box/mask supervision & 3410 & \rankcell{4}{54.79} & \rankcell{4}{56.12} & \rankcell{3}{57.02} \\
    Box/mask/keypoint supervision & 3410 & \rankcell{1}{58.02} & \rankcell{1}{59.92} & \rankcell{1}{59.42} \\
    \bottomrule
  \end{tabular}
  \end{lrbox}
  \includegraphics[width=\linewidth,keepaspectratio]{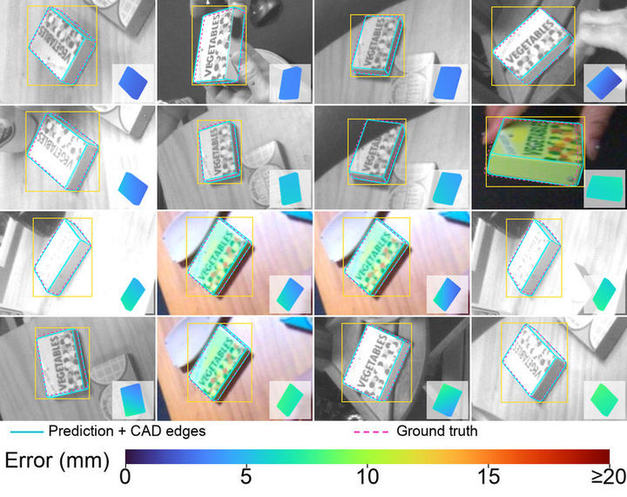}
  \par\vspace{2pt}
  \ifdim\wd\validationtablebox>\linewidth
    \resizebox{\linewidth}{!}{\usebox{\validationtablebox}}
  \else
    \usebox{\validationtablebox}
  \fi
\end{minipage}
\end{lrbox}
\ifdim\dimexpr\ht\validationentrybox+\dp\validationentrybox\relax>0.30\textheight
  \resizebox*{!}{0.30\textheight}{\usebox{\validationentrybox}}
\else
  \usebox{\validationentrybox}
\fi
\end{minipage}
\par\vspace{0.012\textheight}\noindent
\begin{minipage}[t][0.30\textheight][t]{0.32\textwidth}
\vspace{0pt}\centering
\begin{lrbox}{\validationentrybox}
\begin{minipage}{\linewidth}
  \vspace{0pt}
  \centering
  \objectlabel{HOT3D}{22}{hot3d}
  \captionsetup{font=scriptsize,skip=2pt,hypcap=false}
  \captionof{table}{Validation AR (\%). 157 validation target frames. ConvNeXt-Small.}
  \label{tab:app_hot3d_22}
  \fontsize{6.3}{7.0}\selectfont
  \renewcommand{\arraystretch}{0.95}
  \setlength{\tabcolsep}{1.6pt}
  \begin{lrbox}{\validationtablebox}
  \begin{tabular}{@{}lrrrr@{}}
    \toprule
    Variant & Labels & Single & Jitter & GT-box \\
    \midrule
    Synthetic base & -- & \rankcell{5}{60.73} & \rankcell{5}{62.42} & \rankcell{5}{61.94} \\
    Static pseudo & 646 & \rankcell{2}{62.29} & \rankcell{2}{65.03} & \rankcell{2}{62.90} \\
    Dynamic pseudo & 6 & -- & -- & -- \\
    Static + dynamic & 647 & \rankcell{4}{61.69} & \rankcell{3}{64.14} & \rankcell{3}{62.87} \\
    Box/mask supervision & 3391 & \rankcell{1}{62.58} & \rankcell{4}{63.92} & \rankcell{4}{62.64} \\
    Box/mask/keypoint supervision & 3391 & \rankcell{3}{61.78} & \rankcell{1}{65.41} & \rankcell{1}{64.39} \\
    \bottomrule
  \end{tabular}
  \end{lrbox}
  \includegraphics[width=\linewidth,keepaspectratio]{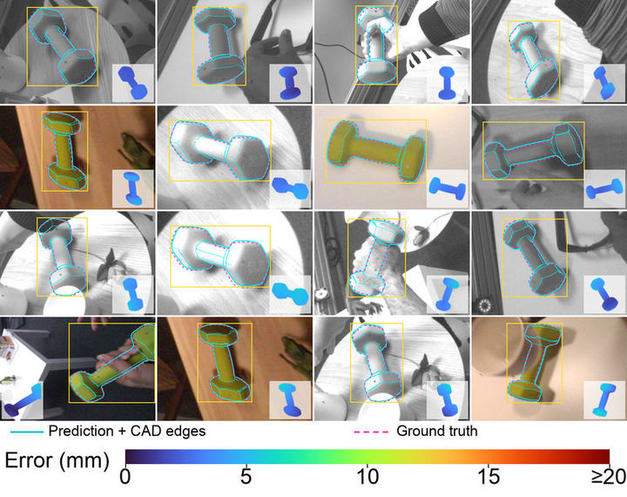}
  \par\vspace{2pt}
  \ifdim\wd\validationtablebox>\linewidth
    \resizebox{\linewidth}{!}{\usebox{\validationtablebox}}
  \else
    \usebox{\validationtablebox}
  \fi
\end{minipage}
\end{lrbox}
\ifdim\dimexpr\ht\validationentrybox+\dp\validationentrybox\relax>0.30\textheight
  \resizebox*{!}{0.30\textheight}{\usebox{\validationentrybox}}
\else
  \usebox{\validationentrybox}
\fi
\end{minipage}
\hfill%
\begin{minipage}[t][0.30\textheight][t]{0.32\textwidth}
\vspace{0pt}\centering
\begin{lrbox}{\validationentrybox}
\begin{minipage}{\linewidth}
  \vspace{0pt}
  \centering
  \objectlabel{HOT3D}{23}{hot3d}
  \captionsetup{font=scriptsize,skip=2pt,hypcap=false}
  \captionof{table}{Validation AR (\%). 298 validation target frames. ConvNeXt-Small.}
  \label{tab:app_hot3d_23}
  \fontsize{6.3}{7.0}\selectfont
  \renewcommand{\arraystretch}{0.95}
  \setlength{\tabcolsep}{1.6pt}
  \begin{lrbox}{\validationtablebox}
  \begin{tabular}{@{}lrrrr@{}}
    \toprule
    Variant & Labels & Single & Jitter & GT-box \\
    \midrule
    Synthetic base & -- & 62.00 & 62.43 & \rankcell{5}{70.30} \\
    Static pseudo & 1835 & \rankcell{2}{64.41} & \rankcell{4}{64.65} & \rankcell{4}{70.60} \\
    Dynamic pseudo & 313 & \rankcell{5}{63.66} & \rankcell{5}{63.71} & \rankcell{2}{70.65} \\
    Static + dynamic & 2136 & \rankcell{3}{64.24} & \rankcell{2}{64.73} & 70.15 \\
    Box/mask supervision & 3300 & \rankcell{4}{64.03} & \rankcell{3}{64.71} & \rankcell{3}{70.62} \\
    Box/mask/keypoint supervision & 3300 & \rankcell{1}{66.71} & \rankcell{1}{67.08} & \rankcell{1}{71.26} \\
    \bottomrule
  \end{tabular}
  \end{lrbox}
  \includegraphics[width=\linewidth,keepaspectratio]{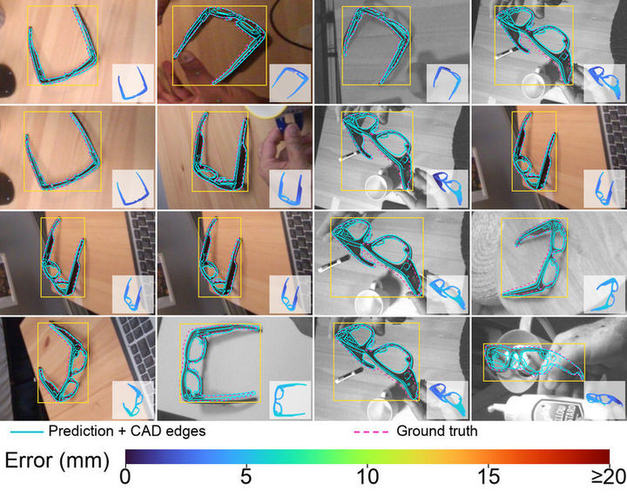}
  \par\vspace{2pt}
  \ifdim\wd\validationtablebox>\linewidth
    \resizebox{\linewidth}{!}{\usebox{\validationtablebox}}
  \else
    \usebox{\validationtablebox}
  \fi
\end{minipage}
\end{lrbox}
\ifdim\dimexpr\ht\validationentrybox+\dp\validationentrybox\relax>0.30\textheight
  \resizebox*{!}{0.30\textheight}{\usebox{\validationentrybox}}
\else
  \usebox{\validationentrybox}
\fi
\end{minipage}
\hfill%
\begin{minipage}[t][0.30\textheight][t]{0.32\textwidth}
\vspace{0pt}\centering
\begin{lrbox}{\validationentrybox}
\begin{minipage}{\linewidth}
  \vspace{0pt}
  \centering
  \objectlabel{HOT3D}{24}{hot3d}
  \captionsetup{font=scriptsize,skip=2pt,hypcap=false}
  \captionof{table}{Validation AR (\%). 293 validation target frames. ConvNeXt-Small.}
  \label{tab:app_hot3d_24}
  \fontsize{6.3}{7.0}\selectfont
  \renewcommand{\arraystretch}{0.95}
  \setlength{\tabcolsep}{1.6pt}
  \begin{lrbox}{\validationtablebox}
  \begin{tabular}{@{}lrrrr@{}}
    \toprule
    Variant & Labels & Single & Jitter & GT-box \\
    \midrule
    Synthetic base & -- & 57.53 & 58.29 & \rankcell{4}{66.98} \\
    Static pseudo & 1999 & \rankcell{4}{61.33} & \rankcell{4}{62.06} & \rankcell{2}{67.20} \\
    Dynamic pseudo & 164 & \rankcell{5}{60.19} & \rankcell{5}{60.55} & 66.64 \\
    Static + dynamic & 2151 & \rankcell{3}{61.93} & \rankcell{3}{63.04} & \rankcell{5}{66.84} \\
    Box/mask supervision & 3356 & \rankcell{2}{63.60} & \rankcell{2}{64.40} & \rankcell{3}{67.12} \\
    Box/mask/keypoint supervision & 3356 & \rankcell{1}{67.99} & \rankcell{1}{68.58} & \rankcell{1}{72.20} \\
    \bottomrule
  \end{tabular}
  \end{lrbox}
  \includegraphics[width=\linewidth,keepaspectratio]{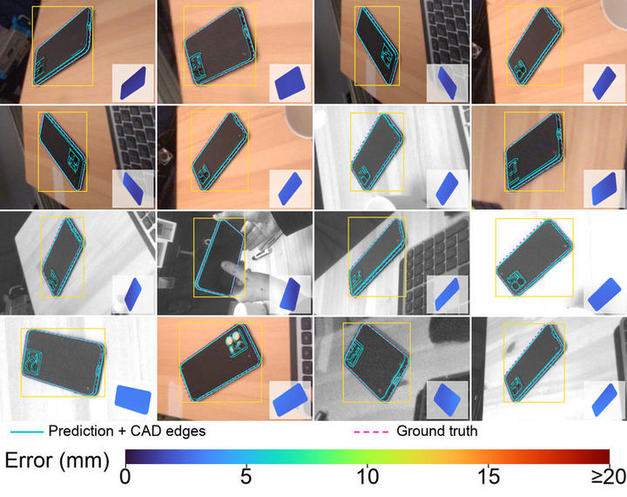}
  \par\vspace{2pt}
  \ifdim\wd\validationtablebox>\linewidth
    \resizebox{\linewidth}{!}{\usebox{\validationtablebox}}
  \else
    \usebox{\validationtablebox}
  \fi
\end{minipage}
\end{lrbox}
\ifdim\dimexpr\ht\validationentrybox+\dp\validationentrybox\relax>0.30\textheight
  \resizebox*{!}{0.30\textheight}{\usebox{\validationentrybox}}
\else
  \usebox{\validationentrybox}
\fi
\end{minipage}
\par\vspace{0.012\textheight}\noindent
\begin{minipage}[t][0.30\textheight][t]{0.32\textwidth}
\vspace{0pt}\centering
\begin{lrbox}{\validationentrybox}
\begin{minipage}{\linewidth}
  \vspace{0pt}
  \centering
  \objectlabel{HOT3D}{25}{hot3d}
  \captionsetup{font=scriptsize,skip=2pt,hypcap=false}
  \captionof{table}{Validation AR (\%). 163 validation target frames. ConvNeXt-Small.}
  \label{tab:app_hot3d_25}
  \fontsize{6.3}{7.0}\selectfont
  \renewcommand{\arraystretch}{0.95}
  \setlength{\tabcolsep}{1.6pt}
  \begin{lrbox}{\validationtablebox}
  \begin{tabular}{@{}lrrrr@{}}
    \toprule
    Variant & Labels & Single & Jitter & GT-box \\
    \midrule
    Synthetic base & -- & \rankcell{5}{38.16} & \rankcell{5}{38.65} & \rankcell{4}{47.98} \\
    Static pseudo & 53 & \rankcell{3}{40.46} & \rankcell{3}{41.84} & \rankcell{3}{48.19} \\
    Dynamic pseudo & 0 & -- & -- & -- \\
    Static + dynamic & 52 & \rankcell{4}{39.54} & \rankcell{4}{40.77} & \rankcell{5}{47.79} \\
    Box/mask supervision & 3257 & \rankcell{2}{48.71} & \rankcell{2}{47.79} & \rankcell{2}{51.41} \\
    Box/mask/keypoint supervision & 3257 & \rankcell{1}{53.13} & \rankcell{1}{51.44} & \rankcell{1}{55.21} \\
    \bottomrule
  \end{tabular}
  \end{lrbox}
  \includegraphics[width=\linewidth,keepaspectratio]{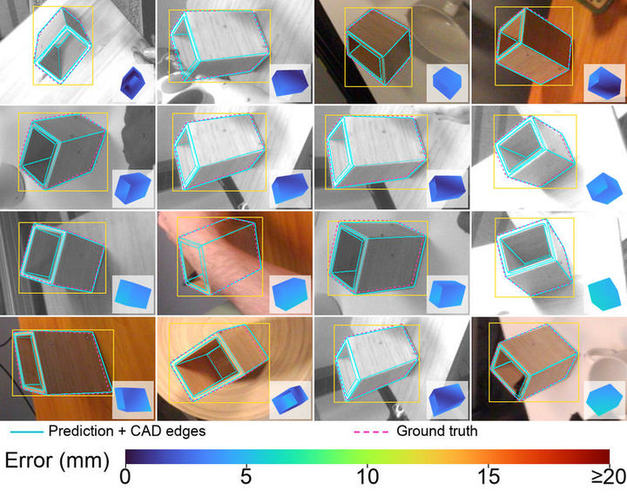}
  \par\vspace{2pt}
  \ifdim\wd\validationtablebox>\linewidth
    \resizebox{\linewidth}{!}{\usebox{\validationtablebox}}
  \else
    \usebox{\validationtablebox}
  \fi
\end{minipage}
\end{lrbox}
\ifdim\dimexpr\ht\validationentrybox+\dp\validationentrybox\relax>0.30\textheight
  \resizebox*{!}{0.30\textheight}{\usebox{\validationentrybox}}
\else
  \usebox{\validationentrybox}
\fi
\end{minipage}
\hfill%
\begin{minipage}[t][0.30\textheight][t]{0.32\textwidth}
\vspace{0pt}\centering
\begin{lrbox}{\validationentrybox}
\begin{minipage}{\linewidth}
  \vspace{0pt}
  \centering
  \objectlabel{HOT3D}{26}{hot3d}
  \captionsetup{font=scriptsize,skip=2pt,hypcap=false}
  \captionof{table}{Validation AR (\%). 156 validation target frames. ConvNeXt-Small.}
  \label{tab:app_hot3d_26}
  \fontsize{6.3}{7.0}\selectfont
  \renewcommand{\arraystretch}{0.95}
  \setlength{\tabcolsep}{1.6pt}
  \begin{lrbox}{\validationtablebox}
  \begin{tabular}{@{}lrrrr@{}}
    \toprule
    Variant & Labels & Single & Jitter & GT-box \\
    \midrule
    Synthetic base & -- & \rankcell{4}{70.48} & \rankcell{4}{70.29} & \rankcell{4}{78.65} \\
    Static pseudo & 73 & \rankcell{1}{71.92} & \rankcell{1}{72.95} & \rankcell{5}{77.47} \\
    Dynamic pseudo & 8 & -- & -- & -- \\
    Static + dynamic & 80 & \rankcell{2}{71.76} & \rankcell{2}{72.63} & \rankcell{3}{79.07} \\
    Box/mask supervision & 3254 & \rankcell{5}{68.27} & \rankcell{5}{70.10} & \rankcell{2}{79.84} \\
    Box/mask/keypoint supervision & 3254 & \rankcell{3}{70.93} & \rankcell{3}{72.34} & \rankcell{1}{81.89} \\
    \bottomrule
  \end{tabular}
  \end{lrbox}
  \includegraphics[width=\linewidth,keepaspectratio]{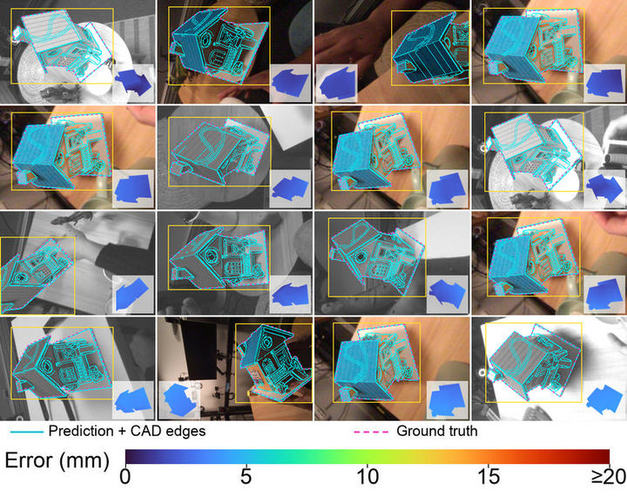}
  \par\vspace{2pt}
  \ifdim\wd\validationtablebox>\linewidth
    \resizebox{\linewidth}{!}{\usebox{\validationtablebox}}
  \else
    \usebox{\validationtablebox}
  \fi
\end{minipage}
\end{lrbox}
\ifdim\dimexpr\ht\validationentrybox+\dp\validationentrybox\relax>0.30\textheight
  \resizebox*{!}{0.30\textheight}{\usebox{\validationentrybox}}
\else
  \usebox{\validationentrybox}
\fi
\end{minipage}
\hfill%
\begin{minipage}[t][0.30\textheight][t]{0.32\textwidth}
\vspace{0pt}\centering
\begin{lrbox}{\validationentrybox}
\begin{minipage}{\linewidth}
  \vspace{0pt}
  \centering
  \objectlabel{HOT3D}{27}{hot3d}
  \captionsetup{font=scriptsize,skip=2pt,hypcap=false}
  \captionof{table}{Validation AR (\%). 152 validation target frames. ConvNeXt-Small.}
  \label{tab:app_hot3d_27}
  \fontsize{6.3}{7.0}\selectfont
  \renewcommand{\arraystretch}{0.95}
  \setlength{\tabcolsep}{1.6pt}
  \begin{lrbox}{\validationtablebox}
  \begin{tabular}{@{}lrrrr@{}}
    \toprule
    Variant & Labels & Single & Jitter & GT-box \\
    \midrule
    Synthetic base & -- & \rankcell{5}{58.45} & \rankcell{5}{60.10} & \rankcell{5}{64.80} \\
    Static pseudo & 2235 & \rankcell{3}{60.49} & \rankcell{3}{60.33} & \rankcell{3}{65.33} \\
    Dynamic pseudo & 781 & \rankcell{1}{63.26} & \rankcell{2}{63.19} & 64.34 \\
    Static + dynamic & 2995 & \rankcell{2}{63.22} & \rankcell{1}{64.38} & \rankcell{4}{65.10} \\
    Box/mask supervision & 3277 & \rankcell{4}{59.57} & \rankcell{4}{60.20} & \rankcell{1}{66.09} \\
    Box/mask/keypoint supervision & 3277 & 57.76 & 59.14 & \rankcell{2}{65.69} \\
    \bottomrule
  \end{tabular}
  \end{lrbox}
  \includegraphics[width=\linewidth,keepaspectratio]{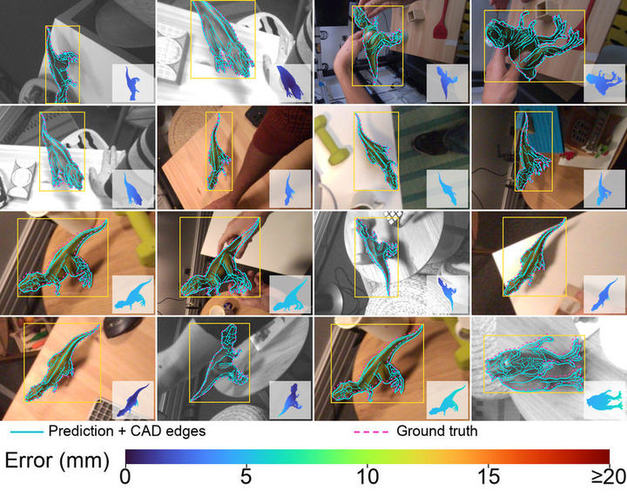}
  \par\vspace{2pt}
  \ifdim\wd\validationtablebox>\linewidth
    \resizebox{\linewidth}{!}{\usebox{\validationtablebox}}
  \else
    \usebox{\validationtablebox}
  \fi
\end{minipage}
\end{lrbox}
\ifdim\dimexpr\ht\validationentrybox+\dp\validationentrybox\relax>0.30\textheight
  \resizebox*{!}{0.30\textheight}{\usebox{\validationentrybox}}
\else
  \usebox{\validationentrybox}
\fi
\end{minipage}
\par
\newpage
\noindent
\begin{minipage}[t][0.30\textheight][t]{0.32\textwidth}
\vspace{0pt}\centering
\begin{lrbox}{\validationentrybox}
\begin{minipage}{\linewidth}
  \vspace{0pt}
  \centering
  \objectlabel{HOT3D}{28}{hot3d}
  \captionsetup{font=scriptsize,skip=2pt,hypcap=false}
  \captionof{table}{Validation AR (\%). 264 validation target frames. ConvNeXt-Small.}
  \label{tab:app_hot3d_28}
  \fontsize{6.3}{7.0}\selectfont
  \renewcommand{\arraystretch}{0.95}
  \setlength{\tabcolsep}{1.6pt}
  \begin{lrbox}{\validationtablebox}
  \begin{tabular}{@{}lrrrr@{}}
    \toprule
    Variant & Labels & Single & Jitter & GT-box \\
    \midrule
    Synthetic base & -- & 42.97 & 44.17 & 75.89 \\
    Static pseudo & 836 & \rankcell{4}{72.50} & \rankcell{4}{72.71} & \rankcell{5}{86.42} \\
    Dynamic pseudo & 131 & \rankcell{5}{72.06} & \rankcell{5}{71.78} & \rankcell{4}{86.91} \\
    Static + dynamic & 964 & \rankcell{3}{76.04} & \rankcell{3}{76.55} & \rankcell{3}{89.05} \\
    Box/mask supervision & 3434 & \rankcell{2}{78.39} & \rankcell{2}{78.75} & \rankcell{2}{89.45} \\
    Box/mask/keypoint supervision & 3434 & \rankcell{1}{80.42} & \rankcell{1}{80.38} & \rankcell{1}{90.11} \\
    \bottomrule
  \end{tabular}
  \end{lrbox}
  \includegraphics[width=\linewidth,keepaspectratio]{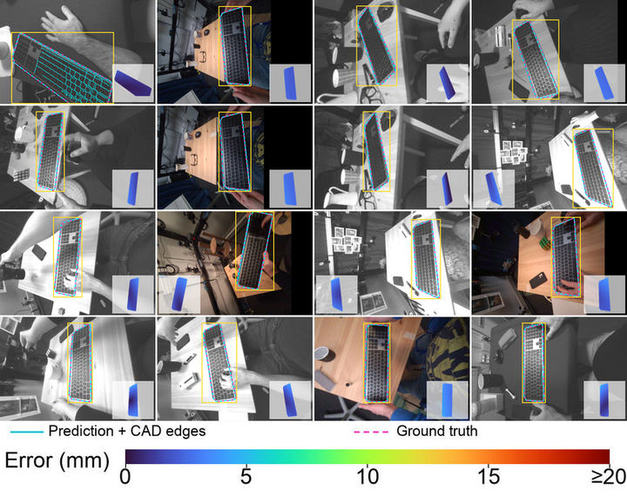}
  \par\vspace{2pt}
  \ifdim\wd\validationtablebox>\linewidth
    \resizebox{\linewidth}{!}{\usebox{\validationtablebox}}
  \else
    \usebox{\validationtablebox}
  \fi
\end{minipage}
\end{lrbox}
\ifdim\dimexpr\ht\validationentrybox+\dp\validationentrybox\relax>0.30\textheight
  \resizebox*{!}{0.30\textheight}{\usebox{\validationentrybox}}
\else
  \usebox{\validationentrybox}
\fi
\end{minipage}
\hfill%
\begin{minipage}[t][0.30\textheight][t]{0.32\textwidth}
\vspace{0pt}\centering
\begin{lrbox}{\validationentrybox}
\begin{minipage}{\linewidth}
  \vspace{0pt}
  \centering
  \objectlabel{HOT3D}{29}{hot3d}
  \captionsetup{font=scriptsize,skip=2pt,hypcap=false}
  \captionof{table}{Validation AR (\%). 227 validation target frames. ConvNeXt-Small.}
  \label{tab:app_hot3d_29}
  \fontsize{6.3}{7.0}\selectfont
  \renewcommand{\arraystretch}{0.95}
  \setlength{\tabcolsep}{1.6pt}
  \begin{lrbox}{\validationtablebox}
  \begin{tabular}{@{}lrrrr@{}}
    \toprule
    Variant & Labels & Single & Jitter & GT-box \\
    \midrule
    Synthetic base & -- & 64.47 & 65.81 & \rankcell{5}{69.14} \\
    Static pseudo & 1405 & \rankcell{3}{67.42} & \rankcell{3}{68.04} & \rankcell{2}{70.88} \\
    Dynamic pseudo & 141 & \rankcell{5}{66.39} & \rankcell{5}{67.18} & \rankcell{4}{70.66} \\
    Static + dynamic & 1537 & \rankcell{4}{67.27} & \rankcell{4}{67.95} & \rankcell{3}{70.75} \\
    Box/mask supervision & 3463 & \rankcell{2}{68.08} & \rankcell{2}{69.05} & \rankcell{3}{70.75} \\
    Box/mask/keypoint supervision & 3463 & \rankcell{1}{70.84} & \rankcell{1}{70.88} & \rankcell{1}{74.60} \\
    \bottomrule
  \end{tabular}
  \end{lrbox}
  \includegraphics[width=\linewidth,keepaspectratio]{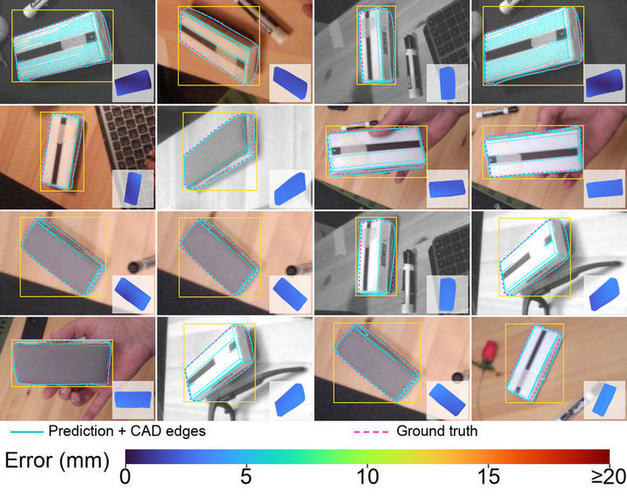}
  \par\vspace{2pt}
  \ifdim\wd\validationtablebox>\linewidth
    \resizebox{\linewidth}{!}{\usebox{\validationtablebox}}
  \else
    \usebox{\validationtablebox}
  \fi
\end{minipage}
\end{lrbox}
\ifdim\dimexpr\ht\validationentrybox+\dp\validationentrybox\relax>0.30\textheight
  \resizebox*{!}{0.30\textheight}{\usebox{\validationentrybox}}
\else
  \usebox{\validationentrybox}
\fi
\end{minipage}
\hfill%
\begin{minipage}[t][0.30\textheight][t]{0.32\textwidth}
\vspace{0pt}\centering
\begin{lrbox}{\validationentrybox}
\begin{minipage}{\linewidth}
  \vspace{0pt}
  \centering
  \objectlabel{HOT3D}{30}{hot3d}
  \captionsetup{font=scriptsize,skip=2pt,hypcap=false}
  \captionof{table}{Validation AR (\%). 285 validation target frames. ConvNeXt-Small.}
  \label{tab:app_hot3d_30}
  \fontsize{6.3}{7.0}\selectfont
  \renewcommand{\arraystretch}{0.95}
  \setlength{\tabcolsep}{1.6pt}
  \begin{lrbox}{\validationtablebox}
  \begin{tabular}{@{}lrrrr@{}}
    \toprule
    Variant & Labels & Single & Jitter & GT-box \\
    \midrule
    Synthetic base & -- & \rankcell{3}{75.28} & \rankcell{4}{75.11} & 75.05 \\
    Static pseudo & 1538 & 74.98 & \rankcell{3}{75.26} & \rankcell{5}{75.28} \\
    Dynamic pseudo & 644 & \rankcell{5}{75.07} & \rankcell{4}{75.11} & \rankcell{2}{76.28} \\
    Static + dynamic & 2181 & \rankcell{2}{75.35} & \rankcell{2}{75.47} & \rankcell{4}{75.96} \\
    Box/mask supervision & 3299 & \rankcell{4}{75.11} & \rankcell{5}{74.81} & \rankcell{3}{76.02} \\
    Box/mask/keypoint supervision & 3299 & \rankcell{1}{76.07} & \rankcell{1}{77.19} & \rankcell{1}{76.72} \\
    \bottomrule
  \end{tabular}
  \end{lrbox}
  \includegraphics[width=\linewidth,keepaspectratio]{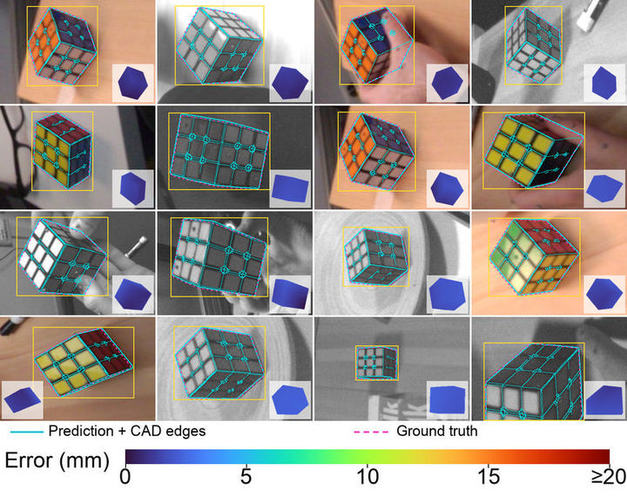}
  \par\vspace{2pt}
  \ifdim\wd\validationtablebox>\linewidth
    \resizebox{\linewidth}{!}{\usebox{\validationtablebox}}
  \else
    \usebox{\validationtablebox}
  \fi
\end{minipage}
\end{lrbox}
\ifdim\dimexpr\ht\validationentrybox+\dp\validationentrybox\relax>0.30\textheight
  \resizebox*{!}{0.30\textheight}{\usebox{\validationentrybox}}
\else
  \usebox{\validationentrybox}
\fi
\end{minipage}
\par\vspace{0.012\textheight}\noindent
\begin{minipage}[t][0.30\textheight][t]{0.32\textwidth}
\vspace{0pt}\centering
\begin{lrbox}{\validationentrybox}
\begin{minipage}{\linewidth}
  \vspace{0pt}
  \centering
  \objectlabel{HOT3D}{31}{hot3d}
  \captionsetup{font=scriptsize,skip=2pt,hypcap=false}
  \captionof{table}{Validation AR (\%). 250 validation target frames. ConvNeXt-Small.}
  \label{tab:app_hot3d_31}
  \fontsize{6.3}{7.0}\selectfont
  \renewcommand{\arraystretch}{0.95}
  \setlength{\tabcolsep}{1.6pt}
  \begin{lrbox}{\validationtablebox}
  \begin{tabular}{@{}lrrrr@{}}
    \toprule
    Variant & Labels & Single & Jitter & GT-box \\
    \midrule
    Synthetic base & -- & 47.54 & 49.14 & 52.64 \\
    Static pseudo & 2329 & \rankcell{4}{51.64} & \rankcell{3}{53.58} & \rankcell{5}{53.88} \\
    Dynamic pseudo & 417 & \rankcell{2}{51.90} & \rankcell{5}{52.86} & \rankcell{2}{55.68} \\
    Static + dynamic & 2735 & \rankcell{5}{51.50} & \rankcell{4}{53.42} & \rankcell{3}{54.46} \\
    Box/mask supervision & 3325 & \rankcell{3}{51.70} & \rankcell{2}{53.68} & \rankcell{4}{54.28} \\
    Box/mask/keypoint supervision & 3325 & \rankcell{1}{54.78} & \rankcell{1}{55.56} & \rankcell{1}{58.80} \\
    \bottomrule
  \end{tabular}
  \end{lrbox}
  \includegraphics[width=\linewidth,keepaspectratio]{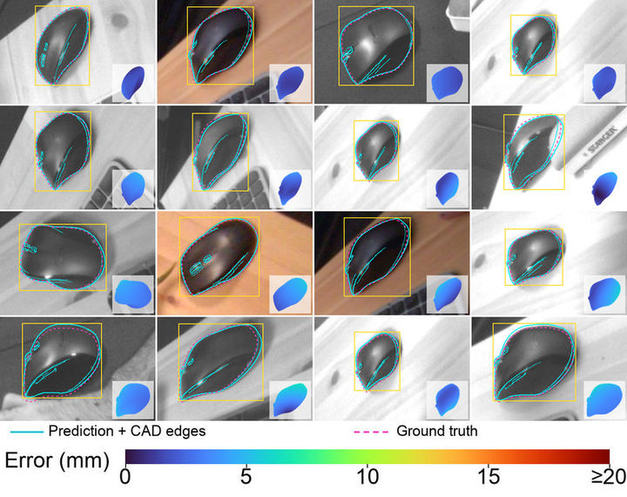}
  \par\vspace{2pt}
  \ifdim\wd\validationtablebox>\linewidth
    \resizebox{\linewidth}{!}{\usebox{\validationtablebox}}
  \else
    \usebox{\validationtablebox}
  \fi
\end{minipage}
\end{lrbox}
\ifdim\dimexpr\ht\validationentrybox+\dp\validationentrybox\relax>0.30\textheight
  \resizebox*{!}{0.30\textheight}{\usebox{\validationentrybox}}
\else
  \usebox{\validationentrybox}
\fi
\end{minipage}
\hfill%
\begin{minipage}[t][0.30\textheight][t]{0.32\textwidth}
\vspace{0pt}\centering
\begin{lrbox}{\validationentrybox}
\begin{minipage}{\linewidth}
  \vspace{0pt}
  \centering
  \objectlabel{HOT3D}{32}{hot3d}
  \captionsetup{font=scriptsize,skip=2pt,hypcap=false}
  \captionof{table}{Validation AR (\%). 221 validation target frames. ConvNeXt-Small.}
  \label{tab:app_hot3d_32}
  \fontsize{6.3}{7.0}\selectfont
  \renewcommand{\arraystretch}{0.95}
  \setlength{\tabcolsep}{1.6pt}
  \begin{lrbox}{\validationtablebox}
  \begin{tabular}{@{}lrrrr@{}}
    \toprule
    Variant & Labels & Single & Jitter & GT-box \\
    \midrule
    Synthetic base & -- & 56.58 & 58.08 & \rankcell{5}{59.46} \\
    Static pseudo & 2164 & \rankcell{1}{60.34} & \rankcell{1}{60.59} & \rankcell{3}{60.29} \\
    Dynamic pseudo & 688 & \rankcell{4}{58.57} & \rankcell{5}{58.73} & \rankcell{1}{60.50} \\
    Static + dynamic & 2850 & \rankcell{5}{58.10} & \rankcell{4}{59.10} & \rankcell{2}{60.34} \\
    Box/mask supervision & 3604 & \rankcell{3}{58.71} & \rankcell{3}{59.30} & \rankcell{1}{60.50} \\
    Box/mask/keypoint supervision & 3604 & \rankcell{2}{58.94} & \rankcell{2}{60.36} & \rankcell{4}{60.27} \\
    \bottomrule
  \end{tabular}
  \end{lrbox}
  \includegraphics[width=\linewidth,keepaspectratio]{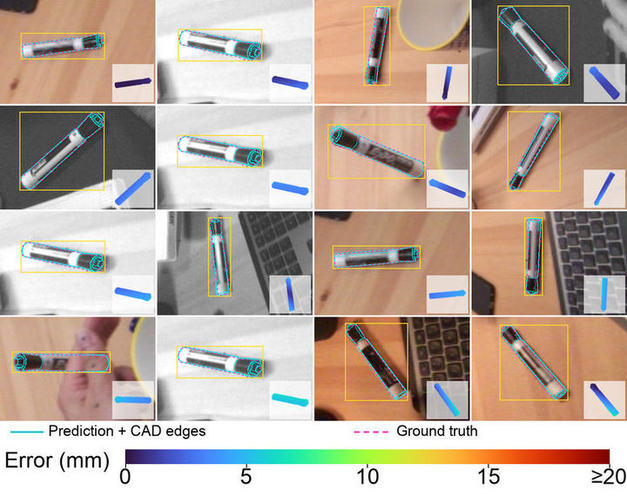}
  \par\vspace{2pt}
  \ifdim\wd\validationtablebox>\linewidth
    \resizebox{\linewidth}{!}{\usebox{\validationtablebox}}
  \else
    \usebox{\validationtablebox}
  \fi
\end{minipage}
\end{lrbox}
\ifdim\dimexpr\ht\validationentrybox+\dp\validationentrybox\relax>0.30\textheight
  \resizebox*{!}{0.30\textheight}{\usebox{\validationentrybox}}
\else
  \usebox{\validationentrybox}
\fi
\end{minipage}
\hfill%
\begin{minipage}[t][0.30\textheight][t]{0.32\textwidth}
\vspace{0pt}\centering
\begin{lrbox}{\validationentrybox}
\begin{minipage}{\linewidth}
  \vspace{0pt}
  \centering
  \objectlabel{HOT3D}{33}{hot3d}
  \captionsetup{font=scriptsize,skip=2pt,hypcap=false}
  \captionof{table}{Validation AR (\%). 124 validation target frames. ConvNeXt-Small.}
  \label{tab:app_hot3d_33}
  \fontsize{6.3}{7.0}\selectfont
  \renewcommand{\arraystretch}{0.95}
  \setlength{\tabcolsep}{1.6pt}
  \begin{lrbox}{\validationtablebox}
  \begin{tabular}{@{}lrrrr@{}}
    \toprule
    Variant & Labels & Single & Jitter & GT-box \\
    \midrule
    Synthetic base & -- & 34.23 & 34.68 & \rankcell{5}{62.14} \\
    Static pseudo & 1885 & \rankcell{3}{48.23} & \rankcell{4}{49.15} & \rankcell{4}{63.63} \\
    Dynamic pseudo & 270 & \rankcell{5}{40.20} & \rankcell{5}{42.22} & \rankcell{3}{63.71} \\
    Static + dynamic & 2141 & \rankcell{2}{48.31} & \rankcell{2}{50.00} & \rankcell{1}{64.64} \\
    Box/mask supervision & 3421 & \rankcell{1}{50.36} & \rankcell{1}{50.77} & \rankcell{2}{64.44} \\
    Box/mask/keypoint supervision & 3421 & \rankcell{4}{47.98} & \rankcell{3}{49.27} & 60.48 \\
    \bottomrule
  \end{tabular}
  \end{lrbox}
  \includegraphics[width=\linewidth,keepaspectratio]{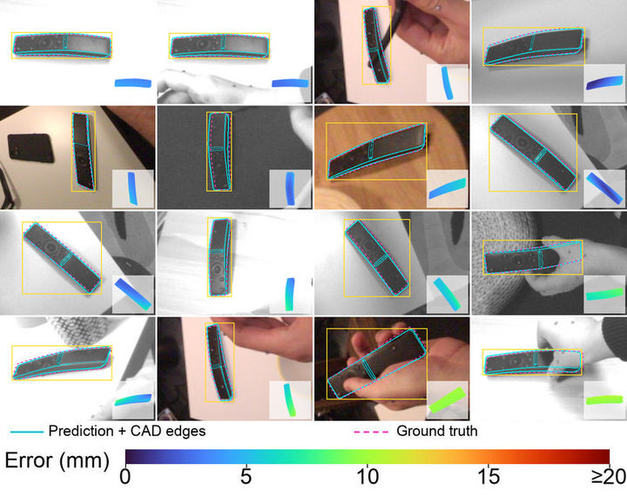}
  \par\vspace{2pt}
  \ifdim\wd\validationtablebox>\linewidth
    \resizebox{\linewidth}{!}{\usebox{\validationtablebox}}
  \else
    \usebox{\validationtablebox}
  \fi
\end{minipage}
\end{lrbox}
\ifdim\dimexpr\ht\validationentrybox+\dp\validationentrybox\relax>0.30\textheight
  \resizebox*{!}{0.30\textheight}{\usebox{\validationentrybox}}
\else
  \usebox{\validationentrybox}
\fi
\end{minipage}
\clearpage
\noindent
\begin{minipage}[t][0.30\textheight][t]{0.32\textwidth}
\vspace{0pt}\centering
\begin{lrbox}{\validationentrybox}
    \begin{minipage}{\linewidth}
    \centering
    \objectlabel{T-LESS}{01}{tless}
    {\scriptsize 818 evaluated instances}\par\vspace{2pt}
    \includegraphics[width=\linewidth,keepaspectratio]{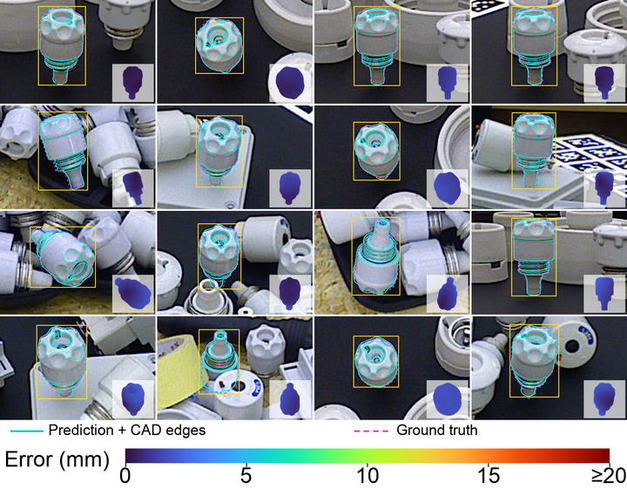}
    \par\vspace{2pt}
    {\fontsize{6.3}{7.0}\selectfont
    \renewcommand{\arraystretch}{0.95}
    \setlength{\tabcolsep}{1.6pt}
    \begin{tabular*}{\linewidth}{@{\extracolsep{\fill}}lrrr@{}}
      \toprule
      Variant & AP & AP$_{\mathrm{MSSD}}$ & AP$_{\mathrm{MSPD}}$ \\
      \midrule
      Synthetic base & \rankcell{2}{43.2} & \rankcell{2}{26.5} & \rankcell{2}{59.8} \\
      Multisensor sup. & \rankcell{1}{51.6} & \rankcell{1}{35.2} & \rankcell{1}{68.0} \\
      \bottomrule
    \end{tabular*}}
    \end{minipage}
    \end{lrbox}
\ifdim\dimexpr\ht\validationentrybox+\dp\validationentrybox\relax>0.30\textheight
  \resizebox*{!}{0.30\textheight}{\usebox{\validationentrybox}}
\else
  \usebox{\validationentrybox}
\fi
\end{minipage}
\hfill%
\begin{minipage}[t][0.30\textheight][t]{0.32\textwidth}
\vspace{0pt}\centering
\begin{lrbox}{\validationentrybox}
    \begin{minipage}{\linewidth}
    \centering
    \objectlabel{T-LESS}{02}{tless}
    {\scriptsize 465 evaluated instances}\par\vspace{2pt}
    \includegraphics[width=\linewidth,keepaspectratio]{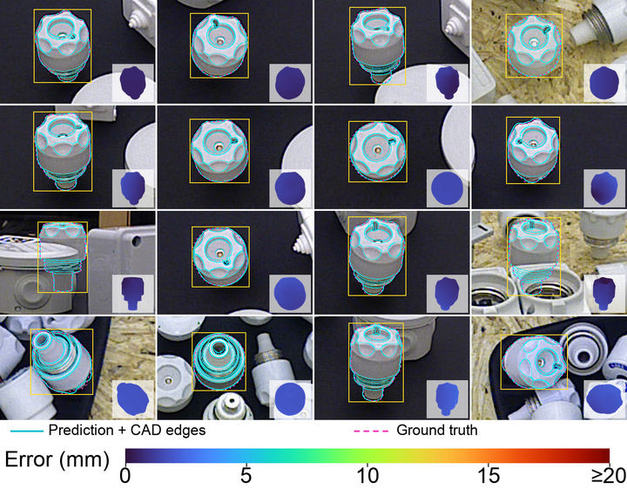}
    \par\vspace{2pt}
    {\fontsize{6.3}{7.0}\selectfont
    \renewcommand{\arraystretch}{0.95}
    \setlength{\tabcolsep}{1.6pt}
    \begin{tabular*}{\linewidth}{@{\extracolsep{\fill}}lrrr@{}}
      \toprule
      Variant & AP & AP$_{\mathrm{MSSD}}$ & AP$_{\mathrm{MSPD}}$ \\
      \midrule
      Synthetic base & \rankcell{2}{47.7} & \rankcell{2}{27.8} & \rankcell{2}{67.7} \\
      Multisensor sup. & \rankcell{1}{54.6} & \rankcell{1}{40.4} & \rankcell{1}{68.7} \\
      \bottomrule
    \end{tabular*}}
    \end{minipage}
    \end{lrbox}
\ifdim\dimexpr\ht\validationentrybox+\dp\validationentrybox\relax>0.30\textheight
  \resizebox*{!}{0.30\textheight}{\usebox{\validationentrybox}}
\else
  \usebox{\validationentrybox}
\fi
\end{minipage}
\hfill%
\begin{minipage}[t][0.30\textheight][t]{0.32\textwidth}
\vspace{0pt}\centering
\begin{lrbox}{\validationentrybox}
    \begin{minipage}{\linewidth}
    \centering
    \objectlabel{T-LESS}{03}{tless}
    {\scriptsize 397 evaluated instances}\par\vspace{2pt}
    \includegraphics[width=\linewidth,keepaspectratio]{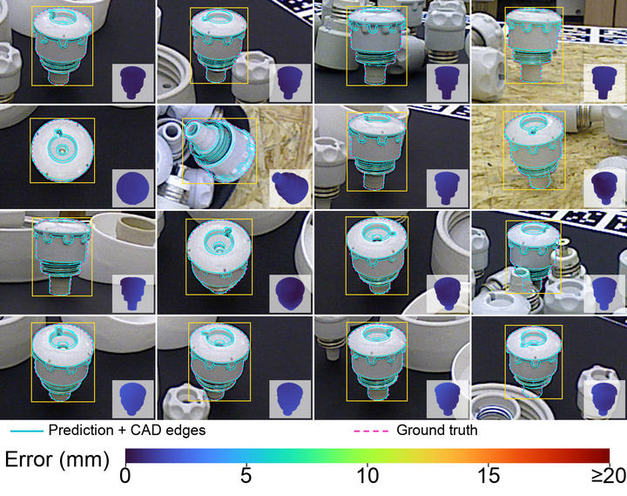}
    \par\vspace{2pt}
    {\fontsize{6.3}{7.0}\selectfont
    \renewcommand{\arraystretch}{0.95}
    \setlength{\tabcolsep}{1.6pt}
    \begin{tabular*}{\linewidth}{@{\extracolsep{\fill}}lrrr@{}}
      \toprule
      Variant & AP & AP$_{\mathrm{MSSD}}$ & AP$_{\mathrm{MSPD}}$ \\
      \midrule
      Synthetic base & \rankcell{2}{61.3} & \rankcell{2}{38.7} & \rankcell{2}{83.9} \\
      Multisensor sup. & \rankcell{1}{75.7} & \rankcell{1}{62.2} & \rankcell{1}{89.1} \\
      \bottomrule
    \end{tabular*}}
    \end{minipage}
    \end{lrbox}
\ifdim\dimexpr\ht\validationentrybox+\dp\validationentrybox\relax>0.30\textheight
  \resizebox*{!}{0.30\textheight}{\usebox{\validationentrybox}}
\else
  \usebox{\validationentrybox}
\fi
\end{minipage}
\par\vspace{0.012\textheight}\noindent
\begin{minipage}[t][0.30\textheight][t]{0.32\textwidth}
\vspace{0pt}\centering
\begin{lrbox}{\validationentrybox}
    \begin{minipage}{\linewidth}
    \centering
    \objectlabel{T-LESS}{04}{tless}
    {\scriptsize 619 evaluated instances}\par\vspace{2pt}
    \includegraphics[width=\linewidth,keepaspectratio]{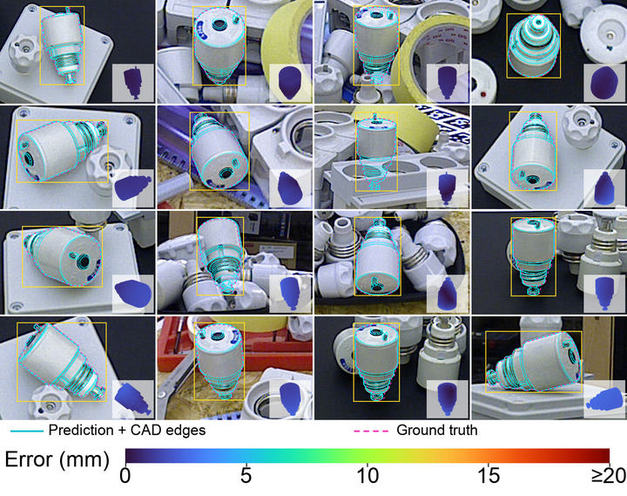}
    \par\vspace{2pt}
    {\fontsize{6.3}{7.0}\selectfont
    \renewcommand{\arraystretch}{0.95}
    \setlength{\tabcolsep}{1.6pt}
    \begin{tabular*}{\linewidth}{@{\extracolsep{\fill}}lrrr@{}}
      \toprule
      Variant & AP & AP$_{\mathrm{MSSD}}$ & AP$_{\mathrm{MSPD}}$ \\
      \midrule
      Synthetic base & \rankcell{2}{44.5} & \rankcell{2}{31.1} & \rankcell{2}{58.0} \\
      Multisensor sup. & \rankcell{1}{57.1} & \rankcell{1}{47.5} & \rankcell{1}{66.8} \\
      \bottomrule
    \end{tabular*}}
    \end{minipage}
    \end{lrbox}
\ifdim\dimexpr\ht\validationentrybox+\dp\validationentrybox\relax>0.30\textheight
  \resizebox*{!}{0.30\textheight}{\usebox{\validationentrybox}}
\else
  \usebox{\validationentrybox}
\fi
\end{minipage}
\hfill%
\begin{minipage}[t][0.30\textheight][t]{0.32\textwidth}
\vspace{0pt}\centering
\begin{lrbox}{\validationentrybox}
    \begin{minipage}{\linewidth}
    \centering
    \objectlabel{T-LESS}{05}{tless}
    {\scriptsize 194 evaluated instances}\par\vspace{2pt}
    \includegraphics[width=\linewidth,keepaspectratio]{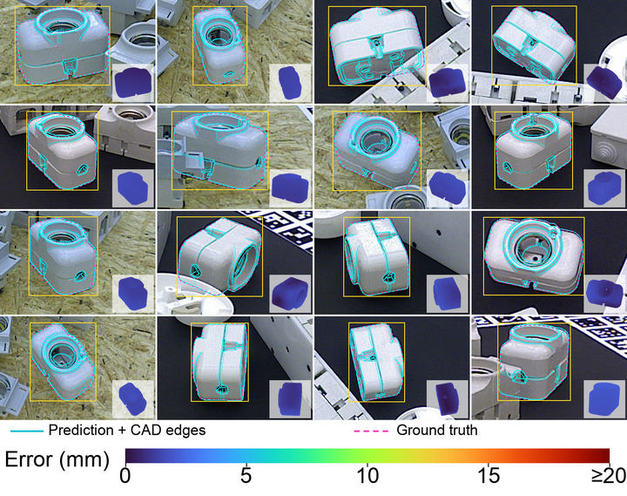}
    \par\vspace{2pt}
    {\fontsize{6.3}{7.0}\selectfont
    \renewcommand{\arraystretch}{0.95}
    \setlength{\tabcolsep}{1.6pt}
    \begin{tabular*}{\linewidth}{@{\extracolsep{\fill}}lrrr@{}}
      \toprule
      Variant & AP & AP$_{\mathrm{MSSD}}$ & AP$_{\mathrm{MSPD}}$ \\
      \midrule
      Synthetic base & \rankcell{2}{88.5} & \rankcell{2}{80.2} & \rankcell{2}{96.8} \\
      Multisensor sup. & \rankcell{1}{93.5} & \rankcell{1}{88.0} & \rankcell{1}{99.0} \\
      \bottomrule
    \end{tabular*}}
    \end{minipage}
    \end{lrbox}
\ifdim\dimexpr\ht\validationentrybox+\dp\validationentrybox\relax>0.30\textheight
  \resizebox*{!}{0.30\textheight}{\usebox{\validationentrybox}}
\else
  \usebox{\validationentrybox}
\fi
\end{minipage}
\hfill%
\begin{minipage}[t][0.30\textheight][t]{0.32\textwidth}
\vspace{0pt}\centering
\begin{lrbox}{\validationentrybox}
    \begin{minipage}{\linewidth}
    \centering
    \objectlabel{T-LESS}{06}{tless}
    {\scriptsize 100 evaluated instances}\par\vspace{2pt}
    \includegraphics[width=\linewidth,keepaspectratio]{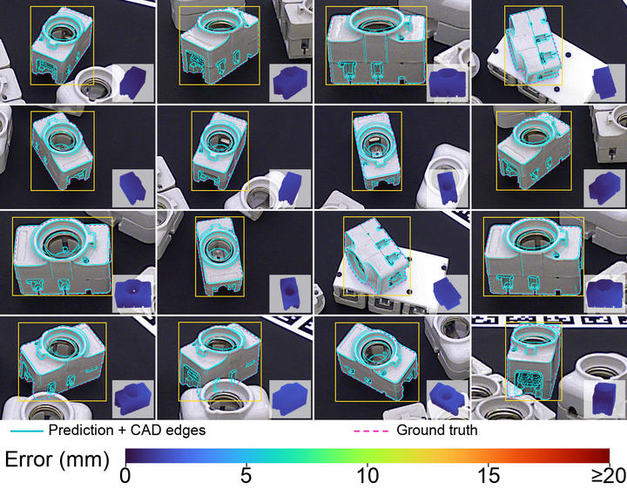}
    \par\vspace{2pt}
    {\fontsize{6.3}{7.0}\selectfont
    \renewcommand{\arraystretch}{0.95}
    \setlength{\tabcolsep}{1.6pt}
    \begin{tabular*}{\linewidth}{@{\extracolsep{\fill}}lrrr@{}}
      \toprule
      Variant & AP & AP$_{\mathrm{MSSD}}$ & AP$_{\mathrm{MSPD}}$ \\
      \midrule
      Synthetic base & \rankcell{2}{86.5} & \rankcell{2}{79.1} & \rankcell{2}{93.9} \\
      Multisensor sup. & \rankcell{1}{94.5} & \rankcell{1}{91.1} & \rankcell{1}{97.8} \\
      \bottomrule
    \end{tabular*}}
    \end{minipage}
    \end{lrbox}
\ifdim\dimexpr\ht\validationentrybox+\dp\validationentrybox\relax>0.30\textheight
  \resizebox*{!}{0.30\textheight}{\usebox{\validationentrybox}}
\else
  \usebox{\validationentrybox}
\fi
\end{minipage}
\par\vspace{0.012\textheight}\noindent
\begin{minipage}[t][0.30\textheight][t]{0.32\textwidth}
\vspace{0pt}\centering
\begin{lrbox}{\validationentrybox}
    \begin{minipage}{\linewidth}
    \centering
    \objectlabel{T-LESS}{07}{tless}
    {\scriptsize 250 evaluated instances}\par\vspace{2pt}
    \includegraphics[width=\linewidth,keepaspectratio]{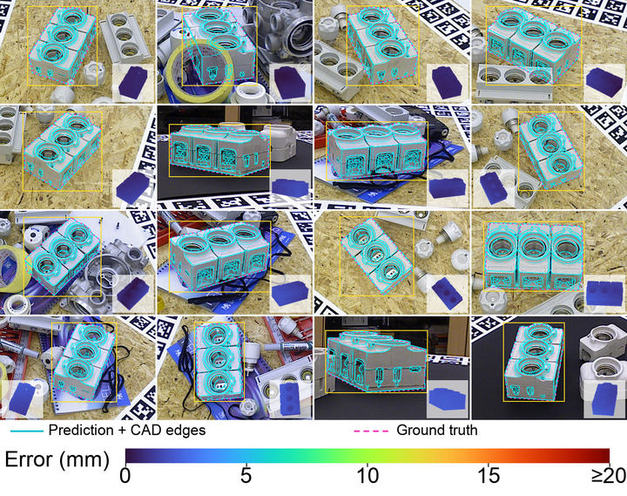}
    \par\vspace{2pt}
    {\fontsize{6.3}{7.0}\selectfont
    \renewcommand{\arraystretch}{0.95}
    \setlength{\tabcolsep}{1.6pt}
    \begin{tabular*}{\linewidth}{@{\extracolsep{\fill}}lrrr@{}}
      \toprule
      Variant & AP & AP$_{\mathrm{MSSD}}$ & AP$_{\mathrm{MSPD}}$ \\
      \midrule
      Synthetic base & \rankcell{2}{59.2} & \rankcell{2}{56.7} & \rankcell{2}{61.7} \\
      Multisensor sup. & \rankcell{1}{94.9} & \rankcell{1}{93.3} & \rankcell{1}{96.6} \\
      \bottomrule
    \end{tabular*}}
    \end{minipage}
    \end{lrbox}
\ifdim\dimexpr\ht\validationentrybox+\dp\validationentrybox\relax>0.30\textheight
  \resizebox*{!}{0.30\textheight}{\usebox{\validationentrybox}}
\else
  \usebox{\validationentrybox}
\fi
\end{minipage}
\hfill%
\begin{minipage}[t][0.30\textheight][t]{0.32\textwidth}
\vspace{0pt}\centering
\begin{lrbox}{\validationentrybox}
    \begin{minipage}{\linewidth}
    \centering
    \objectlabel{T-LESS}{08}{tless}
    {\scriptsize 150 evaluated instances}\par\vspace{2pt}
    \includegraphics[width=\linewidth,keepaspectratio]{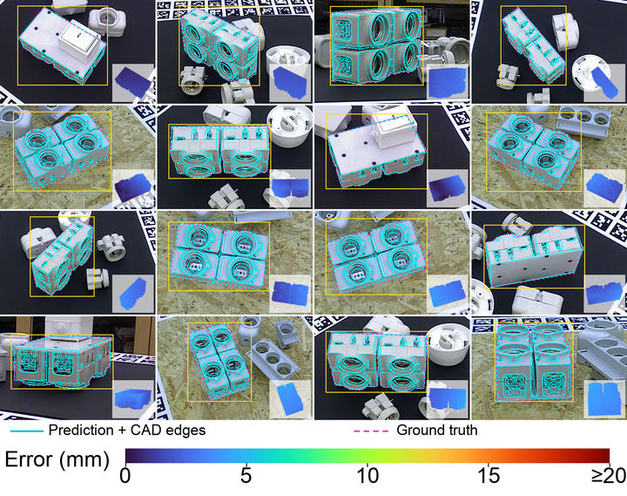}
    \par\vspace{2pt}
    {\fontsize{6.3}{7.0}\selectfont
    \renewcommand{\arraystretch}{0.95}
    \setlength{\tabcolsep}{1.6pt}
    \begin{tabular*}{\linewidth}{@{\extracolsep{\fill}}lrrr@{}}
      \toprule
      Variant & AP & AP$_{\mathrm{MSSD}}$ & AP$_{\mathrm{MSPD}}$ \\
      \midrule
      Synthetic base & \rankcell{2}{80.3} & \rankcell{2}{76.6} & \rankcell{2}{84.1} \\
      Multisensor sup. & \rankcell{1}{94.7} & \rankcell{1}{93.2} & \rankcell{1}{96.3} \\
      \bottomrule
    \end{tabular*}}
    \end{minipage}
    \end{lrbox}
\ifdim\dimexpr\ht\validationentrybox+\dp\validationentrybox\relax>0.30\textheight
  \resizebox*{!}{0.30\textheight}{\usebox{\validationentrybox}}
\else
  \usebox{\validationentrybox}
\fi
\end{minipage}
\hfill%
\begin{minipage}[t][0.30\textheight][t]{0.32\textwidth}
\vspace{0pt}\centering
\begin{lrbox}{\validationentrybox}
    \begin{minipage}{\linewidth}
    \centering
    \objectlabel{T-LESS}{09}{tless}
    {\scriptsize 249 evaluated instances}\par\vspace{2pt}
    \includegraphics[width=\linewidth,keepaspectratio]{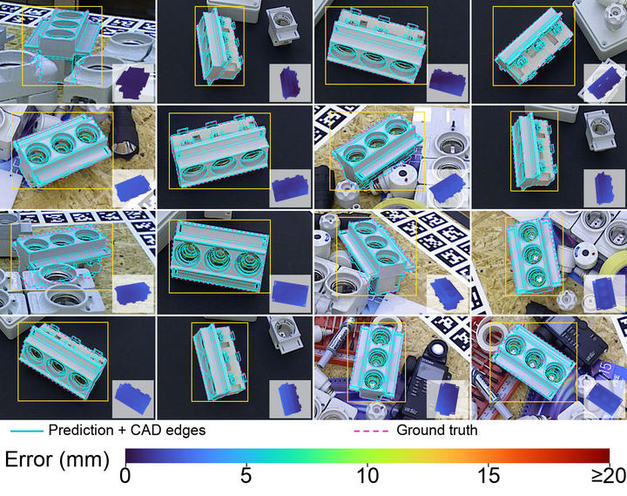}
    \par\vspace{2pt}
    {\fontsize{6.3}{7.0}\selectfont
    \renewcommand{\arraystretch}{0.95}
    \setlength{\tabcolsep}{1.6pt}
    \begin{tabular*}{\linewidth}{@{\extracolsep{\fill}}lrrr@{}}
      \toprule
      Variant & AP & AP$_{\mathrm{MSSD}}$ & AP$_{\mathrm{MSPD}}$ \\
      \midrule
      Synthetic base & \rankcell{2}{89.8} & \rankcell{2}{82.8} & \rankcell{2}{96.7} \\
      Multisensor sup. & \rankcell{1}{93.4} & \rankcell{1}{89.0} & \rankcell{1}{97.9} \\
      \bottomrule
    \end{tabular*}}
    \end{minipage}
    \end{lrbox}
\ifdim\dimexpr\ht\validationentrybox+\dp\validationentrybox\relax>0.30\textheight
  \resizebox*{!}{0.30\textheight}{\usebox{\validationentrybox}}
\else
  \usebox{\validationentrybox}
\fi
\end{minipage}
\par
\newpage
\noindent
\begin{minipage}[t][0.30\textheight][t]{0.32\textwidth}
\vspace{0pt}\centering
\begin{lrbox}{\validationentrybox}
    \begin{minipage}{\linewidth}
    \centering
    \objectlabel{T-LESS}{10}{tless}
    {\scriptsize 145 evaluated instances}\par\vspace{2pt}
    \includegraphics[width=\linewidth,keepaspectratio]{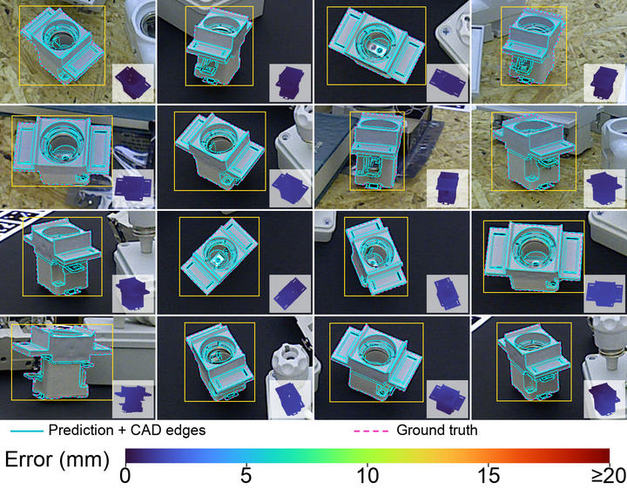}
    \par\vspace{2pt}
    {\fontsize{6.3}{7.0}\selectfont
    \renewcommand{\arraystretch}{0.95}
    \setlength{\tabcolsep}{1.6pt}
    \begin{tabular*}{\linewidth}{@{\extracolsep{\fill}}lrrr@{}}
      \toprule
      Variant & AP & AP$_{\mathrm{MSSD}}$ & AP$_{\mathrm{MSPD}}$ \\
      \midrule
      Synthetic base & \rankcell{2}{91.0} & \rankcell{2}{82.3} & \rankcell{1}{99.6} \\
      Multisensor sup. & \rankcell{1}{91.9} & \rankcell{1}{84.9} & \rankcell{2}{98.8} \\
      \bottomrule
    \end{tabular*}}
    \end{minipage}
    \end{lrbox}
\ifdim\dimexpr\ht\validationentrybox+\dp\validationentrybox\relax>0.30\textheight
  \resizebox*{!}{0.30\textheight}{\usebox{\validationentrybox}}
\else
  \usebox{\validationentrybox}
\fi
\end{minipage}
\hfill%
\begin{minipage}[t][0.30\textheight][t]{0.32\textwidth}
\vspace{0pt}\centering
\begin{lrbox}{\validationentrybox}
    \begin{minipage}{\linewidth}
    \centering
    \objectlabel{T-LESS}{11}{tless}
    {\scriptsize 185 evaluated instances}\par\vspace{2pt}
    \includegraphics[width=\linewidth,keepaspectratio]{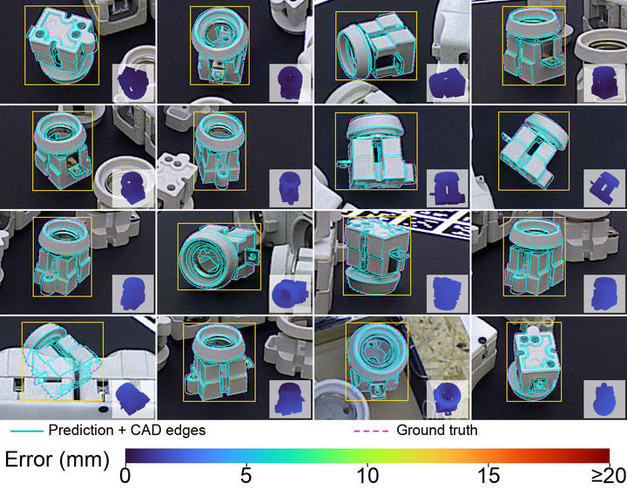}
    \par\vspace{2pt}
    {\fontsize{6.3}{7.0}\selectfont
    \renewcommand{\arraystretch}{0.95}
    \setlength{\tabcolsep}{1.6pt}
    \begin{tabular*}{\linewidth}{@{\extracolsep{\fill}}lrrr@{}}
      \toprule
      Variant & AP & AP$_{\mathrm{MSSD}}$ & AP$_{\mathrm{MSPD}}$ \\
      \midrule
      Synthetic base & \rankcell{2}{75.8} & \rankcell{2}{57.4} & \rankcell{2}{94.2} \\
      Multisensor sup. & \rankcell{1}{86.2} & \rankcell{1}{77.5} & \rankcell{1}{95.0} \\
      \bottomrule
    \end{tabular*}}
    \end{minipage}
    \end{lrbox}
\ifdim\dimexpr\ht\validationentrybox+\dp\validationentrybox\relax>0.30\textheight
  \resizebox*{!}{0.30\textheight}{\usebox{\validationentrybox}}
\else
  \usebox{\validationentrybox}
\fi
\end{minipage}
\hfill%
\begin{minipage}[t][0.30\textheight][t]{0.32\textwidth}
\vspace{0pt}\centering
\begin{lrbox}{\validationentrybox}
    \begin{minipage}{\linewidth}
    \centering
    \objectlabel{T-LESS}{12}{tless}
    {\scriptsize 148 evaluated instances}\par\vspace{2pt}
    \includegraphics[width=\linewidth,keepaspectratio]{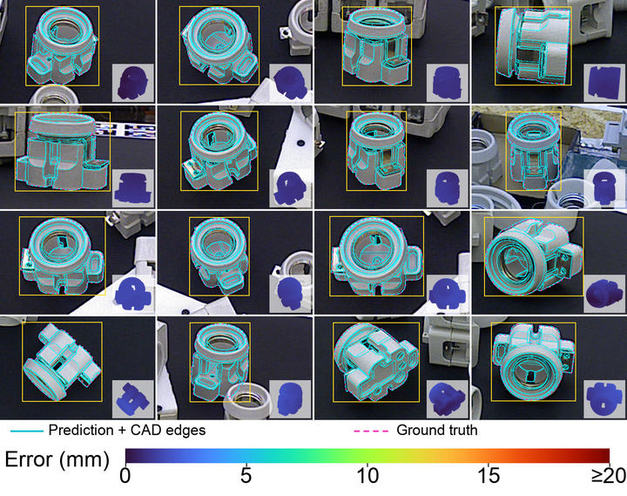}
    \par\vspace{2pt}
    {\fontsize{6.3}{7.0}\selectfont
    \renewcommand{\arraystretch}{0.95}
    \setlength{\tabcolsep}{1.6pt}
    \begin{tabular*}{\linewidth}{@{\extracolsep{\fill}}lrrr@{}}
      \toprule
      Variant & AP & AP$_{\mathrm{MSSD}}$ & AP$_{\mathrm{MSPD}}$ \\
      \midrule
      Synthetic base & \rankcell{2}{83.2} & \rankcell{2}{71.3} & \rankcell{1}{95.0} \\
      Multisensor sup. & \rankcell{1}{87.5} & \rankcell{1}{80.1} & \rankcell{1}{95.0} \\
      \bottomrule
    \end{tabular*}}
    \end{minipage}
    \end{lrbox}
\ifdim\dimexpr\ht\validationentrybox+\dp\validationentrybox\relax>0.30\textheight
  \resizebox*{!}{0.30\textheight}{\usebox{\validationentrybox}}
\else
  \usebox{\validationentrybox}
\fi
\end{minipage}
\par\vspace{0.012\textheight}\noindent
\begin{minipage}[t][0.30\textheight][t]{0.32\textwidth}
\vspace{0pt}\centering
\begin{lrbox}{\validationentrybox}
    \begin{minipage}{\linewidth}
    \centering
    \objectlabel{T-LESS}{13}{tless}
    {\scriptsize 148 evaluated instances}\par\vspace{2pt}
    \includegraphics[width=\linewidth,keepaspectratio]{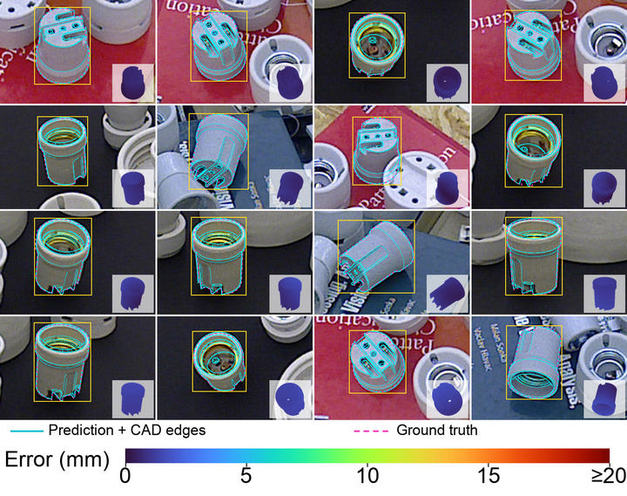}
    \par\vspace{2pt}
    {\fontsize{6.3}{7.0}\selectfont
    \renewcommand{\arraystretch}{0.95}
    \setlength{\tabcolsep}{1.6pt}
    \begin{tabular*}{\linewidth}{@{\extracolsep{\fill}}lrrr@{}}
      \toprule
      Variant & AP & AP$_{\mathrm{MSSD}}$ & AP$_{\mathrm{MSPD}}$ \\
      \midrule
      Synthetic base & \rankcell{2}{69.3} & \rankcell{2}{51.8} & \rankcell{2}{86.8} \\
      Multisensor sup. & \rankcell{1}{76.4} & \rankcell{1}{63.5} & \rankcell{1}{89.3} \\
      \bottomrule
    \end{tabular*}}
    \end{minipage}
    \end{lrbox}
\ifdim\dimexpr\ht\validationentrybox+\dp\validationentrybox\relax>0.30\textheight
  \resizebox*{!}{0.30\textheight}{\usebox{\validationentrybox}}
\else
  \usebox{\validationentrybox}
\fi
\end{minipage}
\hfill%
\begin{minipage}[t][0.30\textheight][t]{0.32\textwidth}
\vspace{0pt}\centering
\begin{lrbox}{\validationentrybox}
    \begin{minipage}{\linewidth}
    \centering
    \objectlabel{T-LESS}{14}{tless}
    {\scriptsize 150 evaluated instances}\par\vspace{2pt}
    \includegraphics[width=\linewidth,keepaspectratio]{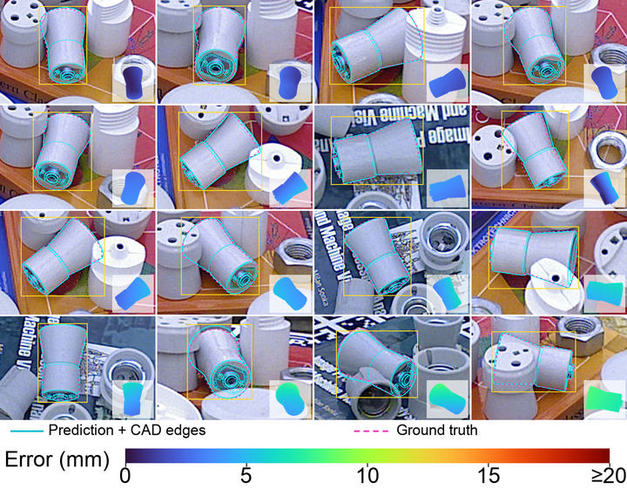}
    \par\vspace{2pt}
    {\fontsize{6.3}{7.0}\selectfont
    \renewcommand{\arraystretch}{0.95}
    \setlength{\tabcolsep}{1.6pt}
    \begin{tabular*}{\linewidth}{@{\extracolsep{\fill}}lrrr@{}}
      \toprule
      Variant & AP & AP$_{\mathrm{MSSD}}$ & AP$_{\mathrm{MSPD}}$ \\
      \midrule
      Synthetic base & \rankcell{2}{22.1} & \rankcell{2}{14.6} & \rankcell{2}{29.5} \\
      Multisensor sup. & \rankcell{1}{24.5} & \rankcell{1}{18.2} & \rankcell{1}{30.8} \\
      \bottomrule
    \end{tabular*}}
    \end{minipage}
    \end{lrbox}
\ifdim\dimexpr\ht\validationentrybox+\dp\validationentrybox\relax>0.30\textheight
  \resizebox*{!}{0.30\textheight}{\usebox{\validationentrybox}}
\else
  \usebox{\validationentrybox}
\fi
\end{minipage}
\hfill%
\begin{minipage}[t][0.30\textheight][t]{0.32\textwidth}
\vspace{0pt}\centering
\begin{lrbox}{\validationentrybox}
    \begin{minipage}{\linewidth}
    \centering
    \objectlabel{T-LESS}{15}{tless}
    {\scriptsize 148 evaluated instances}\par\vspace{2pt}
    \includegraphics[width=\linewidth,keepaspectratio]{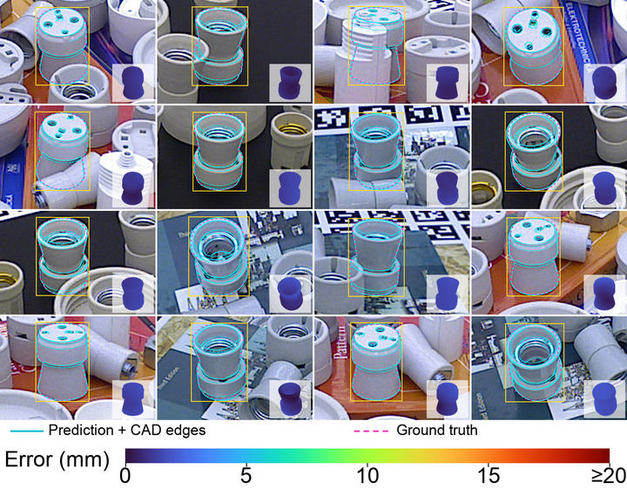}
    \par\vspace{2pt}
    {\fontsize{6.3}{7.0}\selectfont
    \renewcommand{\arraystretch}{0.95}
    \setlength{\tabcolsep}{1.6pt}
    \begin{tabular*}{\linewidth}{@{\extracolsep{\fill}}lrrr@{}}
      \toprule
      Variant & AP & AP$_{\mathrm{MSSD}}$ & AP$_{\mathrm{MSPD}}$ \\
      \midrule
      Synthetic base & \rankcell{2}{61.9} & \rankcell{2}{53.5} & \rankcell{2}{70.4} \\
      Multisensor sup. & \rankcell{1}{70.3} & \rankcell{1}{63.2} & \rankcell{1}{77.3} \\
      \bottomrule
    \end{tabular*}}
    \end{minipage}
    \end{lrbox}
\ifdim\dimexpr\ht\validationentrybox+\dp\validationentrybox\relax>0.30\textheight
  \resizebox*{!}{0.30\textheight}{\usebox{\validationentrybox}}
\else
  \usebox{\validationentrybox}
\fi
\end{minipage}
\par\vspace{0.012\textheight}\noindent
\begin{minipage}[t][0.30\textheight][t]{0.32\textwidth}
\vspace{0pt}\centering
\begin{lrbox}{\validationentrybox}
    \begin{minipage}{\linewidth}
    \centering
    \objectlabel{T-LESS}{16}{tless}
    {\scriptsize 198 evaluated instances}\par\vspace{2pt}
    \includegraphics[width=\linewidth,keepaspectratio]{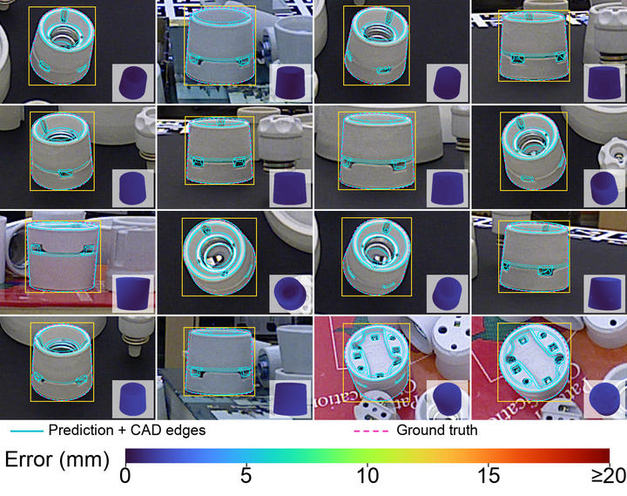}
    \par\vspace{2pt}
    {\fontsize{6.3}{7.0}\selectfont
    \renewcommand{\arraystretch}{0.95}
    \setlength{\tabcolsep}{1.6pt}
    \begin{tabular*}{\linewidth}{@{\extracolsep{\fill}}lrrr@{}}
      \toprule
      Variant & AP & AP$_{\mathrm{MSSD}}$ & AP$_{\mathrm{MSPD}}$ \\
      \midrule
      Synthetic base & \rankcell{2}{78.8} & \rankcell{2}{66.8} & \rankcell{2}{90.8} \\
      Multisensor sup. & \rankcell{1}{84.0} & \rankcell{1}{74.5} & \rankcell{1}{93.5} \\
      \bottomrule
    \end{tabular*}}
    \end{minipage}
    \end{lrbox}
\ifdim\dimexpr\ht\validationentrybox+\dp\validationentrybox\relax>0.30\textheight
  \resizebox*{!}{0.30\textheight}{\usebox{\validationentrybox}}
\else
  \usebox{\validationentrybox}
\fi
\end{minipage}
\hfill%
\begin{minipage}[t][0.30\textheight][t]{0.32\textwidth}
\vspace{0pt}\centering
\begin{lrbox}{\validationentrybox}
    \begin{minipage}{\linewidth}
    \centering
    \objectlabel{T-LESS}{17}{tless}
    {\scriptsize 150 evaluated instances}\par\vspace{2pt}
    \includegraphics[width=\linewidth,keepaspectratio]{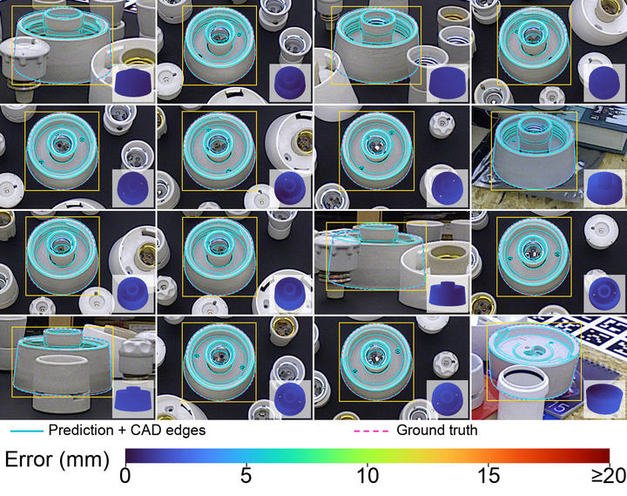}
    \par\vspace{2pt}
    {\fontsize{6.3}{7.0}\selectfont
    \renewcommand{\arraystretch}{0.95}
    \setlength{\tabcolsep}{1.6pt}
    \begin{tabular*}{\linewidth}{@{\extracolsep{\fill}}lrrr@{}}
      \toprule
      Variant & AP & AP$_{\mathrm{MSSD}}$ & AP$_{\mathrm{MSPD}}$ \\
      \midrule
      Synthetic base & \rankcell{2}{89.3} & \rankcell{2}{81.4} & \rankcell{2}{97.2} \\
      Multisensor sup. & \rankcell{1}{92.7} & \rankcell{1}{87.8} & \rankcell{1}{97.7} \\
      \bottomrule
    \end{tabular*}}
    \end{minipage}
    \end{lrbox}
\ifdim\dimexpr\ht\validationentrybox+\dp\validationentrybox\relax>0.30\textheight
  \resizebox*{!}{0.30\textheight}{\usebox{\validationentrybox}}
\else
  \usebox{\validationentrybox}
\fi
\end{minipage}
\hfill%
\begin{minipage}[t][0.30\textheight][t]{0.32\textwidth}
\vspace{0pt}\centering
\begin{lrbox}{\validationentrybox}
    \begin{minipage}{\linewidth}
    \centering
    \objectlabel{T-LESS}{18}{tless}
    {\scriptsize 149 evaluated instances}\par\vspace{2pt}
    \includegraphics[width=\linewidth,keepaspectratio]{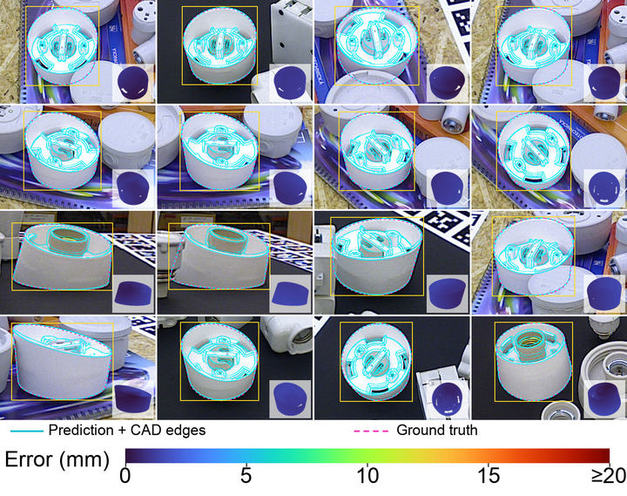}
    \par\vspace{2pt}
    {\fontsize{6.3}{7.0}\selectfont
    \renewcommand{\arraystretch}{0.95}
    \setlength{\tabcolsep}{1.6pt}
    \begin{tabular*}{\linewidth}{@{\extracolsep{\fill}}lrrr@{}}
      \toprule
      Variant & AP & AP$_{\mathrm{MSSD}}$ & AP$_{\mathrm{MSPD}}$ \\
      \midrule
      Synthetic base & \rankcell{2}{86.8} & \rankcell{2}{80.4} & \rankcell{2}{93.3} \\
      Multisensor sup. & \rankcell{1}{93.0} & \rankcell{1}{90.3} & \rankcell{1}{95.6} \\
      \bottomrule
    \end{tabular*}}
    \end{minipage}
    \end{lrbox}
\ifdim\dimexpr\ht\validationentrybox+\dp\validationentrybox\relax>0.30\textheight
  \resizebox*{!}{0.30\textheight}{\usebox{\validationentrybox}}
\else
  \usebox{\validationentrybox}
\fi
\end{minipage}
\par
\newpage
\noindent
\begin{minipage}[t][0.30\textheight][t]{0.32\textwidth}
\vspace{0pt}\centering
\begin{lrbox}{\validationentrybox}
    \begin{minipage}{\linewidth}
    \centering
    \objectlabel{T-LESS}{19}{tless}
    {\scriptsize 198 evaluated instances}\par\vspace{2pt}
    \includegraphics[width=\linewidth,keepaspectratio]{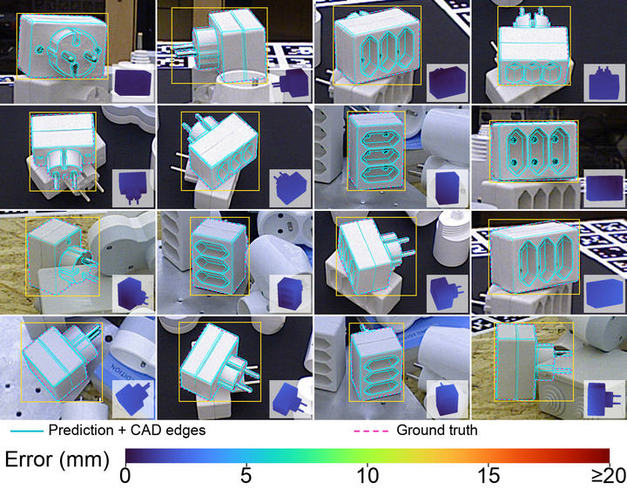}
    \par\vspace{2pt}
    {\fontsize{6.3}{7.0}\selectfont
    \renewcommand{\arraystretch}{0.95}
    \setlength{\tabcolsep}{1.6pt}
    \begin{tabular*}{\linewidth}{@{\extracolsep{\fill}}lrrr@{}}
      \toprule
      Variant & AP & AP$_{\mathrm{MSSD}}$ & AP$_{\mathrm{MSPD}}$ \\
      \midrule
      Synthetic base & \rankcell{2}{80.1} & \rankcell{2}{68.6} & \rankcell{1}{91.6} \\
      Multisensor sup. & \rankcell{1}{81.9} & \rankcell{1}{72.3} & \rankcell{1}{91.6} \\
      \bottomrule
    \end{tabular*}}
    \end{minipage}
    \end{lrbox}
\ifdim\dimexpr\ht\validationentrybox+\dp\validationentrybox\relax>0.30\textheight
  \resizebox*{!}{0.30\textheight}{\usebox{\validationentrybox}}
\else
  \usebox{\validationentrybox}
\fi
\end{minipage}
\hfill%
\begin{minipage}[t][0.30\textheight][t]{0.32\textwidth}
\vspace{0pt}\centering
\begin{lrbox}{\validationentrybox}
    \begin{minipage}{\linewidth}
    \centering
    \objectlabel{T-LESS}{20}{tless}
    {\scriptsize 245 evaluated instances}\par\vspace{2pt}
    \includegraphics[width=\linewidth,keepaspectratio]{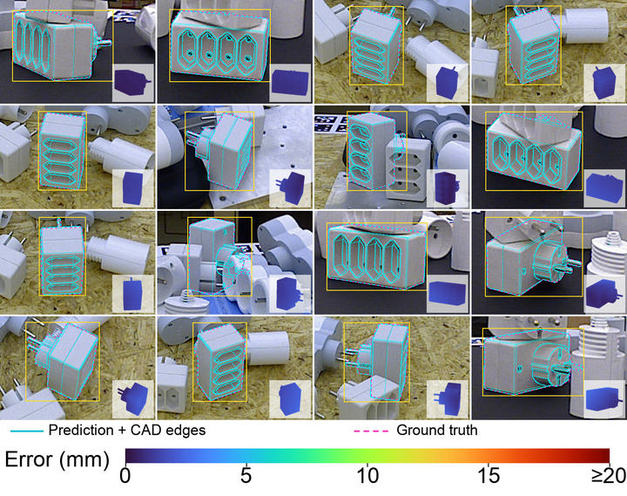}
    \par\vspace{2pt}
    {\fontsize{6.3}{7.0}\selectfont
    \renewcommand{\arraystretch}{0.95}
    \setlength{\tabcolsep}{1.6pt}
    \begin{tabular*}{\linewidth}{@{\extracolsep{\fill}}lrrr@{}}
      \toprule
      Variant & AP & AP$_{\mathrm{MSSD}}$ & AP$_{\mathrm{MSPD}}$ \\
      \midrule
      Synthetic base & \rankcell{2}{79.0} & \rankcell{2}{69.4} & \rankcell{2}{88.5} \\
      Multisensor sup. & \rankcell{1}{80.8} & \rankcell{1}{72.2} & \rankcell{1}{89.4} \\
      \bottomrule
    \end{tabular*}}
    \end{minipage}
    \end{lrbox}
\ifdim\dimexpr\ht\validationentrybox+\dp\validationentrybox\relax>0.30\textheight
  \resizebox*{!}{0.30\textheight}{\usebox{\validationentrybox}}
\else
  \usebox{\validationentrybox}
\fi
\end{minipage}
\hfill%
\begin{minipage}[t][0.30\textheight][t]{0.32\textwidth}
\vspace{0pt}\centering
\begin{lrbox}{\validationentrybox}
    \begin{minipage}{\linewidth}
    \centering
    \objectlabel{T-LESS}{21}{tless}
    {\scriptsize 184 evaluated instances}\par\vspace{2pt}
    \includegraphics[width=\linewidth,keepaspectratio]{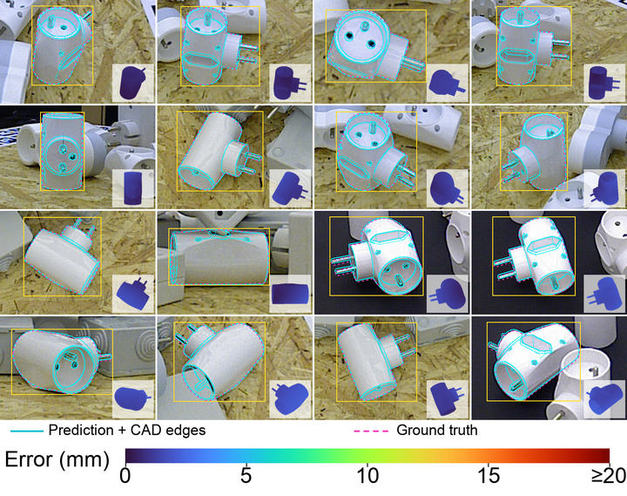}
    \par\vspace{2pt}
    {\fontsize{6.3}{7.0}\selectfont
    \renewcommand{\arraystretch}{0.95}
    \setlength{\tabcolsep}{1.6pt}
    \begin{tabular*}{\linewidth}{@{\extracolsep{\fill}}lrrr@{}}
      \toprule
      Variant & AP & AP$_{\mathrm{MSSD}}$ & AP$_{\mathrm{MSPD}}$ \\
      \midrule
      Synthetic base & \rankcell{2}{67.2} & \rankcell{2}{59.1} & \rankcell{2}{75.2} \\
      Multisensor sup. & \rankcell{1}{75.9} & \rankcell{1}{69.3} & \rankcell{1}{82.6} \\
      \bottomrule
    \end{tabular*}}
    \end{minipage}
    \end{lrbox}
\ifdim\dimexpr\ht\validationentrybox+\dp\validationentrybox\relax>0.30\textheight
  \resizebox*{!}{0.30\textheight}{\usebox{\validationentrybox}}
\else
  \usebox{\validationentrybox}
\fi
\end{minipage}
\par\vspace{0.012\textheight}\noindent
\begin{minipage}[t][0.30\textheight][t]{0.32\textwidth}
\vspace{0pt}\centering
\begin{lrbox}{\validationentrybox}
    \begin{minipage}{\linewidth}
    \centering
    \objectlabel{T-LESS}{22}{tless}
    {\scriptsize 195 evaluated instances}\par\vspace{2pt}
    \includegraphics[width=\linewidth,keepaspectratio]{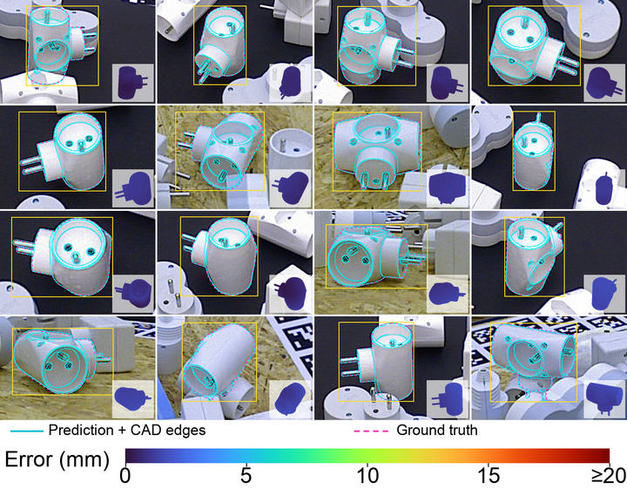}
    \par\vspace{2pt}
    {\fontsize{6.3}{7.0}\selectfont
    \renewcommand{\arraystretch}{0.95}
    \setlength{\tabcolsep}{1.6pt}
    \begin{tabular*}{\linewidth}{@{\extracolsep{\fill}}lrrr@{}}
      \toprule
      Variant & AP & AP$_{\mathrm{MSSD}}$ & AP$_{\mathrm{MSPD}}$ \\
      \midrule
      Synthetic base & \rankcell{2}{59.5} & \rankcell{2}{51.8} & \rankcell{2}{67.2} \\
      Multisensor sup. & \rankcell{1}{63.1} & \rankcell{1}{57.7} & \rankcell{1}{68.5} \\
      \bottomrule
    \end{tabular*}}
    \end{minipage}
    \end{lrbox}
\ifdim\dimexpr\ht\validationentrybox+\dp\validationentrybox\relax>0.30\textheight
  \resizebox*{!}{0.30\textheight}{\usebox{\validationentrybox}}
\else
  \usebox{\validationentrybox}
\fi
\end{minipage}
\hfill%
\begin{minipage}[t][0.30\textheight][t]{0.32\textwidth}
\vspace{0pt}\centering
\begin{lrbox}{\validationentrybox}
    \begin{minipage}{\linewidth}
    \centering
    \objectlabel{T-LESS}{23}{tless}
    {\scriptsize 250 evaluated instances}\par\vspace{2pt}
    \includegraphics[width=\linewidth,keepaspectratio]{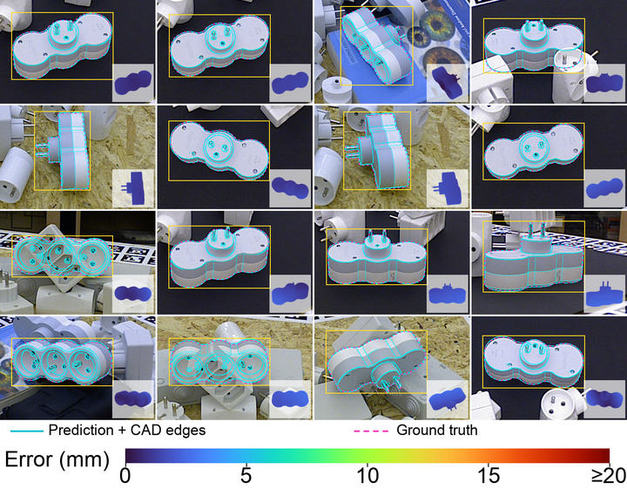}
    \par\vspace{2pt}
    {\fontsize{6.3}{7.0}\selectfont
    \renewcommand{\arraystretch}{0.95}
    \setlength{\tabcolsep}{1.6pt}
    \begin{tabular*}{\linewidth}{@{\extracolsep{\fill}}lrrr@{}}
      \toprule
      Variant & AP & AP$_{\mathrm{MSSD}}$ & AP$_{\mathrm{MSPD}}$ \\
      \midrule
      Synthetic base & \rankcell{2}{86.9} & \rankcell{2}{77.3} & \rankcell{2}{96.5} \\
      Multisensor sup. & \rankcell{1}{91.9} & \rankcell{1}{87.2} & \rankcell{1}{96.7} \\
      \bottomrule
    \end{tabular*}}
    \end{minipage}
    \end{lrbox}
\ifdim\dimexpr\ht\validationentrybox+\dp\validationentrybox\relax>0.30\textheight
  \resizebox*{!}{0.30\textheight}{\usebox{\validationentrybox}}
\else
  \usebox{\validationentrybox}
\fi
\end{minipage}
\hfill%
\begin{minipage}[t][0.30\textheight][t]{0.32\textwidth}
\vspace{0pt}\centering
\begin{lrbox}{\validationentrybox}
    \begin{minipage}{\linewidth}
    \centering
    \objectlabel{T-LESS}{24}{tless}
    {\scriptsize 195 evaluated instances}\par\vspace{2pt}
    \includegraphics[width=\linewidth,keepaspectratio]{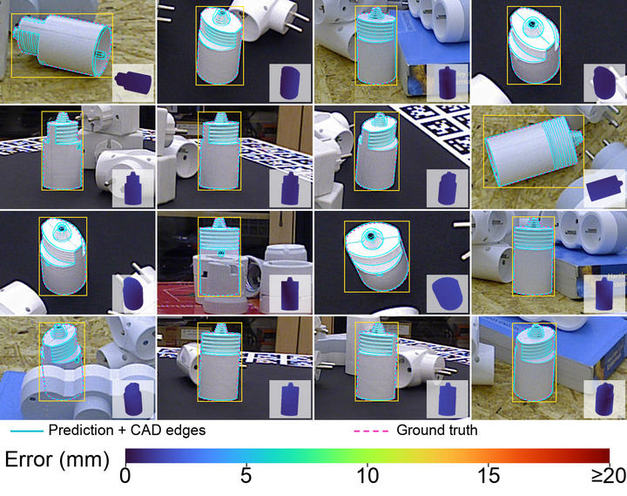}
    \par\vspace{2pt}
    {\fontsize{6.3}{7.0}\selectfont
    \renewcommand{\arraystretch}{0.95}
    \setlength{\tabcolsep}{1.6pt}
    \begin{tabular*}{\linewidth}{@{\extracolsep{\fill}}lrrr@{}}
      \toprule
      Variant & AP & AP$_{\mathrm{MSSD}}$ & AP$_{\mathrm{MSPD}}$ \\
      \midrule
      Synthetic base & \rankcell{2}{90.5} & \rankcell{2}{82.4} & \rankcell{1}{98.6} \\
      Multisensor sup. & \rankcell{1}{91.3} & \rankcell{1}{84.7} & \rankcell{2}{97.8} \\
      \bottomrule
    \end{tabular*}}
    \end{minipage}
    \end{lrbox}
\ifdim\dimexpr\ht\validationentrybox+\dp\validationentrybox\relax>0.30\textheight
  \resizebox*{!}{0.30\textheight}{\usebox{\validationentrybox}}
\else
  \usebox{\validationentrybox}
\fi
\end{minipage}
\par\vspace{0.012\textheight}\noindent
\begin{minipage}[t][0.30\textheight][t]{0.32\textwidth}
\vspace{0pt}\centering
\begin{lrbox}{\validationentrybox}
    \begin{minipage}{\linewidth}
    \centering
    \objectlabel{T-LESS}{25}{tless}
    {\scriptsize 98 evaluated instances}\par\vspace{2pt}
    \includegraphics[width=\linewidth,keepaspectratio]{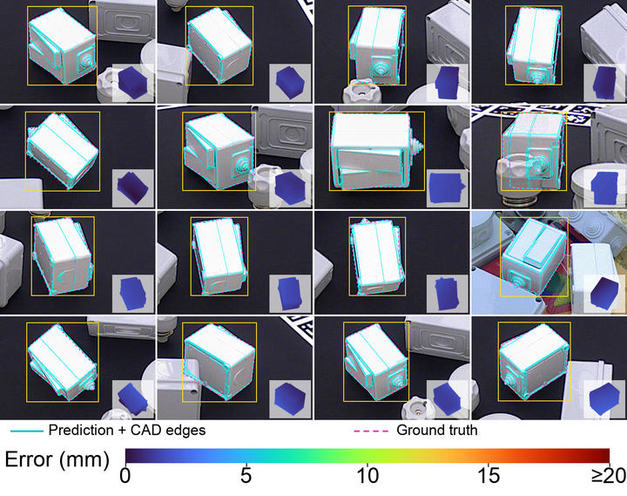}
    \par\vspace{2pt}
    {\fontsize{6.3}{7.0}\selectfont
    \renewcommand{\arraystretch}{0.95}
    \setlength{\tabcolsep}{1.6pt}
    \begin{tabular*}{\linewidth}{@{\extracolsep{\fill}}lrrr@{}}
      \toprule
      Variant & AP & AP$_{\mathrm{MSSD}}$ & AP$_{\mathrm{MSPD}}$ \\
      \midrule
      Synthetic base & \rankcell{2}{89.4} & \rankcell{2}{83.2} & \rankcell{1}{95.6} \\
      Multisensor sup. & \rankcell{1}{90.6} & \rankcell{1}{86.5} & \rankcell{2}{94.7} \\
      \bottomrule
    \end{tabular*}}
    \end{minipage}
    \end{lrbox}
\ifdim\dimexpr\ht\validationentrybox+\dp\validationentrybox\relax>0.30\textheight
  \resizebox*{!}{0.30\textheight}{\usebox{\validationentrybox}}
\else
  \usebox{\validationentrybox}
\fi
\end{minipage}
\hfill%
\begin{minipage}[t][0.30\textheight][t]{0.32\textwidth}
\vspace{0pt}\centering
\begin{lrbox}{\validationentrybox}
    \begin{minipage}{\linewidth}
    \centering
    \objectlabel{T-LESS}{26}{tless}
    {\scriptsize 100 evaluated instances}\par\vspace{2pt}
    \includegraphics[width=\linewidth,keepaspectratio]{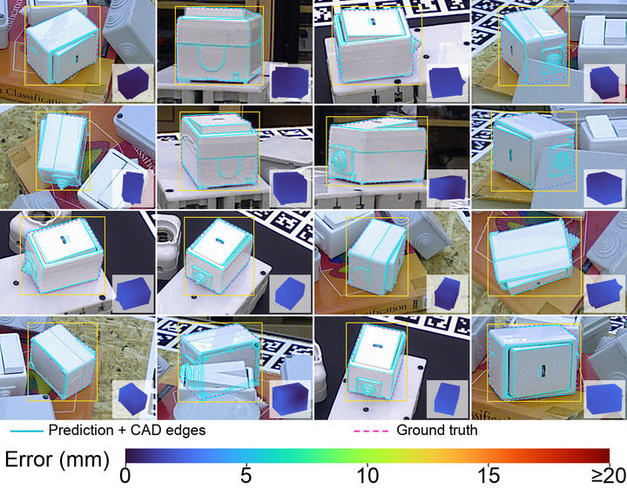}
    \par\vspace{2pt}
    {\fontsize{6.3}{7.0}\selectfont
    \renewcommand{\arraystretch}{0.95}
    \setlength{\tabcolsep}{1.6pt}
    \begin{tabular*}{\linewidth}{@{\extracolsep{\fill}}lrrr@{}}
      \toprule
      Variant & AP & AP$_{\mathrm{MSSD}}$ & AP$_{\mathrm{MSPD}}$ \\
      \midrule
      Synthetic base & \rankcell{1}{94.0} & \rankcell{1}{89.4} & \rankcell{1}{98.7} \\
      Multisensor sup. & \rankcell{2}{92.7} & \rankcell{2}{88.2} & \rankcell{2}{97.3} \\
      \bottomrule
    \end{tabular*}}
    \end{minipage}
    \end{lrbox}
\ifdim\dimexpr\ht\validationentrybox+\dp\validationentrybox\relax>0.30\textheight
  \resizebox*{!}{0.30\textheight}{\usebox{\validationentrybox}}
\else
  \usebox{\validationentrybox}
\fi
\end{minipage}
\hfill%
\begin{minipage}[t][0.30\textheight][t]{0.32\textwidth}
\vspace{0pt}\centering
\begin{lrbox}{\validationentrybox}
    \begin{minipage}{\linewidth}
    \centering
    \objectlabel{T-LESS}{27}{tless}
    {\scriptsize 96 evaluated instances}\par\vspace{2pt}
    \includegraphics[width=\linewidth,keepaspectratio]{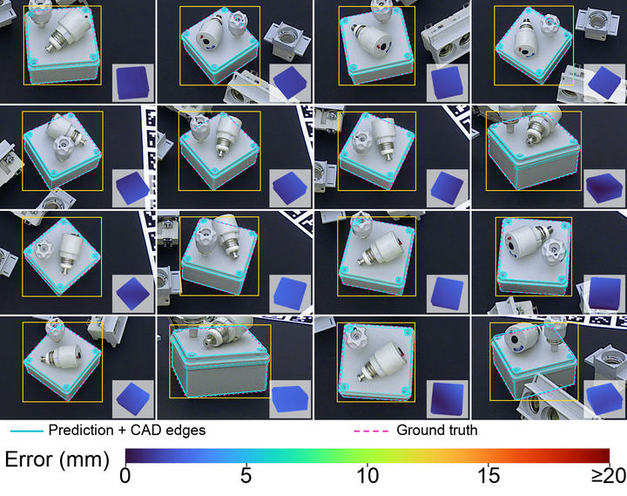}
    \par\vspace{2pt}
    {\fontsize{6.3}{7.0}\selectfont
    \renewcommand{\arraystretch}{0.95}
    \setlength{\tabcolsep}{1.6pt}
    \begin{tabular*}{\linewidth}{@{\extracolsep{\fill}}lrrr@{}}
      \toprule
      Variant & AP & AP$_{\mathrm{MSSD}}$ & AP$_{\mathrm{MSPD}}$ \\
      \midrule
      Synthetic base & \rankcell{2}{83.8} & \rankcell{2}{79.3} & \rankcell{2}{88.4} \\
      Multisensor sup. & \rankcell{1}{87.3} & \rankcell{1}{84.5} & \rankcell{1}{90.2} \\
      \bottomrule
    \end{tabular*}}
    \end{minipage}
    \end{lrbox}
\ifdim\dimexpr\ht\validationentrybox+\dp\validationentrybox\relax>0.30\textheight
  \resizebox*{!}{0.30\textheight}{\usebox{\validationentrybox}}
\else
  \usebox{\validationentrybox}
\fi
\end{minipage}
\par
\newpage
\noindent
\begin{minipage}[t][0.30\textheight][t]{0.32\textwidth}
\vspace{0pt}\centering
\begin{lrbox}{\validationentrybox}
    \begin{minipage}{\linewidth}
    \centering
    \objectlabel{T-LESS}{28}{tless}
    {\scriptsize 195 evaluated instances}\par\vspace{2pt}
    \includegraphics[width=\linewidth,keepaspectratio]{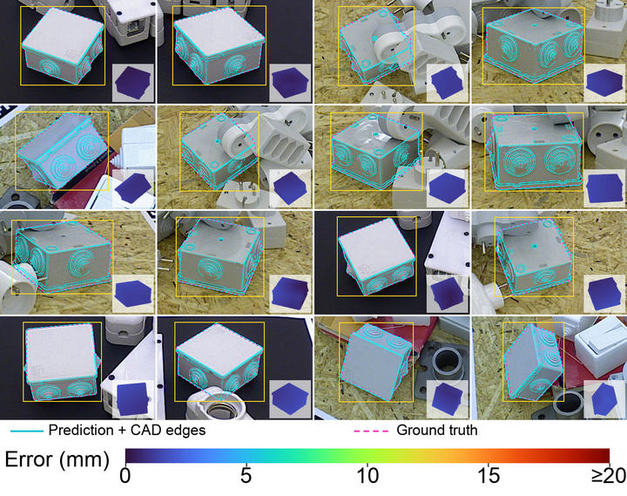}
    \par\vspace{2pt}
    {\fontsize{6.3}{7.0}\selectfont
    \renewcommand{\arraystretch}{0.95}
    \setlength{\tabcolsep}{1.6pt}
    \begin{tabular*}{\linewidth}{@{\extracolsep{\fill}}lrrr@{}}
      \toprule
      Variant & AP & AP$_{\mathrm{MSSD}}$ & AP$_{\mathrm{MSPD}}$ \\
      \midrule
      Synthetic base & \rankcell{2}{88.9} & \rankcell{2}{84.5} & \rankcell{2}{93.4} \\
      Multisensor sup. & \rankcell{1}{92.3} & \rankcell{1}{89.2} & \rankcell{1}{95.5} \\
      \bottomrule
    \end{tabular*}}
    \end{minipage}
    \end{lrbox}
\ifdim\dimexpr\ht\validationentrybox+\dp\validationentrybox\relax>0.30\textheight
  \resizebox*{!}{0.30\textheight}{\usebox{\validationentrybox}}
\else
  \usebox{\validationentrybox}
\fi
\end{minipage}
\hfill%
\begin{minipage}[t][0.30\textheight][t]{0.32\textwidth}
\vspace{0pt}\centering
\begin{lrbox}{\validationentrybox}
    \begin{minipage}{\linewidth}
    \centering
    \objectlabel{T-LESS}{29}{tless}
    {\scriptsize 99 evaluated instances}\par\vspace{2pt}
    \includegraphics[width=\linewidth,keepaspectratio]{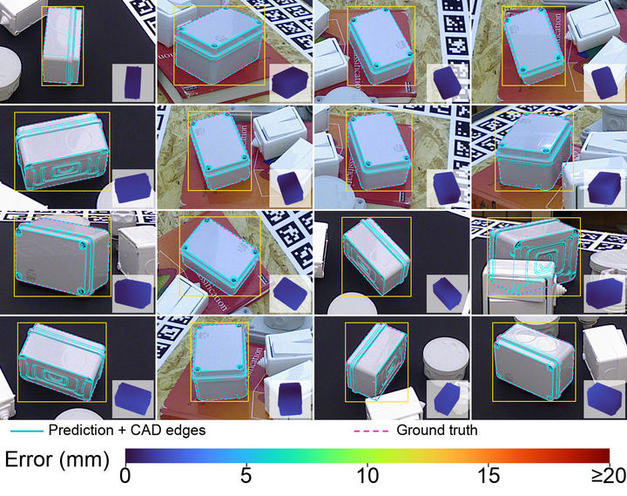}
    \par\vspace{2pt}
    {\fontsize{6.3}{7.0}\selectfont
    \renewcommand{\arraystretch}{0.95}
    \setlength{\tabcolsep}{1.6pt}
    \begin{tabular*}{\linewidth}{@{\extracolsep{\fill}}lrrr@{}}
      \toprule
      Variant & AP & AP$_{\mathrm{MSSD}}$ & AP$_{\mathrm{MSPD}}$ \\
      \midrule
      Synthetic base & \rankcell{2}{89.8} & \rankcell{2}{86.1} & \rankcell{2}{93.5} \\
      Multisensor sup. & \rankcell{1}{95.7} & \rankcell{1}{94.9} & \rankcell{1}{96.5} \\
      \bottomrule
    \end{tabular*}}
    \end{minipage}
    \end{lrbox}
\ifdim\dimexpr\ht\validationentrybox+\dp\validationentrybox\relax>0.30\textheight
  \resizebox*{!}{0.30\textheight}{\usebox{\validationentrybox}}
\else
  \usebox{\validationentrybox}
\fi
\end{minipage}
\hfill%
\begin{minipage}[t][0.30\textheight][t]{0.32\textwidth}
\vspace{0pt}\centering
\begin{lrbox}{\validationentrybox}
    \begin{minipage}{\linewidth}
    \centering
    \objectlabel{T-LESS}{30}{tless}
    {\scriptsize 148 evaluated instances}\par\vspace{2pt}
    \includegraphics[width=\linewidth,keepaspectratio]{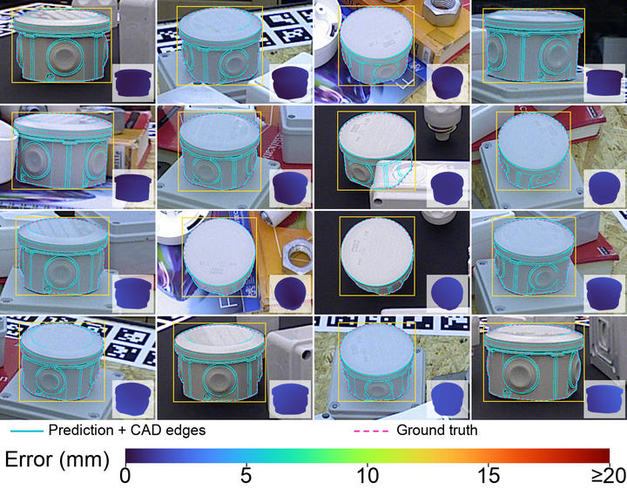}
    \par\vspace{2pt}
    {\fontsize{6.3}{7.0}\selectfont
    \renewcommand{\arraystretch}{0.95}
    \setlength{\tabcolsep}{1.6pt}
    \begin{tabular*}{\linewidth}{@{\extracolsep{\fill}}lrrr@{}}
      \toprule
      Variant & AP & AP$_{\mathrm{MSSD}}$ & AP$_{\mathrm{MSPD}}$ \\
      \midrule
      Synthetic base & \rankcell{2}{69.7} & \rankcell{2}{59.7} & \rankcell{2}{79.7} \\
      Multisensor sup. & \rankcell{1}{80.7} & \rankcell{1}{74.7} & \rankcell{1}{86.7} \\
      \bottomrule
    \end{tabular*}}
    \end{minipage}
    \end{lrbox}
\ifdim\dimexpr\ht\validationentrybox+\dp\validationentrybox\relax>0.30\textheight
  \resizebox*{!}{0.30\textheight}{\usebox{\validationentrybox}}
\else
  \usebox{\validationentrybox}
\fi
\end{minipage}
\par

\par
\twocolumn

\section{Underspecified CAD: per-object results}
\label{sec:appendix_underspecified_cad}

We report the per-object results of the underspecified-CAD experiment in the main paper's underspecified-CAD section.

\noindent\textbf{Training arms.}
For each object and degradation (texture removed, or 20\% of the object removed from one end), we train three doper model from the pretrained DINOv3 ConvNeXt-Small backbone with seeds 42, 43, and 44.
\emph{CAD alone} uses 12{,}500 BlenderProc renders of the degraded CAD model.
\emph{CAD composites} replace half of these renders with composites of the degraded CAD model, rendered onto real HANDAL backgrounds.
\emph{NuRec composites} replace the same half with composites of the NuRec reconstruction at the same poses and on the same backgrounds.

\begin{figure}[!ht]
  \centering
  \includegraphics[width=\columnwidth]{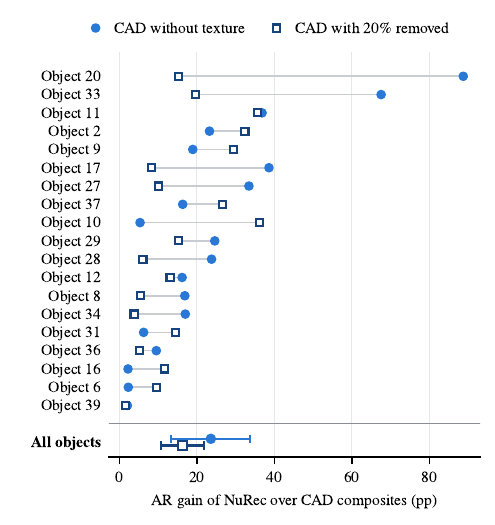}
  \caption{Gain of NuRec composites over CAD composites per HANDAL object (AR, mean of three seeds), with the CAD model untextured (circles) or with 20\% removed (squares). All objects: mean and 95\% CI of the mixed model.}
  \label{fig:appendix_nurec_gain}
\end{figure}

\begin{figure*}[!ht]
  \centering
  \includegraphics[width=0.92\textwidth]{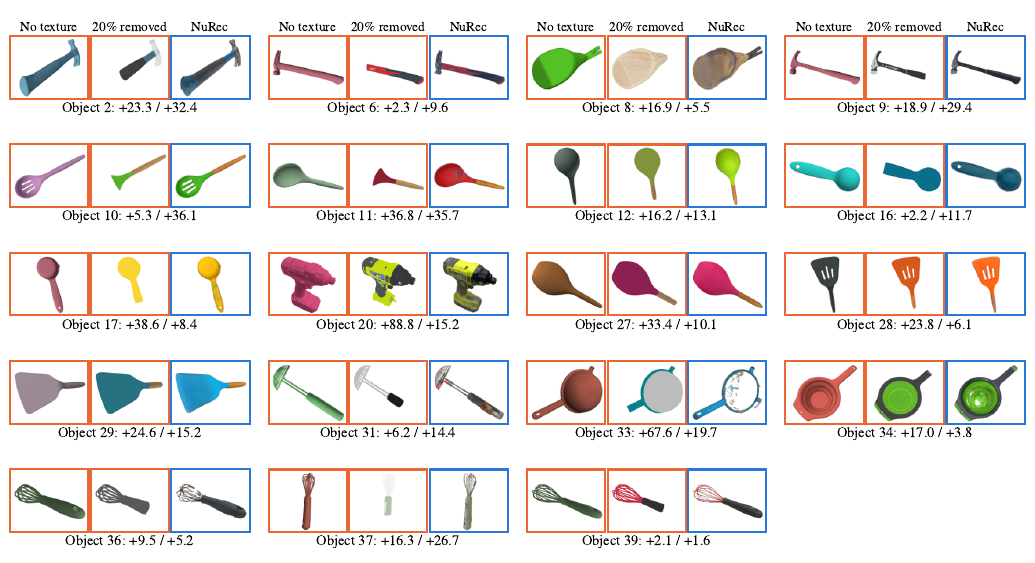}
  \caption{Training objects of the underspecified-CAD experiment for all 19 objects, cut out with each composite's object mask. Each triple shows the same composite frame with the untextured CAD model, with 20\% of the object removed, and with the NuRec reconstruction.
  }
  \label{fig:appendix_nurec_frames}
\end{figure*}

\noindent\textbf{Analysis.}
We score each model on the HANDAL validation images with DOPER detections and report AR, the mean of MSSD and MSPD AR.
The estimated gain is +23.6 [13.4, 33.8] AR without texture and +16.3 [10.8, 21.9] AR with 20\% removed (mean and 95\% CI).
The standard deviation of the object effect is 21.4 and 9.6 AR.
Figure~\ref{fig:appendix_nurec_gain} shows the gain per object, and Figure~\ref{fig:appendix_nurec_frames} shows a training frame for every object.

\raggedbottom
\section{Keypoint count and placement: additional results}
\label{sec:keypoint_placement_details}

\begin{figure}[!ht]
  \centering
  \includegraphics[width=\columnwidth]{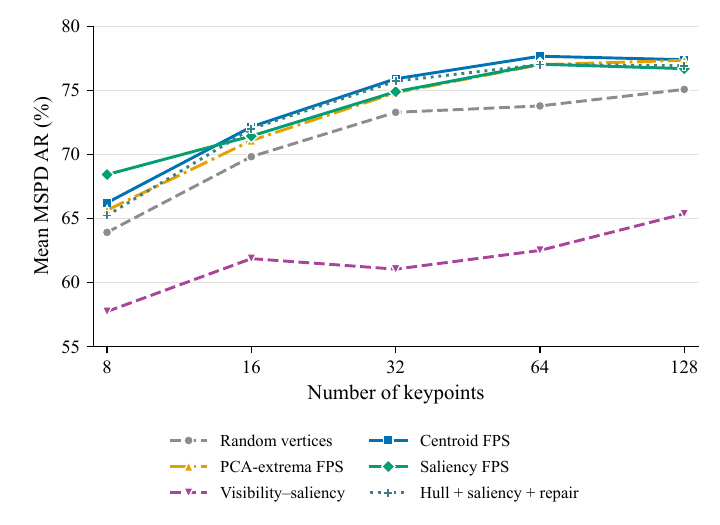}
  \caption{Keypoint placement and count. MSPD AR is averaged equally over HANDAL~8, HOT3D~2, HOPEv2~24, and T-LESS~9, using the same values as the main paper's keypoint-placement table. The first three objects use validation images; T-LESS uses test images. These are selected-object recall scores, not BOP detection AP or the three-metric localization AR.}
  \label{fig:keypoint_placement_curves}
\end{figure}

\begin{figure*}[!tp]
  \centering
  \includegraphics[width=\textwidth,height=0.88\textheight,keepaspectratio]{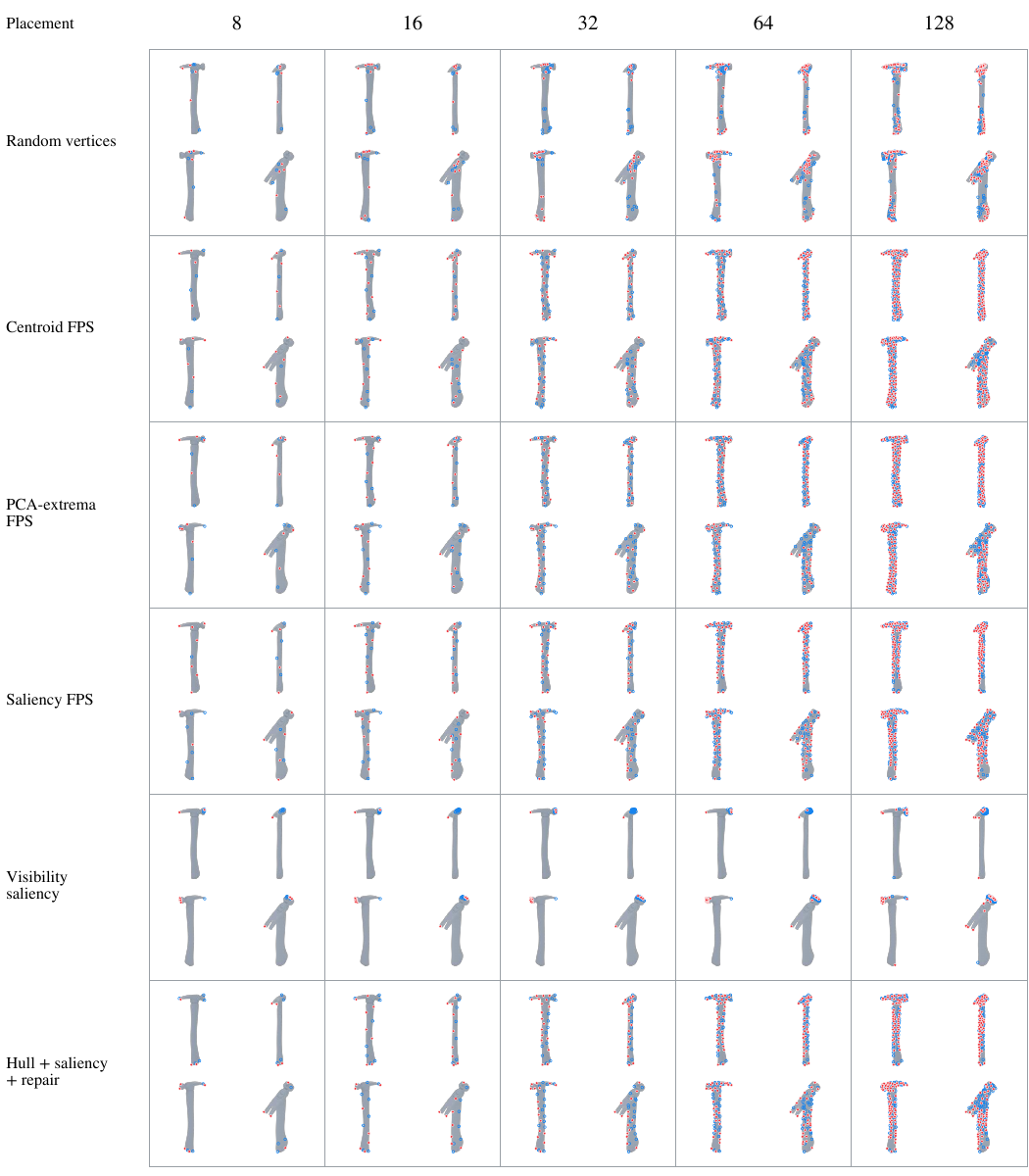}
  \caption{Keypoint count and placement for HANDAL object 8 (Table~\ref{tab:keypoint_placement}). Each method/count entry shows the same saved layout from four views: front and side on the top row, rear and top on the bottom row. Solid red markers indicate visible keypoints; hollow blue rings show self-occluded keypoints as overlays on the opaque CAD model. Column headings give the total 3D keypoint count. Cameras and framing are fixed across layouts, and all image aspect ratios are preserved.}
  \label{fig:keypoint_matrix_handal}
\end{figure*}

\begin{figure*}[!tp]
  \centering
  \includegraphics[width=\textwidth,height=0.88\textheight,keepaspectratio]{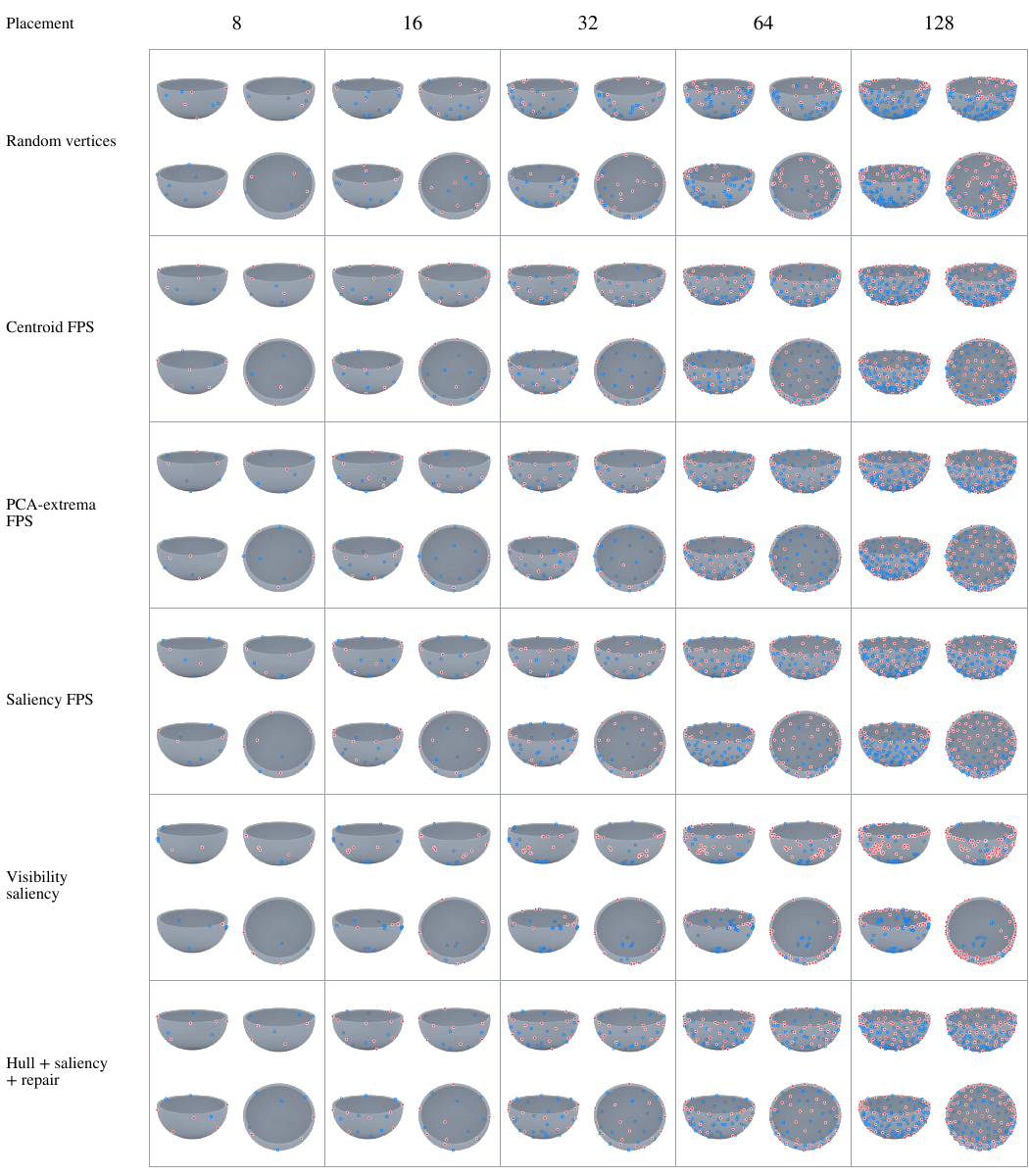}
  \caption{Keypoint count and placement for HOT3D object 2 (Table~\ref{tab:keypoint_placement}). Each method/count entry shows the same saved layout from four views: front and side on the top row, rear and top on the bottom row. Solid red markers indicate visible keypoints; hollow blue rings show self-occluded keypoints as overlays on the opaque CAD model. Column headings give the total 3D keypoint count. Cameras and framing are fixed across layouts, and all image aspect ratios are preserved.}
  \label{fig:keypoint_matrix_hot3d}
\end{figure*}

\begin{figure*}[!tp]
  \centering
  \includegraphics[width=\textwidth,height=0.88\textheight,keepaspectratio]{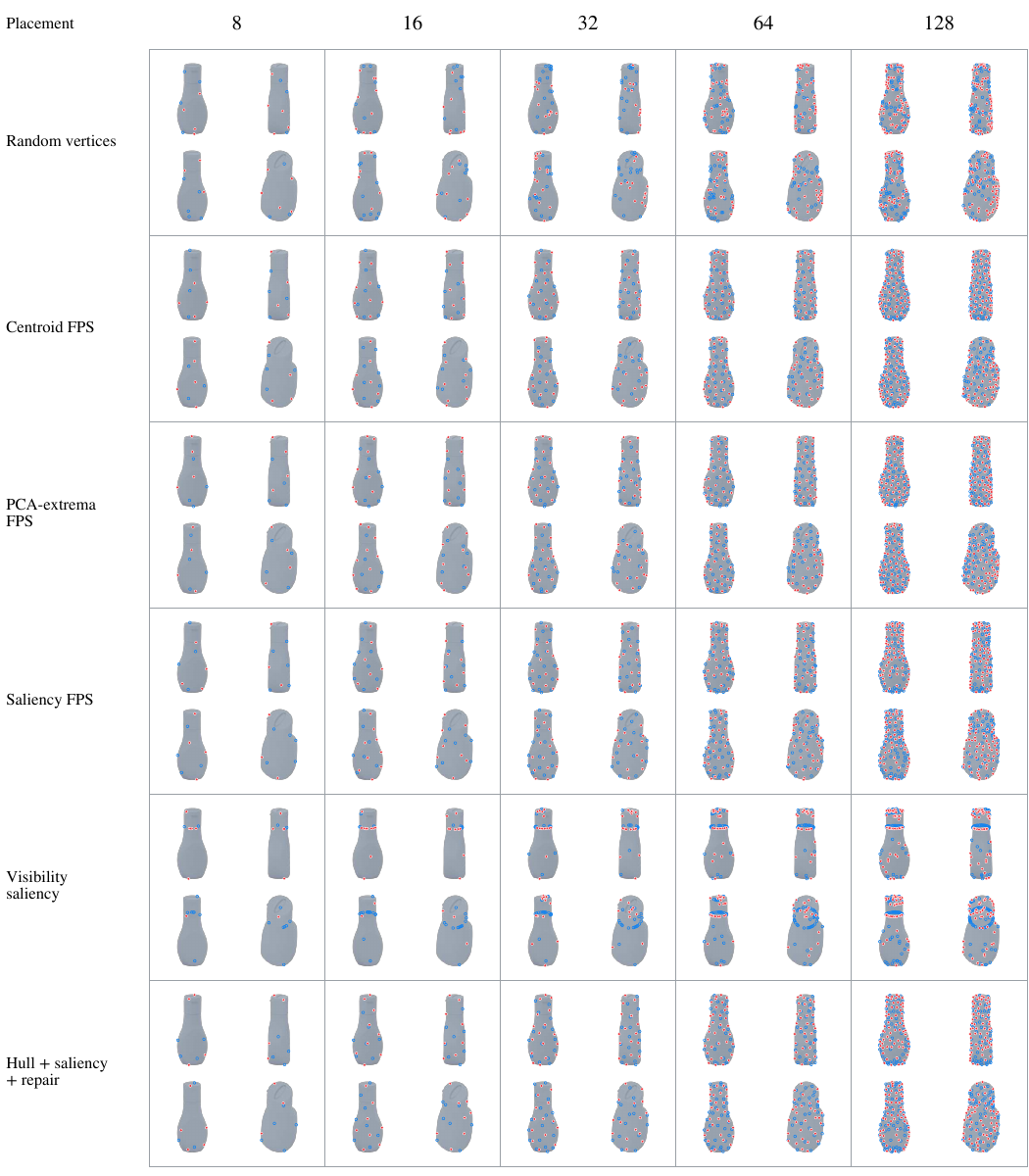}
  \caption{Keypoint count and placement for HOPEv2 object 24 (Table~\ref{tab:keypoint_placement}). Each method/count entry shows the same saved layout from four views: front and side on the top row, rear and top on the bottom row. Solid red markers indicate visible keypoints; hollow blue rings show self-occluded keypoints as overlays on the opaque CAD model. Column headings give the total 3D keypoint count. Cameras and framing are fixed across layouts, and all image aspect ratios are preserved.}
  \label{fig:keypoint_matrix_hope}
\end{figure*}

\begin{figure*}[!tp]
  \centering
  \includegraphics[width=\textwidth,height=0.88\textheight,keepaspectratio]{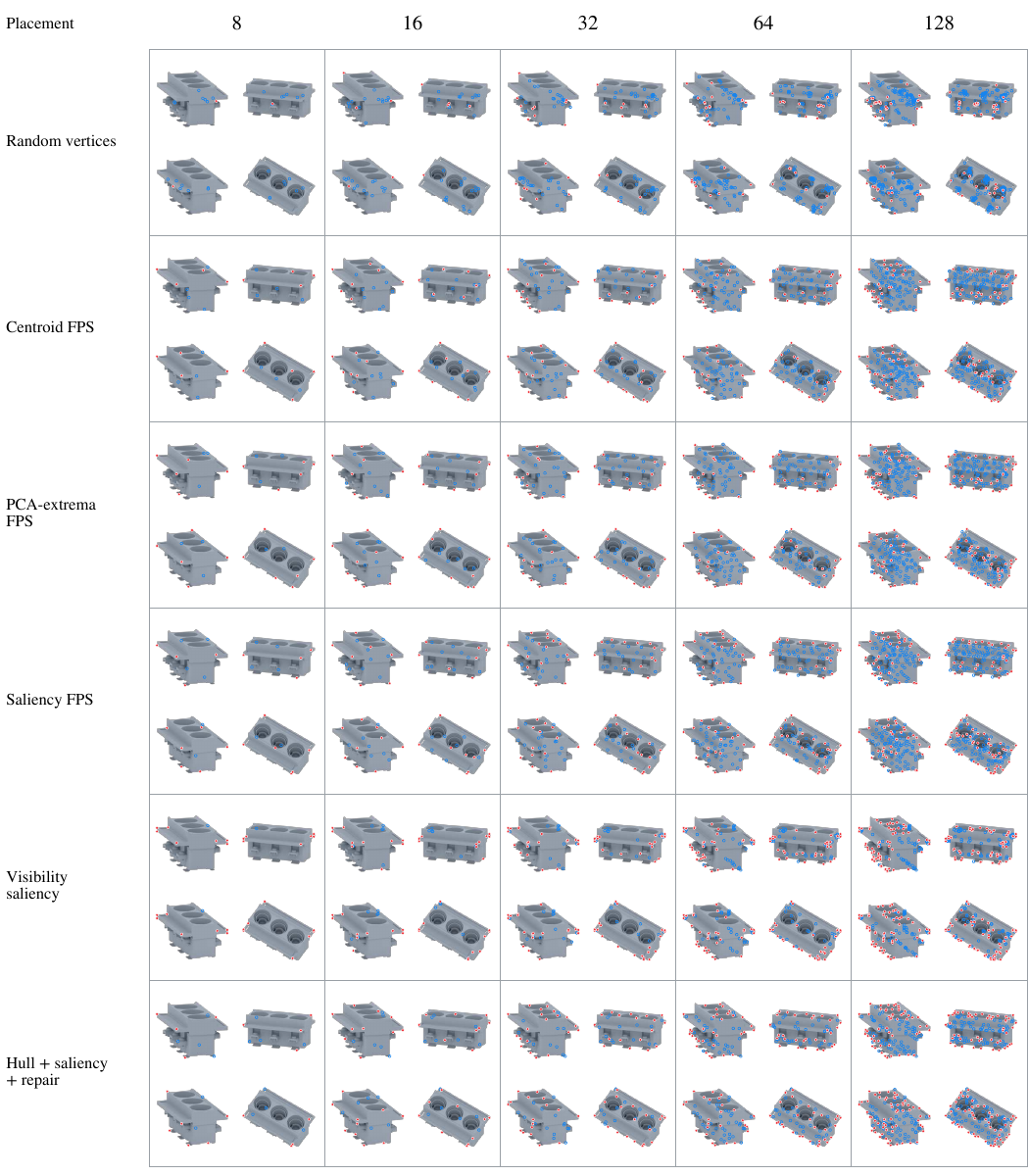}
  \caption{Keypoint count and placement for T-LESS object 9 (Table~\ref{tab:keypoint_placement}). Each method/count entry shows the same saved layout from four views: front and side on the top row, rear and top on the bottom row. Solid red markers indicate visible keypoints; hollow blue rings show self-occluded keypoints as overlays on the opaque CAD model. Column headings give the total 3D keypoint count. Cameras and framing are fixed across layouts, and all image aspect ratios are preserved.}
  \label{fig:keypoint_matrix_tless}
\end{figure*}

The main paper's keypoint-placement section defines the six placement strategies and reports their mean accuracy across HANDAL~8, HOT3D~2, HOPEv2~24, and T-LESS~9.
Figure~\ref{fig:keypoint_placement_curves} plots these means against keypoint count.
Here we give the training protocol and per-object results.
Figures~\ref{fig:keypoint_matrix_handal}--\ref{fig:keypoint_matrix_tless} show each saved layout from four views: front, side, rear, and top oblique.

\noindent\textbf{Training and evaluation.}
All variants use a DINOv3-pretrained ConvNeXt-Tiny backbone and train on the same frozen PBR images and instances for 25 epochs with seed 42.
We retain the RGB images, masks, boxes, poses, calibration, and split membership, changing only the keypoint annotations and head count.
The recipe uses batch size 16 with four-step accumulation, learning rate $5\times10^{-4}$, a backbone learning-rate multiplier of 0.1, 200 warmup steps, 640-pixel detection images, and 512-pixel crops.
We evaluate the final checkpoint with confidence threshold 0.15, at most five detections per image, unchanged P$n$P settings, and no box jitter.
Inference covers all 2,208 HANDAL, 1,285 HOT3D, 50 HOPEv2, and 1,000 T-LESS images, including images without the selected object.
Table~\ref{tab:keypoint_placement} reports symmetry-aware MSPD recall.

\begin{table*}[!t]
  \centering
  \caption{Keypoint placement and count ablation. Entries are detector-based MSPD AR (\%, higher is better); columns give keypoint counts. \ranklegend{} ranks placement strategies within each object and count. T-LESS uses test images; the other objects use real validation images. These are single-object results, not full-dataset AP.}
  \label{tab:keypoint_placement}
  \footnotesize
  \setlength{\tabcolsep}{3pt}
\begin{minipage}[t]{0.485\textwidth}
\vspace{0pt}
\resizebox{\linewidth}{!}{%
\begin{tabular}{@{}l|rrrrr@{}}
\toprule
Placement & 8 & 16 & 32 & 64 & 128 \\
\midrule
    \multicolumn{6}{@{}l}{\textit{HANDAL object 8 (validation; 233 instances)}} \\
    Random vertices & \rankcell{5}{60.64} & \rankcell{5}{68.58} & \rankcell{5}{70.77} & \rankcell{5}{62.32} & \rankcell{5}{67.77} \\
    Centroid FPS & \rankcell{2}{66.65} & \rankcell{2}{72.88} & \rankcell{3}{72.62} & \rankcell{2}{70.30} & \rankcell{3}{69.61} \\
    PCA-extrema FPS & \rankcell{4}{62.40} & \rankcell{4}{69.01} & \rankcell{2}{73.30} & \rankcell{1}{71.07} & \rankcell{1}{71.24} \\
    Saliency FPS & \rankcell{1}{73.39} & \rankcell{3}{69.23} & \rankcell{4}{71.20} & \rankcell{4}{68.67} & \rankcell{2}{70.34} \\
    Visibility--saliency & 30.00 & 38.03 & 32.45 & 30.21 & 35.97 \\
    Hull + saliency + repair & \rankcell{3}{65.84} & \rankcell{1}{73.39} & \rankcell{1}{74.42} & \rankcell{3}{70.17} & \rankcell{4}{69.14} \\
    \midrule
    \multicolumn{6}{@{}l}{\textit{HOT3D object 2 (validation; 265 instances)}} \\
    Random vertices & \rankcell{2}{7.77} & \rankcell{5}{16.34} & \rankcell{5}{26.87} & \rankcell{5}{36.79} & \rankcell{5}{37.17} \\
    Centroid FPS & \rankcell{3}{7.58} & \rankcell{2}{20.04} & \rankcell{1}{34.83} & \rankcell{1}{43.70} & \rankcell{1}{44.23} \\
    PCA-extrema FPS & \rankcell{5}{6.75} & \rankcell{3}{18.98} & \rankcell{4}{29.09} & \rankcell{4}{40.08} & \rankcell{2}{41.77} \\
    Saliency FPS & \rankcell{1}{7.81} & \rankcell{1}{21.13} & \rankcell{2}{31.74} & \rankcell{2}{42.94} & \rankcell{4}{40.83} \\
    Visibility--saliency & \rankcell{4}{6.87} & 13.09 & 13.36 & 22.60 & 28.45 \\
    Hull + saliency + repair & 2.87 & \rankcell{4}{17.85} & \rankcell{3}{31.70} & \rankcell{3}{41.55} & \rankcell{3}{41.70} \\
\bottomrule
\end{tabular}}
\end{minipage}
\hfill
\begin{minipage}[t]{0.485\textwidth}
\vspace{0pt}
\resizebox{\linewidth}{!}{%
\begin{tabular}{@{}l|rrrrr@{}}
\toprule
Placement & 8 & 16 & 32 & 64 & 128 \\
\midrule
    \multicolumn{6}{@{}l}{\textit{HOPEv2 object 24 (validation; 10 instances)}} \\
    Random vertices & \rankcell{2}{99.00} & \rankcell{1}{100.00} & \rankcell{1}{100.00} & \rankcell{1}{100.00} & \rankcell{1}{100.00} \\
    Centroid FPS & \rankcell{2}{99.00} & \rankcell{1}{100.00} & \rankcell{1}{100.00} & \rankcell{1}{100.00} & \rankcell{1}{100.00} \\
    PCA-extrema FPS & \rankcell{1}{100.00} & \rankcell{1}{100.00} & \rankcell{1}{100.00} & \rankcell{1}{100.00} & \rankcell{1}{100.00} \\
    Saliency FPS & \rankcell{1}{100.00} & \rankcell{1}{100.00} & \rankcell{1}{100.00} & \rankcell{1}{100.00} & \rankcell{1}{100.00} \\
    Visibility--saliency & \rankcell{2}{99.00} & \rankcell{1}{100.00} & \rankcell{1}{100.00} & \rankcell{1}{100.00} & \rankcell{1}{100.00} \\
    Hull + saliency + repair & \rankcell{2}{99.00} & \rankcell{1}{100.00} & \rankcell{1}{100.00} & \rankcell{1}{100.00} & \rankcell{1}{100.00} \\
    \midrule
    \multicolumn{6}{@{}l}{\textit{T-LESS object 9 (test; 250 instances)}} \\
    Random vertices & 88.24 & \rankcell{5}{94.36} & 95.48 & 96.00 & 95.36 \\
    Centroid FPS & \rankcell{5}{91.64} & \rankcell{3}{95.68} & \rankcell{5}{96.16} & \rankcell{3}{96.64} & \rankcell{4}{95.72} \\
    PCA-extrema FPS & \rankcell{2}{93.40} & \rankcell{2}{96.32} & \rankcell{2}{96.84} & \rankcell{2}{96.84} & \rankcell{3}{96.36} \\
    Saliency FPS & \rankcell{4}{92.48} & \rankcell{4}{95.36} & \rankcell{4}{96.64} & \rankcell{4}{96.56} & \rankcell{5}{95.60} \\
    Visibility--saliency & \rankcell{1}{95.04} & \rankcell{2}{96.32} & \rankcell{1}{98.36} & \rankcell{1}{97.20} & \rankcell{1}{97.04} \\
    Hull + saliency + repair & \rankcell{3}{93.20} & \rankcell{1}{96.76} & \rankcell{3}{96.72} & \rankcell{5}{96.32} & \rankcell{2}{96.84} \\
\bottomrule
\end{tabular}}
\end{minipage}
\end{table*}

\section{HccePose training details}
\label{sec:hccepose_training}

We adapt the released HccePose(BF)~\cite{wang2025hccepose} ResNet34 training recipe.
We train one object-specific network for each of the 101 objects across HOPEv2, HANDAL, and HOT3D, using the same frozen per-object PBR image and instance populations as our DOPER baselines, without real target-dataset supervision.
Each encoder is initialized from ImageNet-pretrained ResNet34 weights.
Supervision comprises visible-object masks and hierarchical encodings of rendered front- and back-surface coordinates, preserving the original CAD coordinate frames, millimeter units, and camera calibration.

Input crops are resized to $256\times256$, with $128\times128$ output maps and a bounding-box padding factor of $1.5$.
We retain the released dynamic zoom-in augmentation, comprising scale jitter in $[0.75,1.25]$ and center shifts of up to $25\%$ of the bounding-box dimensions, together with its GDR-Net-style photometric augmentation and ImageNet normalization in BGR channel order.
Training uses Adam with a constant learning rate of $2\times10^{-4}$, $\beta=(0.9,0.999)$, and zero weight decay, with batch size 24, automatic mixed precision (AMP), and dynamic loss scaling on a single NVIDIA L40 per model.
The released adaptive hierarchical losses are retained, with objective $\mathcal{L}=3\mathcal{L}_{\mathrm{front}}+3\mathcal{L}_{\mathrm{back}}+\mathcal{L}_{\mathrm{mask}}$.

Each model receives a fixed budget of 50,000 training iterations, successful optimizer updates and AMP-skipped steps are recorded separately.
We use seed 42 and evaluate a deterministic held-out synthetic subset every 500 iterations to estimate P$n$P combination priors.

\medskip
\noindent\textbf{DOPER detections and box jitter.}
We keep the 101 fixed 50,000-attempt HccePose endpoints and replace CNOS/FastSAM boxes with detections from the frozen DOPER model bank, retaining the original detection scores.
DOPER keypoints and P$n$P do not filter these proposals.
Both variants scan every test image with every object model, using a confidence threshold of 0.15 and at most five detections per object and image.
The jitter variant revisits missing poses or poses with mean front-code reprojection error above 8 pixels.
It perturbs the box center by $\{-0.10,-0.05,0,0.05,0.10\}$ of its width and height and its scale by $\{0.85,0.90,1,1.10,1.15\}$, removing duplicate crops.
Among valid candidates that reduce reprojection error, it selects the largest front-code inlier fraction, breaking ties by reprojection error and then grid order; otherwise, a valid original pose is retained.
Each original detection produces at most one pose per variant.

\FloatBarrier

\section{Inference performance details}
\label{sec:appendix_inference_performance}
We time one DOPER model for T-LESS object~1 on 24 test images from six scenes.
For multi-view inference, each image is combined with three other views of the same scene.
Images are decoded and resident on the GPU, model loading and the initial upload are excluded.
After a warm-up pass over all images, we time each image four times and report the mean latency.
We benchmark on an NVIDIA DGX Station with one GB300 GPU, using PyTorch~2.13 and CUDA~13.0 at PyTorch's default precision (TF32 in convolutions).

\end{document}

%% file: sec/experiments_qualitative_figure.tex
\begin{figure}[t]
  \centering
  \includegraphics[width=\linewidth]{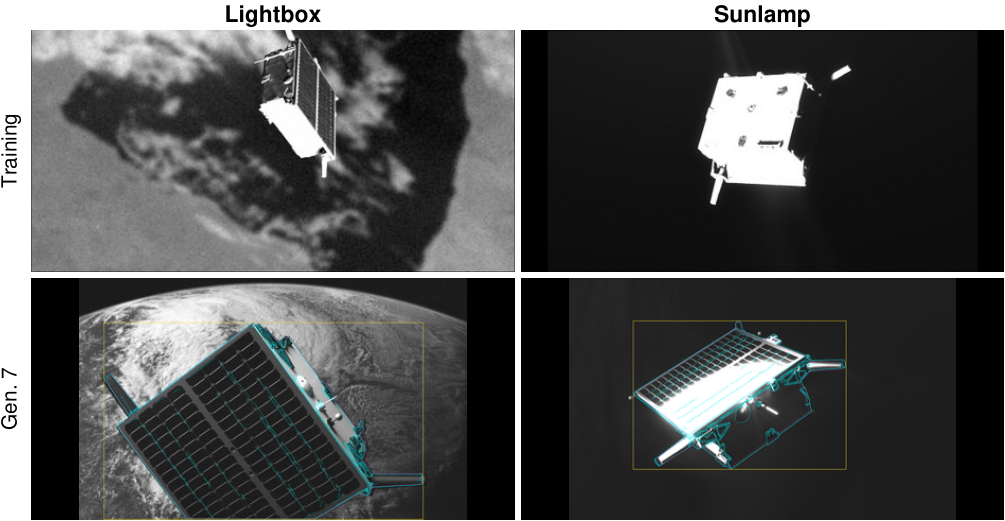}
  \setlength{\abovecaptionskip}{3pt}
  \caption{SPEED+ qualitative examples. Top: render and flare examples from the shared synthetic training pool. Bottom: generation~7 detections and poses on Lightbox (left) and Sunlamp (right).}
  \label{fig:speedplus_qualitative}
\end{figure}

%% file: sec/experiments_comparison_table.tex
\begin{table}[t]
  \centering
  \caption{SPEED+. \ranklegend{} marks the five lowest distinct scores per column among baselines and DOPER gen.\ 7; ties share a color.
  }
  \label{tab:speedplus_recent}
  \small
  \setlength{\tabcolsep}{4pt}
  \begin{tabular}{@{}lrr@{}}
    \toprule
    Method & Lightbox $\downarrow$ & Sunlamp $\downarrow$ \\
    \midrule
    \multicolumn{3}{l}{\textit{Detection-and-pose baselines}} \\
    TangoUnchained~\cite{esa2022lightbox,esa2022sunlamp} & \rankcell{4}{0.07350} & 0.08999 \\
    VPU~\cite{esa2022lightbox,esa2022sunlamp} & \rankcell{5}{0.10143} & \rankcell{5}{0.06119} \\
    lava1302~\cite{esa2022lightbox,esa2022sunlamp} & 0.16465 & \rankcell{2}{0.05886} \\
    u3s\_lab~\cite{esa2022lightbox,esa2022sunlamp} & 0.22401 & 0.14093 \\
    haoranhuang\_njust~\cite{esa2022postlightbox,esa2022postsunlamp} & \rankcell{3}{0.06525} & \rankcell{3}{0.05890} \\
    EagerNet~\cite{ulmer2023eagernet} & \rankcell{1}{0.03902} & \rankcell{4}{0.05909} \\
    SPNv2 + ODR~\cite{park2024spnv2} & 0.122\phantom{00} & 0.197\phantom{00} \\
    \midrule
    \multicolumn{3}{l}{\textit{DOPER training stages}} \\
    DOPER: synthetic & 0.2013\phantom{0} & 0.2018\phantom{0} \\
    DOPER + 10k renders & 0.12273 & 0.10342 \\
    DOPER + adaptation, gen.\ 0 & 0.07771 & 0.08223 \\
    DOPER + adaptation, gen.\ 4 & 0.05045 & 0.05683 \\
    DOPER + adaptation, gen.\ 6 & 0.04821 & 0.05673 \\
    \textbf{DOPER: gen.\ 7} & \rankcell{2}{0.04720} & \rankcell{1}{0.05775} \\
    \bottomrule
  \end{tabular}
\end{table}